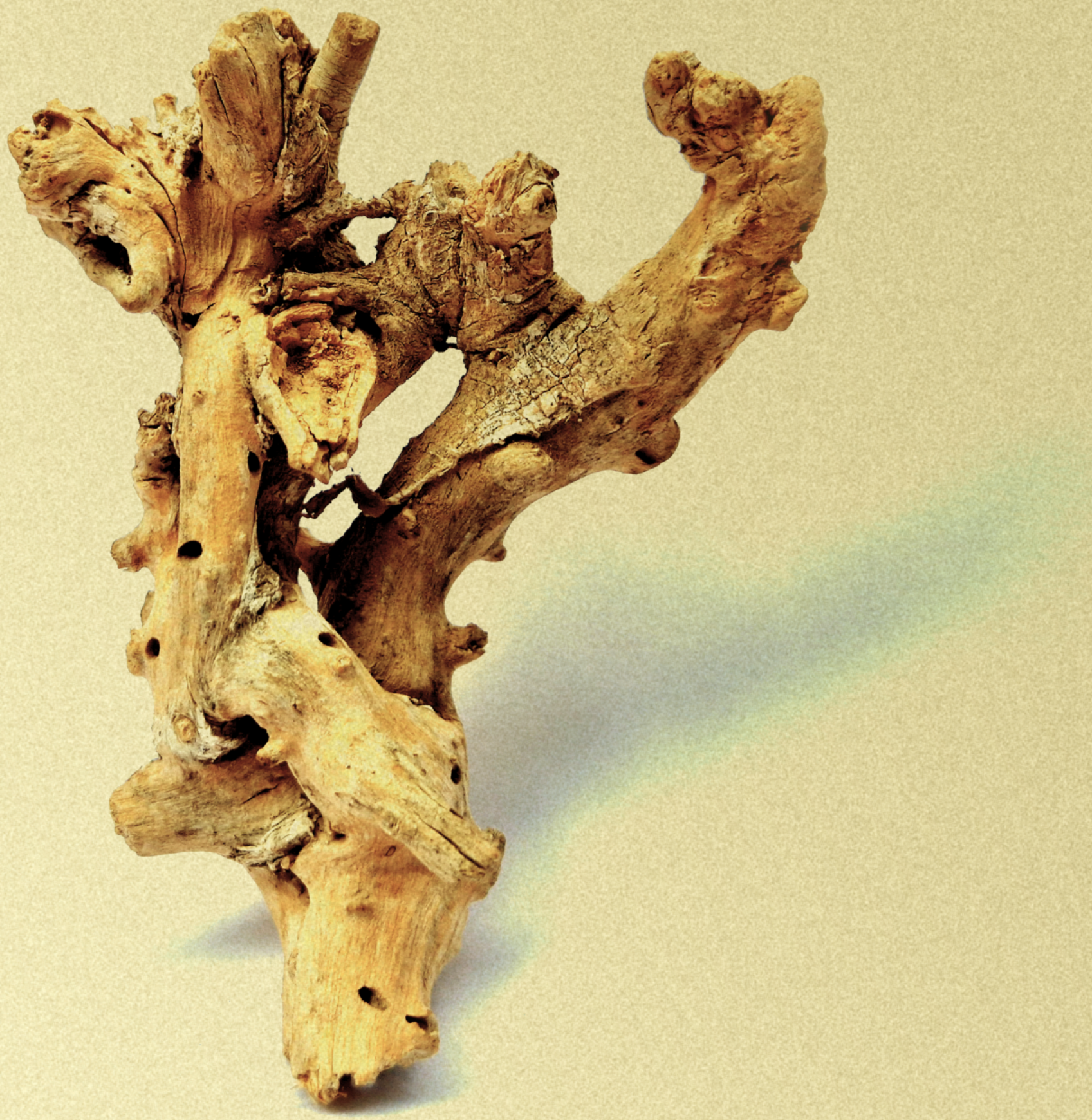

# Finding Structure in Language Models

JAAP JUMELET

# Finding Structure in Language Models

Jaap Jumelet

INSTITUTE FOR LOGIC, LANGUAGE AND COMPUTATION

For further information about ILLC-publications, please contact

Institute for Logic, Language and Computation
Universiteit van Amsterdam

| | |
|---|---|
| ADDRESS | Science Park 107<br>1098 XG Amsterdam |
| PHONE | +31-20-525 6051 |
| E-MAIL | illc@uva.nl |
| HOMEPAGE | illc.uva.nl |

# Finding Structure in Language Models

ACADEMISCH PROEFSCHRIFT

ter verkrijging van de graad van doctor
aan de Universiteit van Amsterdam
op gezag van de Rector Magnificus
prof. dr. ir. P.P.C.C. Verbeek
ten overstaan van een door het College voor Promoties ingestelde commissie,
in het openbaar te verdedigen in de Aula der Universiteit
op dinsdag 10 december 2024, te 14.00 uur

door

JAAP JUMELET

geboren te Utrecht

## Promotiecommissie

### Promotor

| | |
|---|---|
| Dr. W.H. Zuidema | Universiteit van Amsterdam |

### Co-promotor

| | |
|---|---|
| Prof. dr. R. Fernández Rovira | Universiteit van Amsterdam |

### Overige leden

| | |
|---|---|
| Prof. dr. R. Frank | Yale University |
| Dr. A. Alishahi | Tilburg University |
| Prof. dr. K. Sima'an | Universiteit van Amsterdam |
| Prof. dr. F. Roelofsen | Universiteit van Amsterdam |
| Dr. S. Pezzelle | Universiteit van Amsterdam |

Faculteit der Natuurwetenschappen, Wiskunde en Informatica

*Thus they went on living in a reality that was slipping away, momentarily captured by words, but which would escape irremediably when they forgot the values of the written letters.*

Gabriel García Márquez, *One Hundred Years of Solitude*

# Contents

*Reis ver, drink wijn, denk na*
*Lach hard, duik diep, kom terug*

Spinvis

# Acknowledgements

**A Brief Reflection** I have been incredibly fortunate to be given the opportunity of a fully open-ended PhD position by the ILLC. This position has allowed me to explore an intersection of topics that deeply fascinate me, and enabled me to connect the various directions that have been part of my studies in Utrecht and Amsterdam in exciting and novel ways. Before expressing my immense gratitude to the many people that supported me over the past years, I want to first briefly reflect on my academic path so far.

It is in Utrecht, during my bachelor in AI, that my fascination for the structure of language arose. Long before even knowing what a *recurrent neural network* was, we were already being exposed to exotic theories of syntax and semantics, involving complex proof deduction systems and set theoretic operations. And whereas most study programs in AI will nowadays focus solely on teaching programming in Python, Utrecht still spent far more time teaching us Prolog and Haskell instead. This formal approach to representing linguistic structure immediately clicked with me, but I remained puzzled by the extent to which research in this direction stayed separated from a line of research that gained a lot of traction during that time: *deep learning*.

During my master in Amsterdam—which focused much more on deep learning—I kept therefore looking for research directions that aimed to create a connection between 'old-school', *symbolic* approaches to language and novel, *sub-symbolic* approaches that were beginning to take off around 2017. I wonder where I would be today if Dieuwke Hupkes had not decided to discuss the paper of Linzen et al. (2016) in one of the NLP courses, and if I had not reached out to her to work in this direction, which resulted in a wonderful BlackboxNLP publication. Dieuwke's research area was exactly what I was looking for: deep learning models were now the main focus, but linguistic theory still played a vital role in *interpreting* these models. The guidance of Dieuwke and Jelle in the final year of my master has been fundamental in shaping my love for research and I am immensely grateful for that.

At that time, around 2018, we could not have anticipated how important *language models* would become in only a few years time. The models we worked with then were over a thousand times smaller than current models like ChatGPT, and the field of interpretability made up only a small niche of the NLP community. Through the introduction of models like BERT and GPT, language models rapidly took the center stage in NLP and all of a sudden interpretability research became one of the hottest topics in AI. Being a PhD candidate during these years has been a blessing and a curse. While it has been great how important this research direction has become and how much public interest there has been in AI, it has become challenging to keep up with incessant stream of novel papers, and advancements in the field seem to be driven more by the commercial interests of big tech companies instead of ensuring that AI is being deployed and developed in an ethical and fair way. Nonetheless, I am excited to continue my academic journey in this direction and remain optimistic that AI research will be of tremendous help to provide answers for fundamental questions in linguistics and other scientific disciplines.

**A Word of Thanks** This thesis would not have been possible without the incredible guidance of my supervisor. **Jelle**, thank you for your wisdom, your patience, your kindness, and for encouraging me to tackle challenging topics that were not always appreciated by our lovely reviewers, but resulted in a body of work that I can look back on proud and fondly. When the ILLC offered me this PhD position, I didn't hesitate for a minute who my supervisor should be, and I am glad our collaboration turned out exactly as I envisioned it to be back then. I hope our paths will cross many times in the future, and would love to go on a sailing trip with the group some time on a day where there is finally some wind! **Raquel**, I am very glad we got to work on a project together in the first half of my PhD: I learned an immense amount from your guidance back then and I am very proud of the paper that came out of it. Thank you for your co-supervision, and for being such a kind and inspiring presence within the ILLC. Thank you to my defence committee: **Afra**, **Floris**, **Khalil**, **Robert**, and **Sandro**, thank you all for taking the time and effort to assess my work. A special thanks to Khalil, who was also part of the committee that offered me my PhD position, and whose course I have been able to assist three years in a row.

The work in this thesis would also not have been possible without the many fantastic collaborators I have worked with over the past few years. A big thanks to **Jakub**: while you never were my supervisor in any official capacity, we still ended up collaborating for the first and last paper of my PhD and I learned an immense amount from your expertise. Thank you for inviting me over to live in Trento for three months: being able to conduct research in such a wonderful environment has been one of the highlights of my PhD! And thank you for introducing me to Milica and Shane! **Milica**, thank you being part of my first big collaboration back in 2020: your expertise has been vital for shaping the ideas in that lovely paper. **Shane**, I really cherished our online meetings (and thanks for making me jealous many times about your awe-inspiring alpine trips); I am super happy we managed to set up such a great collaboration despite the 9 hour time difference. Of course this would not have been possible without the effort that **Abhinav** put in: thank you for that and I look forward to finally meeting each other in person in Miami! **Janie**, thank you so so much for having been an integral part of my PhD: our many late-night brainstorms have been hugely important for shaping many of the research ideas in this thesis, and it has been an absolute joy working together with you. **Lisa**, thank you for joining my project with Jakub and being such a great and inspiring collaborator, I hope we will get to work together on many future projects! **Lucas**, thank you for involving me in your research projects: your ideas have had a big impact on my own research and I am glad I was able to work alongside you. A big thanks to **Elia** and **Dieuwke** as well for this: it was great that we still got to collaborate this way over the past few years. And Dieuwke, I can not thank you enough for your guidance in the first few years of my academic journey: this thesis and PhD would not have been possible without you.

My PhD would have been a rather lonesome endeavour if it weren't for my many lovely colleagues at the ILLC. **Oskar**, I am so glad we started our PhDs at the exact same time: I have been really blessed with such a nice and fun colleague. Our trips to Singapore and Lisbon together have been some of the absolute highlights of my PhD! **Marianne**, thank you for being such a loving and thoughtful presence within the CLC lab, and for inspiring me to focus more on CogSci-related topics in my work. **Charlotte**, you have been a fantastic addition to the CLC group and I am really excited that we get to travel to Miami together soon! I want to thank my (former) colleagues at the CLC lab for making it such a nice and pleasant place to work in: **Bas**, **Samira**, **Tom**, and **Erman**, thank you for all the great discussions and fun lab outings. **Joris**, thank you for all the inspiring chats over the years, and for being a great travelling partner. Thank you to **David** for that as well, that trip to Oman the three of us had together is something I will never forget and I am glad I was able to give it a special place on the cover of this thesis. **Ece**, thank you for giving me the honour of being your paranymph and for all the lovely chats we have had over the years, it was great having a fellow guitar player at the ILLC! **Seth**, even though our PhDs overlapped only for less than a year, I am very glad I found another colleague with a similar love for nature and outdoor sports; I hope our paths will cross outside of a conference or office as well at some point!

Thank you to **Rochelle**, **Mario**, **Michael**, **Esam**, **Vera**, **Anna**, **Adi**, **Alina**, **Marcel**, **Zhi** and all the other PhDs and post-docs at the ILLC for being such awesome colleagues and making it a joy to come into work every day. Thank you to **Katia**, **Wilker**, **Jelke**, **Katrin**, **Raquel**, **Robert** and all the other ILLC staff for making the institute such an incredible and inspiring place to work at. Thank you to **Sabijn**, **Jan**, **Anna**, **Angela**, **Zoë**, **Tom**, and **Sylke** for choosing to do their thesis with me: I hope my enthusiasm for your research topics made your thesis a pleasant and inspiring experience! Thank you to **Gabriele** and **Hosein** for all the great chats and interpretability insights: I am glad we got to collaborate on a paper together and I am sure we will be co-authors on a future paper again. Thank you to all the awesome people from outside the ILLC I have been fortunate to meet in Abu Dhabi, Singapore, Malta, Barcelona, and Lisbon: I look forward to meeting you all again at whatever exotic conference location ACL comes up with next. Thank you to the lovely people at CIMeC who made my stay in Trento a fantastic experience: **Roberto**, **Raffaella**, **Leonardo**, and **Davide**, I would love to visit Trento again! Special thanks to **Alexey**: it was great sharing the apartment in Trento with you and I hope we can find a time to go climbing together again at some point. Thank you to **Clara** for inviting me over for a talk in Zurich! Thank you to my co-organisers of **BlackboxNLP** for making my first experience with workshop organisation such a valuable and fun experience. Thank you to **Reviewer #2** for ruining my day many times, but ultimately forcing me to look back critically on my research ideas and shaping them into something better. And finally a very special thank you to **Arianna** for providing me with the next big step in my career, I am incredibly excited to start our collaboration soon.

Next, I would like to thank my incredible friends and family for their support over the past few years. I'll start with two dear friends who I have known for almost my whole life, and have agreed to be my paranymphs. **Emile** van Krieken et al. (2024), thank you for being an amazing friend, for having someone to complain to about the state of academia, and for the hundreds of concerts and festivals we've got to visit over the years. **Jop**, thank you for all the fantastic bike rides, the inspiring conversations about life *onder het genot van een pintje*, and all the fun we've had during our many trips abroad. To both of you: I realise how special it is to have known two great friends for such a long time, and I am excited to see where life will take us next.

Thank you to my dear friends from **Incognito** for keeping me sane. Thank you to **Tjeu** for coming over to visit me in Trento, for all the boulder sessions and beers afterwards, and for always being in the mood for a wack adventure. Thank you to **Pim** for all the great bouldering as well: getting to hang out with you and Tjeu weekly was a great way to de-stress from all the academic nonsense. Thank you to **Aafke**, **Mayon**, **Ted**, and **Eline** for being the most loving group of friends I could wish for, and for all the amazing *pretpark*-related trips over the years. Thank you to **Tariq** for all the classy train rides home, and for that amazing trip to Gijón. Thank you to **Tim** for being "*out of his element*", and without whom this thesis would probably be less littered with Big Lebowski quotes. Thank you to **Mark** for all the loud gigs we've attended together and all the great chats on AI. Thank you to **Perry** and **Thomas** for letting me win that beer mile, I'm sure you'll get your revenge soon. Thank you to **Stijn**, **Niels**, and **Mark** for all the band rehearsals: *Kadavermout* will live forever. And finally, a very special thank you to **Richard**, **Kambiz**, **Melissa**, **Nienke**, **Nina**, **Lara**, **Niels**, **Lydia**, and **Merel** for all the amazing vacations over the years, I hope we can keep up this tradition for a long, long time.

Thank you to my amazing friends from high school! **Olaf** and **Luc** Kootte (2023), thank you for all the amazing alpine adventures, I look forward to climbing many more incredible mountains with you. **Lennart**, thank you for all the great years together under a roof in gloomy Lunetten. Thank you to **Maaike** for all the lovely festivals and concerts we attended together! And a massive thank you to **Jurriaan**, **Martin** van der Schelling et al. (2024), **Joris**, **Nino**, and **Marc** for all the incredible vacations and ski trips, I cherish our travels together immensely.

Next, I want to thank my family for all their support over the past few years. To **Hans** and **Marianne**, and all my dear **aunts**, **uncles** and **cousins**: thank you for all your interest in my work, checking in with me how it's all going, and your support during 2022. Thank you to **Vida** for being the cutest dog alive. Thank you to **Marijn** for all the laughs and support, it has been inspiring to see you thrive in all those wonderful communities abroad. Thank you to my loving **parents** for their guidance, wisdom and support, the career advice, the life advice, the trips to Zeeland and Vlieland, and, most importantly, providing me an intellectual and emotional base that allowed me to tackle the challenge a PhD can sometimes be. And finally, finishing off with the most important one of all, thank you to **Simone** for all your care and support. Having you by my side enabled me to grow and make the best out of my PhD, and above all, showed me that love is far more fulfilling than academic success can ever be.

Utrecht, Jaap Jumelet
October 2024

*That rug really tied the room together, did it not?*

Walter Sobchak

# 1 Finding Structure in Language Models

When we speak, write or listen, we continuously make predictions based on our knowledge of a language's grammar. Children acquire grammatical rules within just a few years, and we can almost instantly judge whether a sentence is well-formed. These grammatical intuitions are fundamental for making sense of the world and play a crucial role in our cognitive processes when speaking, reading, or listening to language. The central question this thesis explores is whether artificial language models possess similar notions of grammatical structure and how such structures can be identified within these models.

For decades, the field of Natural Language Processing (NLP) has sought to develop systems that understand language with human-like proficiency. Early approaches primarily focused on rule-based systems, where sentence interpretation was achieved through a series of symbolic transformations. In these systems, grammatical structure was represented *explicitly*, often using hierarchical *parse trees* from which the meaning of a sentence could be derived. While the interpretable nature of these linguistic representations proved invaluable for constructing large-scale models of grammar, they required extensive expert annotation and struggled to generalise to unseen constructions.

The current paradigm in NLP no longer relies heavily on expert annotation, embracing an *unsupervised* and *distributed* approach to language representations through neural networks, most notably in the form of large language models. This approach has led to significant performance gains, but it has come at the cost of transparency: grammatical structure is now represented *implicitly* and is acquired by a model without any explicit supervision. This shift raises an interesting question: do these latent notions of grammar align with the explicit grammatical representations of the previous paradigm?

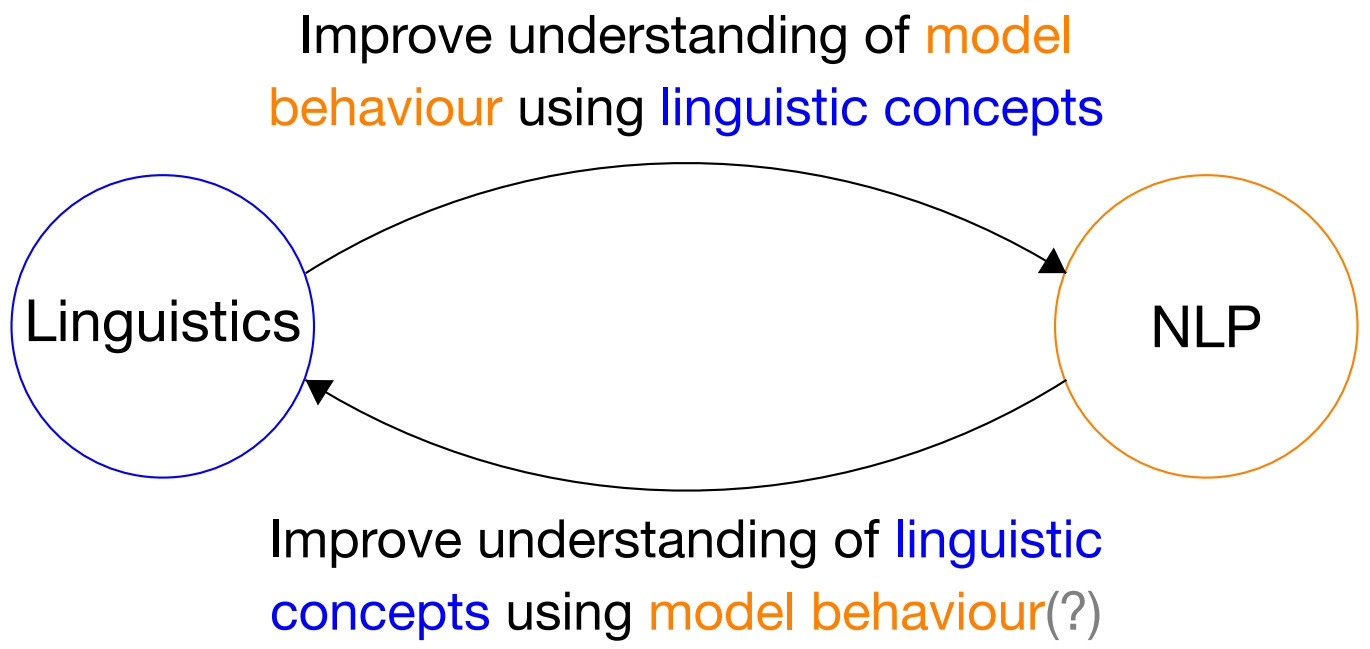

**Figure 1.1:** The interplay between the fields of linguistics and NLP: linguistic concepts are highly useful for interpreting model behaviour, but to what extent can we reverse this direction?

The remarkable success of **language models** in recent years has led to an intriguing interplay between NLP and the cognitive sciences (Figure 1.1). On the one hand, linguistic theory has been crucial for analysing language model behaviour (Opitz et al., 2024), as explored extensively throughout this thesis. The opposite direction, in which findings from NLP are used to improve our understanding of concepts from cognitive science, is slowly taking shape (Baroni, 2022). However, the extent to which artificial models can inform theories of human language processing remains an open question. In this thesis, we offer some modest suggestions on how to advance in this direction. The interplay between these fields not only deepens our understanding of how language models operate but also challenges existing theories in cognitive science by offering new perspectives and tools (Kallini et al., 2024). As we explore these connections, it becomes increasingly important to examine whether the patterns and structures learned by language models reflect those inherent in human cognition, or if they are merely artefacts of the data-driven methods used in NLP.

In the remainder of this chapter we provide a background to various core topics of this thesis. First, we present a brief overview on language models (§1.1), which will be our main object of study. Next, we describe the field of interpretability (§1.2), which attempts to develop methods for gaining a better understanding of the inner workings of neural network models. Finally, we present an overview of related work that attempts to find linguistic structure in language models (§1.3), which will provide a context for the experiments described in this thesis. We conclude with a general overview of the thesis (§1.4).

## 1.1 Language Models

Until around 2018, the field of NLP deployed different kinds of models for solving the different tasks that belong to NLP, such as machine translation, semantic parsing, and named entity recognition. This changed with the advent of *pre-trained* language models (LMs), in particular with the introduction of ELMo and BERT (Peters et al., 2018; Devlin et al., 2019). Instead of optimising a task from scratch using supervised learning, we first perform an unsupervised pre-training procedure that allows a model to gain a general understanding of language, before *fine-tuning* it on a task of interest. This pre-training is done using a language modelling objective, which is the task of *next-word prediction*: given a *context* the model has to compute a probability distribution over the next words. Language models exist in two flavours.

**Masked** language models predict a *masked* token with access to the full sentence:

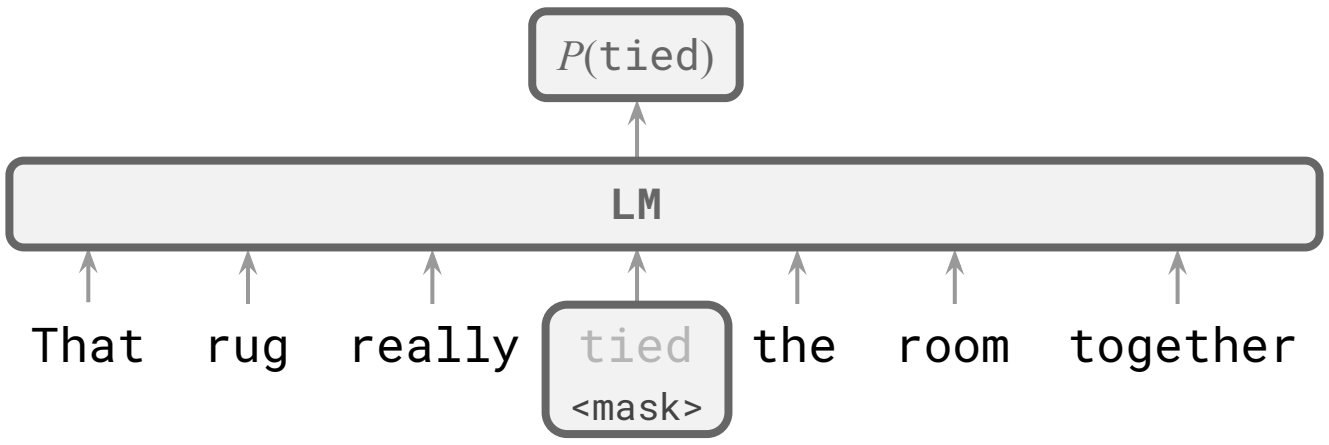

Masked LMs are useful for performing sentence-level tasks, for example when predicting the sentiment of a sentence. A downside of masked LMs, however, is that the masked word prediction task in itself does not find many purposes, and does not allow for *text generation*. Furthermore, the computational overhead that fine-tuning requires has led this model type to become less popular in NLP research in recent years. Nonetheless, we will investigate masked LMs in Chapter 2 and 8, as their ability to encode a full sentence provides an interesting test bed for investigating the structure building abilities of artificial models.

**Causal** language models, also referred to as *left-to-right* or *auto-regressive* language models, predict the next-word only based on the sentence seen up to that point:

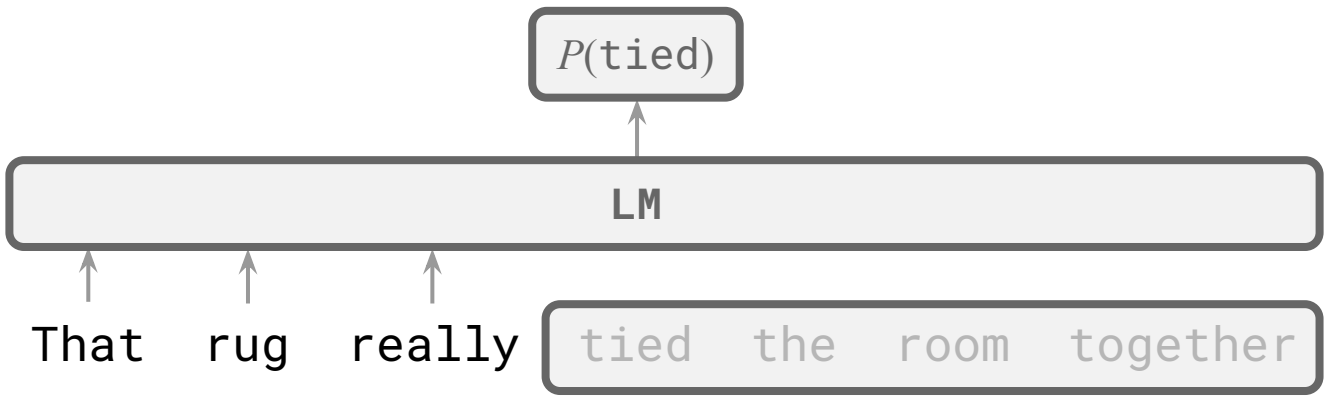

By predicting the next word one step at a time, causal LMs are able to *generate* text by sampling a word from its predicted distribution over the next word, using that sampled word as input, computing the next distribution, and then iteratively repeating this procedure. Causal LMs have become so powerful in recent years that many NLP tasks are nowadays framed as next-word prediction tasks as well. In this framework, we no longer need to fine-tune a model in a supervised way for it to perform a specific task, but instead we can formulate the task directly as a *prompt* to the model (Brown et al., 2020). For harder tasks we can also provide a number of demonstrations to the model as part of the prompt for it to gain an understanding of what task should be performed: this is called *in-context learning* (Dong et al., 2023). Another important development is *model alignment*, which has two purposes: i) the output of a model is aligned with human values by preventing them from generating undesirable content using *reinforcement learning from human feedback* (Ouyang et al., 2022), and ii) the model is trained to solve complex tasks by a supervised learning procedure called *instruction tuning* (Mishra et al., 2022). This procedure is performed after the pre-training phase, and has been fundamental for creating the strong performance of dialogue-driven language models such as ChatGPT. The majority of the experiments in this thesis are performed on causal LMs, also in part because the incremental processing nature of these model is more similar to human language processing.

The success of language models has gone hand in hand with the introduction of a neural network architecture called the **Transformer** (Vaswani et al., 2017). Before this, the most common model architectures were **recurrent** neural networks (RNNs), such as the Long Short Term Memory network (LSTM, Hochreiter and Schmidhuber, 1997). RNNs process input incrementally by writing to and reading from a high-dimensional *hidden state*, which is a dense representation of the input seen so far. Transformers, on the other hand, are build around a *self-attention* mechanism that allows a model to attend back to all input features directly, enabling them to model *long-distance dependencies* more accurately:

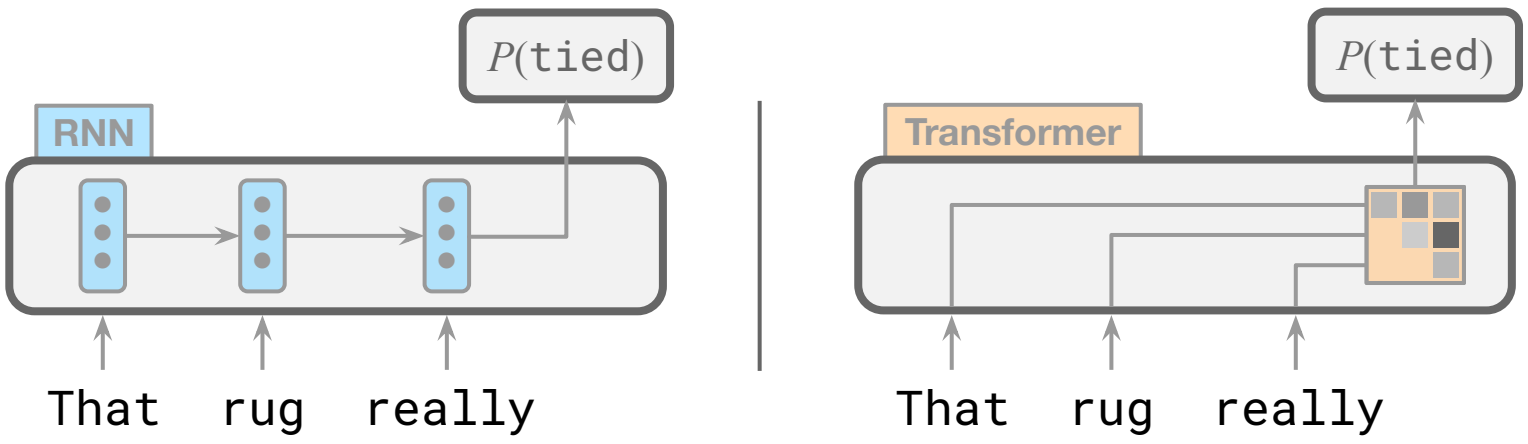

Another benefit of this approach is that it allows model predictions to be computed in parallel, whereas recurrent predictions must be computed sequentially. Although the recurrent processing mechanism is arguably more *cognitively plausible* (Beinborn and Hollen-stein, 2023), Transformer models nowadays have set state-of-the-art performance in most NLP tasks. Most of our experiments focus on Transformer models, but we investigate and compare LSTM models in Chapter 5, 6, and 7.

As can be seen in these examples already, predicting the probability of a next word is a task that depends on a complex interaction of *syntactic*, *semantic*, and *pragmatic* factors. Syntactically, a model must determine whether a word fits grammatically in the sentence, and whether constraints such as number and gender agreement are satisfied. Semantically, a model must determine whether the meaning of a predicted word aligns with the context: cats are more likely to *meow* than *bark*. Pragmatically, a model must be able to take contextual cues into account, and understand the intended meaning of a speaker. These factors will all be a topic of investigation in this thesis, with a specific focus on the syntactic abilities of language models. We provide an overview of related work in this direction in §1.3.

## 1.2 Interpretability

With neural networks having become the dominant modelling paradigm in most AI-related disciplines, a new research field has been formed that aims increase the **interpretability** of these models. Neural networks are notoriously hard to interpret: their high-dimensional dense representations are highly opaque, and require additional tools to make their content interpretable. A wide range of such tools have been introduced in recent years, and significant progress has been booked in reducing the *black box* nature of these models. This need for transparency comes from a wide range of **desiderata** (Doshi-Velez and Kim, 2017; Lipton, 2018). To ensure the *fairness* of a model, for example, we need to know whether and how stereotypical biases are encoded. To guarantee the *trustworthiness* of a model that is used in production, we need know if it behaves *robustly* on unseen data. And to ensure the *faithfulness* of a model, we need to make sure it bases its decisions on the right rationale, i.e. it should not be *right for the wrong reasons*. The need for interpretability in this thesis comes mainly from an analytical perspective: in order to determine a model's comprehension of a particular phenomenon or concept, we need tools that accurately capture and translate that comprehension to human-interpretable terms.

The desideratum of **faithfulness** plays not only a role on the side of the target model: the generated explanations should in turn be faithful to the model's underlying reasoning (Jacovi and Goldberg, 2020). An important distinction to be made here is that an explanation's *plausibility* does not imply whether it is faithful: to a human observer a particular rationale might seem highly plausible, while in fact the models actual reasoning relied on different cues. It is highly challenging, however, to determine whether an explanation is truly faithful: if we would know its inner workings there would not be a need for interpretability, after all. Various approaches have been introduced in recent years to quantify the faithfulness of interpretability methods, by introducing explanation techniques that are better rooted in concepts from the *causality* literature (Elazar et al., 2021) or by training models in a more controlled environment (Bastings et al., 2022). Nonetheless, it remains an open question under what conditions an explanation will be adequately faithful, which is exemplified by the large degree of disagreement between related explanation methods (Atanasova et al., 2020). Our concerns regarding the faithfulness of interpretability techniques will be discussed in more detail in Part 3 of this thesis, where we present a methodology for quantifying the faithfulness of feature interaction methods.

### 1.2.1 Interpretability Methods

The highly complex nature of neural models should be approached by explanations at differing levels of abstraction, analogous to the study of human cognition (Zednik, 2021). We can take inspiration from cognitive science in explaining complex systems: Marr's Tri-Level Hypothesis (Marr and Poggio, 1976) famously introduces three levels at which information processing systems can be analysed. At the *computational* level we examine what a system *does*, what problems does it solve or overcome? Then, at the *algorithmic* level we ask *how* does a system do what it does, and what representations does it use for this? Finally, at the *implementational* level we ask how a system is physically realised, and what neural circuitry it uses to implement this. We can use these three levels as guidance for forming an overview of the different levels of abstraction at which we can explain artificial models (Ferrando et al., 2024).

**Behavioural Interpretability** By observing the predictions of a model on carefully constructed stimuli we can already gain a comprehensive understanding of its general behaviour (Ribeiro et al., 2020). Evaluations at this *computational* level require no access to model internals, and are therefore highly useful for comparing models with different kinds of internal architectures. Behavioural tests are commonly used in the analysis of the linguistic abilities of language models. The next-word prediction objective of LMs can be leveraged directly to generate *acceptability judgements*, by comparing its predictions on minimal pairs of a grammatical and ungrammatical phrase:

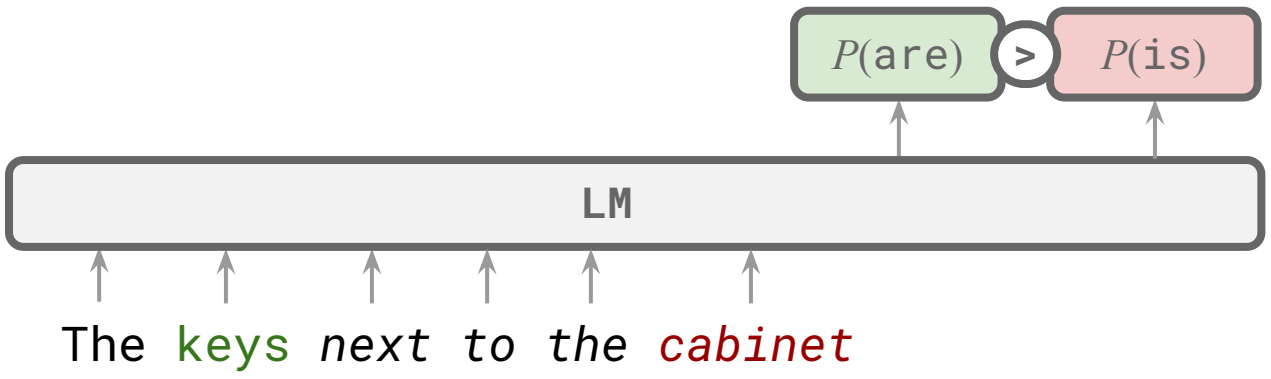

In this example we are testing a model's notion of subject-verb agreement by comparing its predicted probability of a verb that agrees with the subject number (*are*) and the probability of a verb of the incorrect number (*is*). This type of analysis is called *Targeted Syntactic Evaluations* (TSE, Marvin and Linzen, 2018) and will be covered in more detail in §1.3. Behavioural interpretability plays an important role in this thesis, and in both Part 1 and 2 we will make extensive use of TSE analyses.

**Probing** Behavioural tests inform us of *what* a model can do, but do not explain *how* a model performs a task. When interpreting LMs at the algorithmic level, we need tools that explain model behaviour at a more granular level. Various approaches at this level exist, we highlight two of these here. Staying at the example of subject-verb agreement, we may ask how a model has *encoded* its notion of grammatical number: if it is able to predict the verb of the correct number, it must have encoded the number of the subject in its representations. Probing methods, also known as *diagnostic classifiers* (Hupkes et al., 2018), train auxiliary classifiers on top of model representations to uncover a model's notions of more abstract concepts such as gender, number, or part-of-speech information (Belinkov, 2022):

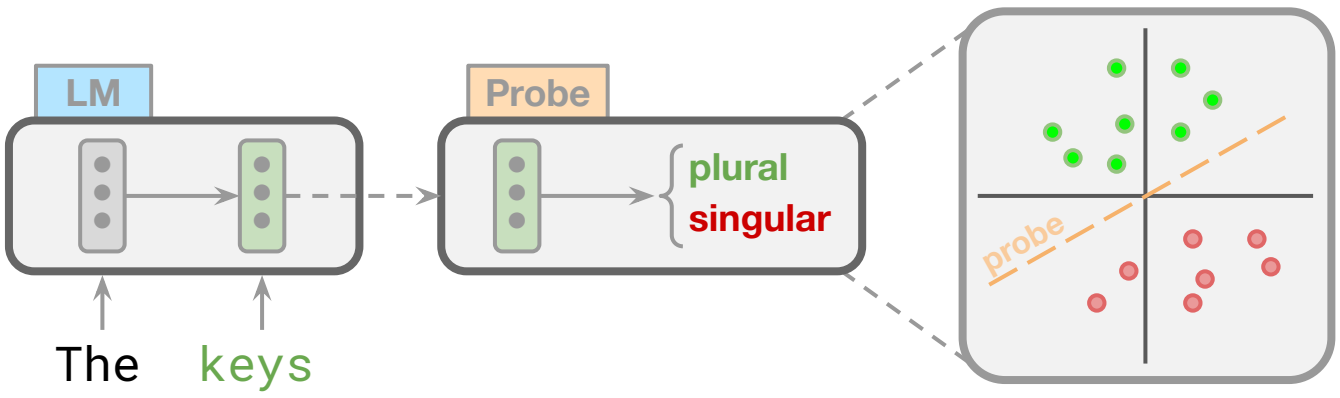

In this example, we would train a linear classifier on a collection of plural and singular noun representations to predict their number. The resulting decision boundary then shows us which *units* in the representation were most important for encoding grammatical number.

An important caveat of probing methods is that high predictive accuracy does not imply that a model is actively using that concept in its predictions: classification alone does not provide a *causal* account and only demonstrates that a concept *can* be decoded. Various procedures have been introduced that uncover the extent to which a concept plays a role in model predictions, by *erasing* that concept from the representations (Giulianelli et al., 2018; Elazar et al., 2021; Belrose et al., 2024). We introduce a related method in Chapter 6, in which we 'verbalise' the decision boundary of a probe by computing its similarity scores to the decision boundaries of the LM vocabulary.

**Attributions** Another common interpretability technique at the algorithmic level are *feature attributions* that explain a model prediction in terms of the strongest contributing input features. These kinds of methods allow us to locate the source of a prediction more precisely and provides insight into *why* a model gave a particular prediction. Feature contributions are often expressed as relative importance scores, that signify whether an input feature boosted or suppressed the final prediction:

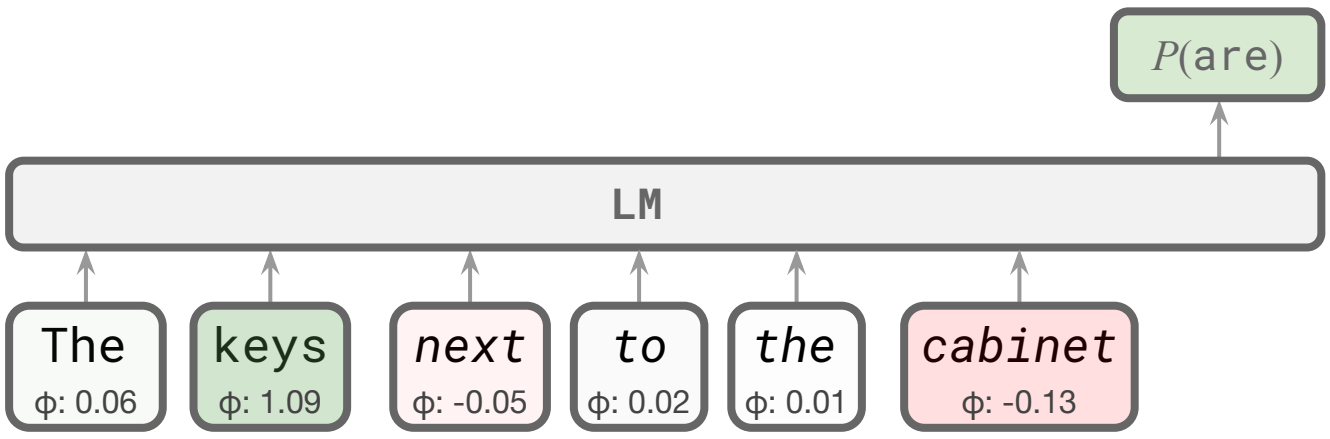

In this example, we can see that the subject *keys* had a positive contribution to the prediction of the verb, whereas the intervening prepositional object *cabinet* had a negative contribution.

A wide range of feature attribution methods has been introduced, each based on different intuitions about how to express feature importance (Ancona et al., 2018). The majority of these methods depend on quantifying the impact of *feature removal* (Covert et al., 2021): how much does a prediction change when we remove a feature from the input? This procedure is often done by replacing input features with a neutral baseline, or based on specific *counterfactuals* that explain behaviour in terms of "Why did the model predict *X instead of Y*?" (Yin and Neubig, 2022). The differences between attribution methods, however, have led to a pressing issue of *disagreement* between explanations (Atanasova et al., 2020; Neely et al., 2022): how do we determine the 'right explanation' when two explanations disagree? Furthermore, questions have been raised to what extent a complex model decision can be explained in terms of a simple sum of importance scores, or whether we need richer explanation structures that take feature interactions into account (Janizek et al., 2021). We will investigate these issues in more detail in Chapter 7.

**Mechanistic Interpretability** At Marr's implementational level we can investigate a system in terms of its 'physical' components. Research in this direction has become known as *mechanistic interpretability* (Olsson et al., 2022), and explains models at the most fine-grained level. The components of Transformer models that are leveraged for explanations of this kind include the self-attention mechanism (Abnar and Zuidema, 2020; Mohebbi et al., 2023), the *residual stream* (Belrose et al., 2023; Merullo et al., 2024), and the feed-forward layers (Meng et al., 2022; Geva et al., 2023). Another popular explanation method is that of *circuit finding* (Olah et al., 2020), in which small sub-circuits are isolated that are responsible for a particular model behaviour:

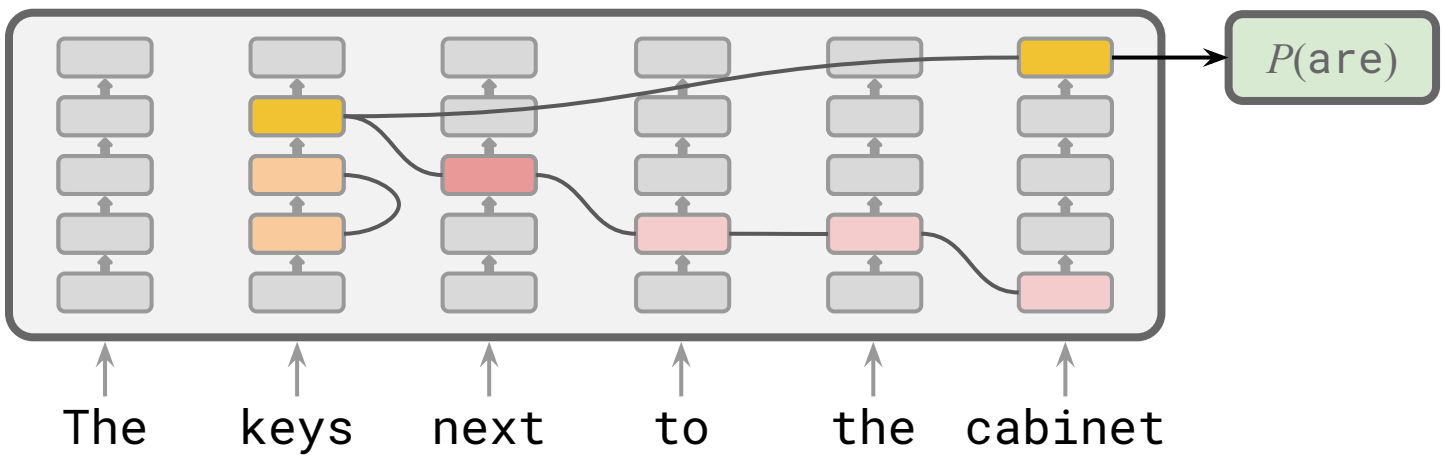

In this (fictional) example we would locate a circuit for subject-verb agreement that identifies the subject number and suppresses number information of intervening distractor nouns. Finding such a circuit is a computationally challenging task, but various procedures have been proposed to automate this (Conmy et al., 2023). The circuits for a wide range of phenomena have been found in language models, such as indirect object identification (Wang et al., 2023), greater-than arithmetic (Hanna et al., 2023a), and the encoding of gender bias (Chintam et al., 2023). Mechanistic interpretability methods do not play an important role in this thesis, but present an exciting direction for future work in linguistic investigations of language models.

## 1.3 Linguistic Structure in Language Models

In the previous section we presented various interpretability techniques that have been introduced for gaining a better understanding of model behaviour. In this section we dive deeper into the various linguistic investigations that have been conducted on language models, and present an overview of the linguistic phenomena that have been covered. Note that this section does not aim to be exhaustive: a more extensive overview is provided by Chang and Bergen (2024).

The increasing capacities of neural language models in recent years have led to a surge in research into their representation of language on a fine-grained linguistic level (Alishahi et al., 2019; Tenney et al., 2019a; Rogers et al., 2020, i.a.). A common approach to examining language models is to consider them as '*psycholinguistic subjects*'; by testing hypotheses derived from psycholinguistics we are able to determine to what extent language models process language similarly to humans (Futrell et al., 2019; Ettinger, 2020; Linzen and Baroni, 2021).

To assess the linguistic knowledge of LMs, a range of tools have been deployed. For instance, through probing experiments we can investigate whether LMs encode certain linguistic properties such as number information (Conneau et al., 2018), part-of-speech tags (Tenney et al., 2019b), number information (Hanna et al., 2023b), or constructional information (Madabushi et al., 2020; Li et al., 2022; Weissweiler et al., 2023). Syntactic dependencies have been shown to be decodable from model representations with reasonable accuracy using *structural probes* (Hewitt and Manning, 2019; White et al., 2021) that reconstruct a minimum spanning tree from linearly transformed representation distances. Structural probes were originally developed for recovering dependency tree information, but constituency structure has also been shown to be decodable (Arps et al., 2022; Zhao et al., 2023), as well as incremental parsing strategies (Eisape et al., 2022). Alternative techniques have also been developed to recover constituency tree information without auxiliary probes, by targeted masking of linguistic information (Wu et al., 2020).

Behavioural approaches to assessing linguistic competence of LMs have focused on Targeted Syntactic Evaluations, as described in the previous section, in which the LM's output probabilities are compared on a minimally different pair of a grammatical and an ungrammatical sentence (Marvin and Linzen, 2018; Hu et al., 2020). This procedure makes it possible to investigate a model's knowledge of specific linguistic phenomena without probing the model's internal representations. A wide range of linguistic phenomena have been tackled this way, such as subject-verb agreement (Linzen et al., 2016; Gulordava et al., 2018; Lakretz et al., 2019), negative polarity items (Jumelet and Hupkes, 2018; Warstadt et al., 2019a; Bylinina and Tikhonov, 2022; DeCarlo et al., 2023), and filler-gap dependencies (Wilcox et al., 2018; Suijkerbuijk et al., 2023). Various benchmark suites have been developed that provide a broad coverage approach to these evaluations, such as BLiMP (Warstadt et al., 2020) and SyntaxGym (Gauthier et al., 2020).

Results from probing, Targeted Syntactic Evaluations and other evaluation paradigms can certainly be viewed as providing converging evidence that modern neural LMs learn non-trivial structural, linguistic knowledge and do not just memorise fragments of texts from the data and simple sequential dependencies. However, although converging, the evidence is not yet conclusive: each of these evaluation paradigms has also been found to occasionally produce false positives. In probing, for instance, a well-known risk is that probes pick up information represented in the internal states of the language model, but not causally involved in the predictions of the model (Hewitt and Liang, 2019a; Voita and Titov, 2020). In Targeted Syntactic Evaluations, the strength of the evidence depends on the quality of the set of alternative explanations that is considered, which ultimately is a matter of judgements and differs for different linguistic constructions (Vamvas and Sennrich, 2021; Newman et al., 2021). Several studies have provided other challenges, including work pointing out the indifference of LMs towards word order (Sinha et al., 2021a, i.a., although this has been contested by Abdou et al. (2022)), their reliance on spurious heuristics (Lovering et al., 2021), and their difficulty in dealing with negation (Ettinger, 2020; Kassner and Schütze, 2020). These studies demonstrate the need for a rigorous evaluation of the interpretability methods themselves as well.

A recent line of work focuses more strongly on the connection between the training data distribution and subsequent linguistic generalisations (McCoy et al., 2020; Yedetore et al., 2023; Mueller and Linzen, 2023; Wilson et al., 2023). We highlight two studies here as example cases. Misra and Mahowald (2024) investigate the acquisition of rare linguistic constructions in LMs, focusing in particular on the English Article + Adjective + Numeral + Noun construction (*a beautiful five days*). By ablating specific phrases from the training data, they are able to pinpoint with precision the conditions that are necessary for acquiring an understanding of such rare constructions. Leong and Linzen (2024) use a similar methodology for investigating the passive construction in English, zooming in on the acquisition rare exceptions (**One hour was lasted by the meeting*). In Chapter 5 and 6, we present two studies that rely on a similar *causal* approach for investigating linguistic behaviour. Ultimately, we hope that such approaches not only provide deeper insights into the linguistic competence of LMs, but may also shed light on fundamental questions in human language acquisition.

## 1.4 Thesis Overview

The work described in this thesis comes from seven publications that have been split into three parts (an overview of the work published during my PhD can be found in Appendix B). While all three parts describe approaches to finding linguistic structure in language models using various interpretability methods, we have decided to create three splits based on the way a connection is made between training data and model performance. We provide a brief overview of these parts here; more detailed descriptions can be found at the start of each part in the thesis.

**Part I: Linguistic Structure in Model Predictions** The first part takes on a behavioural perspective for finding structure in language models, taking inspiration from psycholinguistic research that investigates this direction for human language processing. In **Chapter 2** we introduce a methodology for measuring *structural priming* in language models, an abstract language processing mechanism that has been influential in demonstrating the presence of abstract linguistic structure in human language processing. We develop a large-scale evaluation corpus for measuring priming and investigate various linguistic factors that play a role in this behaviour. We build on these findings in **Chapter 3**, where we conduct a more precise investigation into the factors that drive structural priming, and find that priming effects in LMs are impacted by similar factors as found in human studies.

**Part II: Connecting Data Statistics to Model Behaviour** In the second part we present three studies that focus on the impact of the training distribution on model behaviour. In **Chapter 4**, we present an investigation into the phenomenon of adjective order (e.g. *big red car* vs. *red big car*) and create a detailed connection between model behaviour and *n*-gram corpus statistics for large-scale LMs. Then, in **Chapter 5**, we introduce a technique called Filtered Corpus Training that allows us to investigate the linguistic generalisation capabilities of LMs by measuring their performance on corpora in which specific linguistic constructions have been removed. We build on this work in **Chapter 6** by zooming in on a specific linguistic phenomenon—*monotonicity* and its connection to *negative polarity items*—and we show how this abstract semantic property plays a role in LM predictions.

**Part III: Linguistic Investigations in a Controlled Environment** In the final part we investigate linguistic structure in a fully controlled environment using synthetic data. In **Chapter 7** we focus on the importance of *feature interactions* for modelling hierarchical structure, and examine models trained on synthetic toy languages. By evaluating our models on such a restrictive synthetic domain we obtain stronger guarantees regarding the faithfulness of our evaluation methods. Finally, in **Chapter 8**, we present a methodology for evaluating LMs trained on synthetically generated data at a scale that is more reflective of the complexity of natural language.

**Conclusion** In this final chapter we provide a summary of the contributions of this thesis, and we reflect back on the general implications of our findings. Several directions for future work are laid out, and we establish various broader connections between the three parts of the thesis.

# Part I

# Linguistic Structure in Model Predictions

I

# Overview

Examining the presence of abstract structure in language models is a challenging task that can be approached from many different angles. Research within psycholinguistics has addressed this question in human language processing for decades, and may therefore provide valuable inspiration for linguistic investigations of language models. An important paradigm in this direction is called *structural priming* (Bock, 1986), which has been one of the dominant research areas for uncovering linguistic structure in human language production and comprehension for the past 40 years. In the next two chapters we present a methodology for investigating structural priming patterns in language models, which provides a first step in *finding structure in language models*.

**Chapter 2** We investigate the extent to which modern, neural language models are susceptible to structural priming, the phenomenon whereby the structure of a sentence makes the same structure more probable in a follow-up sentence. We explore how priming can be used to study the potential of these models to learn abstract structural information, which is a prerequisite for good performance on tasks that require natural language understanding skills. We introduce a novel metric and release Prime-LM, a large corpus where we control for various linguistic factors which interact with priming strength. We find that Transformer models indeed show evidence of structural priming, but also that the generalisations they learned are to some extent modulated by semantic information. Our experiments also show that the representations acquired by the models may not only encode abstract sequential structure but involve certain level of hierarchical syntactic information. More generally, our study shows that the priming paradigm is a useful, additional tool for gaining insights into the capacities of language models and opens the door to future priming-based investigations that probe the model's internal states.

**Chapter 3** We explore which linguistic factors—at the sentence *and* token level—play an important role in influencing language model predictions, and investigate whether these are reflective of results found in humans and human corpora (Gries and Kootstra, 2017). We make use of the structural priming paradigm, where recent exposure to a structure facilitates processing of the same structure. We don't only investigate whether, but also *where* priming effects occur, and what factors predict them. We show that these effects can be explained via the *inverse frequency effect*, known in human priming, where rarer elements within a prime increase priming effects, as well as lexical dependence between prime and target. Our results provide an important piece in the puzzle of understanding how properties within their context affect structural prediction in language models.

## Author Contributions

The content of this part is based on the following publications:

1. Arabella Sinclair*, Jaap Jumelet*, Willem Zuidema, and Raquel Fernández. 2022. Structural persistence in language models: Priming as a window into abstract language representations. *Transactions of the Association for Computational Linguistics*, 10:1031–1050

2. Jaap Jumelet, Willem Zuidema, and Arabella Sinclair. 2024b. Do language models exhibit human-like structural priming effects? In *Findings of the Association for Computational Linguistics ACL 2024*, pages 14727–14742, Bangkok, Thailand and virtual meeting. Association for Computational Linguistics

Arabella proposed the idea to investigate structural priming in the context of language models. For the first study, Arabella focused on the development of the Prime-LM corpus, while I developed the priming metrics and the computational pipeline. I wrote the initial draft of the paper and subsequent refinements to the paper were done jointly by Arabella and me, with guidance and support from Raquel and Willem. The second study was led by me with extensive help from Arabella in working out the methodology, and supervision from both Arabella and Willem. We provide more detailed author contributions following the Contributor Role Taxonomy (Allen et al., 2019) in Table 1.1.

| | AS | JJ | WZ | RF |
|---|---|---|---|---|
| Conceptualisation | ✓ | ✓ | ✓ | ✓ |
| Methodology | ✓ | ✓ | | |
| Software | | ✓ | | |
| Data Curation | ✓ | | | ✓ |
| Investigation | ✓ | ✓ | | |
| Visualisation | | ✓ | | |
| Analysis | ✓ | ✓ | | |
| Writing - Original Draft | ✓ | ✓ | | |
| Writing - Review & Editing | ✓ | ✓ | ✓ | ✓ |
| Supervision | | | ✓ | ✓ |

Chapter 2: Sinclair* et al. (2022)

| | JJ | WZ | AS |
|---|---|---|---|
| Conceptualisation | ✓ | ✓ | ✓ |
| Methodology | ✓ | | ✓ |
| Software | ✓ | | |
| Data Curation | ✓ | | ✓ |
| Investigation | ✓ | | |
| Visualisation | ✓ | | |
| Analysis | ✓ | | ✓ |
| Writing - Original Draft | ✓ | | ✓ |
| Writing - Review & Editing | ✓ | ✓ | ✓ |
| Supervision | | ✓ | ✓ |

Chapter 3: Jumelet et al. (2024b)

**Table 1.1:** Author contributions for the work described in Chapters 2 and 3.

*This will not stand, ya know, this aggression will not stand, man!*

The Dude being *primed* by George H. W. Bush

# 2

# Structural Persistence in Language Models: *Priming as a Window into Abstract Language Representations*

It has become increasingly clear that modern, neural language models (LMs) are capable of representing and learning a broad range of linguistic phenomena, as explained in detail in §1.3. However, many open questions remain about the extent to which specific LMs have indeed acquired specific linguistic constructions, about whether these models encode an abstract notion of structure in their representations, and about the best ways to even assess the syntactic abilities of these models. A rich literature has emerged in the last few years addressing these questions, often taking inspiration from methodologies developed in theoretical linguistics, psycholinguistics, neuro-linguistics and language acquisition research (Futrell et al., 2019; Ettinger, 2020; Boleda, 2020; Gauthier et al., 2020; Baroni, 2022), where the same questions have been asked about the human mind/brain for centuries. Building on this tradition, this chapter turns to **structural priming** to investigate the degree to which LMs encode abstract structural information independent from the concrete words that make up sentences.

## 2.1 Introduction

The phenomenon of structural priming refers to the fact that humans are more likely to produce—or to more easily comprehend—a sentence of a certain structure $X$ (the *target*) when they have been exposed before to a sentence of a similar structure $X$ (the *prime*), than if they had been prompted with a sentence of a different structure $Y$. For example, a native speaker of English will be more inclined to produce the target sentence with a prepositional object in (2-a) after having read sentence (1-a) instead of (1-b), and, vice versa, be more inclined to produce the double-object target sentence (2-b) after having read (1-b) instead of (1-a). Similar effects are also observed in language comprehension.

(1) a. *A teacher cooked a chicken for a worker*
  b. *A teacher cooked a worker a chicken*

(2) a. *The guest threw the pot to the lady*
  b. *The guest threw the lady the pot*

Evidence for structural priming—to the extent that it can be shown to be independent from lexical overlap and other confounds—is taken as evidence for a linguistic structural level of representation that abstracts away from the surface form of sentences. Thus whether or not language models display structural priming can provide insights as to their structural awareness, which is necessary for downstream tasks requiring natural language understanding skills. Previous experiments designed to test structural encoding in LMs are inconclusive. On the one hand, studies on structural probing (Hewitt and Manning, 2019) and on syntactic evaluation tasks (Warstadt et al., 2020) have yielded evidence for its presence. On the other hand, other sets of experiments have indicated that current LMs are surprisingly indifferent to word order (Hessel and Schofield, 2021; Pham et al., 2021; Sinha et al., 2021a) and rely on superficial heuristics when resolving downstream tasks (McCoy et al., 2019; Sinha et al., 2021b). Such unresolved tensions between results—and the active debate about them—highlights the need for developing additional methodologies that isolates structure from the lexico-semantic cues given to the model. In this chapter, we leverage findings from structural priming in human language processing to develop a systematic experimental pipeline with the aim of assessing the extent to which pre-trained neural language models learn representations that encode structural information—a prerequisite for their good performance on natural language understanding tasks.

We use the term 'structural priming' (Pickering and Ferreira, 2008) rather than 'syntactic priming' (first described in Katryn Bock's *Syntactic persistence in language production*, 1986) because it comprises priming of abstract structural information that is not restricted to syntactic hierarchical rules, such as the linear positions of semantic roles or the sequential order of parts of speech. In this chapter, we focus mostly on the latter and touch upon syntactic rules in Section 2.6.5.

In Section 2.3, we define an efficient novel metric for measuring the effect of priming. For our experiments, we create **Prime-LM**, a large-scale corpus for examining structural priming consisting of ∼1.3M prime-target sentence pairs, as we describe in Section 2.4. Earlier work on priming in LMs by Prasad et al. (2019) operationalised priming as adaptation or implicit learning and thus fine-tuned the model weights in between prime and target. While our priming effect metric is compatible with priming as adaptation, our experiments in this chapter concentrate on priming after recent exposure to linguistic context without updating the model weights. This allows us to assess the structural representational abilities acquired by the models during training and investigate to what extent such structural information remains active at inference time.

In Section 2.5.2 and 2.6 we use our corpus and priming paradigm to answer three main research questions: (1) Are modern neural language models susceptible to structural priming? (2) Which factors influence the strength of the priming effect? And: (3) what is the nature of the structural representations acquired by those models? Our results show that Transformer language models *do* exhibit structural priming. This finding provides evidence that abstract structural information is encoded by the models to some degree and persists as a model makes predictions about upcoming sentences. The strength of the priming effect is influenced by several factors, including the semantic similarity and the proximity between prime and target, as well as the amount of exposure to a given structure during prompting. Our final experiment moreover reveals that the structural representations encoded by the model may not only be sequential but involve a certain level of hierarchical syntactic structure.

## 2.2 Structural Priming

### 2.2.1 Structural Priming in Humans

Priming is the dominant paradigm in psycholinguistics for investigating the extent to which human language processing involves a level of structural representation independent from other types of linguistic knowledge. The rationale behind this paradigm is that if speakers are sensitive to sentence structure independently from sentence content, then it is reason-able to assume that such structural information is an integral part of the representations built during processing.

In human language processing, structural priming effects are well attested both in comprehension and production (Bock, 1986; Pickering and Branigan, 1998; Bock and Griffin, 2000; Pickering and Ferreira, 2008; Goldwater et al., 2011; Pickering et al., 2013; Reitter and Moore, 2014; Tooley and Bock, 2014, among others). Several studies have shown that the strength of the priming effect increases after repeated exposure to a given structure (Kaschak et al., 2011; Jaeger and Snider, 2013) and tends to decay if material intervenes between prime and target (Reitter et al., 2011). Other experiments have shown that ungrammatical and semantically incongruent sentences (e.g., *the waitress brunks the book to the monk*) lead to similar priming effects as well-formed sentences (Ivanova et al., 2012, 2017), which suggests that structural persistence effects are robust enough in the absence of semantic and lexical cues.

Yet, structural priming has been found to be affected by various aspects. For example, priming effects are stronger with lower-frequency than higher-frequency constructions (e.g., Scheepers, 2003; Bernolet and Hartsuiker, 2010; Pickering et al., 2013). Similarly, some types of lexical repetition between prime and target have been shown to enhance structural priming, suggesting that there is a lexical component involved (Pickering and Branigan, 1998; Cleland and Pickering, 2003). Semantic relatedness between prime and target also has a boosting effect, albeit smaller than the lexical repetition boost (Cleland and Pickering, 2003; Mahowald et al., 2016).

In the present study, we take inspiration from this tradition to investigate the priming behaviour of neural language models, which in turn depends on them encoding structural information. Two (not necessarily exclusive) mechanisms have been proposed to account for structural priming in humans: short-term residual activation of structural information across utterances (e.g., Branigan et al., 1999; Wheeldon and Smith, 2003) and long-term adaptation or implicit learning involving changes in the probability of a given structure (Bock et al., 2007; Kaschak et al., 2011; Fine and Jaeger, 2013). Here we focus on the ability of large pre-trained LMs to encode structural information given in the preceding context, similarly to residual activation in humans.

### 2.2.2 Structural Sensitivity of Neural LMs

In Section 1.3 we explained the various approaches that have been proposed to examine linguistic structure in language models. We pointed out various challenging factors to this, demonstrating that the debate about the abilities of language models to learn structural information is far from over. The research we present in this chapter starts from the observation that structural priming may provide a much needed, complementary methodology that, like Targeted Syntactic Evaluations, examines the behaviour of a model, but also, like probing, informs us about the nature of the internal states. We will assess a model's representation of a sentence by measuring its consequences in processing the next sentence. Instead of examining how the model deals with specific syntactic properties within a sentence, such as number agreement, we measure its encoding of abstract structure at the overall sentence level and the consequences this has for upcoming sentences. In the next section we explain our approach in detail.

## 2.3 Measuring Priming in LMs

We capture the effects of priming by measuring the difference in log probability of a target sentence $\text{T}^{\text{X}}$ given a prime sentence $\text{P}^{\text{X}}$ of the same syntactic structure X, vs. $\text{T}^{\text{X}}$ given $\text{P}^{\text{Y}}$, a sentence of the exact same semantic and lexical content as $\text{P}^{\text{X}}$ but differing in syntactic structure Y. We call this metric the ***Priming Effect*** *(PE)*:

$$\log P(\text{T}^{\text{X}} \mid \text{P}^{\text{X}}) - \log P(\text{T}^{\text{X}} \mid \text{P}^{\text{Y}}) \quad [2.1]$$

By measuring priming based on a fixed prime-target pair our method is akin to structural priming in comprehension. We condition a target sentence on a prime sentence by concatenating them, separated by a period. The log probability is computed as the sum of token log probabilities of the LM:

$$\log P(\text{T}^{\text{X}} \mid \text{P}^{\text{X}}) = \sum_i \log P_{\text{LM}}(\text{T}^{\text{X}}_i \mid \text{P}^{\text{X}}, \text{T}^{\text{X}}_{<i}) \quad [2.2]$$

For example, the Priming Effect of the example in the introduction would be computed as follows:

$$PE(\text{T}^{\text{PO}}) = \log P(\text{T}^{\text{PO}} \mid \text{P}^{\text{PO}}) - \log P(\text{T}^{\text{PO}} \mid \text{P}^{\text{DO}})$$
$$PE(\text{T}^{\text{DO}}) = \log P(\text{T}^{\text{DO}} \mid \text{P}^{\text{DO}}) - \log P(\text{T}^{\text{DO}} \mid \text{P}^{\text{PO}})$$

(where $\text{P}^{\text{PO}}$, $\text{P}^{\text{DO}}$, $\text{T}^{\text{PO}}$, $\text{T}^{\text{DO}}$ denote sentences 1a, 1b, 2a, and 2b). To ensure our estimates of the priming effect are robust, we incorporate the procedure of Newman et al. (2021) by pairing each target sentence in a corpus with 10 different prime sentences.

**Definition 2.3.1** (*Priming Effect (PE)*). Measures the effect of priming as the difference in log probabilities:

$$PE(\textsc{t}^{X}) = \frac{1}{|\mathcal{P}|} \sum_{\textsc{p}^{X} \in \mathcal{P}(\textsc{t}^{X})} \left[\log P(\textsc{t}^{X} \mid \textsc{p}^{X}) - \log P(\textsc{t}^{X} \mid \textsc{p}^{Y})\right]$$

where $\mathcal{P}(\textsc{t}^{X})$ denotes the set of prime sentences that can be matched with target $\textsc{t}^{X}$. In our experiments, we report the mean of this metric, taken over large-scale corpora of semantically diverse sentences.

Our Priming Effect method is related to various other metrics that are used in the context of priming and statistics in general. When the conditional probabilities are close to zero—as is the case for our corpora with a mean sentence probability around $10^{-18}$—this metric approaches the log odds ratio that is used by Mahowald et al. (2016). This allows our scores to be directly comparable to their results on human priming. A more general connection can be made between our metric and Bayes factors (Jeffreys, 1961; Kass and Raftery, 1995), which determine the strength of evidence and are, similar to our metric, also defined as a log probability ratio.

Prasad et al. (2019) model priming as an implicit learning procedure (Chang et al., 2000), instantiated as a fine-tuning-based adaptation process (van Schijndel and Linzen, 2018). The adaptation effect is then obtained by comparing the impact of a single prime structure on two target sentences of opposing structure, comparing their perplexity before and after fine-tuning:

$$PP(\textsc{t}^{X}) - PP(\textsc{t}^{X} \mid \textsc{p}^{X}) > PP(\textsc{t}^{Y}) - PP(\textsc{t}^{Y} \mid \textsc{p}^{X})$$

The authors also identify a problem: this metric is strongly proportional to the prior perplexities $PP(\textsc{t}^{X})$ and $PP(\textsc{t}^{Y})$, which may override priming effects. They resolve the issue by regressing out this relationship. This procedure, however, is based on assumptions that do not always hold, namely, that the relationship between the priming metric and the prior perplexities of the two targets is linear and homoscedastic. In our experiments we found neither assumption to hold empirically, and hence we opted to directly compare the impact of two prime sentences on a single target sentence. This way we do not need to regress out confounding effects of prior probabilities, since we are comparing the same quantity (the target sentence) to two primes. The contrast between these metrics is illustrated by the diagrams in Figure 2.1.

Note that our Priming Effect metric could be applied to the priming-as-adaptation

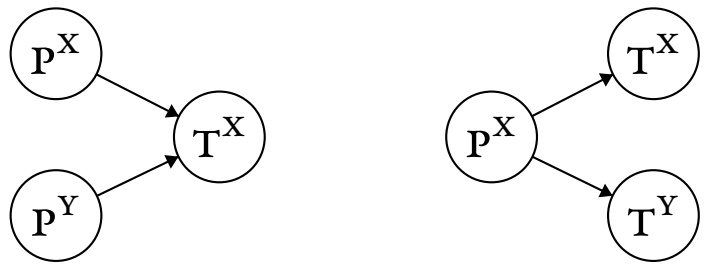

*Priming Effect* (Eq. 2.1) Prasad et al. (2019)

**Figure 2.1:** Our Priming Effect metric compares the impact of two prime sentences with different structures on a single target exhibiting one of the structures. Prasad et al. (2019) examine the impact of a single prime structure on two target sentences.

paradigm as well, by comparing the target sentence probabilities of two fine-tuned models. In the experiments presented in this chapter, we focus on priming as residual activation and thus do not update the model weights, which makes the approach more computationally efficient.

## 2.4 The Prime-LM Corpus

We create a large-scale set of corpora designed to examine the priming behaviour of LMs.

### 2.4.1 Syntactic Alternations

In the current experiments, we focus on two types of syntactic alternations, *dative* and *transitive*, which allow for the same content to be expressed by two different structures. The dative alternation includes ditransitive verbs whose complements can be expressed by a double-object (DO) structure or a prepositional-object (PO) structure (e.g., *The boss gave the dog a bone* vs. *The boss gave a bone to the dog*). The transitive alternation includes transitive verbs within an active (ACT) or a passive (PASS) structure (e.g., *the actor followed the student* vs. *the student was followed by the actor*).

(3) **Dative**

DO: Det $N_{agent}$ V Det $N_{recipient}$ Det $N_{patient}$

PO: Det $N_{agent}$ V Det $N_{patient}$ Pr Det $N_{recipient}$

(4) **Transitive**

ACT: Det $N_{agent}$ V Det $N_{patient}$

PASS: Det $N_{patient}$ Aux V by Det $N_{agent}$

In the transitive case, the active structure is dominant in English (Bock, 1986; Merriam-Webster, 1989). The proportion of use between structures for the dative alternation is less marked, with different studies showing a preference for the direct-object structure (e.g., Bock, 1986; Bresnan et al., 2007).

### 2.4.2 Corpus Construction

We construct a set of corpora by filling in the templates in (3) and (4) above. For the content words (nouns and verbs), we exploit the vocabulary present in the University of South Florida (USF) free association norms dataset (Nelson et al., 2004), which contains pairs of cue-target words with their association strength.[1] This allows us to control for the degree of semantic association between prime and target sentences. To minimise any effects stemming from word frequency factors, we only include USF content words which appear in the top 5000 most common words according to the COCA corpus (Davies, 2009).

We identify transitive and ditransitive verbs using vocabulary lists targeted at English language learners,[2] keeping those that are present in USF and meet the frequency constraints (around 80 verbs in total). The ditransitive verbs were manually labelled for the preposition to be used in the PO structure (*to/for*) and the transitive verbs were annotated with their past participle form to be used in the passive construction. In addition, all verbs were manually labelled for some of the noun categories they can take as arguments (e.g., the transitive verb *wash* was annotated as accepting agents of category `person` and patients of category `person` or `object`). Following the same frequency constraints, a set of nouns fulfilling these categories was selected from USF using the WordNet closure categories of *person*, *social_group*, *social_control*, *institution*, *physical_entity*, and *object*, which we further hand split into *non-edible*, *edible*, and *drinkable*.[3] This yielded 119 nouns in total.

From this vocabulary, we are able to generate many realisations of our sentence templates through sampling, respecting the grammaticality of the sentences produced. Three native speakers of English manually examined a subset of sentences for each verb and syntactic alternation to confirm that the sentences produced are well formed. This resulted in the elimination of a few ditransitive verbs for which the DO structure was considered awkward. The final corpus contains 48 transitive and 16 ditransitive verbs.

Using this template-based method, we create a series of corpora that satisfy various semantic and lexical constraints. For each of these corpora we specify a corpus size of 15,000 prime-target pairs per syntactic target structure (DO, PO, ACT, PASS), which are obtained by pairing 1,500 different target sentences with 10 semantically different primes.[4] In total, Prime-LM contains ∼1.3M prime-target pairs.

[1] Corresponding to the percentage of human participants who produced the target word when asked to come up with words related to the cue (`w3.usf.edu/FreeAssociation`).

[2] `englishpost.org/transitive-verbs-list`.

[3] To ensure compatibility with the indefinite article *a/an* (see §2.4.3), uncountable nouns were discarded.

### 2.4.3 The Core Corpus

Prime-LM consists of a *core* corpus and a set of variants over this core. In the core corpus, we ensure that prime and target sentences (1) include different determiners, either *a/an* or *the*, (2) do not share any nouns nor verbs, and (3) only contain nouns and verbs that are not semantically associated across prime and target according to the USF free association norms dataset.[5] For the PO structure, we additionally make sure that prime and target differ in preposition (*to* vs. *for*), which makes all the prime and target sentences in the dative alternation lexically fully disjoint. For the transitive alternation, this is not possible since the preposition *by* must appear in the PASS structure. Other than that, we completely limit lexical overlap for transitive constructions by using alternate auxiliary verb forms (*is* vs. *was*) for the passive prime and target, and create their active counterparts by using the corresponding tense of the auxiliary to maintain semantic equivalence. All sentences in the dative alternation are in the past simple tense.

As an illustration, below we show two examples from the *Core* condition following the scheme in Figure 2.1, where $P$ denotes a prime sentence and $T$ the target:

(5) $P_{\text{PO}}$: *A pilot bought a pie for an attorney*
$P_{\text{DO}}$: *A pilot bought an attorney a pie*
$T_{\text{PO}}$: *The professor sent the tea to the band*

(6) $P_{\text{ACT}}$: *The nurse purchased the beer*
$P_{\text{PASS}}$: *The beer was purchased by the nurse*
$T_{\text{PASS}}$: *An engine is wrapped by a colonel*

We create different variants of the core corpus that isolate specific aspects shown to influence structural priming in human sentence processing. They are described in Section 2.6 together with the corresponding experiments. Example sentences for each of our corpora can be found in Table 2.1.

[4] The corpus size of 15,000 was determined based on Cochran's Formula for sample size determination (Cochran, 1977), with a $p$-value and margin of error of 0.01.

[5] The average cosine similarity across pairs of words in prime and target computed with `word2vec` embeddings by Fares et al. (2017) is 0.2 for both nouns and verbs.

## 2.5 Priming Effects in LMs

In this section we describe our first set of experimental results on a range of language models that are tested on the *Core* condition of Prime-LM.

### 2.5.1 Language Models

We focus our experiments on the class of *auto-regressive*[6] LMs, which are trained to predict the next token, in line with human incremental language processing. Our methodology can be applied to masked LMs as well; we briefly reflect on this in the discussion (§2.7). The main focus of our analysis is directed upon on Transformer models (Vaswani et al., 2017), that constitute the current state of the art in language modelling, and have been shown to produce representations that correlate strongly with human brain signals (Schrimpf et al., 2020). This is the set of models we consider:

- **GPT2**, in its four sizes (small, medium, large, xl; Radford et al., 2019), and its *distilled* version (Sanh et al., 2019);
- **DialoGPT**, three GPT2 models of increasing size that have been fine-tuned on dialogue data (Zhang et al., 2020);
- **GPT-Neo** in three sizes (125m, 1.3b, 2.7b; Black et al., 2021), which is based on GPT3 (Brown et al., 2020).

All Transformer LMs are imported with the `transformers` library (Wolf et al., 2020). The extraction of the model probabilities is done using the `diagNNose` library (Jumelet, 2020), which provides support for efficient activation extraction. Our implementation allows our priming procedure to be efficiently tested on any kind of language model and to be easily reproducible.

**Why should LMs exhibit structural priming?** Since structural repetition is present in human language use and common in corpora (Dubey et al., 2008), LMs have, in theory, the potential to learn such structural dependencies during training. It is, however, not reasonable to expect that models which have been trained on shuffled sentences will exhibit priming, because such models will not be able to adequately carry over a linguistic signal

[6] Also known as *causal* or *left-to-right* language models, predicting the probability of the next token solely on prior context.

(structural or otherwise) from the prime sentence to the target.[7] As mentioned in the introduction and in Section 1.2, several studies have suggested that structural information is being encoded by large language models; yet, other studies showing that LMs are often insensitive to permutations in word order (e.g., Kodner and Gupta, 2020; Sinha et al., 2021b) cast doubt on these results. Thus, while there is potential for LMs pre-trained on unshuffled data to encode structural dependencies that are detectable with our priming paradigm, whether they will in fact do so remains an open question, since the language modelling objective (next word prediction) contains no explicit cues for structural information. This is precisely the question we address in this work.

**Priming behaviour** To interpret our results we distinguish between three types of behaviour: i) *symmetrical priming* occurs when a model obtains positive Priming Effects for both constructions within an alternation: the model has fully picked up on the structural congruence between prime and target; ii) *asymmetrical priming* occurs when a model obtains a positive Priming Effect for one construction, and a Priming Effect close to zero for its counterpart;[8] and iii) *biased priming* occurs when a model obtains a positive Priming Effect for one construction, but a negative Priming Effect for its counterpart. A priming bias indicates that a prime of the preferred structure is more likely to boost any subsequent target that we consider, regardless of its structural congruence with the prime. Hence, we take symmetrical and, to some extent, asymmetrical priming behaviour to represent evidence for the structural priming effect we are interested in.[9]

### 2.5.2 Core Priming Results across LMs

We initially test all LMs described in the previous section on our *core* corpus, designed to control for lexical overlap and semantic similarity. This provides a clean experimental setup, where the only element shared between prime and target is the abstract sequential structure. The results are reported in Figure 2.2, split by the structure type of the target sentence. It can be seen that across many models a positive Priming Effect is present. We will now discuss these results in more detail.

[7] In our experiments we had initially incorporated two LSTM LMs (Józefowicz et al., 2016; Gulordava et al., 2018), and indeed due to their shuffled training corpus we did not observe any notable Priming Effect. We are not aware of any available LSTM LM trained on unshuffled data.

[8] Such asymmetries are common in humans (Bock, 1986; Gries, 2005; Segaert et al., 2016).

[9] This is analogical to, for example, subject-verb agreement: a model that always prefers a plural verb, regardless of the subject number, can't be said to understand the task. A model that scores 100% on plural verb prediction, but randomly for singular verbs, has an *asymmetric* understanding of the task.

2

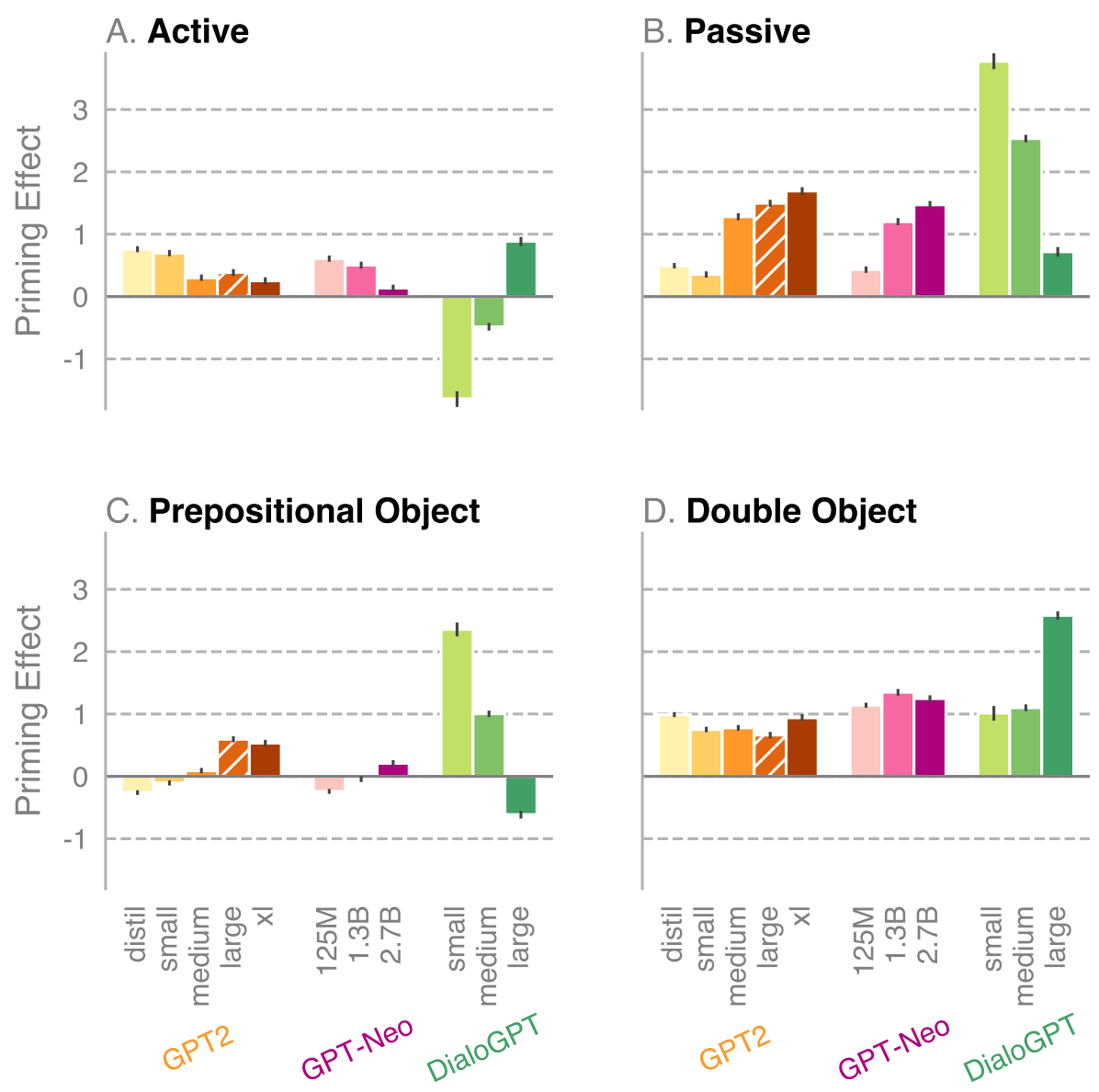

**Figure 2.2:** *Priming Effect* results of all models on the *core* corpus, across the four syntactic structures. Error bars denote 99% confidence intervals of the mean. The GPT2-LARGE model that will be explored in more detail in §3.6 has been highlighted.

There are two models that exhibit symmetrical priming for both transitive and dative alternations: GPT2-LARGE and GPT2-XL. The other GPT2 models exhibit symmetrical priming for transitive as well, but exhibit moderate asymmetrical priming behaviour for dative, with priming occurring only for double-object structure. DialoGPT-SMALL exhibits biased priming for transitive constructions: a negative Priming Effect on active constructions, but a large positive Priming Effect for passive constructions. This shows that for this particular model a passive prime boosts the probability of an active target more than an active prime does, resulting in a negative effect.

**Model size** We can investigate the impact of model size by comparing the results of the different sizes of the models we consider.[10] Larger models may have more potential for encoding finer-grained structural information (see e.g., Hu et al., 2020). If model size were to have a positive effect on structural priming this might manifest itself in two ways: either (1) the Priming Effect increases for both structural alternatives, or (2) the priming bias towards one structure decreases. We do not see evidence of (1). As for (2) regarding bias, results differ between transitive and dative. For the GPT2 models the asymmetrical priming towards double objects is decreased, resulting in symmetrical priming for both GPT2-LARGE and GPT2-XL. For the DialoGPT results on transitive we can see that the severe bias towards passive decreases as model size is increased, resulting in symmetrical priming behaviour for DialoGPT-LARGE. For dative constructions, however, the larger model size gives rise to a priming bias towards double objects: in this case increasing model size actually has a detrimental effect on the model's priming behaviour. From this we conclude that sensitivity to structural priming is partly driven by model size, but is likely to depend on a more intricate combination of factors related to model architecture and training data, which needs to be investigated further in future work.

[10]Note that the different sizes of a model are trained on the same amount of data; only the number of parameters is affected.

| Corpus | Condition | Prime (*active*) | Target (*passive*) |
|---|---|---|---|
| **Core** | — | *The boy judged the adult.* | *A cousin is forgotten by a secretary.* |
| **Semantic Similarity** | Verb Only | *The chief **struck** the mayor.* | *A bishop was **beaten** by a hero.* |
| | All Nouns | *An **actor** broke a **glass**.* | *The **bottle** was wrapped by the **actress**.* |
| | All Words | *The **student drank** the **wine**.* | *A **beer** was **prepared** by a **professor**.* |
| **Lexical Overlap** | Random Noun | *The girl smelled the **chicken**.* | *A **chicken** was prepared by a pilot.* |
| | Main Verb | *A woman **used** a computer.* | *The iron was **used** by the father.* |
| | Function Words | ***The** soldier wanted **the** pie.* | ***The** book was carried by **the** manager.* |
| | All Nouns | *The **king** smelled the **wine**.* | *A **wine** was drunk by a **king**.* |
| **Implausible Prime** | — | *The **newspaper grabbed** the **pot**.* | *A key is removed by an attorney.* |
| **Structural Complexity** | Prime Complex | *A lady **with a red bag** chased a minister.* | *The juice was purchased by the child.* |
| | Target Complex | *The physician judged the leader.* | *A **rich** school was embraced by a business.* |
| | Both Complex | *The **bad** adult **with the hat** raised the knife.* | *A son was helped by an author **from Cuba**.* |

**Table 2.1:** Example sentences for the core corpus and each condition described in §2.6.1, §2.6.2 and §2.6.5. The same manipulations illustrated here for the ACT and PASS also hold for the dative alternation.

**Best model** The models that exhibit more susceptibility to structural priming across all four construction types are GPT2-LARGE and GPT-2-XL. For GPT2-LARGE the congruent conditional probability $P(T_X|P_X)$ was larger than the incongruent one $P(T_X|P_Y)$ 60.5% of the time for active, 81.0% for passive, 65.4% for prepositional object, and 72.1% for double object. In the subsequent experiments we will focus our analysis on GPT2-LARGE and use more specialised experimental conditions within the priming paradigm to dig deeper into the potential of the model for encoding structural information.

## 2.6 Impact of Specific Factors

The next battery of experiments isolates various factors that have been shown to be of influence to priming in human language processing. For each experimental condition, we present a specialised corpus followed by an analysis of the priming effects exhibited by GPT2-LARGE on this data, comparing them to the model's behaviour on the core corpus. Examples from the core and specialised conditions can be found in Table 2.1.

### 2.6.1 Lexical Dependence

In the *core* corpus, prime and target sentences are semantically unrelated, which ensures that priming effects cannot stem from the model assigning higher probabilities to words that are similar or identical to those present in the prime sentence. In the following two experiments we relax this constraint to investigate the extent to which lexical semantic similarity and lexical repetition across prime and target have an impact on structural priming effects.

#### Semantic Similarity

We create versions of the *core* corpus where prime and target sentences have different degrees of lexical semantic similarity. Concretely, a pair of words sharing the same semantic role in the prime and target is considered semantically similar if they (a) are associated according to the USF norms, and (b) have a cosine similarity (computed with embeddings from Fares et al., 2017) equal or higher than the 90%-percentile of the distribution of similarities in the core corpus.[11]

In human experiments, semantic similarity has been found to boost priming (Goldwater et al., 2011), both in nouns (Cleland and Pickering, 2003), and in verbs (Pickering and Branigan, 1998). We isolate the effects of verb and noun similarity by creating conditions where (1) only the verb, (2) all nouns, or (3) all content words are semantically similar across prime and target sentences. These additional constraints result in a more limited set of possible sentence pairs for condition (3), and thus in a reduced corpus of 228 (transitive) and 1648 (dative) prime-target pairs rather then 15,000.[12]

**Results** We find greater Priming Effect across constructions in this setup compared to the core corpus, although this is less pronounced for the PO structure. As can be seen in Figure 2.3A, a semantically similar verb in prime and target leads to an increase of the Priming Effect, comparable to the condition where all nouns are similar. With the exception of DO, we do not observe an additive effect: when all content words are similar, the Priming Effect is not substantially higher than when only the verb is similar.

[11] This results in a cosine similarity threshold of ∼0.4.

[12] In this case, to maximise the number of unique pairs, we allow a varying number of primes to target, rather than observing the 10-to-1 prime-target setup of the other corpora.

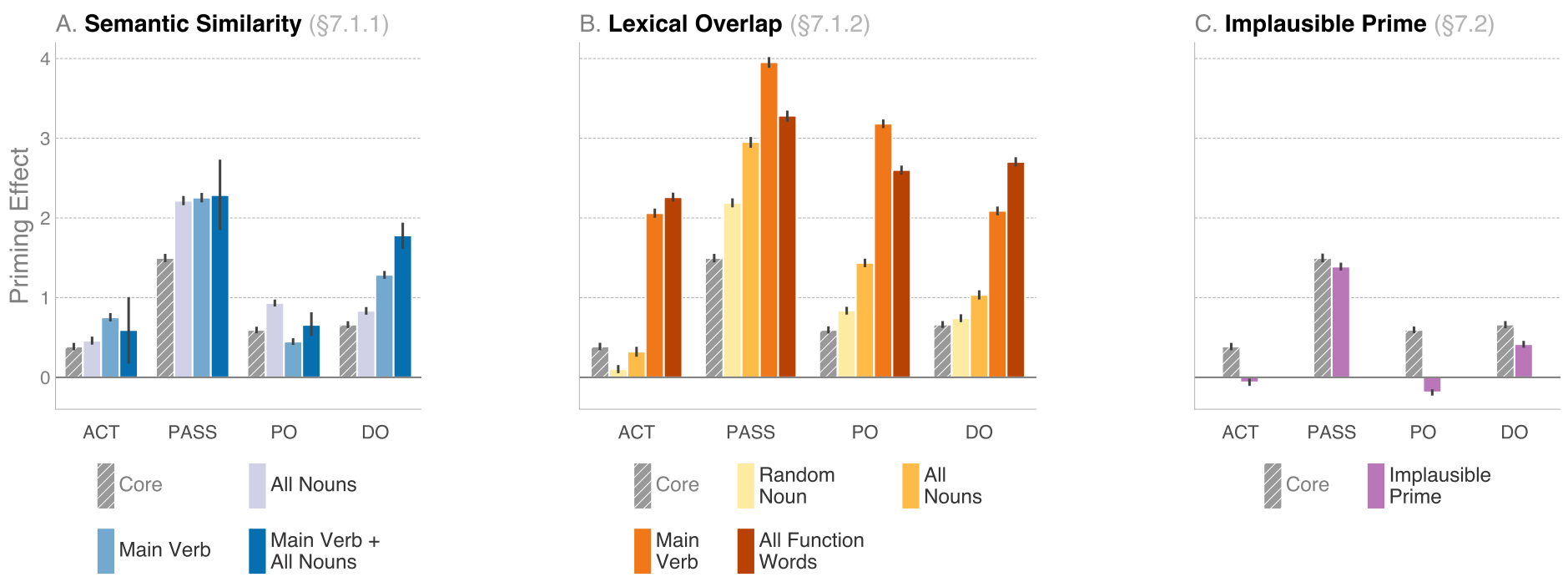

**Figure 2.3:** Results for GPT2-LARGE on the experiments described in and §2.6.1 and §2.6.2: **A.** measures the impact of semantic similarity between prime and target, **B.** the impact of lexical overlap between prime and target, and **C.** whether priming is affected by the semantic implausibility of the prime.

### Lexical Overlap

Lexical overlap between prime and target in the core corpus was avoided in both content and function words. Here we systematically introduce lexical repetition across prime and target sentences. We create versions of the core corpus where lexical overlap takes place with respect to only (1) one of the nouns at random but with the same semantic role across prime and target (*agent, patient, recipient*, see §2.4.1), (2) all nouns, (3) the verb, and (4) all function words (i.e., any determiners, prepositions, and auxiliary verbs are shared across prime and target, without content words being shared).

**Results** As can be seen in Figure 2.3B, overall the presence of lexical overlap greatly boosts structural priming effects. For all constructions, verb overlap leads to higher priming effects than repeating one noun or even all nouns. Surprisingly, overlap of function words has the highest boosting effect for ACT and DO.[13] To place these results into context, we calculate the Priming Effect when prime and target are identical sentences. Language models are known to fall prone to repeatedly generating the same sentence (Foster and White, 2007; Fu et al., 2021); hence this value can be considered a ceiling. We obtain a PE of 2.5 for ACT, 7.2 for PASS, 9.2 for PO, and 10.1 for DO constructions. None of the lexical overlap conditions we consider reaches the corresponding ceiling.

[13]This contrasts with psycholinguistic evidence suggesting that structural priming is not led by function-word priming in humans (Bock, 1989; Tree and Meijer, 1999).

### 2.6.2 Semantic Implausibility

In this experiment, we test whether the effects found in the core corpus are robust to manipulations concerned with the semantic plausibility of the sentences used as stimuli. This helps to diagnose to what extent any structural information encoded by the model is autonomous from semantics. To this end, we construct a version of the corpus where the prime sentences are specifically designed to be semantically implausible. Gulordava et al. (2018) employed a similar method in their study of long-distance agreement dependencies, finding that RNN's ability to predict number agreement was robust to nonsensical sentences. The authors interpret this result as evidence that the networks track abstract structure, in line with Chomsky's (1957) proposal that grammaticality is distinct from meaningfulness in the human language faculty. Here we further test this hypothesis by analysing whether the LM is susceptible to structural priming effects when the prime sentence is nonsensical. As mentioned in §2.2.1, humans do exhibit structural priming effects when prompted with incongruent sentences (Ivanova et al., 2012, 2017). We construct semantically implausible primes via sampling nouns at random among noun categories that do not respect the verb selectional restrictions. This results in grammatically correct, yet nonsensical sentences such as *'the iron threw the hero to the chocolate'*. The same constraints regarding absence of semantic similarity and lexical overlap between prime and target present in the core corpus apply here as well.

**Results** The results of this experiment are shown in Figure 2.3C. We find here that the Priming Effect exhibits asymmetrical priming behaviour, indicating that the prime structure itself is more likely to boost any subsequent target regardless of shared structural properties. The Priming Effect disappears and becomes negative for the ACT and PO constructions, while for PASS and DO it decreases when compared to the results on the core corpus, but remains positive. While some degree of abstract structural information present in the nonsensical sentences may be exploited to predict the target construction, the asymmetrical behaviour suggests that structural encoding is not fully independent from semantic plausibility.

### 2.6.3 Recency Effects

In the following two experiments, we test whether structural priming effects are affected by the proximity of prime to target and by increased exposure to the priming structure. We maintain the strict setting of our core corpus, where prime and target are semantically and lexically unrelated, thus testing to what extent the activation of abstract structural information across sentences is affected by recency and cumulativity factors.

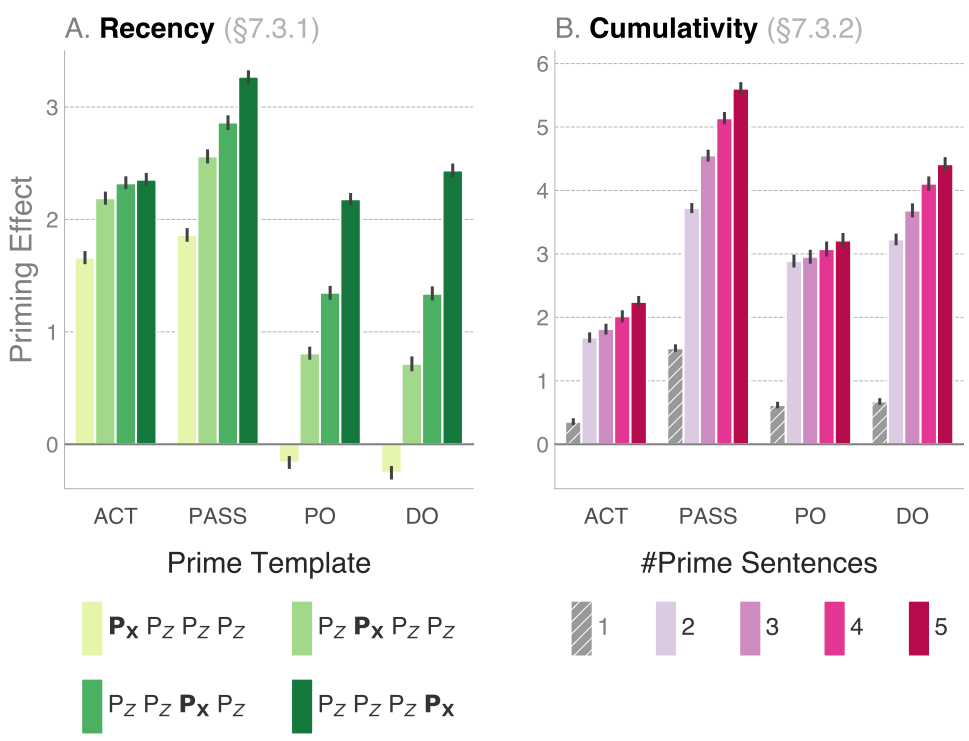

**Figure 2.4:** Results for GPT2-LARGE on the experiments described in §2.6.3: **A.** recency effect on priming, by increasing the distance between prime and target with additional intransitive sentences: each bar denotes a different position of the prime ($P_X$), surrounded by intervening sentences ($P_Z$); **B.** cumulative effects on priming, by increasing the number of prime sentences before a target.

To vary the proximity of prime to target, we create a set of *padding* sentences, using intransitive verbs, personal pronouns, and different auxiliary verbs to those used in our core corpus, including modal auxiliary verbs (e.g., *you might come, he did remain, they should appear*). These sentences were designed to contain frequent vocabulary with no lexical overlap nor semantic similarity to the prime and target sentences in the core corpus. A context in this setting consists of a sequence of 4 sentences, within which the priming sentence will vary in position relative to the target. This setup ensures that any priming observed is not influenced by the total length of the context, but solely by the position of the prime. In this condition, the Priming Effect is computed as follows:

$$\log P(T_X | P_Z^* \, P_X \, P_Z^*) - \log P(T_X | P_Y) \qquad [2.3]$$

where $P_Z$ denotes the sequence of intransitive padding sentences.

**Results** The results of this experiment are shown in Figure 2.4A, which shows that increasing the proximity between prime and target has a highly positive impact on the strength of priming. Interestingly, the PE for the transitive cases is still relatively high even when the distance between prime and target is at its largest, whereas for the dative cases the PE has dropped drastically. This may indicate that the syntactic configuration of a transitive sentence is not corrupted as much by the intermediate intransitive sentence as the configu-ration of a dative sentence.

### 2.6.4 Cumulativity Effects

To investigate the effect of cumulativity, we create a version of the core corpus where for each target sentence we sample multiple primes and concatenate them, resulting in priming contexts which vary between 1 and 5 sentences in length. All prime sentences in the prompt satisfy the semantic constraints with respect to the target that were outlined in §2.4. In this case, the Priming Effect is measured as follows:

$$\log P(T_X|P_X^+) - \log P(T_X|P_Y) \qquad [2.4]$$

in other words, the Priming Effect of a sequence of congruent primes $P_X^+$ is expressed with relation to the log probability of a *single* incongruent prime sentence $P_Y$.

**Results** As shown in Figure 2.4B, for all constructions the Priming Effect increases monotonically as the number of congruent prime sentences increases. This resonates with the potential of large LMs for few-shot learning: the multiple priming sentences appear to act as "demonstrations" (in the sense of Brown et al., 2020) of a given structure, which presumably increases the activation of that type of structural information. This result is a yet another indication of structural information being encoded by the model and remaining active across sentences, as the main feature that is repeated across the multiple primes is the shared abstract structure.

### 2.6.5 Structural Complexity

Finally, we test whether the priming effects present in the core corpus are robust to different degrees of structural complexity between prime and target. In our core corpus, congruent prime and target sentences are constructed from the same sequence of parts of speech (see §2.4.1). Results by Reitter and Keller (2007) suggest that, for humans, short-term priming via residual activation is better explained by assuming hierarchical representations. In this experiment, we test whether the structural information encoded by the model is limited to sequential abstract structure or rather involves hierarchical syntactic representations.

To gain more insight on the nature of the structural information represented by the model, we construct a version of the corpus where some of the noun phrases are more complex than simply "`Dt N`" (e.g., *the awful tea from Spain*). The rationale behind this manipulation is the following: if the structure of a sentence is represented in terms of some-thing akin to a hierarchical phrase-structure rule such as $VP \rightarrow NP\ NP$ or $VP \rightarrow NP\ PP$ rather than as a sequence of part-of-speech categories, then it should not matter whether prime and target differ with respect to the internal structure of the sub-constituents – we should observe a similar degree of priming whether the noun phrases are complex or not. Evidence suggests that this is indeed the case for humans (Tree and Meijer, 1999; Pickering and Branigan, 1998; Branigan et al., 2006).

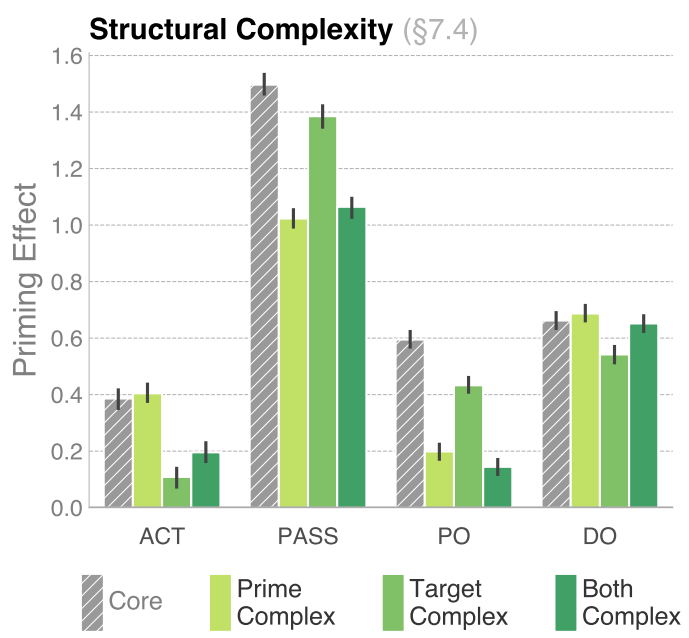

**Figure 2.5:** Results for GPT2-LARGE on the experiment described in §2.6.5, measuring the impact of increasing the complexity of one noun phrase per sentence in prime and target.

We create a version of the core corpus where the noun phrases may contain a prenominal adjective, a prepositional phrase, neither or both in order to introduce varying degrees of complexity. We use a total of 164 adjectives manually labelled for compatibility with the different noun categories. The prepositional phrases are constructed with either *with* or *from*. For the *with* case, we select a set of 27 suitable nouns within the WordNet categories of *clothing*, *device* or *container*. This results in noun phrases such as "`Dt(A)N` *with* `Dt(A)N`". For the *from* case, we use 23 country names, resulting in noun phrases such "`Dt(A)N` *from* `N`". All the additional vocabulary adheres to the same selection procedure as in §2.4, with prime and target being semantically unrelated. We test three conditions: (1) only the prime sentence has a complex NP, (2) only the target sentence does, (3) both prime and target have a complex NP – ensuring different NP structures across prime and target. In all three settings, any semantic role (*agent*, *patient*, or *recipient*) can be modified to become complex and there is at most one complex NP per sentence.

**Results** The results are shown in Figure 2.5. The first thing to note is that the presence of noun phrases of varying complexity across prime and target does not cancel out the Priming Effect: in all cases, the effect remains positive, although there is a decrease for several conditions. We also observe *asymmetrical* priming effects. For example, for transitive with complex prime, e.g., active is unaffected, whereas the Priming Effect for passive is clearly reduced. This suggests that some of the effects observed on the core corpus may be driven by the consistently simple sequential structures present in that data. Yet, the fact that the priming effect remains positive suggests that there is some degree of hierarchical structural information commonly encoded for both simple and complex NPs, which is carried over to influence the prediction of the target.

## 2.7 Discussion and Conclusions

In this chapter, we investigated three main questions: (1) Are modern neural LMs susceptible to structural priming? (2) Which factors influence the strength of the priming effect? And: (3) what is the nature of the structural representations acquired by those models? To answer these questions, we designed a series of carefully curated large-scale corpora, proposed a metric to measure the degree to which a model is susceptible to priming, and ran a series of experiments on several Transformer LMs. This methodology constitutes a new way of assessing the representational abilities of LMs via examining their behaviour in controlled setups, which complements tools like Targeted Syntactic Evaluations and the adaptation-based priming measure by Prasad et al. (2019).

Our results in Section 2.5.2 showed that on our *core* corpus, where we control for lexical overlap and semantic similarity between prime and target, *most* of the language models we test exhibit *some* degree of priming for *most* of the constructions we study. This is important, as it opens up the possibility of using priming to investigate what influences the learned representations of these language models.

In Section 2.6, we focused on GPT2-large to conduct a series of subsequent experiments to dig deeper into the impact of different factors on the model's susceptibility to priming. In line with psycholinguistic accounts of residual activation, we found that the effects of priming decrease with the distance between prime and target and increase with the amount of exposure to a given structure. Our results indicate that the structural information being encoded is not fully autonomous from semantics: the Priming Effect is highly boosted by semantic similarity and lexical overlap between the words used in prime and target. Such boosting effects are well known to be present in human language processing as well. Furthermore, the Priming Effect partly disappears with semantically implausible prime sentences, suggesting that semantic plausibility is an important cue for the encoding of structure, arguably more so than in human language processing. Finally, we showed that priming effects remain positive in the presence of phrases with differing degrees of complex-ity across prime and target. This offers some insight into the nature of the representations learned by the model: it suggests that, in addition to abstract sequential structure, some degree of hierarchical syntactic information is being represented.

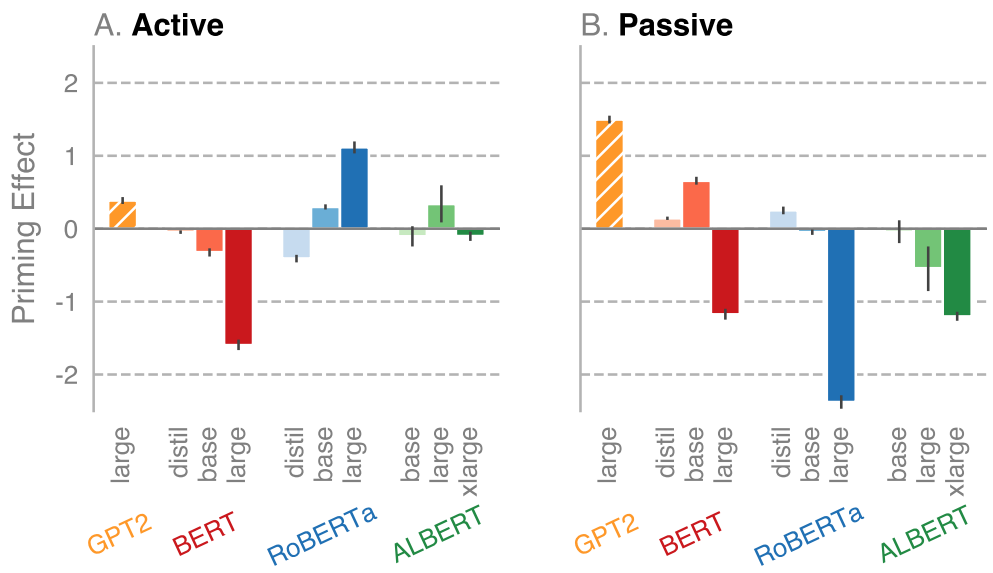

**Figure 2.6:** Priming Effects for three masked language models on the *core* corpus: BERT (Devlin et al., 2019), RoBERTa (Liu et al., 2019), and ALBERT (Lan et al., 2020). To compute sentence probabilities we utilise the pseudo-log-likelihood of Salazar et al. (2020), masking out one token at a time. Results for dative yield a similar pattern.

The current work does not reveal, for the various conditions tested, what the mechanics of the boosting or suppressing effects are. For example, we do not know whether the boosts from lexical overlap or semantic similarity are the result of an improved match with the same structural representations, or of independent factors that influence priming behaviour. Similarly, the precise interplay between semantic plausibility and structural encoding remains unclear. Overall, the pattern of results calls for further investigation using interpretability methods, such as probing and feature attributions, which we plan to pursue in future work.

An additional aspect that requires further study is the role of the training data and its statistics, for example regarding the frequency of the different constructions under investigation and the impact this may have on priming asymmetries within an alternation, and on priming behaviour more generally. An important future step to disentangle the factors that may give rise to priming behaviour would involve training a range of different model types on the same data. This way it becomes possible to interpret the role that model architecture, model size, training objective, and corpus statistics play in shaping the behaviour of the model. An important class of models to include in such studies are Masked Language Models. We conducted a preliminary experiment on three such models, which resulted in biased priming behaviour for all (see Figure 2.6). We postulate that these models may rely less on the structure of a prime because their bi-directional nature allows them to take the entire target sentence into account. However, in order to adequately determine that this is entirely due to their training objective, and not due to external factors stemming from corpus statistics, future work could control for this with newly trained models.

Our study reveals novel details about the potential of LMs to represent structural information and the persistence of this information when making predictions about upcoming sentences. But more generally, we believe our findings also demonstrate the usefulness of the priming paradigm for investigating such questions. Even more generally, they illustrate the benefits of repurposing experimental paradigms from psycholinguistics to investigate the knowledge acquired by large neural language models. In that sense, the current chapter complements exciting recent work that borrows other paradigms from linguistics and psycholinguistics, including grammaticality judgements, few shot learning, and Cloze tests (Gauthier et al., 2020; Brown et al., 2020; Baroni, 2022; Lovering et al., 2021). That is, while syntactic priming offers one window into abstract language representations in neural language models, linguistics offers a whole row of windows that are starting to reveal an exciting vista.

*If you torture the data long enough, it will confess to anything.*

Ronald Coase *(no data were tortured in the making of this thesis)*

# 3 Do Language Models Exhibit Human-like Structural Priming Effects?

In the previous chapter we demonstrated that structural priming effects can be measured in language model predictions. Our approach was behavioural in nature, and focused on average priming effects on a corpus level. In this chapter we take a more precise approach, zooming in on priming effects on an *item level*, as well as a linear mixed-effects modelling methodology to determine the various factors that drive the strength of a priming effect.

## 3.1 Introduction

Structural priming is well attested in humans, for both language production (Mahowald et al., 2016) and comprehension (Tooley, 2023). Interestingly, it has also been shown to occur in large language models, as discussed in the previous chapter and demonstrated by Prasad et al. (2019) and Michaelov et al. (2023). Here, structural priming can be viewed as a simple form of 'in-context learning' (Dong et al., 2023), where the *task* is to generate a sentence (or compute its likelihood) with the target grammatical structure, influenced by the *demonstration* (the *prime* presented to the LLM before processing the target).

Priming effects in humans are typically stronger when there are shared words between prime and target, and when the prime is more unusual, or less frequent. This is the *inverse frequency* effect; it extends to other properties of structures themselves, and it is one of the main phenomena we focus on in this chapter. To explain these effects without direct access to the underlying training data, we turn to factors known to predict priming effects from corpus linguistics (e.g. Gries, 2005; Jaeger and Snider, 2013), which highlight surprisal and structural preference as key factors, and demonstrate the importance of a more fine-grained method of measuring priming.

A second focus of this chapter is examining the relationship between lexico-semantic overlap and the asymmetry of the priming effects observed. We examine priming at the token level, discovering that *where* priming takes place is important for understanding *how* lexico-semantic factors affect priming and for analysing the mechanisms underlying priming in models. Finally, we find that models' structural predictions are highly influenced by specific lexical items, and that they incorporate systematic properties of human production preferences learnt from the training data. We demonstrate that models, like humans, exhibit inverse frequency effects in terms of surprisal and verb preference, and that these are predictive of priming.

## 3.2 Structural Priming

Structural priming in humans is part of a rich literature on factors that impact human language processing, both in controlled experiments of production (Mahowald et al., 2016) and comprehension (Tooley, 2023), and analyses from corpus linguistics (e.g., Gries and Kootstra, 2017). We provide a brief theoretical background on structural priming in §3.2.1, and priming in language models in §3.2.2.

### 3.2.1 Properties of Priming

**Production vs. Comprehension** Structural priming has been shown to manifest in both language production and comprehension. Although recent work has shown that the underlying mechanisms for these two areas may not be as different as originally assumed (Segaert et al., 2013; Tooley and Bock, 2014) and are intricately related (Dell and Chang, 2014), numerous works have uncovered distinct differences in the factors that play a role for each modality (Ziegler and Snedeker, 2019). We therefore take the explicit stance in this paper that language models are likely to follow patterns found in human production, since they are exposed solely to human produced data, and for the factors we consider, we find this to be the case. While LMs are not necessarily expected to align directly with factors found in comprehension studies, arguably there may be similar acquisition mechanisms (e.g. error-based learning) that result in comprehension aligned behaviour. In this back-ground, we focus on production- and corpus-based analyses of structural priming, unless unless explicitly mentioned otherwise.

**Inverse Frequency** One influential theory on the mechanism behind priming in humans is the implicit learning theory by Chang et al. (2006). This theory predicts that our expectation for a particular structure is proportional to degree of ***surprisal*** of having encountered this structure before. This effect —the *inverse frequency effect* (a *rarer* prime will boost priming more) — has indeed been confirmed experimentally to be a strong predictor of priming behaviour. Specifically, in language production in humans it has been found that highly surprising primes (as measured by language models) will have higher priming effects (Gries and Wulff, 2005; Jaeger and Snider, 2008, 2013; Fazekas et al., 2024).

Relatedly, ***structural preference***—which expresses within which structure a verb is most likely to occur—is another important factor when predicting priming behaviour: verbs that are strongly associated with one construction are more likely to be primed by that construction as well (Gries and Wulff, 2005; Gries et al., 2005; Bernolet and Hartsuiker, 2010). From this it then follows that priming effects are stronger when the prime sentence was of a less preferred structure: a prime containing the verb *gave*, for example, will prime subsequent targets more strongly when it is encountered in its *dispreferred* structure (PO) (Pickering and Branigan, 1998; Zhou and Frank, 2023). There exists an extensive line of work into determining the factors that govern this structural preference, which is driven by various complex syntax-semantic interactions (Green, 1974; Thompson and Koide, 1987; Gropen et al., 1989; Bresnan et al., 2007). Inspired by this literature, we find evidence in §3.6 of a verb-mediated inverse frequency effect in modern LLMs.

**Lexical Dependence** Many findings in production and corpus studies have shown that priming effects of sounds, words, meanings and structures interact: prime sentences and target sentences with shared words (***lexical overlap***), or words that share semantics (***semantic overlap***), boost structural priming (Hare and Goldberg, 1999; Jones et al., 2006; Hartsuiker et al., 2008; Snider, 2009; Gerard et al., 2010), and similar findings have been found in comprehension studies as well (Chiarello et al., 1990; Traxler et al., 2014). A common explanation is that words in the prime that are identical or similar to words in the target already activate the relevant abstract syntactic frames. These frames, in turn, are most closely associated with verbs, or the syntactic head of the primed structure (Pickering and Branigan, 1998; Pickering and Ferreira, 2008; Reitter et al., 2011).

Lexical overlap effects in human experiments typically do not consider effect of prepo-sition or determiner overlap, rather focusing on the content words. Findings have shown that structural priming does not depend on the repetition of function words, thus in humans there is a clear difference between content-word and function word repetition (Bock, 1989; Tree and Meijer, 1999; Pickering and Ferreira, 2008).

### 3.2.2 Structural Priming in Language Models

Structural priming has been used to investigate abstract language representations in language models. A number of (early) papers used fine-tuning on a small sample of items of a particular structure, and measured its impact on related items (van Schijndel and Linzen, 2018; Prasad et al., 2019). Our approach in Chapter 2 measured the impact of congruent and incongruent prime sentences on a subsequent target, paralleling approaches in psycholinguistics that view priming as resulting from *residual activations* (Branigan et al., 1999). Using this approach, LMs are shown to exhibit priming effects that are cumulative, susceptible to recency effects, boosted by lexico-semantic overlap, and persisting in cross-lingual settings (Michaelov et al., 2023; Xiao et al., 2024). Concurrent to our work, Zhou et al. (2024) present a similar setup that connects priming effects to inverse frequency effects, demonstrating similarities between priming behaviour and in-context learning.

One key finding of the previous chapter is that priming effects are often asymmetric: when comparing alternative structures in the dative and transitive data, they remark that some of these structures are more susceptible to priming than their alternatives. In §6.5 we confirm this observation for the dative; the strength and direction of the asymmetry are a surprising result, given priming effects are typically higher for the *opposite* alternation in humans (Bock, 1989; Kaschak et al., 2011; Reitter et al., 2011). We show that this finding extends to a wide range of state-of-the-art LLMs, and is predictable via other inverse frequency effects.

## 3.3 Measures, Data & Models

### 3.3.1 Measuring Priming Effects

To measure structural priming, we again make use of the PE metric of Section 2.3, which has recently also been adapted by Sinha et al. (2023) and Michaelov et al. (2023). To contrast this measure with the measure from Equation 3.2, we will refer to it as the **sentence-** **level Priming Effect**, *s*-PE:

$$s\text{-}PE(\textsc{x}) = \log P(\textsc{t}^{\textsc{x}}|\textsc{p}^{\textsc{x}}) - \log P(\textsc{t}^{\textsc{x}}|\textsc{p}^{\textsc{y}}) \quad [3.1]$$

where $\textsc{t}^{\textsc{x}}$ denotes a *target* sentence of structure x, which is preceded by a *prime* of congruent or incongruent structure ($\textsc{p}^{\textsc{x}}/\textsc{p}^{\textsc{y}}$).

The Priming Effect metric of Eq. 3.1 shows whether a target sentence is primed by structural congruence as a whole, but does not provide insight into *which* tokens within the target were most responsible for such an effect. To investigate this, we introduce the **token-level Priming Effect** metric (*w-PE*), which expresses priming effects for each individual target token $\textsc{t}_i^{\textsc{x}}$:

$$\begin{aligned} w\text{-}PE(\textsc{x}, i) &= \log P(\textsc{t}_i^{\textsc{x}}|\textsc{p}^{\textsc{x}}, \textsc{t}_{<i}^{\textsc{x}}) \\ &\quad - \log P(\textsc{t}_i^{\textsc{x}}|\textsc{p}^{\textsc{y}}, \textsc{t}_{<i}^{\textsc{x}}) \end{aligned} \quad [3.2]$$

Note that the sentence-level PE decomposes into a sum of *w*-PE scores; as such *w*-PE expresses the relative contribution of each target token to *s*-PE:

$$s\text{-}PE(\textsc{x}) = \sum_i w\text{-}PE(\textsc{x}, i)$$

### 3.3.2 The Prime-LM Corpus

We use the dative constructions from the Prime-LM corpus of Section 2.4. This subset of sentences is convenient for our purposes, because we can select both prime-target pairs with *no* lexical overlap and minimal semantic similarity between nouns and verbs, as well as pairs with varying degrees of overlap and varying degrees of semantic similarity. The datives thus allow us to not only measure structural priming, but also inspect the role of lexical overlap in more detail. We briefly explain the subsets we select for our experiments (each containing 15.000 prime/target pairs), as well as two additional sub-conditions we introduce to the lexical overlap category.

**Core** contains a) no lexical overlap exists between prime and target sentences, not even between function words, and b) no semantic association exists between prime and target exists in the USF free association norms dataset (Nelson et al., 2004). In our experiments, we use the Core condition as a baseline.

**Semantic Similarity** contains explicit pairwise semantic similarity between prime and target, where similarity is assessed by a non-zero human association from the USF dataset or a minimum cosine similarity of at least 0.4 based on GPT2-large embeddings. We consider three conditions : i) all nouns are semantically similar, ii) the verbs are similar, iii) all nouns *and* verbs are similar.

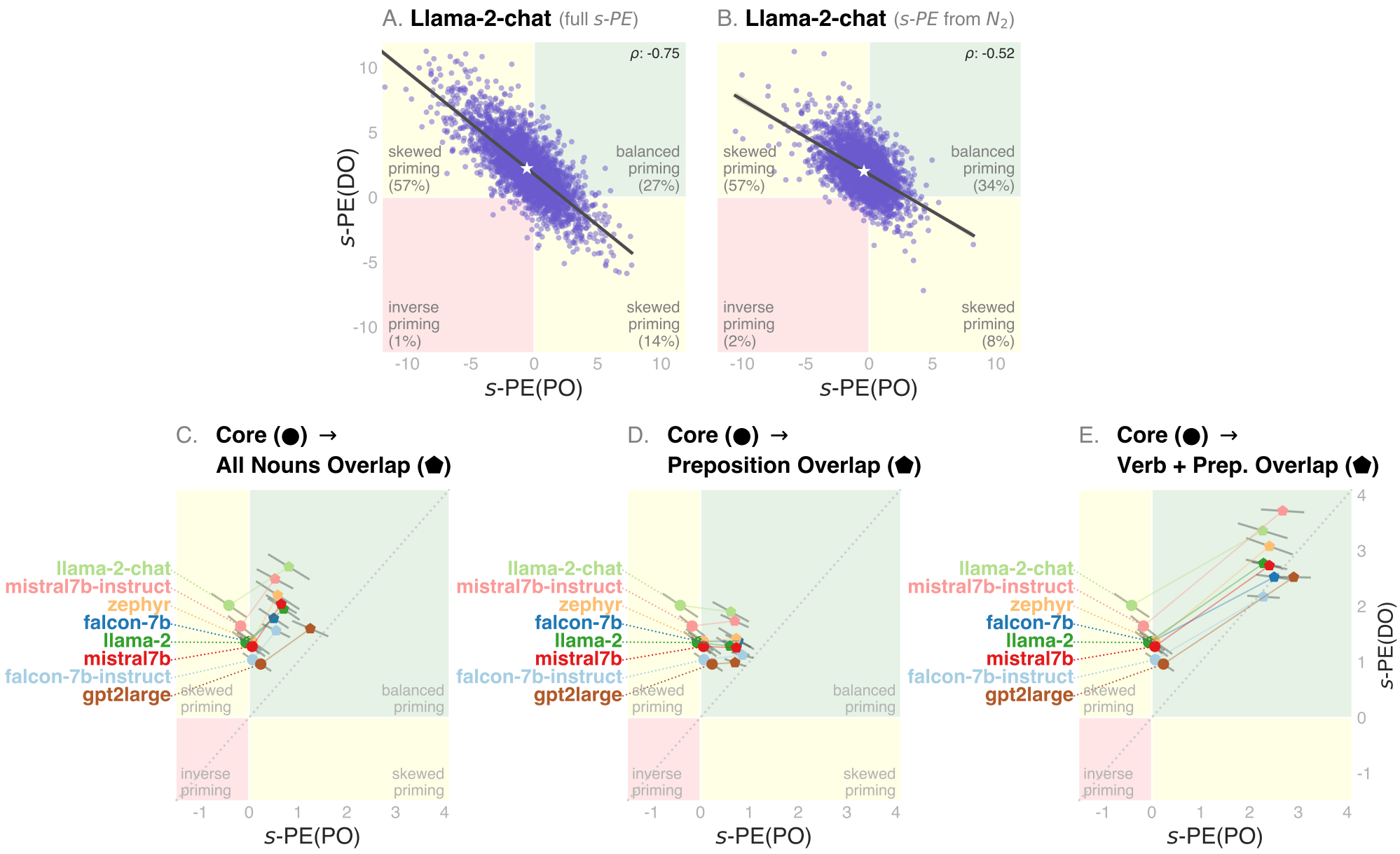

**Figure 3.1:** We plot PE results against one another. The four quadrants in this '*PE space*': *balanced priming* where the PE is positive in both directions, *skewed priming* where it's only positive in one, and *inverse priming* when the PE is negative in both directions. There exists a strong negative correlation between priming effects of opposite structures (A). Only a small portion of the data is primed in both directions for Core. Priming becomes more balanced when measured from the point of divergence in the target (B, §3.5), or when lexical overlap is increased (C–E).

**Lexical Overlap** ensures lexical items are shared across prime and target. We consider three such conditions : i) all nouns overlap, ii) determiners and prepositions overlap, iii) verbs overlap. We create two additional conditions, iv) determiner overlap and v) preposition overlap. This allows us to separately measure the impact of determiners and prepositions, since *verb overlap* necessitates preposition overlap.

### 3.3.3 Models

We consider the following (auto-regressive) LLMs. For models with an * we also test their *aligned* versions. PE scores are computed using the `diagnnose` library (Jumelet, 2020).

**GPT2-large** (Radford et al., 2019): This is the impactful 2019 model from OpenAI, with 774M parameters, trained only on a (causal) language modelling objective.

*****Llama-2-7b** (Touvron et al., 2023): We consider both the 7B base model and the RLHF/PPO aligned *chat* model (Ouyang et al., 2022).

*****Falcon-7b** (Almazrouei et al., 2023): We consider both the 7B base, and the *instruction-tuned* variant fine-tuned on dialogue data taken from ChatGPT (OpenAI, 2023).

*****Mistral-7b** (Jiang et al., 2023): This 7B model is the current state-of-the-art in this size bracket. We also consider the instruction-tuned variant, trained on similar data to Falcon-7b.

*****Zephyr** (Tunstall et al., 2023): An aligned version of Mistral-7b using Direct Preference Optimization (Rafailov et al., 2023).

## 3.4 Experiment 1: Measuring Structural Priming

We aim to better understand the asymmetrical priming effects observed in §2.2, to gain a more detailed picture of how lexical overlap affects this asymmetry. We start our experimental setup with their sentence-level approach, considering a wider range of large, contemporary LLMs. In the next section we then examine priming effects at a more fine-grained level.

**Priming Effects are skewed and correlated** We compute the *s*-PE scores for the models of §3.3.3 on the Core condition of Prime-LM. We observe there exists a strong negative correlation between the PE scores of the prepositional object and the double object. In Figure 3.1A we plot those results as a scatter plot in the space formed by PE score for one construction against PE score for the alternative construction. This representation highlights that, for Llama-2-Chat and all the other models we consider, the *s*-PE of PO constructions is *negatively* correlated with that of DO constructions ($\rho$: -0.72 to -0.77), and that only for a fraction of sentences there exists a positive priming effect in both directions (26 to 38%).

This correlation and skew towards one of the two constructions were already observed in Chapter 2 for GPT2-large and other relatively small LMs. Interestingly, correlation and skew do also exist in the newest, large LMs, and, moreover, are even more pronounced. Figure 3.1C shows the mean *s*-PE in the same *PE space* for all the LLMs we considered. GPT2-large shows the least skew, Llama-2-chat the most (for completeness, the plot also shows the strength of the correlation between the PE(PO) and PE(DO) scores, as well as the spread of the distribution.) Note that this observed behaviour of *large* LMs is less consistent and far more asymmetric than results in the human literature, where priming effects, while typically asymmetric (Bock, 1989) , are generally observed to be positive for both structures.

**Lexical overlap balances Priming Effects** Next, we investigate the priming effects of the LMs where either the semantic similarity or lexical overlap between prime and target is increased. The results for lexical overlap are shown in Figure 3.1C–E (additional plots regarding semantic similarity are in Appendix A.1.1). The plots show that an increase in lexical overlap of any type moves all models more solidly into the upper-right quadrant of the PE-space. That is, it pushes all LM priming behaviours to become both stronger and more balanced (less skewed towards one or the other construction). This is especially prevalent for the overlap in verbs and function words. We will explore the impact of these factors in more detail in the next section.

## 3.5 Experiment 2: Locating Structural Priming

Sentence-level analysis does not allow us to investigate individual token-level predictions, making it impossible to examine *where* in the target sentence priming effects are at their strongest. To better understand the *s*-PE results of §3.4 we thus compute the token-level *w*-PE scores for the same conditions.

**Structural Divergence** Figure 3.3 shows the average *w*-PE scores for Llama-2 on the *Core* condition, which exhibits much higher sentence-level priming effects for the DO sentence than for the PO sentence. The *token-level* scores show that the treatment of the two sentences starts to diverge from the position of the *second* noun onwards (the second noun is is the patient for PO, whereas it is the recipient for DO). Prior to that, the target sentence is, in fact, the same for both PO/DO alternations and as such will be inversely proportional to each other: scores up till this point merely show a target's bias towards a prime of a particular structure, regardless of structural congruence.

This provides a partial explanation for the strong negative correlation between *s*-PE scores that was observed in §3.4: since (roughly) half of *s*-PE score is made up of *w*-PE scores that have a perfectly negative correlation of -1, the overall correlation of *s*-PE scores is strongly affected by this. Based on this insight we compute a modified *s*-PE score that is

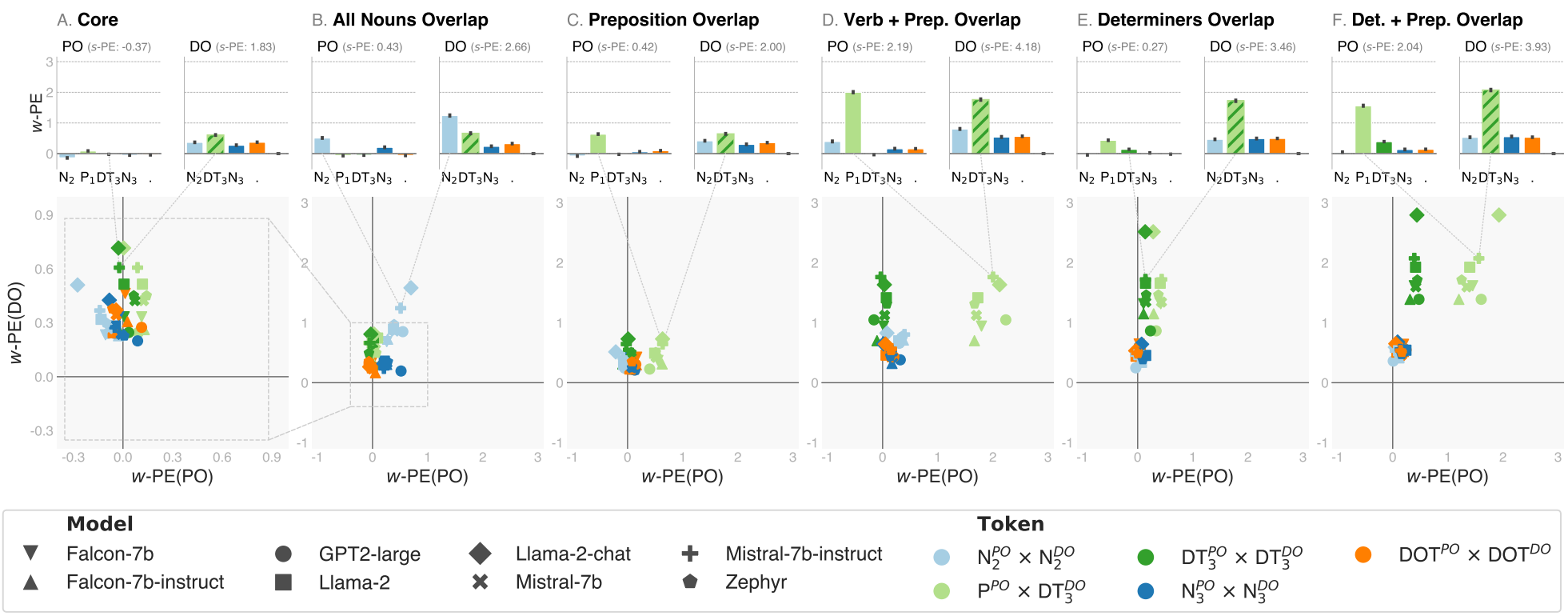

**Figure 3.2:** The *w*-PE scores for the *Core* and *Lexical Overlap* conditions. Scores are grouped by token (based on colour) and model (based on shape). To exemplify how these *Priming Space* coordinates map to a bar chart, we show the Mistral-7b-instruct scores at the top of each plot. Note that the *Core* results are plotted at a different scale than the other conditions. PO: *The girl gave the* **ball** to the **boy** . DO: *The girl gave the* **boy** the **ball** .

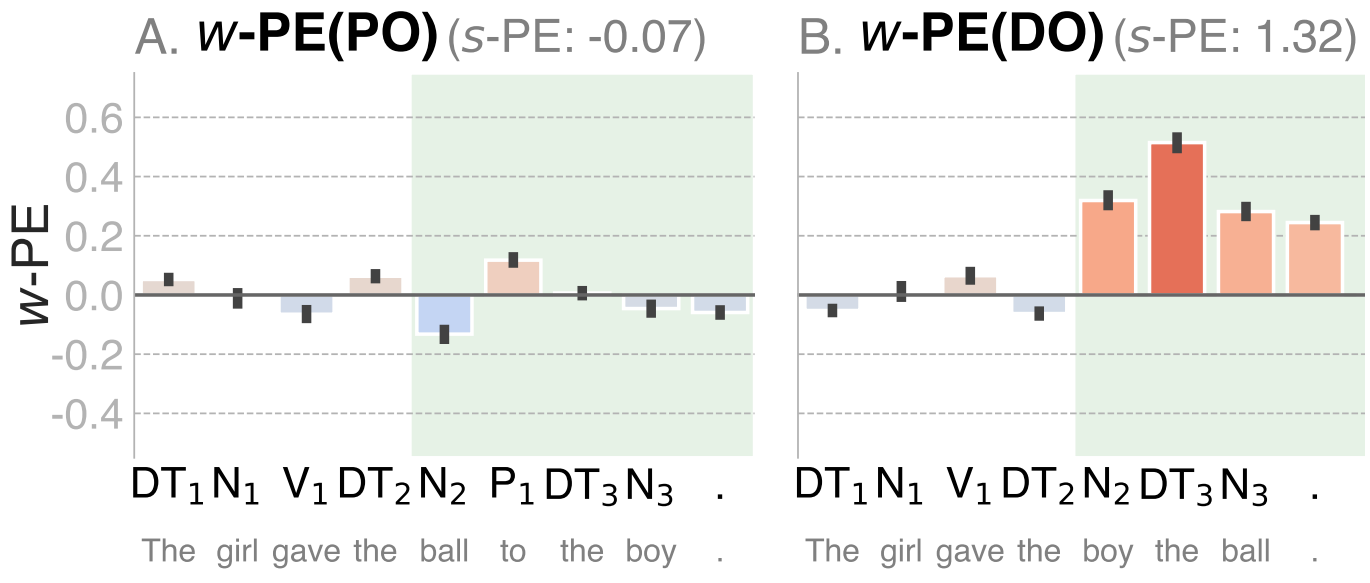

**Figure 3.3:** The token-level Priming Effect reveals which token predictions in the target sentence contributed the most to the overall sentence-level Priming Effect, here averaged for Llama-2 over the *Core*. It is inversely correlated up to the point of divergence between the two structures, at the position of the second noun ($N_2$). Only after that point can the congruence between prime and target start playing a role.

only measured from the point of target sentence divergence ($s_\eth$):

$$s_\eth\text{-}PE(\mathbf{x}) = \sum_{i>4} w\text{-}PE(\mathbf{x}, i) \qquad [3.3]$$

This allows us to confirm that the negative correlation between PO and DO decreases with $s_\eth$-*PE* ($\rho$: -0.76 to -0.52 for Llama-2-chat; Figure 3.1B).

**Lexical Dependence** We focus our analysis on *lexical overlap* (Figure 3.2), which showed the strongest balancing effects in §3.4.[1] Priming behaviour could be distributed in two ways across the target: either uniformly or peaked at a particular token, and either balanced or skewed towards one structure. From the point of structural divergence, we have the noun ($N_2$) of the first noun phrase, followed by the function word ($P_1$:PO, $DT_3$:DO) that marks the start of the second noun phrase of the construction. Priming effects for $N_2$ stem from cues with respect to the *semantic role* (e.g. gave the ball$_{PO}$ | boy$_{DO}$). Numerous works in production have shown priming to already take place at this location (Pickering and Branigan, 1998; Cleland and Pickering, 2003). We would therefore expect to find some evidence of *balanced* priming from the $N_2$ within the core condition. However, the *skewed* priming we observe in *Core* (3.2A) suggests that the semantic role of the noun does not play as important a part in structural prediction for models. Indeed, we observe the most consistent and balanced priming effects from the start of the second NP (*w*-PE(PO, $P_1$), *w*-PE(DO, $DT_3$)), suggesting that models only narrow their structural predictions later on within a sentence.

[1] Results for semantic similarity are provided in Appendix A.1.2.

Next, we observe the local impact of lexical overlap between prime and target. For overlapping nouns, we can see that the *w*-PE for both $N_2$ and $N_3$ has increased significantly for both PO and DO. The other tokens, on the other hand, are not impacted by this overlap at all: the priming boost manifests itself solely at the position of the overlapping token. For verb overlap, we show that the increased *s*—PE scores here stem from the verb as well as the prepositional overlap (necessary when sharing the same verb and preserving semantics) resulting in significantly larger *w*-PE(PO, P) scores (Figures 3.2C and D). Interestingly, verb overlap also leads to a boost in $N_2$ and $N_3$, compared to the *Core*. This shows that, under this condition, the model *is* aware of the expected semantic role in the $N_2$ position: the verb overlap has primed the model in the DO case to expect an animate entity here (and inanimate for PO). Unlike findings in the human literature for both production (Bock, 1989) and comprehension (Traxler, 2008), we observe prepositional overlap strongly boosting priming effects in the language models we investigate.

## 3.6 Experiment 3: Explaining Structural Priming

We now take a closer look at the factors that impact Priming Effects by conducting a regression analysis inspired by factors from corpus linguistics and production studies (Gries, 2005; Jaeger and Snider, 2013). Following Gries (2011), we make use of linear mixed effects models to determine salient word and sentence level factors that predict priming, to discover whether models display consistent behaviour with respect to one another and to human patterns of priming in production they may have learnt. We first describe the fac-tors we use in our regression analysis in §3.6.1, and present the results in §3.6.2.

### 3.6.1 Priming Factors

We investigate the two broad categories of factors discussed in §3.2: *lexical dependence*, making use of the various conditions of the Prime-LM corpus (§3.3.2), and *inverse frequency*, choosing to focus on sentence-level surprisal and the structural preference of the verbs used.

**Lexical Dependence** We include pairwise token-level ***semantic similarity*** across prime and target content words, measured as the cosine similarity of the word embeddings taken from GPT2-large. We also include sentence-level similarity, based on the sentence embeddings of MPNet (Song et al., 2020), a high-performance sentence encoder. Here, we compute the cosine similarity between the PO prime and target embeddings. We add ***lexical overlap*** as a binary factor per token to our analysis. This allows us to separate overlap effects in conditions where multiple tokens overlap, which is not possible in corpus-level experiments.

**Surprisal** We include the ***surprisal*** of the congruent and incongruent prime and target, based on the negative log likelihood of the language model itself. Surprisal gives us a measure of how predictable or expected the sentences are as a whole, encompassing within-sentence collocation frequency effects.

**Structural Preference** Whereas corpus-based analysis of preferences is often based on normalised frequency statistics (Gries and Stefanowitsch, 2004), we base preference on the average probability difference of a verb in two alternating structures:

$$\text{PO-}\mathit{pref}(v) = \frac{1}{|\mathcal{V}|} \sum_{s \in \mathcal{V}} \log P(s^{\text{PO}}) - \log P(s^{\text{DO}}) \quad [3.4]$$

where $\mathcal{V}$ is the set of sentences containing verb $v$. This score expresses a verb's preference towards a particular structure on a scale from DO to PO. Hawkins et al. (2020) and Veenboer and Bloem (2023) provide a similar methodology for measuring structural preferences in LMs. For computing these scores we make use of the prime sentences from the *Core* condition of PrimeLM.

The majority of verbs have a preference towards PO structure (as an example, Figure 3.4 contains the preferences of GPT2-large and Llama-2). This is not in line with dative usage preferences found in English, although it varies across vernaculars: some production and corpus studies suggest American English has a 2:1 preference towards DO constructions Bock and Griffin (2000); Grimm and Bresnan (2009), whereas Australian English has a PO preference (Bresnan and Ford, 2010). Preference towards PO in LMs may be confounded by transitive verb phrases followed by a prepositional modifier.

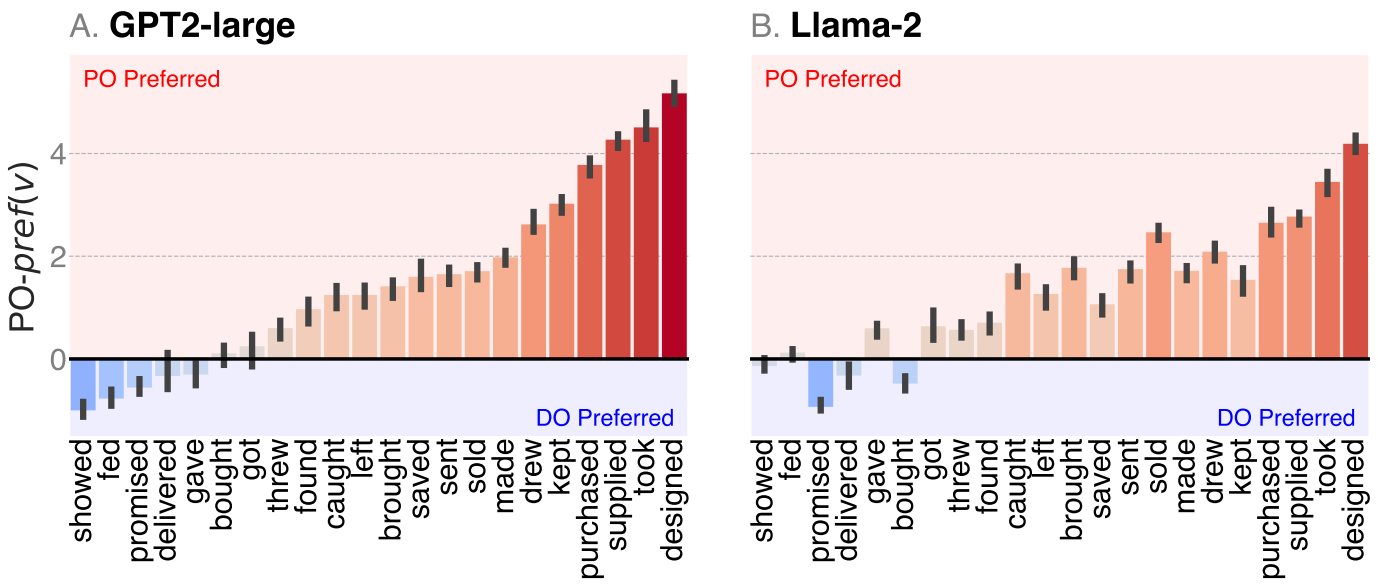

**Figure 3.4:** Structural preferences for GPT2-large and Llama-2, expressing the preference of a ditransitive verb with respect to a prepositional object versus a double object construction. The verb order of Llama-2 is based on the sorted order of GPT2-large.

We also compute the Spearman correlations between the preference orders of all LMs and humans (Gries and Stefanowitsch, 2004), which reveals that there exists a high degree of variance across models and low correlation across models and human preference order (full figure in Appendix A.1.4). We leave a more thorough investigation of these differences open for future work that can take inspiration from established linguistic findings (Gropen et al., 1989; Arnold et al., 2000).

### 3.6.2 Modeling Priming Effects

**Linear Mixed Model** We fit a linear mixed model (LMM) using the factors of §3.6.1 that are added as fixed effects (Baayen et al., 2008). We fit two LMMs: one for predicting $s_{\vartheta}$-PE(PO) and one for $s_{\vartheta}$-PE(DO), which will provide insights whether different factors predict these effects. Fitting is done based on 30.000 items: 15.000 items are sampled for the *Core* condition, and 15.000 items are sample from the *Semantic Similarity* and *Lexical Overlap* conditions. This provides a balanced dataset of the *Core* and conditions that diverge from this baseline. We add a by-LM random intercept to account for individual model biases, akin to by-speaker random effects in human priming studies (Gries, 2011; Jaeger and Snider, 2013). All factors are centred and scaled to unit variance.

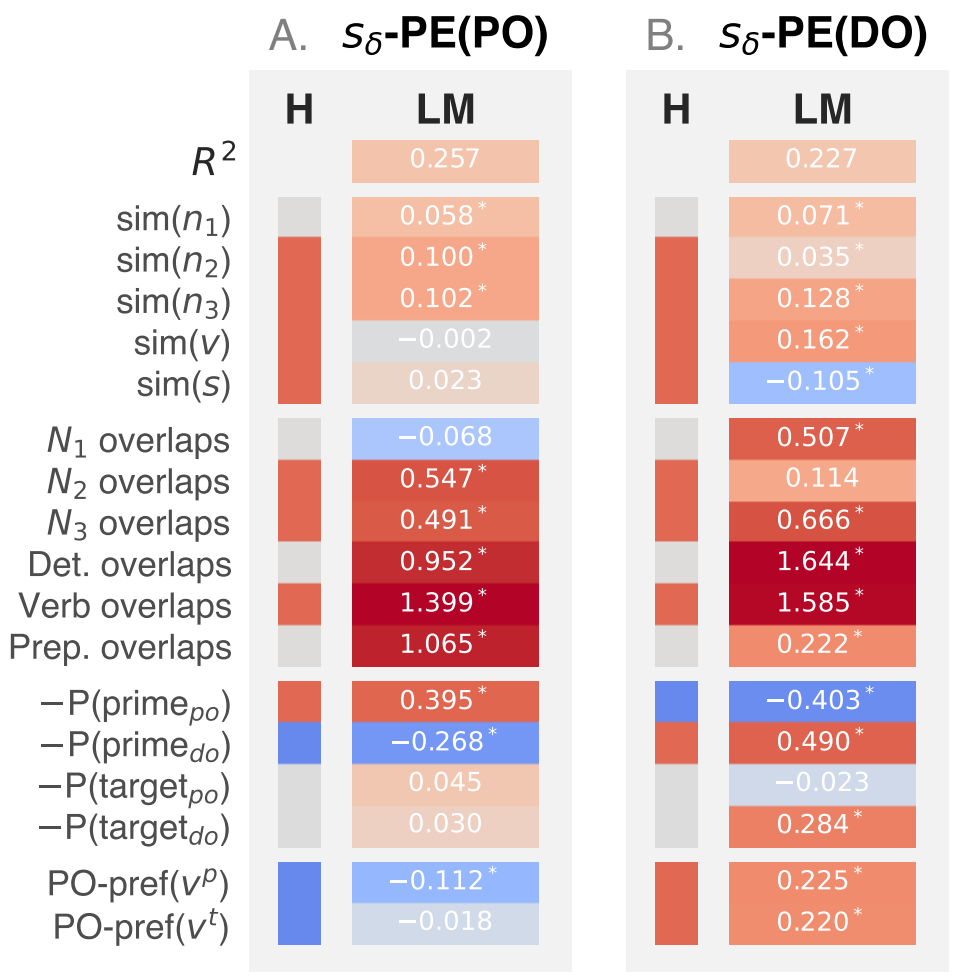

**Figure 3.5:** LMM coefficients for (A) predicting $s_\delta$-PE(PO) and (B) $s_\delta$-PE(DO), shown side-by-side with reported effects for predicting human priming in production- and corpus-based studies. Significant LLM coefficients ($p < 10^{-3}$) are denoted by an asterisk.

**Results** We report full LMM with coefficients in Figure 3.5, next to the reported effects found in the literature on human priming ($z$-scores and standard errors in Appendix A.1.5). The LMM reaches an $R^2$ of 0.257 (PO) and 0.227 (DO), which indicates that a large fraction of PE could still be predicted based on other factors and more complex interactions. We leave a more extensive exploratory analysis for future work, and focus on confirmatory hypothesis testing for now (Tukey, 1980; Barr et al., 2013).

**Semantic Similarity** As in human production and corpus linguistics findings, we observe priming effects are predicted by semantic similarity between prime and target words (Snider, 2009), with the most consistent effects across structures for noun similarity (Cleland and Pickering, 2003), although the effect is relatively weak compared to other factors. Sentence similarity however, was not a predictive factor. In part this could be due to the sentence encoder not being sensitive to the structurally similar PO items that we use for computing sentence-level similarity. Incorporating the feature-based Grower similarity employed by Snider (2009) could be an alternative to explore the relation between sentence-level semantic similarity and priming.

**Lexical Overlap** We observe that lexical overlap is the strongest predictor of priming behaviour, in particular for primes sharing the same verbs, prepositions and determiners as the targets. This is in line with human findings: the meta analysis of Mahowald et al.

(2016) shows lexical overlap is 'the most consistent moderator of syntactic priming'. We also observe that shared determiner overlap is consistently of high importance when predicting model PE, something observed but given far less attention in the human literature. Contrary to human findings in both production (e.g. Bock, 1989) and comprehension (e.g. Traxler, 2008), prepositional overlap is one of the strongest priming predictors. This indicates that priming in LMs is strongly driven by lexical cues, tying in with our observation in §3.5 that priming effects are highly influened by single token prediction, and this is driven more strongly by function than content words.

**Surprisal** Similar to Jaeger and Snider (2013) and Fazekas et al. (2024), we find that priming effect is predicted by prime surprisal, in both directions for PO and DO. This is evidence for an inverse frequency effect: a less frequent/plausible prime leads to an *increase* in priming effect. Target surprisal is less significant: only DO surprisal is a significant predictor.

**Structural Preference** We find that verb preference plays a highly predictive role, which again provides evidence for inverse frequency effects. A verb that has a structural preference for PO will lead to a higher DO priming effect, and vice versa, in line with results observed in human data (Gries and Wulff, 2005; Gries et al., 2005). This provides further explanation for the DO skewed priming effects that models display: for most of DO targets, their primes will *not* be in the preferred structure, thus boosting priming effects.

## 3.7 Discussion & Conclusion

In this chapter, we seek to better understand the mechanisms that may underlie structural priming behaviour in LLMs. Borrowing insights from empirical and theoretical work on priming in humans, we investigate how, where and why a range of modern LLMs assign higher or lower probabilities to target sentences depending on preceding context, allowing us to investigate the extent to which language models are influenced by structure and semantics when making upcoming predictions.

**Do models demonstrate structural priming?** We find, in line with Chapter 2, that models exhibit asymmetrical priming effects, and that this is even more pronounced in newer, larger LMs. By introducing a token-level priming effect we are able to locate more precisely potential sources of this asymmetry. We observe the direction of the asymmetry in PE is consistently *inverse* to priming effects in humans: where humans consistently display higher PE for the PO alternation, rather than DO, which we observe in models. We speculate that the verb preference effects we find in §3.6, which are predictive of PE as in humans, may play a role in this. Finally, through observing priming at the token level, we observe that balanced priming in models is only visible from later on within the target, later than we may expect the same effects to be observed in humans.

**How do humans and models compare?** In predicting model PE using human factors known to predict priming in humans and in human corpora, we observe that lexico-semantic and frequency-related predictors of priming in humans also predict priming in LLMs. However, although we find lexico-semantic overlap of content words to be a reliable predictor of priming, we find that *function word overlap* plays a surprisingly predictive role, which has not been shown to be the case for humans with dative constructions (Bock, 1989; Pickering and Ferreira, 2008). Similar to human findings, we observe—consistently across models—inverse frequency effects of prime surprisal and verb preference (e.g. Gries and Wulff, 2005; Jaeger and Snider, 2008). This demonstrates that models are able to pick up on highly abstract factors influencing language predictions in humans from corpora, and highlights how influential seemingly small properties of the context are when it comes to upcoming model predictions.

**What are the implications of implicit learning?** We showed that priming is driven by similar inverse frequency effects observed in human priming. From a broader perspective, this is a striking finding. Inverse frequency effects have been argued to stem from an error-based implicit learning procedure (Chang et al., 2006): we adapt future predictions proportionally to recent predictive errors. This cognitive mechanism then leads to detectable patterns in human-produced corpus data (Jaeger and Snider, 2013), on which LLMs are trained. LLMs are thus able to pick up on this highly abstract pattern, which shows that their priming behaviour is far more intricate than a simple repetition-based mechanism. An interesting endeavour for future work would be to test this finding in a setting with control over data distribution (e.g. Jumelet and Zuidema (2023b)), to ensure that inverse frequency effects do not stem from some other indirect effect of language learning.

**Comprehension and production in LLMs** We build on the work described in Chapter 2, where we design the priming effect metric to measure *comprehension* in LLMs forcing its prediction on a fixed target without allowing for open-ended production. It is important to remember, however, that through the way LLMs are trained, their predictions will be driven by human *production* patterns in the training data, thus motivating our choice to base our predictions on findings from corpus and production studies. Although the mechanisms for comprehension and production in LLMs are highly linked—they rely on the same output distribution—it would be interesting to investigate priming in LLMs in a generation-based setting as well. A thorough investigation in this direction may provide deeper insights into the relation between LLM behaviour and cognitive theories of human language processing (e.g., Dell and Chang, 2014).

**Outlook** Language models as cognitive models can potentially aid in discovering important properties of human linguistic behaviour (Futrell et al., 2019; Linzen and Baroni, 2021; Wilcox et al., 2023); we thus view our results as a contribution to defining the border where human patterns are replicated in models. Looking outwards from this detailed analysis, future studies could investigate the extent to which these priming effects influence structural repetition patterns in generation, complementing existing work finding priming-like lexical repetition effects in LLM generation (Molnar et al., 2023). Furthermore, a more detailed investigation in the exact nature of the (potentially) hierarchical representations that underlie priming behaviour could take inspiration from parsing-based theories of priming (Prasad and Linzen, 2024), or deploy techniques from interpretability research to uncover hierarchical structure (Murty et al., 2023; Jumelet and Zuidema, 2023a). Not only will such investigations provide deeper insights into the cognitive plausibility of LLMs (Beinborn and Hollenstein, 2023), but it may also yield a better understanding of the mechanisms underlying in-context learning (Min et al., 2022; Han et al., 2023).

# Part II

# Connecting Data Statistics to Model Behaviour

3

# Overview

In the first chapter of this thesis we showed how abstract linguistic structure plays a role in model predictions. In Chapter 3 we then connected these patterns to various factors derived from data statistics. This approach, however, was *observational* in nature and did not connect model behaviour to the actual distribution of the data it was trained on. In the next three chapters we present various approaches in which we create more explicit connections between the data a model was trained on, and how the data distribution shapes subsequent model predictions.

**Chapter 4** In English and other languages, multiple adjectives in a complex noun phrase show intricate ordering patterns that have been a target of much linguistic theory. These patterns offer an opportunity to assess the ability of language models (LMs) to learn subtle rules of language involving factors that cross the traditional divisions of syntax, semantics, and pragmatics. We review existing hypotheses designed to explain Adjective Order Preferences (AOPs) in humans and develop a setup to study AOPs in LMs: we present a reusable corpus of adjective pairs and define AOP measures for LMs. With these tools, we study a series of LMs across intermediate checkpoints during training. We find that all models' predictions are much closer to human AOPs than predictions generated by factors identified in theoretical linguistics. At the same time, we demonstrate that the observed AOPs in LMs are strongly correlated with the frequency of the adjective pairs in the training data and report limited generalisation to unseen combinations. This highlights the difficulty in establishing the link between LM performance and linguistic theory. We therefore conclude with a road map for future studies our results set the stage for, and a discussion of key questions about the nature of knowledge in LMs and their ability to generalise beyond the training sets.

**Chapter 5** This chapter introduces **Fi**ltered **C**orpus **T**raining, a method that trains language models (LMs) on corpora with certain linguistic constructions filtered out from the training data, and uses it to measure the ability of LMs to perform linguistic generalisation on the basis of indirect evidence. We apply the method to both LSTM and Transformer LMs (of roughly comparable size), developing filtered corpora that target a wide range of linguistic phenomena. Our results show that while transformers are better qua LMs (as measured by perplexity), both models perform equally and surprisingly well on linguistic generalisation measures, suggesting that they are capable of generalising from indirect evidence.

**Chapter 6** We investigate the semantic knowledge of language models (LMs), focusing on (1) whether these LMs create categories of linguistic environments based on their semantic *monotonicity* properties, and (2) whether these categories play a similar role in LMs as in human language understanding, using negative polarity item licensing as a case study. We introduce a series of experiments consisting of probing with diagnostic classifiers (DCs), linguistic acceptability tasks, as well as a novel *DC ranking* method that tightly connects the probing results to the inner workings of the LM. By applying our experimental pipeline to LMs trained on various filtered corpora, we are able to gain stronger insights into the semantic generalisations that are acquired by these models.[2]

## Author Contributions

The contents of this part are based on the following publications:

1. Jaap Jumelet, Lisa Bylinina, Willem Zuidema, and Jakub Szymanik. 2024a. Black big boxes: Do language models hide a theory of adjective order? Under Review

2. Abhinav Patil*, Jaap Jumelet*, Yu Ying Chiu, Andy Lapastora, Peter Shen, Lexie Wang, Clevis Willrich, and Shane Steinert-Threlkeld. 2024. Filtered corpus training (FiCT) shows that language models can generalize from indirect evidence. *Transactions of the Association for Computational Linguistics*

3. Jaap Jumelet, Milica Denic, Jakub Szymanik, Dieuwke Hupkes, and Shane Steinert-Threlkeld. 2021. Language models use monotonicity to assess NPI licensing. In *Findings of the Association for Computational Linguistics: ACL-IJCNLP 2021*, pages 4958–4969, Online. Association for Computational Linguistics

The first paper was led by me, with supervision from Lisa, Willem, and Jakub. Jakub proposed the idea for investigating adjective order in language models. Lisa conducted various experiments on context dependence and helped out with writing the original draft.

[2]Note that these three chapters are not presented in the order of their publication dates. Chapter 6 was published three years prior to Chapter 5, and was the initial inspiration for this chapter. Due to this, some of the approaches in Chapter 6 might feel slightly outdated, in particular our sole focus on LSTMs.

The second paper was led by Abhinav, Shane, and me. The FiCT methodology was proposed by Shane and me. Abhinav focused on model training and corpus creation, whereas I conducted various experiments for analysing the results. Yu Ying, Andy, Peter, Lexie, and Clevis helped out designing some of the FiCT filters. The first draft of the paper was written by Abhinav as part of his master thesis, which Shane and I took as a starting point for writing it up as a journal paper. The third paper was led by me, with supervision from the four other authors. We provide more detailed author contributions following the Contributor Role Taxonomy (Allen et al., 2019) in Table 3.1.

| | JJ | LB | WZ | JS |
|---|---|---|---|---|
| Conceptualisation | ✓ | ✓ | ✓ | ✓ |
| Methodology | ✓ | | | ✓ |
| Software | ✓ | | | |
| Data Curation | ✓ | | | |
| Investigation | ✓ | ✓ | | |
| Visualisation | ✓ | | | |
| Analysis | ✓ | | | |
| Writing - Original Draft | ✓ | ✓ | | |
| Writing - Review & Editing | ✓ | ✓ | ✓ | ✓ |
| Supervision | | ✓ | ✓ | ✓ |

Jumelet et al. (2024a)

| | AP | JJ | SST |
|---|---|---|---|
| Conceptualisation | ✓ | ✓ | ✓ |
| Methodology | ✓ | ✓ | ✓ |
| Software | ✓ | ✓ | ✓ |
| Data Curation | ✓ | ✓ | ✓ |
| Investigation | ✓ | | |
| Visualisation | ✓ | ✓ | |
| Analysis | ✓ | ✓ | ✓ |
| Writing - Original Draft | ✓ | ✓ | ✓ |
| Writing - Review & Editing | ✓ | ✓ | ✓ |
| Supervision | | ✓ | ✓ |

Patil* et al. (2024)

| | JJ | MD | JS | DH | SST |
|---|---|---|---|---|---|
| Conceptualisation | ✓ | ✓ | ✓ | ✓ | ✓ |
| Methodology | ✓ | ✓ | ✓ | ✓ | ✓ |
| Software | ✓ | | | | |
| Data Curation | ✓ | | | | |
| Investigation | ✓ | | | | |
| Visualisation | ✓ | | | | |
| Analysis | ✓ | | | | |
| Writing - Original Draft | ✓ | ✓ | ✓ | ✓ | ✓ |
| Writing - Review & Editing | ✓ | ✓ | ✓ | ✓ | ✓ |
| Supervision | | | ✓ | ✓ | ✓ |

Jumelet et al. (2021)

**Table 3.1:** Author contributions for the work described in Chapters 4, 5, and 6.

*Even a man's exact imitation of the song of the nightingale displeases us when we discover that it is a mimicry, and not the nightingale.*

Immanuel Kant

# 4 Black Big Boxes: Do Language Models Hide a Theory of Adjective Order?

Linguistic capabilities of language models have been subject to a lot of research and benchmarking (e.g., Marvin and Linzen 2018; Hu et al. 2020; Gauthier et al. 2020). While LMs show impressive fluency in language generation and often score high on linguistic benchmarks, the mechanisms underlying their linguistic proficiency are hard to uncover. To what extent do LMs generalise beyond mere memorisation of training data? Do they acquire abstract linguistic rules and constraints that don't boil down to basic co-occurrence patterns? How similar are linguistic generalisations in LMs to those that humans have?

## 4.1 Introduction

These questions have been asked and partially answered for different constructions and grammatical phenomena. The focus of such studies has mostly been on strict and binary grammatical constraints, where a violation results in an ungrammatical sequence. There has been less attention to how 'softer' linguistic constraints are captured by LMs. Such constraints can give rise to slight and defeasible preferences rather than sharp acceptability contrasts. One example is adjective order in noun phrases where multiple adjectives modify one noun, e.g. *large wooden box*. Changing the order of adjectives does not necessarily lead to unacceptability, but can result in a slight decrease in naturalness (Dyer et al., 2023) ('>'

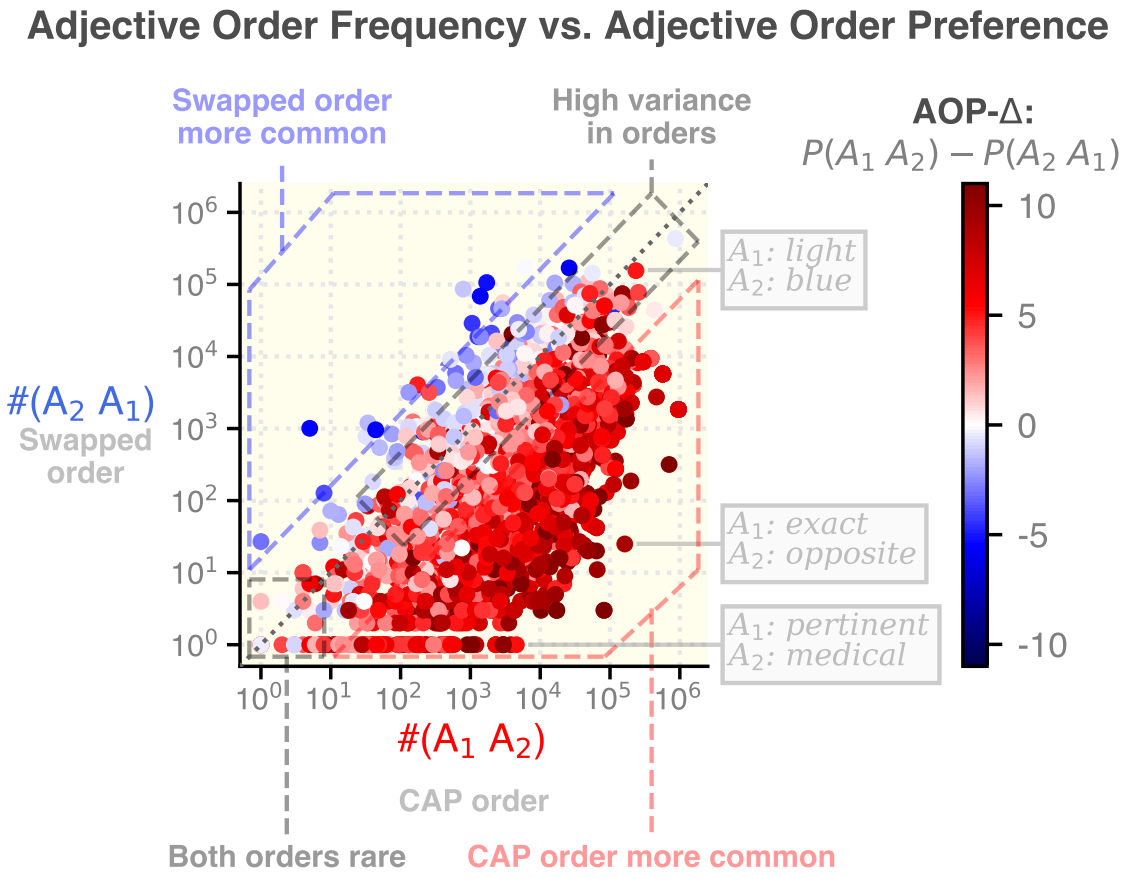

**Figure 4.1:** We connect the **adjective order preferences** (AOP-Δ, §4.3.1) of language models (here Pythia-12b) to the **adjective order frequencies** of the corpus they have been trained on (The Pile). We highlight various regions of interest: adjective pairs for which both orders are rare and that require the LM to generalize from other adjective orders; pairs for which one particular order is far more common that can be resolved from frequency alone; and orders with high variance that rely more strongly on **context**.

marks preference):

*large wooden box* > *wooden large box*

Moreover, these preferences are context-dependent: often, a context can be found where preferences switch – for instance, when a contrasting property is introduced (Teodorescu, 2006).

*Take a **wooden large box**, not a plastic one.*

\> ... *large wooden box* ...

These adjective order preferences (**AOP**s) have been subject to a lot of interest in theoretical linguistics, but proved difficult to capture due to their graded and context-sensitive nature (see Levshina et al. 2023 for arguments in favor of a systematically gradient approach to word order). A number of factors ranging from superficial properties like adjective length to semantic and pragmatic properties to information-theoretic properties of a sentence have been identified as predictors for AOP. But even with all those factors combined together, a large portion of AOPs remains unaccounted for: the best model combining different predictors explains only 71% of human AOPs (Dyer et al., 2023).

Adjective order offers an opportunity to assess the ability of LMs to learn subtle linguistic preferences involving implicit, hard-to-identify factors. This paper is the first study of AOP in LMs. We put together metrics and data, and conduct a series of experiments that explore AOP in LMs, its relation to properties of training data and the extent to which LMs learn more abstract principles behind AOP, generalising to unseen items. The contributions of our paper are the following:

- We introduce a reusable Corpus of Adjective Pairs (CAP) and a series of metrics to measure AOP in LMs (§4.3).

- We show that LMs are impressively good predictors of naturally occurring adjective orders, with AOP prediction accuracy as high as 94.1%. We identify three stages of AOP acquisition in LMs that lead to their performance at the end of the training (§4.4).

- We demonstrate that linguistic and cognitive predictors only partly account for LM performance in AOP prediction. We identify the gap between linguistic theory and LM performance in the AOP domain that needs to be explained (§4.5).

- To pinpoint the role of training data co-occurrence statistics in LM AOP predictions, we study *the Pile* (Gao et al., 2021) – a large-scale corpus that Pythia models were trained on. Simple bigram statistics turn out to be a powerful AOP predictor for naturally occurring adjective orders on their own, at 90.3% accuracy (§4.6). Figure 4.1 summarises a close relation between count statistics and LM predictions, questioning whether LM predictions are significantly driven by anything beyond *n*-gram counts.

- At the same time, we find two ways in which LMs go beyond simple collocational statistics: 1) We test LM predictions for adjectival combinations not seen *during* training (a challenging setup due to the large size of the Pile; we overcome this obstacle by turning to intermediate checkpoints and corresponding portions of training data). We find some—but limited—generalisation to new data (§4.6); 2) We study the effect of context on AOP. We conclude that the sentence helps LMs make better AOP predictions compared to noun phrase with adjectives taken in isolation or in a set of random contexts, which indicates that AOP mechanisms in LMs go beyond *n*-gram counts in training data (§4.7).

## 4.2 Background

### 4.2.1 Adjective Order Theory

The mechanisms that drive adjective order have been a topic of research for decades (see Dyer et al. 2023 for overview). Theories have been proposed as far back as the 19th century (Sweet, 1898), and refinements to the theory are being made to this day. Early theories proposed lexical hierarchies that described adjective order based on abstract classes such as *dimension* or *colour* (Dixon, 1982; Sproat and Shih, 1991; Cinque, 1996; Scott, 2002). Adjectives were expected to be ordered based on the class they belong to: e.g. dimension goes in front of colour, which gives rise to the order of a phrase like *the big red house*.

While these hierarchies are powerful predictors of adjective order, they are not really explanatory: the question remains how such hierarchies are formed by the underlying properties of adjectives. Furthermore, corpus studies have shown that many counterexamples against various versions of these hierarchies can be found (Truswell, 2009), demonstrating the need for a more fine-grained analysis (Trotzke and Wittenberg, 2019).

One line of work looks for semantic driving forces of AOP, placing emphasis on adjectival properties like *intersectivity* and *subsectivity* (Truswell, 2009) or identifying conditions under which adjective order constraints can be lifted, such as *focus* (Vendler, 1968; Teodorescu, 2006).

Other approaches take a broader perspective, leveraging insights from quantitative linguistics, distributional semantics and information theory, and connecting this to psycholinguistic theories of human language processing. Early analyses focus on properties such as *length*—shorter adjectives tend to precede longer ones (Behaghel, 1930)—and *frequency*: more frequent adjectives tend to appear earlier (Martin, 1969; Ney, 1981; Wulff, 2003). This follows from a broader theory of language production, according to which more accessible and familiar material is produced first (Bock, 1982; Ferreira and Dell, 2000) and is comprehended faster (Arnon and Snider, 2010). Collocational patterns of adjectives and nouns have been shown to be highly predictive of adjective order: adjectives most closely connected to the noun in term of co-occurrence appear closer to it (Sweet, 1898; Byrne, 1979; Lapata et al., 1999). This intuition has been operationalized through pointwise mutual information (*PMI*) (Hahn et al., 2018; Futrell et al., 2020a), which aligns with more general theories of word order arguing that distance between words that closely predict each other tends to be small (Gibson, 1998; Futrell et al., 2020b).

Predictors themselves can be complex, based on collocational or semantic factors or com-binations thereof: *modification strength* of adjectives, for instance, is a predictor based on adjectival compositional potential (Vecchi et al., 2013, 2017); other predictors proposed in the literature include *information gain* that an adjective provides for a noun (Dyer et al., 2021), and *subjectivity* of adjectives, with more subjective adjectives occurring first (Hill, 2012; Scontras et al., 2017, 2019; Franke et al., 2019). Although the majority of these approaches focus on English, various cross-lingual accounts have demonstrated their universal nature: adjective order tends to be driven by the same principles across languages (Wulff and Gries, 2015; Leung et al., 2020; Scontras, 2023; Dyer et al., 2023).

Finally, various papers highlight the need for a multi-factorial account of adjective order: it is unlikely adjective order is driven by a single factor and therefore it's the result of various competing pressures (Wulff, 2003; Trotzke and Wittenberg, 2019; Dyer et al., 2023). Relatedly, a recent position paper (Levshina et al., 2023) argues for a fundamentally gradient theory of word order that would be able to capture its hybrid and multi-faceted nature.

In our work, we compare LM predictions to various adjective order theories. Additionally, we examine the utility of LMs for investigating the role that *context* plays on word order: since almost all theories focus on adjectives in isolation, LMs may provide a fruitful starting point for shaping a contextual account of adjective order.

### 4.2.2 Word Order in Language Models

With the increasing capabilities of language models, their linguistic abilities have been under much scrutiny in recent years. A wide range of procedures have been introduced to uncover their understanding of grammaticality and notions of linguistic structure (Linzen and Baroni, 2021), as outlined in §1.3. To our knowledge, no prior work has focused on adjective order in particular: the closest related phenomenon to this would be Misra and Mahowald (2024)'s analysis of adjective-numeral constructions in LMs.

The importance of word order in LMs has been a topic of debate, with various works claiming that downstream performance is not affected by scrambled inputs (Malkin et al., 2021; Sinha et al., 2021a), although it has been shown that LMs are able to retain a notion of word order through their positional embeddings (Abdou et al., 2022). It has been argued that LMs acquire an abstract notion of word order that goes beyond mere *n*-gram co-occurrence statistics (Futrell and Levy, 2019; Kuribayashi et al., 2020; Merrill et al., 2024), a claim that we in this paper assess for large-scale LMs in the context of adjective order. Finally, numerous works have investigated the trade-off between memorisation and generalisation in LMs: it has been shown that larger LMs are able to memorise entire passages from the training data (Biderman et al., 2023a; Lesci et al., 2024; Prashanth et al., 2024), but generalisation patterns for grammatical phenomena have also been shown to follow human-like generalisation (Dankers et al., 2021; Hupkes et al., 2023; Alhama et al., 2023).

## 4.3 Methods

### 4.3.1 Measuring Word Order Preference

We measure a language model's AOP by comparing its log probability of the *natural* order (the order in which a pair of adjectives appears in the sentence from our corpus) with a *swapped* order. In our experiments we measure AOP in two settings, either by considering the phrase in *isolation*[1], or by taking the *context* of the phrase into account. For a double adjective phrase $\textsc{a1a2n}$ that is extracted from a corpus with context $\textsc{c}$, we compute a model's preferred order as follows:

$$\text{AOP-}\Delta(\textsc{a1a2n}) = P(\textsc{a1a2n}) - P(\textsc{a2a1n}) \quad [4.1]$$

$$\text{AOP-}\Delta(\textsc{a1a2n} \mid \textsc{c}) = P(\textsc{a1a2n} \mid \textsc{c}) - P(\textsc{a2a1n} \mid \textsc{c}) \quad [4.2]$$

All probabilities are in log space. The AOP-Δ metric represents the *magnitude* of a models' preference. In our experiments we also consider an accuracy metric based on AOP-Δ, expressed as the number of items for which AOP-Δ is positive:

$$\text{AOP-\%}(\mathcal{C}) = \frac{\{\varphi \in \mathcal{C} \mid \text{AOP-}\Delta(\varphi) > 0\}}{|\mathcal{C}|} \quad [4.3]$$

Where $\mathcal{C}$ is an evaluation corpus of double adjective phrases, as described in the following section. Misra and Mahowald (2024) consider a similar metric for measuring properties of word order preferences in LMs, based on corrupted word orders.

### 4.3.2 CAP🧢: Corpus of Adjective Pairs

To evaluate adjective order preferences across a wide range of adjective pairs and contexts, we extract a novel set of double adjective phrases from a corpus. Earlier work has extracted adjective pairs from the Universal Dependencies treebank (Dyer et al., 2023), but in order to collect a larger range of adjective pairs we collect novel constructions ourselves. We release this evaluation corpus under the name of CAP🧢: Corpus of Adjective Pairs.

[1] We compute these isolated probabilities by prepending the noun phrase with '*The*'.

**Procedure** As a starting point, we use the BabyLM 10M corpus (Warstadt et al., 2023), which consists of a mix of various English language sources including Wikipedia articles, books, and transcribed dialogue data. From this corpus we make a pre-selection of sentences containing multiple adjectives by filtering all sentences containing at least two adjectives, based on a list of 11,000 adjectives released by Futrell et al. (2020a). Next, we generate dependency parse trees for each sentence using spaCy and select all sentences in which a NOUN contains two AMOD relations to an ADJ that is part of the adjective list. This procedure results in 2678 double adjective sentences. We provide a sample of CAP sentences in Appendix A.2.1.

From these sentences we also generated corrupted sentences in which the adjective order is swapped. In case the first adjective is preceded by an indefinite article we ensure the article is changed accordingly to satisfy morpho-phonological constraints. Here is an example of an extracted phrase and its swapped counterpart:

+ *Peter received **a** full athletic scholarship [..]*

− *Peter received **an** athletic full scholarship [..]*

### 4.3.3 Models

For our experiments we make use of the Pythia suite of language models (Biderman et al., 2023b). The models are released in increasing sizes (from 70M to 12B parameters) and all trained on The Pile corpus (Gao et al., 2021). Importantly, all intermediate checkpoints are available for these models in logarithmic intervals, which allows us to investigate the learning dynamics at a fine-grained level.

## 4.4 Experiment 1: AOP in LLMs

In this first set of experiments we evaluate a series of pretrained LLMs on CAP using the AOP-Δ and AOP-% metrics of §4.3.1.

**Model Size** For the eight Pythia models of increasing size we plot the AOP-% performance in Figure 4.2A, and the AOP-Δ performance in Figure 4.2B. It can be seen that performance improves with model size: contextual AOP-% performance increases from 89.6% for the 70m model to 94.1% for the 12b model. AOP-% does not increase consistently with size, however, from the 1b model onward it reaches a plateau. Interestingly, AOP-Δ *does* increase consistently with size: this shows that larger models become more *certain* about their AOP judgement.

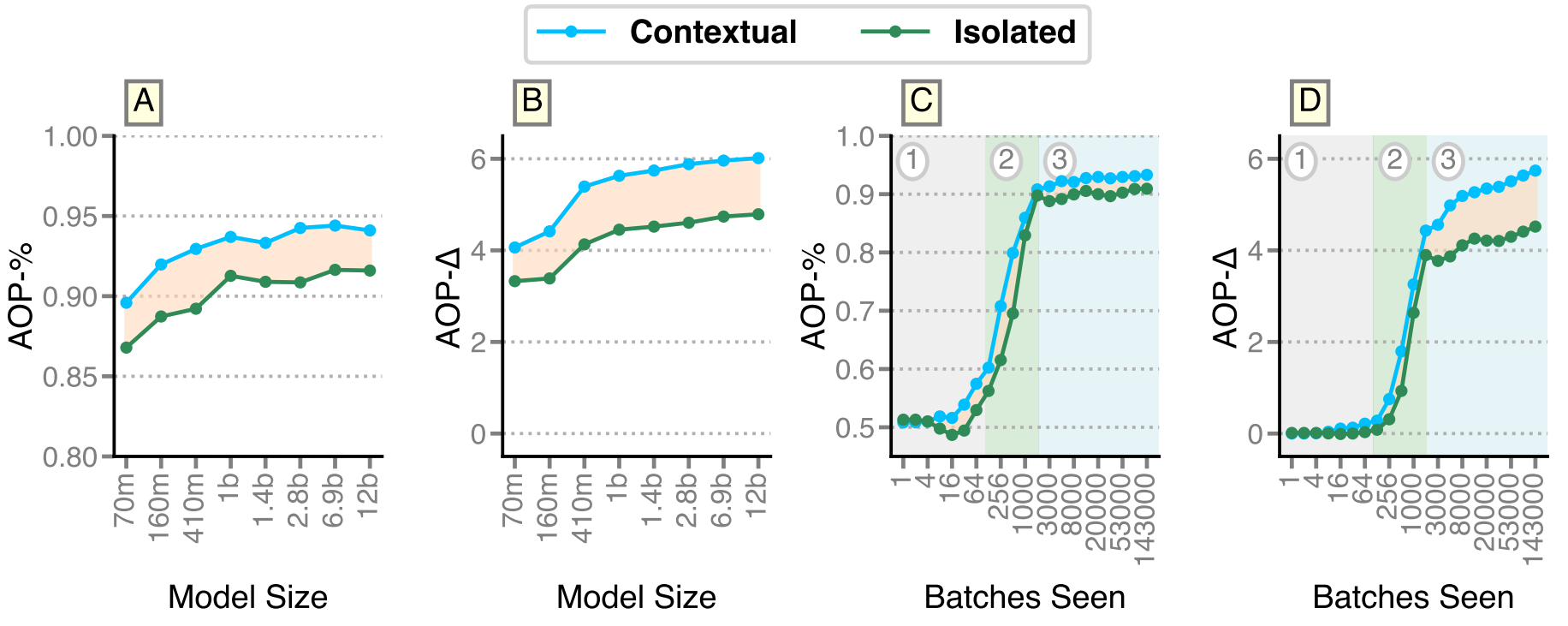

**Figure 4.2:** A–B: AOP-% and AOP-Δ scores for Pythia models of increasing size. C–D: AOP-% and AOP-Δ scores for Pythia-1.4b during training. We highlight the three learning phases: 1) initialisation, 2) acquisition, and 3) consolidation.

**Learning Dynamics** To investigate how AOP develops during training, we compute scores for the 1.4b model on all its intermediate checkpoints in Figure 4.2C and 4.2D. From these plots we can identify three distinct learning phases: 1) an *initialisation* phase in which AOP judgements have not been formed yet, 2) an *acquisition* phase in which AOP judgements are rapidly being formed, and 3) a *consolidation* phase in which AOP judgements remain stable and only the role of context is being reinforced. A similar three-phase learning dynamic has been observed by Van Der Wal et al. (2022) and Chen et al. (2023). Note that the consolidation phase is reached relatively quickly: after around 2000 batches, which is ~1.4% of the total model training. This demonstrates that adjective order is a relatively low-level linguistic feature that is acquired early during training. Furthermore, it highlights the importance of investigating learning dynamics at a logarithmic scale: with linear checkpoints we would not obtain this fine-grained level of precision.

**Localising AOP** In our experiments so far we have measured AOP based on the probabilities of the full adjective-noun triplet. Preference order can arise at different locations of the phrase. The most natural location for this is the probability of the second adjective after having seen the first adjective, but context could also play a role for preferring which adjective comes at the first position, and a noun could be more probable after having seen two adjectives in their natural order.

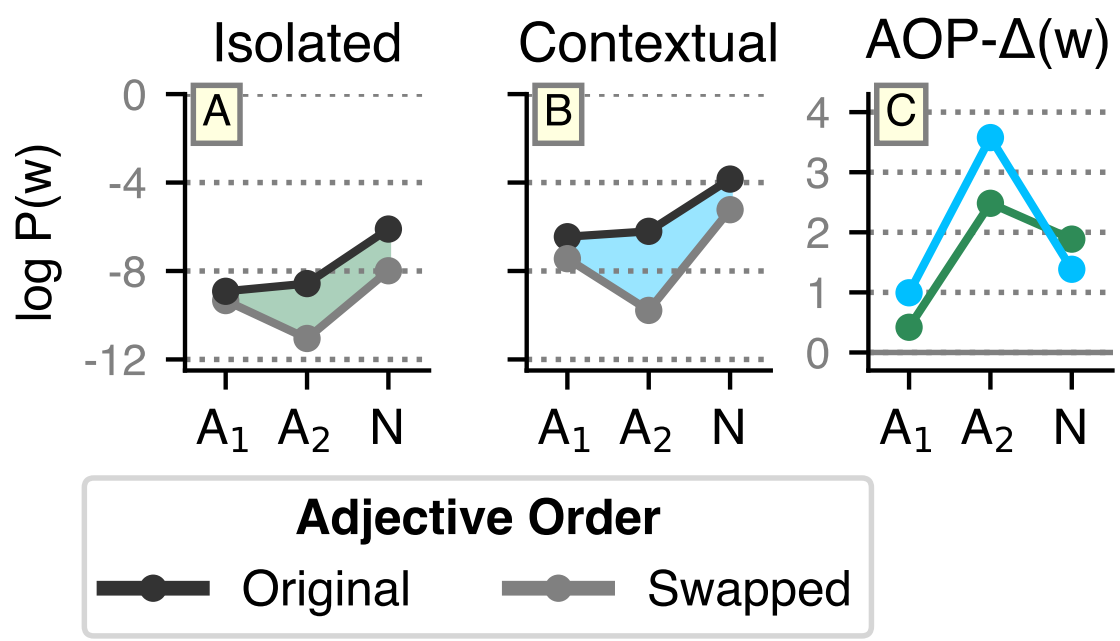

**Figure 4.3:** Average token probabilities for the original and swapped adjective orders on Pythia-12b, without and with sentence context (A–B), as well as the token-level differences (C) that correspond to the difference between the curves in (A) and (B).

In Figure 4.3 we plot the average token probabilities on CAP for Pythia-12b, in isolation and contextual. We observe that, as expected, the largest probability difference occurs at the position of the second adjective: the model is highly surprised by encountering the second swapped adjective after having seen the first. The noun probability is strongly affected by swapping adjective order as well: a noun is more likely to follow adjectives in natural order. Furthermore, we can see that context plays a role for the probability of the first adjective already: whereas in the isolated case both adjectives in first position receive equal probability, order preference already manifests here when the model has access to context.

**Conclusion** The Pythia LMs acquire a strong sense of adjective order that is learned early in training at around 1.5% of the total amount of training. We showed that sentence context has a positive effect on AOP, bumping up AOP-% from 91% to around 94%. By localising the AOP scores on the token-level we show how these order preferences manifest at different points in a phrase. In the next experiments we will investigate these results in more detail, connecting LM AOP to cognitive predictors and training data statistics.

## 4.5 Experiment 2: Cognitive Predictors

As discussed in §4.2.1, a range of cognitively plausible predictors have been proposed to explain adjective order. In the previous section we have shown that LMs are highly capable of predicting adjective order, and in this section we investigate how their predictions are related to various theories on adjective order. We evaluate the performance of three cognitive predictors—Length, PMI, and Subjectivity— and correlate them against AOP-Δ

| Metric | AOP-% | $\rho$(AOP-$\Delta(\bullet)$) | $\rho$(AOP-$\Delta(\bullet\|c)$) |
|---|---|---|---|
| AOP-$\Delta(\bullet)$ | 91.6 | – | 0.71 |
| AOP-$\Delta(\bullet\|c)$ | **94.1** | 0.71 | – |
| $\|$A$\|$ | 49.6 | 0.16 | 0.10 |
| PMI(A;N) | 62.4 | 0.27 | 0.20 |
| Subj(A) | **69.3** | 0.07 | 0.08 |
| %(A1) | 60.5 | 0.21 | 0.18 |
| %(A1A2) | **90.3** | 0.58 | 0.53 |
| %(A1A2N) | 87.6 | 0.54 | 0.53 |

**Table 4.1:** Predictive accuracy of adjective order on CAP for various metrics: AOP-Δ for Pythia-12b, 3 cognitive predictors (§4.5), and relative n-gram counts from The Pile (§4.6). The highest accuracy for each group is **bolded**. For each metric we provide the Spearman correlations with respect to the AOP-Δ scores of Pythia-12b.

scores across training.

### 4.5.1 Cognitive Predictors

For each predictor we compute its score for a particular adjective pair. An adjective order predictor $f$ returns a graded score for adjective order, akin to our formulation of AOP-Δ.

**Length** For Length—which hypothesizes that the first adjective is the shortest—the score for adjectives A1 and A2 is expressed as

$$f(\text{A}_1, \text{A}_2) = |\text{A}_2| - |\text{A}_1|$$

We also considered expressing length as the number of tokenizer subwords, since this is the only notion of length LMs have access to, but found that this led to similar results as string length.

**PMI** We follow the procedure of Futrell et al. (2020a) for computing PMI, which is based on the distribution over adjective relations in dependency parse trees. It is defined as

$$\text{PMI}(\text{A}; \text{N}) = P(\text{A}, \text{N})/(P(\text{A}) \cdot P(\text{N}))$$

$P(\text{A}, \text{N})$ is computed as the normalised number of occurrences of adjective A and noun N in `amod` relation. $P(\text{A})$ and $P(\text{N})$ are obtained by marginalising over this joint distribution.

PMI hypothesises that adjectives closer to the noun have a higher PMI:

$$f(\text{A}_1, \text{A}_2, \text{N}) = \text{PMI}(\text{A}_2; \text{N}) - \text{PMI}(\text{A}_1; \text{N})$$

**Subjectivity** We leverage the 450 subjectivity ratings provided by Dyer et al. (2023), to compute subjectivity scores for the subset of adjective pairs that contain a rating. This procedure retains 37.5% of the adjective pairs in CAP (1005 items). Subjectivity as predictor hypothesises that more subjective adjectives occur first (Scontras et al., 2017):

$$f(\text{A}_1, \text{A}_2) = \text{Subj}(\text{A}_1) - \text{Subj}(\text{A}_2)$$

### 4.5.2 Predictive Accuracy

We present the predictive accuracy of each predictor in Table 4.1. Subjectivity is the strongest predictor with 69.3% accuracy, followed by PMI (62.4%) and Length (49.6%). While Subjectivity reaches the highest accuracy, its correlation with the LM's AOP-Δ scores is surprisingly the lowest of the three. We plot these correlations over time in Appendix A.2.2: all three correlations develop in the acquisition phase and remain stable in the consolidation phase.

We also compute the overlap between these three predictors on the 1005 items for which we have subjectivity scores:

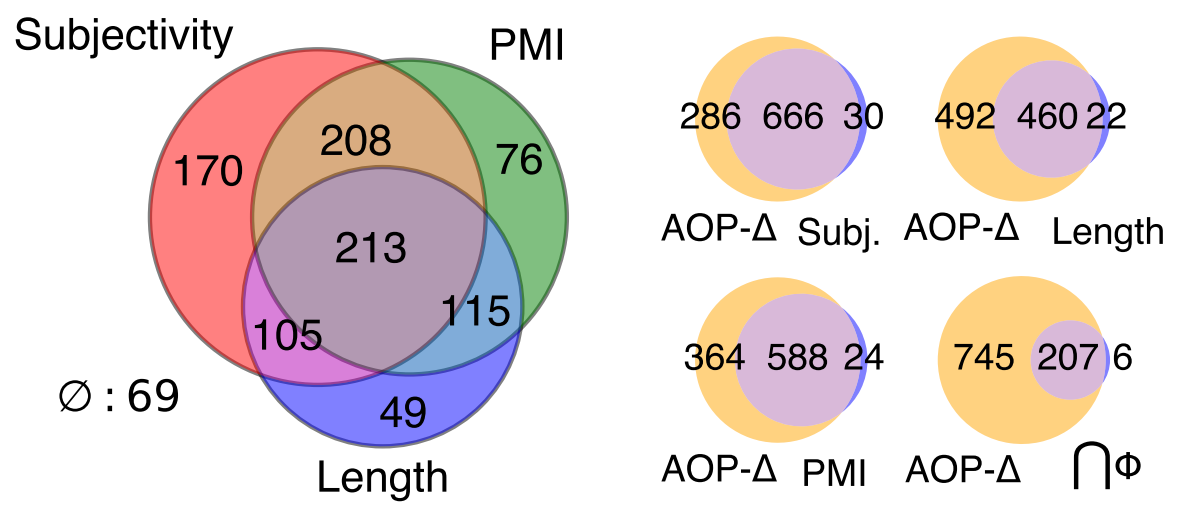

The first Venn diagram makes clear that the cognitive predictors have low agreement: only 213 orders are predicted correctly by all three. The broad coverage (93.1%) of the union, however, shows that most of the CAP items are explained by at least one of these predictors. In the four diagrams on the right we show the overlap of the predictors with the AOP-Δ scores of Pythia-12b. All three predictors and their intersection are almost fully subsumed by the AOP-Δ predictions. This demonstrates that AOP-Δ's unique contribution goes significantly above that of the ensemble of the cognitive factors, raising the ques-tion whether LMs 'hide' a meaningful linguistic theory that extends beyond established factors.

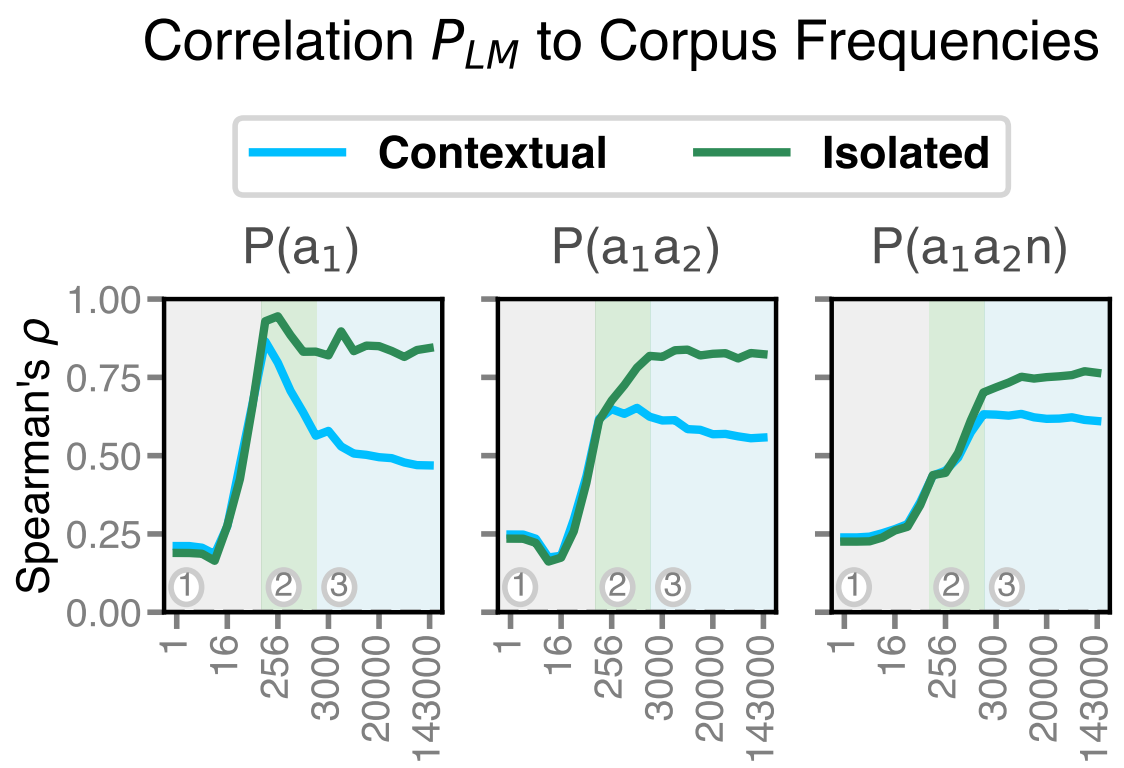

**Figure 4.4:** Correlations during training of LM probabilities for single adjectives, adjective pairs, and adjective-noun triplets with respect to their frequency in The Pile.

## 4.6 Experiment 3: Is AOP Driven by Data Statistics?

In order to truly understand what LMs base their adjective order preferences on, we need to get a better insight into how double adjective constructions are distributed in the training data. The way that contemporary LMs are trained nowadays makes this challenging: many models are released without open access to their training data, and the scale of training corpora that *are* available poses a challenge for targeted analysis.

### 4.6.1 WIMBD

To overcome this, we make use of the WIMBD API (*What's In My Big Data*, Elazar et al., 2024), which provides a fully indexed search engine over various large-scale corpora such as The Pile, on which the Pythia models have been trained. WIMBD makes it possible to retrieve the exact $n$-gram statistics that LMs have been trained on, and as such provides a detailed insight into the collocational factors that play a direct role in shaping AOP.

We retrieve the uni-, bi- and trigram counts for all the adjective-noun triplets in CAP for The Pile, as well as for the swapped orders. To provide an analogue to AOP-Δ, we express the n-gram counts as log count differences. We express these relative n-gram counts as

$$\%(\text{A}_1\text{A}_2\text{N}) = \log \#(\text{A}_1\text{A}_2\text{N}) - \log \#(\text{A}_2\text{A}_1\text{N})$$

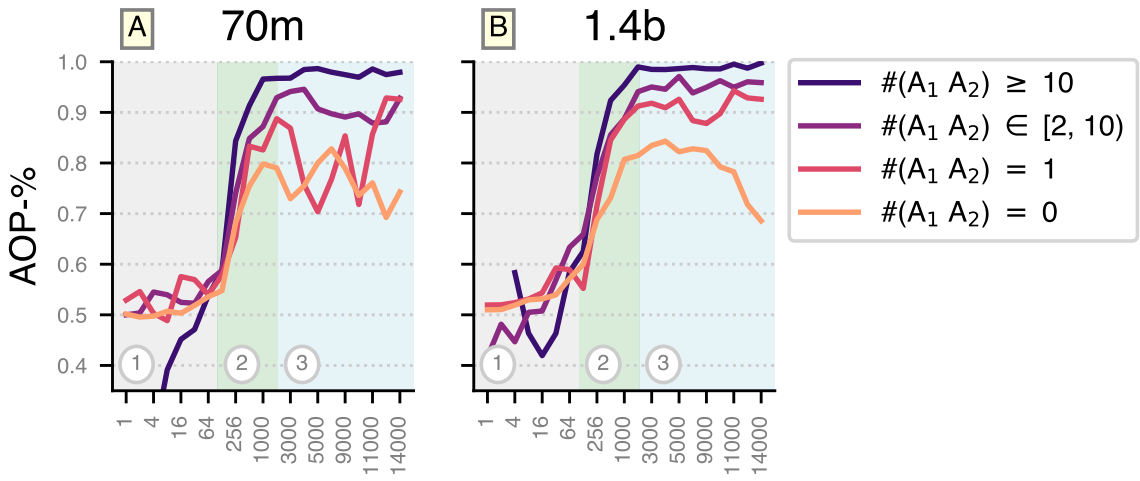

**Figure 4.5:** The contextual AOP-% performance for Pythia-70m and 1.4b across training, split out for items that have been seen 0, 1, 2 to 10, and more than 10 times at each specific checkpoint.

with similar formulations for adjective bigrams (%($A_1A_2$)) and unigrams (%($A_1$)).

**Predictive Accuracy** We present the predictive accuracy of these relative n-gram occurrences in Table 4.1. %($A_1$) predicts adjective order with 60.5%, which provides weak evidence that more frequent adjectives tend to occur first (Martin, 1969; Trotzke and Wittenberg, 2019). %($A_1A_2$) is the strongest predictor of adjective order—except for the LMs—at 90.3%. This shows that a large majority of the items in CAP can be resolved based on bigram statistics alone, which is leveraged by the LMs during training. However, n-gram collocations alone do not fully explain LM performance: a large amount of trigrams are not part of the Pile and performance for %($A_1A_2N$) drops to 87.6%.

**Correlations** To investigate how n-gram statistics play a role in the formation of AOP-Δ during training, we compute the Spearman correlation between various log-transformed n-gram counts and a LM's AOP-Δ on intermediate checkpoints. We plot this in Figure 4.4, for contextual and isolated AOP-Δ. The trends we observe here align closely with the three learning phases that we outlined in §4.4. The initialisation phase (1) is marked by a rapid acquisition of unigram and bigram statistics, that remain unaffected by context at this point. Then, at the start of the acquisition phase (2), in which AOP performance rapidly improves, the model starts to incorporate linguistic context in its uni- and bigram predictions. Finally, in the consolidation phase (3), trigram predictions start being affected by context as well, whereas unigram and bigram predictions remain relatively stable. The fully trained model strongly reflects these correlations to data statistics, which explains the strong relation we observed in Figure 4.1 between bigram frequencies and AOP-Δ.

### 4.6.2 Generalising to Unseen Adjective Orders

4

Having access to the n-gram distribution a model was trained on allows us to investigate its performance on unseen adjective orders: is the model able to generalise from an abstract signal of adjective order to novel orders? However, due to the enormous size of The Pile (300B tokens) it turns out that number of unseen adjective pairs in CAP is too small to draw significant conclusions from (only 4 out 2678 pairs). This leaves us with two options: either we collect an additional sample of rare adjective pairs that are not part of the Pile, or we investigate the model at intermediate checkpoints where a greater amount of adjective pairs has not been encountered yet. An issue with collecting highly rare pairs is that their canonical order will be much harder to determine, for which we would require extensive human judgements on the most natural order. We therefore go for the second option, and set up a procedure to measure the data statistics during training.

**Intermediate *n*-gram Counts** Since WIMBD only provides *n*-gram statistics for The Pile as a whole, we need to collect the intermediate statistics ourselves. Using the Pythia batch viewer (Biderman et al., 2023b) this becomes possible: our implementation is able to process the first 10% of The Pile (up to batch 14,000; 29.4B tokens) in around 3 hours. We collect the bi-gram counts of all adjective pairs in CAP on the batch level, allowing us to determine exactly when and how often a specific adjective pair has been seen by the model.

**Unseen Adjective Pairs** Based on these intermediate *n*-gram counts, we compute the AOP-% scores for various splits based on the number of times an adjective pair has been encountered. We focus on the cases where the swapped order has *never* been encountered: this allows us to isolate the moment a model encounters an adjective combination for the first time, and measure its 'zero-shot' performance on cases that are not already skewed towards to swapped order. For each model checkpoint we collect the *n*-gram counts at that particular point in training, and split the CAP items based on the number of times an A1A2 bi-gram has been seen. We provide the size of these splits across training in Figure 4.6.

We present the results for Pythia-70m and 1.4b in Figure 4.5. The performance on unseen pairs is remarkably high, reaching around 85% at the highest point for both models. Interestingly, the gap between unseen pairs and pairs seen once is far greater for the 1.4b model than the small 70m model. This shows that this larger model is able to acquire adjective order preference faster: a single exposition to an order immediately leads to an increase in AOP-Δ. We provide a more detailed item-level analysis for these items in Appendix A.2.3: while it is clear on the split level that a single occurrence improves AOP-%, we do not observe such a clear effect on the item level. Interestingly, as the number of unseen adjective pairs decreases, the performance on unseen pairs decreases as well. This shows that the predicting the order for the set of unseen pairs around batch 2,000 (6% of CAP) could be better inferred from a general signal of adjective order than the set at batch 14,000 (1.3% of CAP).

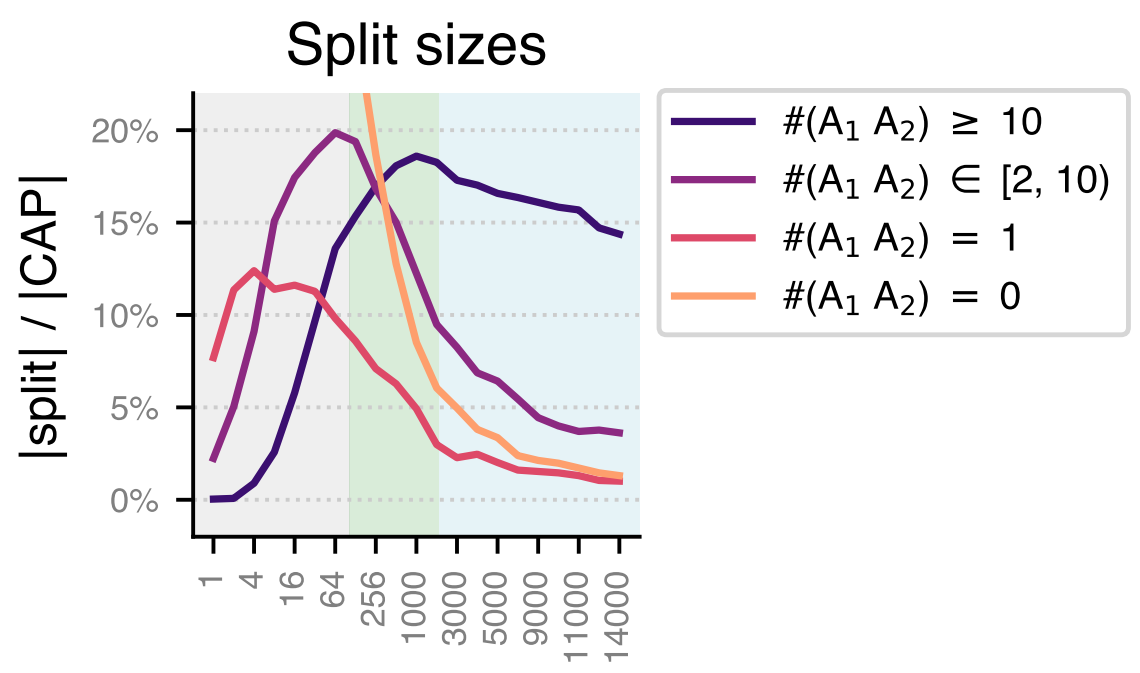

**Figure 4.6:** Relative sizes of the 4 splits of CAP based on the number of bigram occurrences seen at each point in training. Note that an additional constraint on each split is that the swapped adjective order has not been seen at all, which explains why the 4 splits do not sum up to 1.

**Conclusions** The question we set out to answer in this set of experiments is whether AOPs are driven by data statistics. We approached this question both in relation to the *n*-gram distribution of the Pile and to the shifting *n*-gram distribution across training. On the one hand, AOP is strongly driven by data statistics: we find a strong correlation between AOP-Δ and count-based AOP scores, which is also visualised in Figure 4.1. However, the relatively strong performance on the set of unseen adjective pairs during training indicates that LMs *are* able to infer order from a more general signal for adjective pairs it has not encountered before.

## 4.7 Experiment 4: The Role of Context

In §4.4 we established that LMs have a thorough grasp of adjective order that manifests itself early during training. We showed that having access to sentence context boosts performance, and we now investigate the role of context in more detail.

**Contextual Improvements** To gain a better insight how much items are impacted by context, we investigate AOP-Δ scores on an item-level with and without context. In Fig-ure 4.7A, we plot the AOP-Δ scores in isolation against the contextual scores, and mark the four quadrants of items that result in a positive or negative score after adding context. In the majority of cases (89.4%), AOP-Δ is positive both with and without context: the model is able to determine adjective order based on the adjective-noun triplet alone. In 4.8% of the cases, however, adding context flips a negative AOP-Δ score to a positive one. Adding context has a boosting effect in general: in 65.1% of the cases, AOP-Δ is increased by adding context. We provide an overview of how these ratios are affected by model size in Appendix A.2.4: the number of cases where context improves AOP-% is actually *larger* for small models, but so is the number of cases where context worsens it.

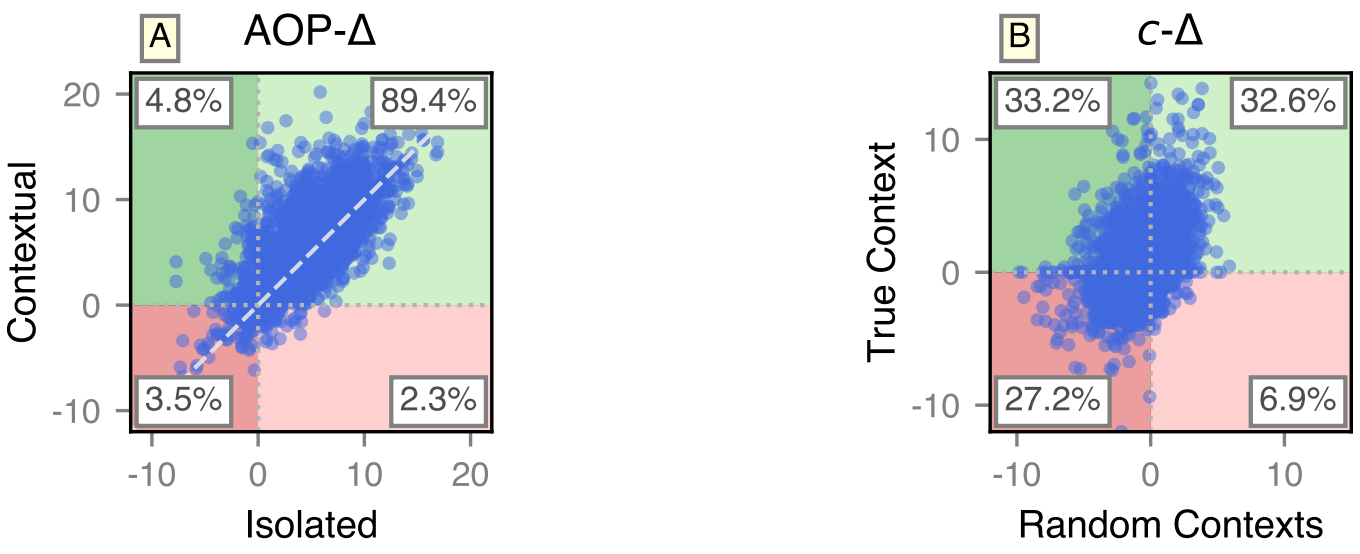

**Figure 4.7:** A: Item-level AOP-Δ scores with and without sentence-context for Pythia-12b. We provide a sample of sentences from each quadrant in Appendix A.2.1. B: The relative impact ($c$-Δ, Eq. 4.5) of an item's true context compared to an expectation over random contexts (Eq. 4.4).

**Random Contexts** The fact that context improves AOP-Δ raises the question what information in the context is responsible for this improvement. Does context contain relevant semantic information that plays a meaningful role for determining adjective order, or is any linguistic signal sufficient for this? To test this, we compute an expectation over AOP-Δ scores with *random contexts*, in which an adjective pair is preceded by a randomly sampled context from another adjective pair in CAP:

$$\text{AOP-}\Delta(\text{A}_1\text{A}_2\text{N} \mid \mathcal{C}) = \mathbb{E}_{c\sim\mathcal{C}}\left[\text{AOP-}\Delta(\text{A}_1\text{A}_2\text{N} \mid c)\right] \qquad [4.4]$$

We compute the expectation over a random sample of 500 contexts, taken from CAP ($\mathcal{C}$). To express the *relative* impact of adding context, we regress out a model's isolated AOP-Δ score by subtracting it from the contextual scores, which we call $c$-Δ:

$$c\text{-}\Delta(\bullet|c) = \text{AOP-}\Delta(\bullet|c) - \text{AOP-}\Delta(\bullet) \qquad [4.5]$$

By computing this score for both the original context and for the expectation over random contexts we can quantify how much more important the original context is for AOP-Δ.

We plot the results for this experiment in Figure 4.7B. The true context improves isolated AOP-Δ in 65.1% of the cases, whereas random contexts lead to an improvement for only 39.6%. Furthermore, for 33.2% the true context leads to an improvement whereas random contexts don't: this subset can serve as a valuable starting point for investigating what properties of the context are important for adjective order.

**Conclusions** AOP in Pythia is **context-sensitive**. Providing the model with the sentence where the adjective combination appears increases the model's preference towards the target adjective order – both compared to the noun phrase in isolation and to the noun phrase in a random context. This result points in the direction of a more linguistically interesting mechanism for AOP than reliance on *n*-gram counts in training data alone. In particular, it shows a non-trivial interaction between the noun phrase and bigger context that turns out relevant for adjective placement. Establishing this opens up a possibility of studying the forms and types of this contextual interaction – a potential novel contribution to linguistic theories of adjective order.

We leave a systematic discovery of types of AOP context-dependence open for future work, and focus on a small-scale qualitative study for now. An inspection of context-sensitive cases from CAP reveals at least two ways in which context matters for AOP:

**1. Structural parallelism** The 'canonical' word order can be overridden when context contains a parallel noun phrase with adjectives in non-canonical order (the example shows contextual AOP according to Pythia-1.4b):

a) *The* ***green big cars*** AOP-Δ 4.2

b) *The story is not about* small red houses, *it's about* ***big green cars*** AOP-Δ 6.63

**2. Property saliency** The property salient in the context brings the adjective describing this property to the left edge of the noun phrase. If the sentence is about colour, colour adjective will come first; if the sentence is about size, it will bring size adjectives to the left (again, orders preferred by Pythia-1.4b):

a) *They asked my favourite colour of houses and I said I liked* ***red small houses*** AOP-Δ 0.32

b) *They asked my favourite size of houses and I said I liked* ***small red houses*** AOP-Δ 1.61

These observations, of course, barely scratch the surface of possible contextual AOP interactions, but exemplify the power of LMs in incorporating contextual information into word order judgments.

## 4.8 Discussion

**Abstract Notions of Adjective Order** In this paper, we looked for AOP generalisation in LMs. We were interested in whether abstract driving forces are at play in LM predictions beyond simple *n*-gram counts in training data. From our experiments we conclude that LMs show limited AOP generalisation. Here, we want to highlight an important question that is rarely brought up in linguistic generalisation discussions in NLP: How general and abstract is human linguistic behaviour?

Linguistic literature suggests that human language production, including word order preferences, is guided by a mix of memorisation and abstract principles (O'Donnell, 2015). Speakers tend to recycle more routinised and therefore more accessible combinations (MacDonald, 2013), guided by sequence frequencies in linguistic input that, in turn, form sequential associations that lead to production automatising (Bybee, 2010; Diessel, 2019). This is compatible with a picture of language acquisition where knowledge of grammar is formed based on encountered exemplars of usage, which are stored in memory (Bybee, 2010) and serve as basis of generalisation (Goldberg, 2006), but can also serve as patterns for routinised language production. So, human linguistic knowledge and language production is not totally abstract and is not based exclusively on generalisation. This has been discussed across linguistic phenomena and has been suggested for word order (MacDonald, 2013; Levshina et al., 2023), although not for adjective order specifically.

When investigating generalisation in LMs it is therefore important to make a distinction between the degree to which they create generalisations for a particular phenomenon, and how similar this generalisation is to human behaviour. In principle, one could even find cases of LM *overgeneralisation*, if we are interested in how human-like LMs are linguistically. Importantly, in order to answer the question of human-like generalisation, we would need data on the trade-off in memorisation and generalisation in human linguistic behaviour. Such data does not exist yet for adjective order preferences.

**LMs and Linguistic Theory** Despite the impressive performance of LMs in recent years, their impact on theoretical linguistics has remained minimal (Baroni, 2022). There are various reasons for this. On the one hand, the cognitive plausibility of current LMs remains questionable, both in terms of the scale at which they are trained (Huebner et al., 2021; Warstadt et al., 2023) and their inductive biases that depend strongly on model architecture and are not reflective of human language processing (Wilson and Frank, 2023; Oh et al., 2024). On the other hand, their black box nature makes it highly challenging to extract linguistic patterns that can be tested against existing linguistic theories. While advancements in interpretability have made this latter point less problematic, the number of works explicitly connecting LM behaviour to human language processing remains few and limited to simple syntactic phenomena (e.g., Lakretz et al., 2021).

In this paper, we explore the possibility that LMs may 'hide' a theory of adjective order, by encoding features that are not covered by linguistic theories on adjective order. We show that LMs far exceed the predictive power of prior *cognitive predictors*, which raises the question whether this increase stems from meaningful interpretable features. While a lot of the LM predictions can be explained by collocational patterns—and those statistics alone do not *explain* the distribution of adjective orders—we *do* find some generalisation to unseen adjective combinations and a linguistically meaningful pattern in the way that context boosts AOP. We hope our work can serve as an inspiration for using LMs as tools for measuring context-dependence, which is hard to quantify with traditional quantitative methods.

**Corpus Interventions** Our analysis of AOP abstractness is constrained by the fixed training corpus the models are trained on. More detailed insights into this question can be obtained by an intervention-based approach in which we filter out specific constructions from the training data, an approach that we describe in more detail in Chapter 5. An interesting experiment in this direction would be to remove *all* sentences containing double adjectives from the training corpus, and measure whether the remaining distribution of adjectives contains a sufficient signal to acquire adjective order constraints.

*If It Is To Be Said, So It Be, So It Is.*

Gregory Hirsch

5

# 5 Filtered Corpus Training (FiCT) Shows that Language Models can Generalize from Indirect Evidence

The mechanisms underlying the acquisition of linguistic proficiency in language models remain largely unknown. In particular, the degree that language learning relies on memorisation versus generalisation remains a topic of investigation (Hupkes et al., 2023). The reliance of LMs on large amounts of training data raises the suspicion that they do not generalise in a 'human-like manner' (McCoy et al., 2019; Hu et al., 2020; Oh and Schuler, 2023b), but it is hard to address such questions with traditional evaluation metrics such as perplexity.

## 5.1 Introduction

This chapter introduces *Filtered Corpus Training* (FiCT) as a method for measuring the linguistic generalisation abilities of language models. As depicted in Figure 5.1, FiCT involves training models on corpora that have been filtered to remove specific linguistic constructions, thereby testing the models' ability to generalise beyond their training data. For example: we can train a model on a corpus that has never seen subjects modified by a prepositional phrase (e.g. "A sketch *of lights* { doesn't / *don't }…"), and then ask whether it can judge the grammaticality of such sentences. If a model has learned that verbs must agree with the head noun of the subject noun phrase (NP), and that NPs can be modified by PPs (e.g. from seeing these in object but not subject position), it should be capable of generalising to the unseen PP-modified subjects.

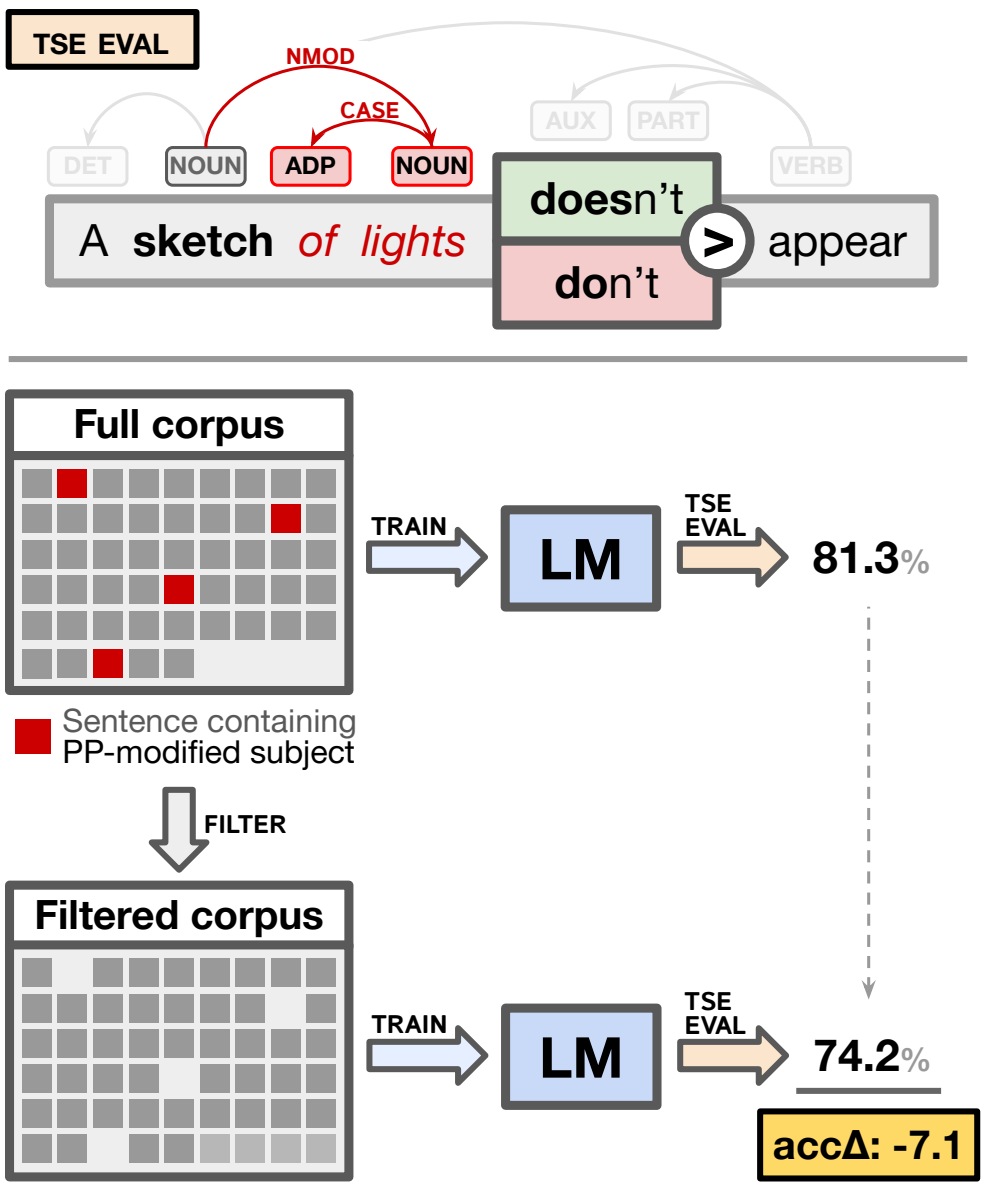

**Figure 5.1:** Overview of the **Fi**ltered **C**orpus **T**raining methodology (FICT). For a linguistic construction of interest (e.g. prepositionally modified subjects), we filter out sentences containing that construction and train a new language model on the filtered corpus. We measure performance on *targeted syntactic evaluations* to assess the capacity of the LM to generalise from related constructions to this novel, unseen construction.

This method enables us to ask whether models can form relevant linguistic generalisations from *indirect evidence*, or whether they require direct evidence (e.g. examples of constructions during training; Warstadt and Bowman, 2022; Mueller and Linzen, 2023). In essence, by *intervening* on patterns in the training data we obtain a more causal account of the relation between training data and model behaviour (Pearl, 2009). Furthermore, by carefully controlling for the number of parameters we can investigate the inductive biases of two major LM architectures, Transformers and LSTMs, which allows us to give more detailed answers about the recent successes of Transformer models on a fine-grained linguistic level.

We apply the FiCT methodology by developing filters targeting a wide range of the linguistic phenomena evaluated by BLiMP (section 5.3; Warstadt et al., 2020) and training both LSTM and Transformer LMs on the resulting corpora (section 5.4). Our results (section 5.5) show that while Transformers are uniformly better qua language models (as measured by perplexity), their linguistic generalisation abilities are not better than that of the LSTMs (as measured by a metric we introduce called *accuracy delta*), demonstrating a dissociation between perplexity and linguistic generalisation. Furthermore, for both models, the impact of filtered corpus training on grammaticality judgements is quite low, suggesting that language models are able to form sophisticated linguistic generalisations on the basis of only indirect evidence (as discussed in section 5.6). These results shed light on the debate between memorization and generalization in language models: by causally intervening on the training data, we ensure that models have never seen instances of their evaluation targets. That they can still make correct grammaticality judgments shows they generalize in subtle and linguistically-relevant ways that go beyond their training data.

## 5.2 Background

### 5.2.1 Surprisal Theory

Language modelling performance can be measured using *perplexity*, indicating a model's fit to a corpus distribution. Intuitively, one might expect that lower perplexity leads to more human-like linguistic behaviour. This connection has been explored in detail in the context of *surprisal theory* (Hale, 2001; Levy, 2008): encountering a highly surprising token results in a longer reading time. Initial findings indicate that lower perplexity, as measured by language models, leads to better reading time predictions (Fossum and Levy, 2012; Goodkind and Bicknell, 2018; Wilcox et al., 2020), although affected by model architecture (Hao et al., 2020), cross-lingual effects (Kuribayashi et al., 2021), and syntactic ambiguity (Arehalli et al., 2022). It has been shown, however, that lower perplexity only results in better predictive power up to around 2 billion training tokens (Oh and Schuler, 2023a): after this point LMs become too accurate at predicting low-frequency constructions and long-distance dependencies (Oh et al., 2024). The present paper also explores the connection between perplexity and human-like linguistic behaviour and will find a dissociation with perplexity.

### 5.2.2 Targeted Syntactic Evaluations

Perplexity should be augmented with other evaluations that specifically target the models' ability to generalise in a human-like way. Such investigations often draw on psycholinguistic paradigms, treating language models as participants in order to learn what such models "know" about specific linguistic phenomena (Futrell et al., 2019; Ettinger, 2020). A common paradigm in this body of literature, usually referred to as "targeted syntactic eval-uations" (Linzen et al., 2016; Jumelet and Hupkes, 2018; Marvin and Linzen, 2018; Gauthier et al., 2020; Newman et al., 2021) involves comparing language models' preferences between minimal pairs of sentences: a model is deemed to understand a phenomenon if it assigns a higher probability to the grammatical alternation.

The benchmark suites with the widest coverage over linguistic phenomena are SyntaxGym (Gauthier et al., 2020) and the Benchmark of Linguistic Minimal Pairs (BLiMP, Warstadt et al., 2020), the latter of which we will use in our experiments. BLiMP consists of 67 different benchmarks, each consisting of 1,000 minimal pairs, which target twelve different linguistic areas, broadly construed, across morphology, syntax, semantics, and the syntax-semantics interface. This is the benchmark we use as a primary means of evaluation in the present investigation, discussed in greater detail in paragraph 5.4.

### 5.2.3 Linguistic Generalisation

While targeted syntactic evaluations give an insight into a model's linguistic competence, it does not show *how* a model acquires this notion of grammaticality. In this chapter we focus on two kinds of linguistic generalisation. *Structural generalisation* (Hupkes et al., 2023) asks: can language models make grammaticality judgements in syntactically more complex constructions than seen during training? One line of work approaches this question from a fine-tuning perspective: by fine-tuning a model on a particular set of constructions we can measure the impact that this has on other linguistic constructions (Prasad et al., 2019; Weber et al., 2024). *Lexical generalisation* asks whether models can generalise a seen construction to new lexical items that it has not seen in that construction (Kim and Linzen, 2020).

In order to gain a *causal* perspective on how the training data influences model performance, we retrain models from scratch on *filtered* corpora. This methodology has been deployed in earlier work to investigate how LMs learn the licensing conditions of negative polarity items from different contexts, as described in Chapter 6 and by Weber et al. (2021). Warstadt (2022) investigates the *poverty of the stimulus* debate through the lens of filtered corpora, focusing on the phenomenon of subject auxiliary inversion. Finally, Misra and Mahowald (2024) investigate rare adjective-noun constructions and manipulate training corpora to investigate how models acquire an understanding of rare constructions. Whereas most of these focus on a particular linguistic construction, our work applies the approach to a wide range of phenomena.

## 5.3 Filtered Corpus Training (FiCT)

| Corpus name | BLiMP benchmark | Example | %BLiMP items targeted | %sentences filtered out | #Tokens as % of full |
|---|---|---|---|---|---|
| FULL | — | — | — | 0.00 | 100.0 |
| AGR-PP-MOD | distractor_agr_relational_noun | _A sketch of lights doesn't/*don't appear_ | 99.5 | 18.50 | 95.80 |
| AGR-REL-CL | distractor_agr_relative_clause | _Boys that aren't disturbing Natalie suffer/*suffers._ | 94.4 | 2.76 | 98.99 |
| AGR-RE-IRR-SV | irregular_plural_subject_verb_agr_1 | _This goose isn't/*weren't bothering Edward._ | 99.4 | 11.29 | 98.59 |
| | irregular_plural_subject_verb_agr_2 | _The woman/*women cleans every public park._ | 97.2 | | |
| | regular_plural_subject_verb_agr_1 | _Jeffrey hasn't/*haven't criticized Donald._ | 99.3 | | |
| | regular_plural_subject_verb_agr_2 | _The dress/*dresses crumples._ | 99.1 | | |
| NPI-ONLY | only_npi_licensor_present | _Only/*Even Bill would ever complain._ | 100 | 0.09 | 99.93 |
| | only_npi_scope | _Only those doctors who Karla respects ever ... / *Those doctors who only Karla respects ever ..._ | 100 | | |
| NPI-SENT-NEG | sentential_negation_npi_licensor_present | _Those banks had not/*really ever lied._ | 100 | 0.45 | 99.82 |
| | sentential_negation_npi_scope | _The turtles that are boring me could not ever ... / *The turtles that are not boring me could ever ..._ | 100 | | |
| NPI-SIM-QUES | matrix_question_npi_licensor_present | _Should I ever join? / *I should ever join._ | 100 | 0.01 | 99.98 |
| QUANTIFIER-SUPERLATIVE | superlative_quantifiers_1 | _No man has revealed more than/*at least 5 forks._ | 98.5 | 7.29 | 97.72 |
| | superlative_quantifiers_2 | _An/*No actor arrived at at most 6 lakes._ | 99.3 | | |
| QUANTIFIER-EXISTENTIAL-THERE | existential_there_quantifiers_1 | _There aren't many/*all lights darkening._ | 99.1 | 1.15 | 99.82 |
| BINDING-C-COMMAND | principle_A_c_command | _A lot of actresses that thought about Alice healed themselves/*herself._ | 96.6 | 0.01 | 100.0 |
| BINDING-CASE | principle_A_case_1 | _Tara thinks that she/*herself sounded like Wayne._ | 100 | 1.54 | 99.54 |
| | principle_A_case_2 | _Anna imagines herself praising/*praises this boy._ | 92.5 | | |
| BINDING-DOMAIN | principle_A_domain_1 | _Carlos said that Lori helped him/*himself._ | 100 | 0.44 | 99.84 |
| | principle_A_domain_2 | _Mark imagines Erin might admire herself/*himself._ | 99.3 | | |
| | principle_A_domain_3 | _Nancy could say every guy hides himself. / *Every guy could say Nancy hides himself._ | 99.5 | | |
| BINDING-RECONSTRUCTION | principle_A_reconstruction | _It's herself who Karen criticized / *criticized Karen._ | 99.1 | 0.01 | 99.99 |
| PASSIVE | passive_1 | _Jeffrey's sons are insulted/*smiled by Tina._ | 96.9 | 2.67 | 99.57 |
| | passive_2 | _Most cashiers are disliked/*flirted._ | 98.9 | | |
| DET-ADJ-NOUN | det_noun_agr_with_adj_1 | _Tracy praises those lucky guys/*guy._ | 95.6 | 1.14 | 99.78 |
| | det_noun_agr_with_adj_2 | _Some actors buy these/*this gray books._ | 93.0 | | |
| | det_noun_agr_with_adj_irregular_1 | _He shouldn't criticize this upset child/*children._ | 92.0 | | |
| | det_noun_agr_with_adj_irregular_2 | _That adult has brought that/*those purple octopus._ | 93.9 | | |
| DET-NOUN | det_noun_agr_1 | _Craig explored that grocery store/*stores._ | 99.7 | 0.47 | 99.95 |
| | det_noun_agr_2 | _Carl cures those/*that horses._ | 99.8 | | |
| | det_noun_agr_irregular_1 | _Phillip was lifting this mouse/*mice._ | 100 | | |
| | det_noun_agr_irregular_2 | _Those ladies walk through those/*that oases._ | 100 | | |

**Table 5.1:** An overview of all the filters, the BLiMP benchmark they target, an example for each benchmark, and number of items targeted by the filter. The rightmost column represents the relative number of tokens in each filtered corpus after they have been downsampled to the same number of lines.

This section first introduces the logic of the FiCT method before detailing the specific filters that we use in our experiments. The final experimental setup is described in section 5.4. Code and data, as well as a link to all models on the HuggingFace Hub, can be found at https://github.com/CLMBRs/corpus-filtering.

### 5.3.1 Logic of the Method

The core methodological basis of this paper is what we call *Filtered Corpus Training*, or FiCT. This involves comparing the performance of otherwise identical learners that are trained on data which differs in some interesting way.

In this paper, the FiCT methodology is primarily used to test whether LMs are capable of extrapolating linguistic rules learned from environments in training data to unseen environments. In order to ensure that the specified environments are not seen in the training data, we use *filters* to remove sentences with the specified environments from a naturalistic corpus. By comparing models trained on the ablated data and models trained on the full, naturalistic corpus, we can potentially determine whether, how, and when language models are able to make such generalisations.

Figure 5.1 illustrates the logic of our method. The sentence pair "A sketch of lights {doesn't / *don't} appear" contains a subject with a prepositional phrase (PP) modifying a noun, itself with a noun that differs in number from the main subject. We filter from the training corpus all sentences with subjects containing PP modifiers, and then compare the ability to make the correct grammaticality judgements on this pair between a model trained on the full corpus and this filtered corpus. This difference in performance we call acc$\Delta$ (formally defined in section 5.4). A model that has not seen PP-modified subjects could still make the correct judgements by forming the following generalisations: verbs agree with the head noun of the subject, and noun phrases with PP modifiers (which can be seen in object, but not subject position) are headed by the main noun. Low acc$\Delta$ would then provide evidence that the model has developed such generalisations.

The filters used in the present investigation are listed in Table 5.1, along with the BLiMP benchmark(s) each targets, and some descriptive summary statistics for each. These filters utilised part-of-speech, morphological features, and syntactic dependency annotations generated via the use of Stanza (Qi et al., 2020), an off-the-shelf package that uses pretrained neural models to generate grammatical annotations within the framework of Universal Dependencies (UD) (Nivre et al., 2017, 2020). We now describe the filters in more detail.

### 5.3.2 Corpus Filters

In general, we favour "stronger" filters, i.e. those that include false positives (and so filter out more training data), since our goal is to ensure that the LM has not seen a given construction during training. In what follows, $x >_z y$ means that there is a dependency from $x$ to $y$ with label $z$.

#### Structural Generalisation

In the following filters, a particular structural configuration has been completely removed from the corpus, and a model must generalise to it from similar/related configurations.

**AGR-PP-MOD** The benchmark targeted by this filter tests subject-verb number agreement in the presence of an intervening *distractor* in a prepositional phrase, as illustrated in Figure 5.1. AGR-PP-MOD filters all sentences containing the dependency structure VERB $>_{\text{nsubj}}$ NOUN $>_{\text{nmod}}$ NOUN $>_{\text{case}}$ ADP. The resulting filtered corpus will still contain PPs modifying nouns in other contexts (e.g. object position). If a learner has formed a general 'rule' for subject-verb agreement, and seen PP-modified objects, it should be able to generalise to agreement with PP-modified subjects, even when it hasn't seen them during training.

**AGR-REL-CL** This filter is similar to the previous one, but targets sentences where the distractor occurs in a relative clause in subject position, removing all sentences containing the structure VERB $>_{\text{nsubj}}$ NOUN $>_{\text{acl:relcl}}$ ADJ, e.g. "The boys that aren't disturbing Natalie dream". A model might generalise again from its general 'rule' for subject-verb agreement, and learn about relative clause structure from relative clauses in object position.

**NPI- filters** We use the list of negative polarity items (NPIs) provided by Jumelet et al. (2021) and filter as follows: NPI-ONLY removes all sentences with an NPI occurring after 'only' (e.g. "Only students have ever complained about morning classes"), NPI-SENT-NEG removes sentences with a negation and an NPI, and NPI-SIM-QUES removes questions with NPIs in them. In each of these cases the model can generalise NPI licensing conditions for a particular environment from other environments that are still present.

**QUANTIFIER-SUPERLATIVE** Superlative quantifiers (e.g., *at least*, *at most*) cannot be embedded under negation: *An actor arrived at at most six lakes* vs. **No actor arrived at at most six lakes*. BLiMP targets this phenomenon in two ways: either by replacing the superlative quantifier under negation with a relative quantifier (e.g. *more than 5*), or by removing the negation. We cannot detect superlative quantifiers based on dependency information alone, so we use morphological feature annotations. Next, we filter all such constructions that appear in object position: VERB $>_{\text{obl/obj/iobj}}$ NOUN $> \cdots >$ QUANTIFIER. It is less clear for this filter how a model can still infer the grammaticality from other constructions that are not covered by the filter.

**QUANTIFIER-EXISTENTIAL-THERE** *Weak* quantifiers can occur in the scope of existential *there* constructions, whereas *strong* quantifiers can not: *There are many people here* vs. **There are all people here* (Milsark, 1974). BLiMP targets this phenomenon in two ways: either replacing a weak quantifier with a strong one, or increasing the scope of a locative *there* such that it becomes existential. We filter all weak quantifiers occurring in subject position under an existential *there*: THERE $<_{\text{expl}}$ ARE $>_{\text{nsubj}}$ NOUN $>$ WEAK-Q. However, we only filter the 5 weak quantifiers occurring in the BLiMP benchmark (*a(n), no, some, few, many*), which still allows a model to generalise from other weak quantifiers to infer the grammaticality conditions. Furthermore, weak vs. strong quantification plays a role in other linguistic phenomena as well, a fact which a learner could leverage.

**BINDING- filters** Four filters, BINDING-C-COMMAND, BINDING-CASE, BINDING-DOMAIN, and BINDING-RECONSTRUCTION target the seven binding-related benchmarks of BLiMP. All seven benchmarks typify various facets of Chomsky (1993)'s Principle A. The implementations of all four filters is generally similar: they target sentences where a reflexive or non-reflexive pronoun occurs in the specific context(s) illustrated by the corresponding benchmarks, narrowly construed, while leaving in sentences where the same or similar principle is applied in a different environment. For example, the BINDING-C-COMMAND filter removes evidence of the use of the c-command relationship in anaphora licensing *in relative clauses*, but not elsewhere, as in sentences like *Mary's brother hurt himself* (but not **Mary's brother hurt herself*).[1] The other three benchmarks operate in similar ways.

**DET-ADJ-NOUN** One of the filters targeting determiner-noun agreement focuses on cases where an adjective occurs between a demonstrative determiner and a noun, e.g. *These/*This red cars*. We create a filter that removes *all* occurrences of a demonstrative determiner followed by an adjective and a noun. A model can then still infer the number agreement from determiner/noun pairs without an intervening adjective.

### Lexical Generalisation

In the following filters we do not filter out an entire configuration, but only do so for a subset of lexical items. This way a model can indirectly generalise to a specific occurrence of the configuration from other occurrences, but no longer rely on direct co-occurrences. These filters focus on lexical generalisation because the BLiMP benchmarks that they target are centred around particular lexical items and not particular syntactic constructions.

**AGR-RE-IRR-SV** The four BLiMP benchmarks targeted by AGR-RE-IRR-SV all test language model performance on subject-verb agreement, targeting "regular" plurals, like *dress/dresses* and "irregular" plurals, like *goose/geese*. The filter removes all sentences with nominal subjects where the noun occurs in any of the four benchmarks. A learner on this filtered corpus can still solve the benchmark if it develops a notion of grammatical number, a representation of the grammatical number of the nouns in the benchmark based on their usage in other contexts, and then generalises the subject-verb agreement it sees for other nouns to these nouns.

[1] BLiMP assumes a straightforward one-to-one relationship between certain names and their grammatical gender. While such a relationship may not actually be borne out in practice today, the corpora used in this investigation likely do adhere to such a formulation.

**DET-NOUN** The other filter besides DET-ADJ-NOUN that targets determiner-noun agreement for demonstrative determiners (e.g. *These/*This books*) does so with the determiner directly adjacent to the noun. We create a filter based on all nouns occurring in the BLiMP benchmark that are preceded by a demonstrative determiner. A model can still infer the number agreement between determiner and noun from other nouns, and learn the number information of the filtered nouns from other agreement tasks like subject-verb agreement.

**PASSIVE** In English, passive constructions can only be formed by transitive verbs. BLiMP targets this phenomenon by replacing transitive verbs in passive constructions by intransitive verbs: *John is insulted by Mary* vs. **John is smiled by Mary*. Much like AGR-RE-IRR-SV and DET-NOUN, the PASSIVE filter operates by removing sentences that contain words on a word list in a specific linguistic environment. Concretely, this word list consists of the verbs that are actually used in these two benchmarks in passive form, and the filter removes sentences where such words appear in passive voice.

## 5.4 Experimental Setup

**Data** The base train, validation, and test corpora are the English Wikipedia corpora released by Gulordava et al. (2018), with the train corpus consisting of 3.05M sentences (83M tokens, with a vocabulary size of 50000 plus an unknown and EOS token). The 15 filtered corpora are derived from this base corpus by discarding all sentences that are targeted by the filter. The number of sentences and tokens discarded by each filter varied from as little as ∼0.1% to as much as ∼18.5%; for specifics, refer to Table 5.1. Then, as an additional control, the fifteen filtered corpora plus the original, FULL training corpus were uniformly downsampled to 2.4M lines, corresponding to ∼80% the size of the original training corpus. It is worth noting that the number of *tokens* did vary by as much as ∼4.2%, as reflected in the rightmost column of Table 5.1: this is explained by the fact that certain filters target longer sentences more often.

**Models** Two architectures are used for the models trained in this investigation: LSTMs (Hochreiter and Schmidhuber, 1997)) and decoder-only Transformers (Vaswani et al., 2017). For each architecture we train separate models on the 16 training corpora for five random seeds each, resulting in a total of 160 models. Model hyperparameters were selected to control for number of parameters as closely as possible. The LSTMs have two layers with embedding and hidden dimension of 1024. Output and embedding layer weights were tied, and we used dropout of 0.1 during training. The Transformers were constructed with feed-forward and hidden layer dimensions of 768, eight attention heads, and eight hidden layers. The LSTMs and the Transformers had 68.0M and 67.1M trainable parameters, respectively.

**Training** Each model was trained on a single A40 GPU for 40 epochs with mixed-precision training, using the AdamW optimisation algorithm (Loshchilov and Hutter, 2019), a linear scheduler with an initial learning rate of $5 \times 10^{-5}$, and a batch size of 32. We evaluated each model at the end of every epoch, and report results for the model with the best validation perplexity. The full hyperparameter set may be found in subsection A.3.1.

**Evaluation** We use four metrics—three standard and one novel—as the primary means of evaluation for all models. The first is perplexity over the (unfiltered) test corpus of Gulordava et al. (2018). The second is accuracy on each of the 67 benchmarks in the BLiMP challenge set (Warstadt et al., 2020). Accuracy on the BLiMP benchmarks was assessed via the "full-sentence" method (Marvin and Linzen, 2018), where a "success", for any minimal pair, is defined by the model assigning a higher probability to the grammatical sentence in the minimal pair ($s^+$) than to the ungrammatical sentence ($s^-$).

However, the FICT methodology's main advantage lies not in looking at the performance of each model in isolation, but on the *difference* in performance between two models that are otherwise identical but for their training data. Thus, for each model and each BLiMP benchmark, a change score (or delta) was calculated with respect to the average performance of all models of the same architecture trained on the FULL corpus (i.e. average over the five seeds).

To be more precise, with $M$ a model type (i.e. $M \in \{\mathsf{LSTM}, \mathsf{Transformer}\}$), $F$ a filter, and $B$ a benchmark, $F(B)$ will refer to the filtered corpus targeting $B$, and $M_F$ will refer to a model trained on $F$. We can then define the accuracy delta by:

$$\mathsf{acc}\Delta(M, F, B) := \mathsf{acc}_B^{M_F} - \overline{\mathsf{acc}_B^{M_{\textsc{full}}}}$$

where $\mathsf{acc}_B^M$ refers to the accuracy of model $M$ on benchmark $B$. We will often be interested in the case where $F = F(B)$, i.e. the benchmark(s) corresponding to the corpus filter, but report others as well.

Our final evaluation metric looks at the *probability deltas* between grammatical and ungrammatical sentences:

$$P\Delta(M,F)(s) = \log P_{M_F}(s^+) - \log P_{M_F}(s^-) \qquad [5.1]$$

$P\Delta$ expresses the magnitude of a model's grammaticality judgment: whereas $\mathsf{acc}\Delta$ only expresses the ratio of items for which a model assigned a higher probability to the grammatical case, $P\Delta$ can be interpreted as the confidence of a model's judgment.

## 5.5 Results

We present our results along the four metrics of §5.4: perplexity (§5.5.1), TSE accuracy (§5.5.2), accuracy delta (§5.5.3), and probability delta (§5.5.4).

### 5.5.1 Perplexity

We found that Transformers uniformly achieve lower perplexities on the test corpus than the LSTMs for all training corpora, as expected. The mean test perplexity across all corpora and random seeds was 47.13 for the Transformers and 53.56 for the LSTMs; a paired $t$-test of mean perplexities per corpus found the difference between the model types to be significant ($t = 270.94$, $p \approx 0$). As noted in paragraph 5.4, while we downsampled all corpora to the same number of lines, the number of tokens varies between different training corpora. Previous research has shown a clear negative relationship between the number of tokens seen in training and test corpus perplexity. This effect is also present in our data, for both architectures (LSTMs: Pearson's $r = -0.970$; Transformers: $r = -0.976$).

We also investigate the perplexity on the BLiMP sentences for the FULL and Filtered models. This provides us insight into the likelihood of these sentences: if the model assigns a relatively low likelihood to them, then grammaticality judgements will be less reliable as well (Newman et al., 2021). In Figure 5.3 we show the scores for this. Surprisingly, the LSTM models yield *lower* perplexity on the BLiMP sentences than the Transformers. This shows that Transformers have shifted their probability mass to other sentence types than found in BLiMP, but where to exactly remains an open question. Nonetheless, the perplexity scores on BLiMP are similar to the average perplexity on the test corpus, which demonstrates that these items are of similar likelihood.

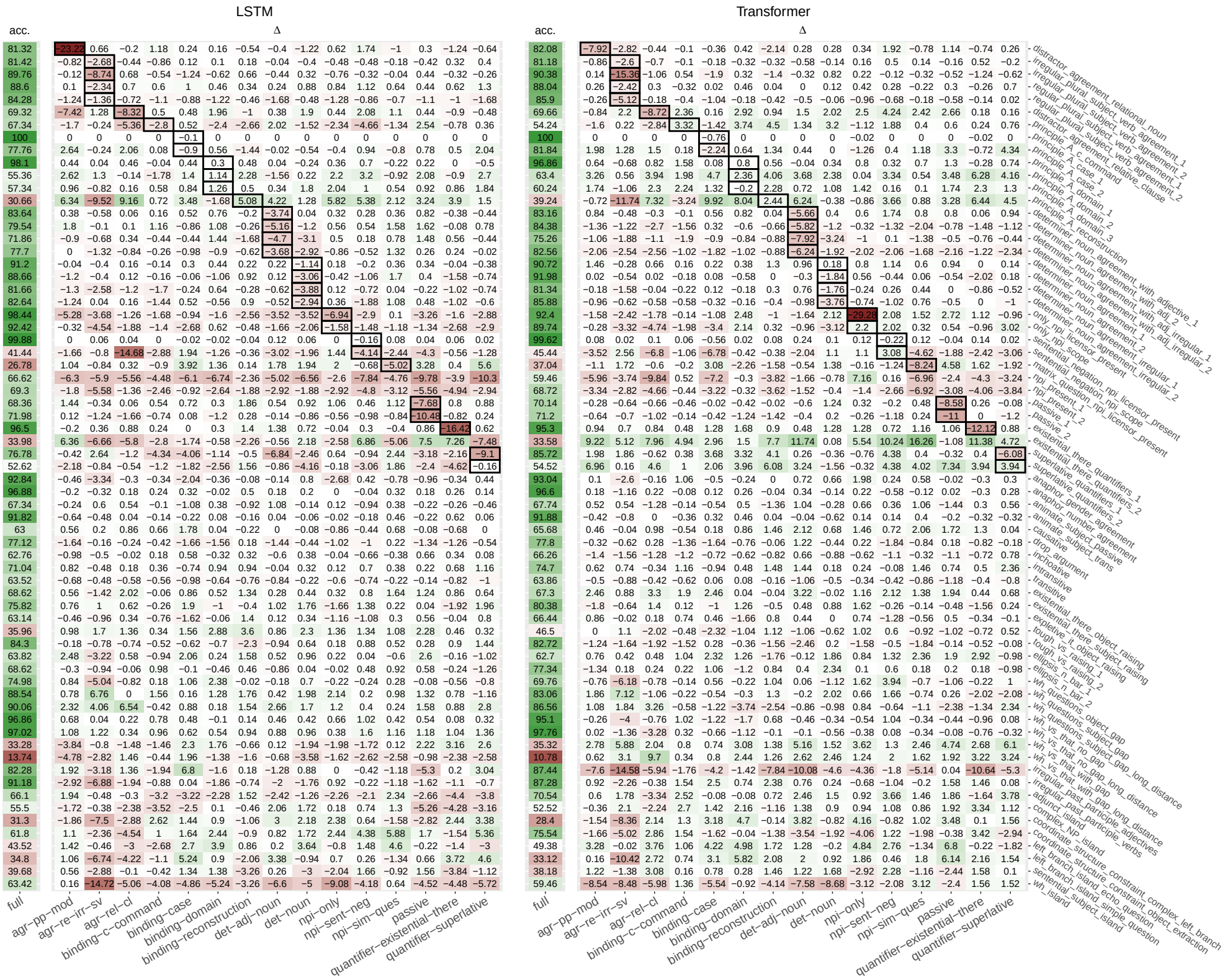

**Figure 5.2:** BLiMP benchmark *accuracy* for the models trained on the full corpus, and *accuracy delta* ($\Delta(M, F, B)$) for the filtered corpora, averaged across seeds. Boxes with bold outlines correspond to benchmarks targeted by the model's corpus filter (i.e. where $F = F(B)$). The accuracy scored by a given model on a given benchmark trained on a filtered corpus can be recovered by adding its delta to the accuracy score in the "full" column of the same row.

### 5.5.2 TSE Accuracy on BLiMP

Mean overall accuracy on all of BLiMP across different training corpora (i.e. $\overline{\text{acc}^{M_F}_{\text{ALL}}}$) was 70.4 for the LSTMs and 71.9 for the Transformers. This result was statistically significant (paired $t = -17.38, p \approx 0$). Figure A.1 in subsection A.3.2 shows all of the accuracies.

We next look only at benchmark accuracy data where the filtered corpus targeted a given benchmark, i.e. where $F = F_B$. Here, the mean is 68.8 for the Transformers and 66.7 for the LSTMs *and this difference is not statistically significant* (paired $t = -1.18, p = 0.258$).

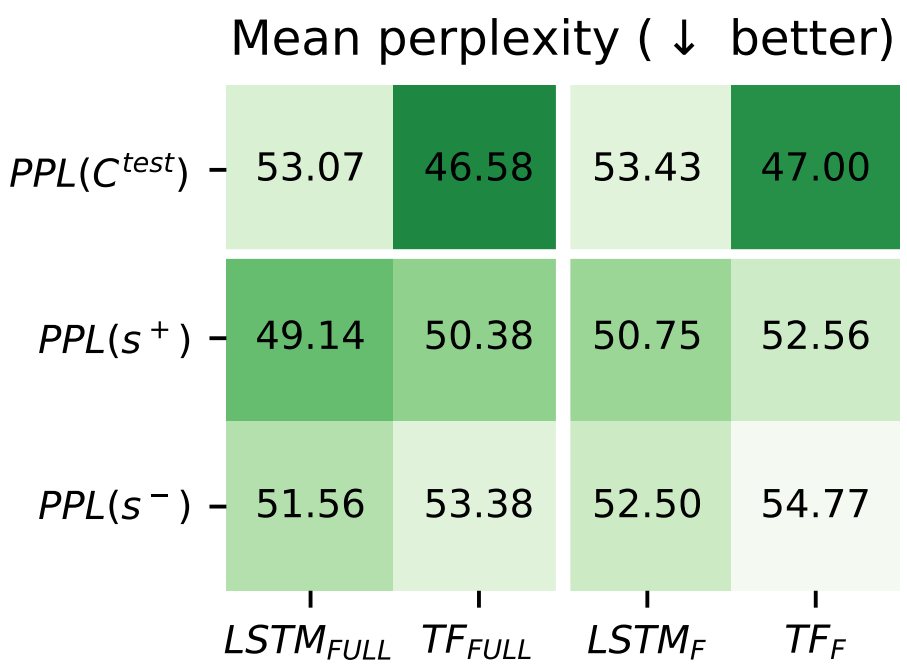

**Figure 5.3:** Perplexity scores on the test corpus ($C^{test}$) and the grammatical and ungrammatical BLiMP sentences ($s^+$ & $s^-$). BLiMP scores for the FULL models are averaged over all benchmarks, and for the Filtered models for their corresponding benchmark.

In other words, we find no difference in the two models' ability to make grammaticality judgements when trained on filtered data that forces them to perform subtle generalisations, despite differences in perplexity.

### 5.5.3 Accuracy Delta

A table of the accuracy deltas, averaged across all random seeds, can be found in Figure 5.2. Mean overall accuracy delta over all benchmarks and across all training corpora (i.e. $\overline{\Delta}(M, F, B)$) was $-0.393$ for the LSTMs and $0.0313$ for the Transformers. This result was statistically significant (paired $t = -5.10, p \approx 0.00013$).

Focusing on the $F = F(B)$ cases (i.e. black-outlined cells in the chart), we note that most deltas are generally negative but fairly close to zero, with a few outliers, such as the models trained on the EXISTENTIAL-THERE, AGR-PP-MOD, and NPI-ONLY corpora. These results suggest that, overall, learners *are* usually able to use various sorts of indirect evidence to acquire correct grammatical generalisations when direct evidence has been made unavailable, as otherwise we could expect much larger deltas across the board.

We may also observe that, for the cases where the absolute value of the deltas was appreciably larger than zero, it is not the case that one architecture is uniformly better than the other. For example, LSTMs perform better than Transformers (that is, their deltas are smaller in magnitude) on the benchmarks associated with the AGR-RE-IRR-SV and the NPI-ONLY corpora, while the converse is true for AGR-PP-MOD and quantifier-existential-there. This is true *even* for phenomena that are seemingly relatively similar; for example, the AGR-PP-MOD and AGR-RE-IRR-SV-AGR filters are extremely similar, in that they both test long distance agreement in the present of a clausal distractor intervening between the subject and the verb; they differ only in the nature of that distractor. Yet, as noted, LSTMs trained on the AGR-RE-IRR-SV corpus have, on average, a less negative delta on the associated benchmarks than the analogous Transformer models ($\overline{\text{acc}\Delta}$(LSTM, AGR-RE-IRR-SV, $F(B)$) $= -3.78$; for the Transformer, $-6.38$); conversely, on the models trained on the AGR-PP-MOD corpus, it is Transformers which have the smaller magnitude delta ($\overline{\text{acc}\Delta}$(LSTM, AGR-PP-MOD, $F(B)$) $= -23.22$; Transformer, $-7.92$).

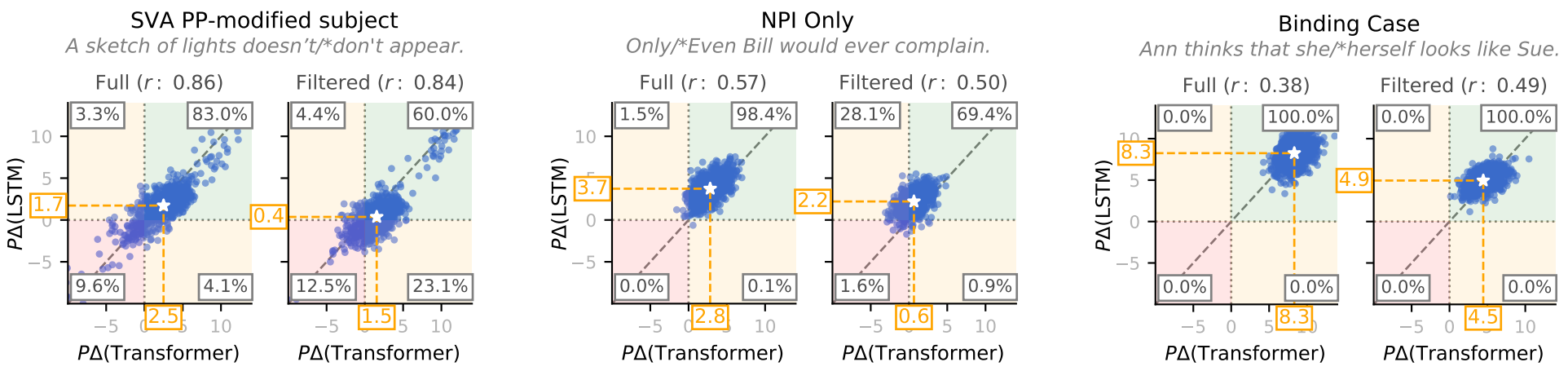

**Figure 5.4:** Log probability differences between grammatical and ungrammatical minimal pairs ($P\Delta(M, F)(s)$), with Transformer performance plotted against LSTM performance. Individual points are the averaged scores across the five model seeds. The four quadrants indicate the cases where i) both architectures got a correct prediction (green), ii) only one architecture got a correct prediction (orange), and iii) neither architecture was right (red). It can be seen that corpus filtering results in probability differences moving closer to the origin, and that the magnitude of the difference of the full models can create a sufficient margin for the model to generalise in the filtered cases as well.

As in the previous section, we can make this precise by analysing all of the accuracy deltas where $F = F_B$. The mean here is $-5.41$ for the LSTMs and $-4.62$ for the Transformers and this difference is not statistically significant (paired $t = -0.562, p = 0.583$). That means that we again find no difference between the two architectures in the extent to which filtering affects their accuracy, despite significant differences in perplexity. This suggests that perplexity *does not* predict the ability of a model to perform linguistic generalizations from indirect evidence.

### 5.5.4 Probability Delta

In order to gain a more fine-grained insight into the impact of corpus filtering, we examine the results at an item-level. For this we make use of the $P\Delta$ metric, which expresses a model's magnitude of a grammaticality judgement. In Figure 5.5A we plot the average $P\Delta$ scores for the FULL models for each BLiMP benchmark, averaged across seeds. It can be seen here that the Transformers and LSTMs result in highly similar $P\Delta$'s ($r = 0.98; p \approx 0$), although the Transformer scores are slightly higher on average those of the LSTMs (2.99 vs. 2.41, respectively), which is in line with the significant difference in TSE accuracy of §5.5.2.

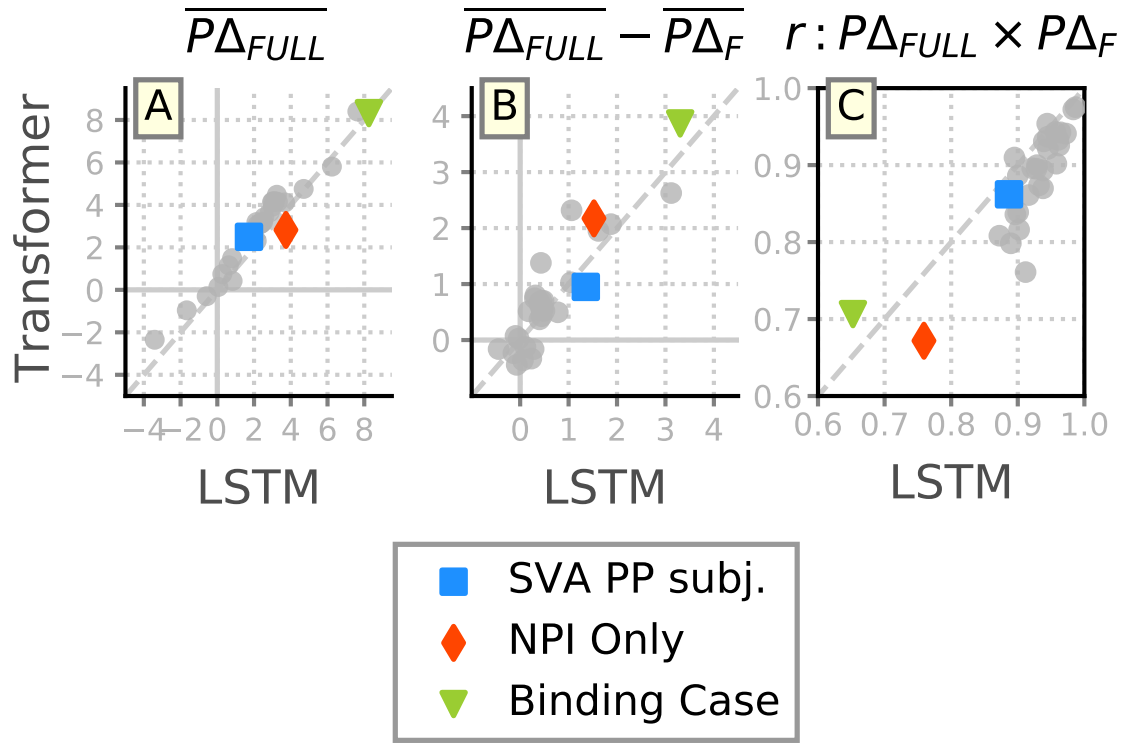

**Figure 5.5:** A: $P\Delta$ scores for the FULL Transformers and LSTMs for each BLiMP paradigm. The more positive this score, the more certain a model is in its grammaticality judgment. B: Paradigm-level differences in $P\Delta$ scores going from the FULL to the Filtered model. The closer to the origin, the less impact the filtering procedure had on model behavior. C: Pearson correlation of $P\Delta$ scores between the FULL and Filtered models. A detailed table with these results per paradigm is provided in Figure A.2 in Appendix A.3.2.

For the sake of brevity, we focus on three salient filters that each yielded distinct results: i) Subject-Verb Agreement for PP-modified subjects, in which LSTMs are more impacted than Transformers (acc$\Delta$: $-23.2$ vs. $-7.9$); ii) NPI Only, in which LSTMs are *less* impacted than Transformers (acc$\Delta$: $-6.9$ vs. $-29.3$); and iii) Binding Case, in which neither architecture is impacted by filtering. In Figure 5.4 we plot the item-level scores of the LSTMs against the Transformers (averaged across seeds). For each benchmark $B$ we plot the results on the FULL model and the $F(B)$ filtered model. This demonstrates that corpus filtering has the effect of moving $P\Delta$ closer to the origin: the model becomes *less certain* in its grammaticality judgement. The resulting acc$\Delta$ score for a benchmark is then dependent on the $P\Delta$ scores of the FULL model: a sufficient margin here makes it robust to the decrease in $P\Delta$ and allows it to correctly assign higher probability to the grammatical item.

To investigate this observation across all benchmarks we plot the difference in $P\Delta$ going from FULL to Filtered in Figure 5.5B. This difference represents the *absolute impact* of filtering on the TSE task. By plotting the Transformer results against the LSTM we gain an insight whether filtering has a similar impact on both architectures. We observe a strong correlation between these $P\Delta$ differences ($r$ : 0.91, $p \approx 0$). Subtle difference are present, however, for a number of filters the $P\Delta$ score *increases* after filtering which is especially prevalent for the Transformer models.

Finally, we examine the *robustness* of a model's grammaticality judgements: does filtering have a significant impact on the distribution of judgements? For this we compute the Pearson correlation of $P\Delta$ before and after filtering for each filter benchmark. A model is robust to filtering if this correlation remains high. In Figure 5.5C we plot the LSTM correlations against the Transformer. A striking difference between the two architectures arises here: the LSTM correlations are systematically larger than those of the Transformer. This shows that LSTMs are less impacted by filtering on an item-level than Transformers.

## 5.6 Discussion

**Perplexity Versus Linguistic Generalisation** Our findings contribute to a growing body of research that suggest a dissociation between perplexity and more targeted evaluations of linguistic competence in artificial learners (Hu et al., 2020). In a carefully controlled setting and for a wide range of phenomena, we demonstrate that the training objective of minimising perplexity does not predict linguistic generalisation. This raises interesting questions on the relation between perplexity and grammaticality judgements (Lau et al., 2017): while Transformers are better at *memorising* the structure of its training data, we show they are less capable than LSTMs of forming robust linguistic generalisations. An interesting step for future work would be to uncover what language modelling aspects Transformers *do* excel at, which allows them to obtain a superior test perplexity (e.g. word frequency, as studied in Wei et al., 2021). Future work should also compare our measure(s) of generalisation with others in the literature, given evidence that these are not always well-correlated with each other (Sun et al., 2023).

We also note that while likelihood judgements do not necessarily directly measure grammaticality, since likelihood is the outcome of many other factors (e.g. semantic plausibility, pragmatic felicity), the use of minimal pairs for BLiMP does help control for this since it reports judgements on sentences which differ on (usually) one word, thus keeping these other components constant between the two sentences. That being said, it would be a worthwhile follow-up to conduct probing experiments to more directly model grammaticality judgements, in the style of e.g. Jumelet et al. (2021) (see the next subsection as well).

**Generalising from Indirect Evidence** Our study also builds on the insights of numerous other works that use artificial learners as models for understanding human language acquisition, and gaining better insights in the inductive biases of such learners (Warstadt and Bowman, 2020a; Mueller and Linzen, 2023; Weber et al., 2024). The present study conducts for a wide range of phenomena what Warstadt (2022) calls a "proof-of-concept [of a] large-scale controlled ablation study on the input to model learners," and finds that direct attestation of linguistic evidence is not strictly necessary for the development of sophisticated linguistic generalisations. Rather, learners can leverage much more indirect sources of evidence to arrive at the correct generalisations.

Where earlier work has focused on specific linguistic constructions, such as subject auxiliary inversion (Warstadt, 2022), relative clauses (Prasad et al., 2019), and negative polarity items (Weber et al., 2021), the results of this paper essentially confirm a similar result for a much wider array of syntactic and semantic phenomena. While in many cases the ablations we performed did clearly negatively affect the performance of our artificial learners on the relevant linguistic evaluations, the magnitude of this effect was generally quite small for all but a small handful of the linguistic phenomena we analysed. In general, even when tested on the specific benchmarks corresponding to the environments that were ablated from their input, models still perform considerably better than chance. Thus, our research provides evidence in favour of the indirect evidence hypothesis.

Notably, we find that this is true not only for filters where there are fairly obvious sources of indirect evidence (as enumerated in section 5.3), but also for filters where potential sources of indirect evidence for a correct generalisation are much less clear (such as the SUPERLATIVE-QUANTIFIER filter). This suggests that there may be complex mechanisms by which certain linguistic generalisations can be derived via highly indirect means. Thus, our results open a door to future research that can provide a more thorough account of the source of these generalisations, with potentially significant ramifications for linguistics.

**Explaining Linguistic Generalisation** As just discussed, the primary contribution of this paper has been the development of the FiCT method and the use of it to demonstrate LMs' successful generalisation from indirect evidence across a *wide range* of linguistic phenomena. This success raises a very natural follow-up question: what explains this successful generalisation behaviour?

While a complete answer to this question must await future work, a detailed look at the NPI cases can provide insight into what an answer may look like. In Chapter 6 we use a filtered corpus method to test LSTM LMs' understanding of negative polarity items, but we also conduct a further analysis to examine the basis upon which the models made their grammaticality judgements. In particular: we find (via probing classifiers) that LMs' were successfully recognising the *monotonicity* of a linguistic environment and (via a novel correlation method) that these judgements of monotonicity were highly correlated with the LMs' judgement of NPI acceptability, reflecting human acceptability judgements (Denić et al., 2021; Chemla et al., 2011).

This example suggests two paths forward for explaining the generalisation observations in the present paper. On the one hand, in the same way that the monotonicity explanation was inspired by human generalisation, detailed explanations of individual cases of generalisation can be developed with human behaviour as an initial inspiration. On the other hand, in the same way that this paper extends the filtered corpus training method to a much wider range of phenomena, one can attempt to generalise these forms of explanation on the breadth axis as well. We leave these exciting pursuits to future work.

## 5.7 Conclusion

We introduced the **Fi**ltered **C**orpus **T**raining methodology and applied it to a wide range of linguistic constructions from the BLiMP benchmark. Our results show that while Transformers are better language models (via perplexity) than comparable LSTMs, the latter generalise equally well (via $\mathsf{acc}\Delta$ and $P\Delta$). The relatively low $\mathsf{acc}\Delta$ scores in general show that all of our LMs exhibit a strong ability to generalise from indirect evidence, even for models of relatively low parameter count trained on relatively small data. In summary, this shows that language model success cannot be attributed solely to memorisation from its training data, since the data has been systematically purged of the evaluation targets. They are, instead, able to form subtle and linguistically-relevant generalisations from indirect evidence. Future work will (i) extend this approach to models of different sizes and pretraining corpora, (ii) perform deeper analyses of the bases on which the models do make their generalisations, and (iii) analyse other forms of lexical and structural generalisation through the lens of filtered corpus training.

*Look, Larry. Have you **ever** heard of Vietnam?*

Walter Sobchak

# 6 Language Models Use Monotonicity to Assess NPI Licensing

In the previous chapter we introduced FiCT, a general methodology for investigating the indirect sources that may drive the acquisition of grammatical proficiency in LMs. While our approach allowed us to demonstrate that many linguistic phenomena can be learned by LMs without having seen them directly during training, a question that remains is whether the generalisation signal is linguistically meaningful. In other words, we ask whether this signal based on underlying abstractions that will be useful for many other linguistic generalisations as well, and whether these abstractions correspond to notions from established linguistic theories. For example, is the generalisation to unseen relative clauses on subject-verb agreement based on a general signal of number agreement, or that of anaphora agreement based on general signal of grammatical gender? To answer such questions, we need to beyond behavioural tests, and investigate the model representations using *probing* techniques. In this chapter, we present an analysis of a semantic property, *monotonicity*, and connect a probing-based procedure to generalisation results from FiCT.

## 6.1 Introduction

Language models have become powerful approximators of human language, making it increasingly important to understand the features and mechanisms underlying their be-haviour. In the past few years, a substantial number of studies have investigated the linguistic capabilities of LMs (see §1.3 for a detailed overview). Such work has focused primarily on *syntactic* properties, while fewer studies have been done on what kind of *formal semantic* features are encoded by language models. In this chapter, we focus explicitly on what LMs learn about a semantic property of sentences, and in what ways their knowledge reflects well-known features of human language processing.

As the topic of our studies, we consider **monotonicity**, a semantic property of linguistic environments that plays an important role in human language understanding and inference (Hoeksema, 1986; Valencia, 1991; Van Benthem, 1995; Icard III and Moss, 2014): the monotonicity of a linguistic environment determines whether inferences from a general to a particular term or vice versa are valid in that environment. For example, the fact that the inference from "*Mary didn't write a paper*" to "*Mary didn't write a linguistics paper*" is valid shows us that the position where "*a paper*" occurs is *downward monotone*: the inference is valid when a more general term ("*a paper*") is replaced with a more specific one ("*a linguistics paper*").

To investigate monotonicity we focus on **negative polarity items** (NPIs): a class of expressions such as *any* or *ever* that are solely acceptable in downward monotone environments (Fauconnier, 1975; Ladusaw, 1979). Psycholinguistic research has confirmed this connection between NPIs and monotonicity: humans judge NPIs acceptable in a linguistic environment if they consider that environment to be downward monotone (Chemla et al., 2011). Previous research has established that LMs are relatively successful in processing NPIs (Warstadt et al., 2019a), but without investigating *how* they came to these successes.

We raise the following research questions:

**RQ1** Do language models encode the monotonicity properties of linguistic environments?

**RQ2** To what extent do they employ this information when processing negative polarity items?

We developed a series of experiments, in which we first evaluate the general capacities of LMs in handling monotonicity and NPIs and then investigate the generalisation heuristics of the LM by doing experiments with modified training corpora. First, we establish that LMs are able to encode a notion of monotonicity by probing them with diagnostic classifiers (DCs, Hupkes et al., 2018) (§6.5.1). In our second experiment we demonstrate that our LMs are reasonably successful with NPI licensing using an NPI acceptability task (§6.5.2). Next, we introduce a novel *DC ranking* method to investigate the overlap between the information that the model uses to make judgements about NPIs and the information that the DCs use to predict monotonicity information, finding that there is a significant overlap (§6.5.3).

We then investigate two potential confounds that may obfuscate our results. First, we consider whether the signal that is picked up by the monotonicity DC is not simply a proxy that tells the model that an NPI may occur at that position (§6.5.4). To assess this, we train new LMs on a corpus from which all sentences with NPIs have been removed, re-run the montonicity probing task, and find that even in the absence of NPI information, LMs are still able to encode a notion of monotonicity. Next, we consider whether an LM bases its NPI predictions on simple co-occurrence heuristics, or if it can extrapolate from a general notion of monotonicity to cases of NPIs in environments in which they have never been encountered during training (§6.5.5). We again train new LMs on modified corpora, this time removing NPIs only in one specific environment, and repeat the NPI acceptability and DC ranking experiments. The results of this setup demonstrate that LMs indeed use a general notion of monotonicity to predict NPI licensing.

**Contributions** With this work, we contribute to the ongoing study of the linguistic abilities of language models in several ways:

- With a series of experiments we demonstrate that LMs are able to acquire a general notion of monotonicity that is employed for NPI licensing.
- We present two novel experimental setups: *filtered corpus training* and *DC ranking*, that can be used to assess the impact of specific information during training and compare the information used by DCs with the information used with the model, respectively.
- By using experimental results from psycho-semantics to motivate hypotheses for LM behaviour, we find that our models reflect behaviour similar to human language processing.

In the remainder of this chapter, we will first provide some linguistic background that helps to situate and motivate our experiments and results (§6.2). We then discuss related work on NPI processing in LMs in §6.3. In §6.4, we discuss our methods and experimental setup. §6.5.1 through §6.5.5 explain and present the results. We conclude in §6.6 with a general discussion and pointers to future work.

## 6.2 Linguistic Background

**Monotonicity** Monotonicity is a property of a linguistic environment which determines what kind of inferences relating general and particular terms are valid in that environment. If inferences from a general to a particular term are valid, the linguistic environment is said to be *downward monotone* (**DM**). If inferences are valid the other way around, from a particular to a general term, the linguistic environment is said to be *upward monotone* (**UM**).

Examples of expressions inducing DM environments are negation and quantifiers like *nobody, no NP*, but also specific types of adverbs and the antecedents of conditional sentences. For instance, (1) below exemplifies that in these environments the inference from a sentence with a general term (*cookies*) to that sentence with a more particular term (*chocolate cookies*) is valid, but not vice versa.

(1) a. Mary did**n't** eat cookies. ⇒
Mary didn't eat *chocolate* cookies.
b. **Nobody** ate cookies. ⇒
Nobody ate *chocolate* cookies.
c. Mary **rarely** ate cookies. ⇒
Mary rarely ate *chocolate* cookies.

Common examples of UM environments are (non-quantified) positive sentences, quantifiers such as *somebody, many NP*, and other kind of adverbs. (2) exemplifies that in these environments the inference from a sentence with a more particular term (*chocolate cookies*) to the same sentence with a general term (*cookies*) is valid, but not vice versa.

(2) a. Mary ate *chocolate* cookies. ⇒
Mary ate cookies.
b. **Everyone** ate *chocolate* cookies. ⇒ Everyone ate cookies.
c. Mary **often** ate *chocolate* cookies. ⇒ Mary often ate cookies.

**NPIs** NPIs are expressions such as the English words *any, anyone, ever*, whose acceptability depends on whether its linguistic environment is downward monotone (Fauconnier, 1975; Ladusaw, 1979; Dowty, 1994; Kadmon and Landman, 1993; Krifka, 1995; Lahiri, 1998; Chierchia, 2006, 2013).[1] While the conditions for NPI acceptability are complex, a good approximation is that NPIs are acceptable (or *licensed*) in the syntactic scope of *NPI*

[1] See however Zwarts, 1995; Giannakidou, 1998; Barker, 2018 for different takes on NPI acceptability generalizations.

| Environment Class | Abbrev. | DM example | UM example |
|---|---|---|---|
| Adverbs | ADV | A lady **rarely** *ever* ... | *A lady sometimes *ever*... |
| Conditionals | COND | **If** the dancers see *any* ... | *While the dancers see *any*... |
| Determiner Negation | D-NEG | **No** teacher says that the students had practiced *at all*. | *Some teacher says that the students had practiced *at all*. |
| Sentential Negation | S-NEG | The dancer was **not** saying that the guy had profited *yet*. | *The dancer was really saying that the guy had profited *yet*. |
| Only | ONLY | **Only** the boys had *ever* ... | *Even the boys had *ever* ... |
| Quantifiers | QNT | **Every** senator who had *ever* ... | *Some senator who had *ever* ... |
| Embedded Questions | QUES | The patients wonder **whether** the lady admires *any* ... | *The patients say that the lady admires *any*... |
| Simple Questions | SMP-Q | Did the boy *ever* listen**?** | *The boy did ever listen. |
| Superlatives | SUP | A lady buys the **oldest** dish that the adult had *ever* ... | *A lady buys the old dish that the adult had *ever* ... |

**Table 6.1:** The nine environment classes of Warstadt et al. (2019a), with an example of a minimal DM/UM pair for each class taken from the corpus.

6

*licensors* that induce a DM environment.[2] If we again consider the DM environment of (1-a) and the UM environment of (2-a), it can be seen that English *any* is an NPI, as it is acceptable when inside the syntactic scope of negation (a DM expression) as in (3-a), and not acceptable when they are in an UM environment as in (3-b).

(3) a. Mary didn't eat (any) cookies.
b. Mary ate (*any) cookies.

Importantly, monotonicity plays a role at the psychological level: human judgments about the monotonicity of a linguistic environment predict their judgments of NPI acceptability in that environment (Chemla et al., 2011; Denić et al., 2021). For example, how plausible someone finds the inference (1-a) predicts how acceptable they find the sentence (3-a). Summing up, NPI licensing has a syntactic component (NPIs must reside in syntactic scope of a licensor) and a semantic component (NPI licensors are DM expressions), that are connected on a psychological level (monotonicity judgments predict NPI acceptability). Our research aims to uncover whether this connection is exhibited by LMs as well.

## 6.3 Language Models and NPIs

In Section 1.3 we provided a general overview of linguistic investigations of language model capacities. The majority of these studies have focused on syntactic ability, whereas in this study we aim to uncover the formal semantic capacities of LMs.

Jumelet and Hupkes (2018) conclude that LSTM language models encode information about the dependency between the NPI and the NPI licensor, although this effect diminishes as the distance between the NPI and its licensor grows. Marvin and Linzen (2018) study NPI judgements of LMs on minimally different sentence pairs (with the NPI licensor either in an appropriate syntactic configuration or not) and find that their models are unable to reliably assign higher probability to sentences in which NPIs are correctly licensed. The syntactic aspect of NPI licensing is also examined by Futrell et al. (2019), who demonstrate that LSTM LMs are susceptible to learning spurious licensing relationships, a finding that Warstadt and Bowman (2020b) demonstrate to also hold for BERT (Devlin et al., 2019). Wilcox et al. (2019) investigate how explicit syntactic supervision of LMs affects their success with syntactic aspects of NPI licensing. The broad linguistic suites of Warstadt et al. (2020) and Hu et al. (2020) also contain a set of tasks related to NPI licensing, demonstrating that it is one of the most challenging tasks for LMs to handle. Weber et al. (2021) investigated the dynamics of NPI learning during training, and connected this to a multi-task learning paradigm, demonstrating that LMs are able to efficiently leverage information from related licensing environments. Bylinina and Tikhonov (2022) connect LM behaviour on NPIs to their notion of *polarity*, which is studied in more detail by Bylinina et al. (2023) using an artificial language learning experiment that shows LMs generalise these formal semantic concepts in a similar way as humans do.

[2] An NPI occurs in the syntactic scope of a licensor if the licensor *c-commands* the NPI. An NPI licensor c-commands an NPI if the NPI is the licensor's sister node or one of its sister's descendants in a constituent tree (Reinhart, 1976).

Lastly, Warstadt et al. (2019a) examine BERT's ability in determining NPI acceptability. They demonstrate that BERT has significant knowledge of the dependency between NPIs and their licensors, but that this success varies widely across different experimental methods. Our study builds on the resources of Warstadt et al. (2019a). Although they demonstrate that BERT is generally successful with NPI licensing, their results do not reveal whether BERT has constructed a more general category of DM expressions that is independent of collocational cues, nor whether it has understood that this category matters for NPI licensing.

## 6.4 Methods

Before getting to the main experimental part of our work, we briefly discuss the training corpus, model architecture and evaluation corpus we consider.

**Training Corpus** The base training corpus we consider in our experiments is the corpus used by Gulordava et al. (2018). This corpus is a collection of sentences from Good and Featured English Wikipedia articles and consists of over 90M tokens. The vocabulary of the corpus consists of the 50.000 most frequent tokens in this corpus; less frequent tokens are mapped to a special <unk> token. We refer to the full training corpus type with the name Full, and to the LMs trained on this corpus as Full LMs. In addition to Full, we use multiple other corpora which are derived from Full by means of filtering. This will allow us to draw conclusions about specific generalization abilities and reliance on collocational cues of LMs; filtered corpora will be introduced in the relevant sections.

**Model Architecture** In our studies, we focus on recurrent language models. More specifically, following Gulordava et al. (2018), we consider two-layer LSTM language models, with an embedding and hidden size of 650. All training runs across our experiments follow the same regime, identical to the regime described by Gulordava et al. (2018): 40 epochs of training with SGD, with a plateau scheduler and an initial learning rate of 20, a batch size of 64, BPTT length of 35, and dropout of 0.1.[3]

**Evaluation Corpus** To assess monotonicity and NPI licensing knowledge of LMs in our experiments, we leverage the NPI corpus of Warstadt et al. (2019a), which consists of a large amount of grammatical and ungrammatical sentences with NPIs. This corpus is divided into 9 distinct **environment classes**, allowing for fine-grained analysis of NPI licensing. Importantly, these nine environment classes come in two versions: a DM version—in which NPIs are grammatically acceptable, and a minimally different UM version—in which they are not. We provide an overview with examples of DM and UM versions of all environment classes in Table 6.1. The full size of the corpus is 106.000 distinct DM sentences, and the division of environment classes is split roughly uniformly.

## 6.5 Experiments and Results

In this section we describe the experimental pipeline in more detail. A graphical overview of our experiments is depicted in Figures 6.1 and 6.4. Each experiment description is directly followed by an analysis of its results.

### 6.5.1 Experiment 1: Do LMs represent monotonicity information?

In our first experiment, we test whether LMs trained on our Full corpus possess a notion of monotonicity. We train five different LMs and test how well they represent monotonicity properties of different environments by training linear **diagnostic classifiers** (DCs,

[3] Models are trained on a *GeForce 1080 Ti* GPU, take around 40 hours to train, and consist of 71M parameters.

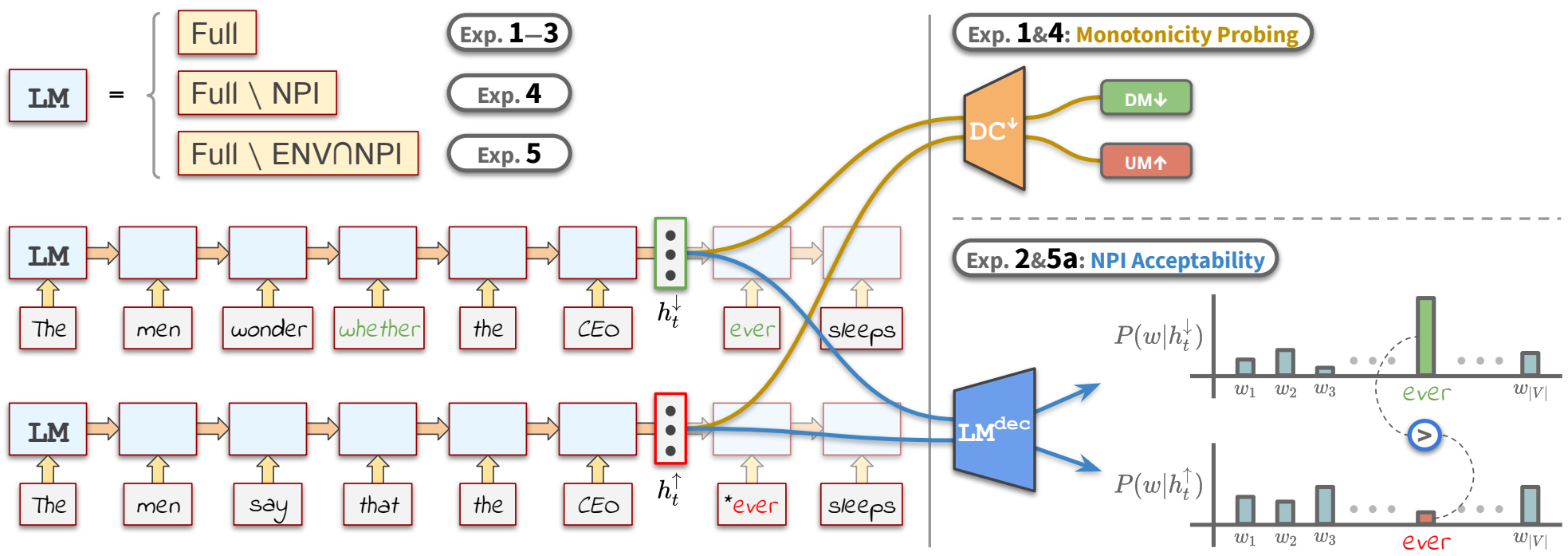

**Figure 6.1:** The pipeline of our experimental setup. We start by computing the hidden states $h_t^{\downarrow}$ (within a DM environment ahead of the NPI) and $h_t^{\uparrow}$ (within a UM environment). These hidden states are then used for training the monotonicity DC (Exp. 1 & 4), and to compare $P_{\text{LM}}(\text{NPI} \mid h_t^{\downarrow}) > P_{\text{LM}}(\text{NPI} \mid h_t^{\uparrow})$ (Exp. 2 & 5a). The task of Experiments 3 and 5a can be found in Figure 6.4. Experiments 4 and 5 consist of the same tasks as the first three experiments, but differ in the language model that is used.

Hupkes et al., 2018) on top of the hidden states of the LM. To create a corpus of monotonicity sentences for training and testing the DCs, we leverage the corpus of Warstadt et al. (2019a), now selecting all DM and UM sentences to build up a balanced corpus of these categories. The nine environment classes in that corpus hence provide a broad spectrum of DM environments and their minimally different UM counterparts.

For training and testing the DCs, we consider the hidden states at the position directly before an NPI occurs (see Figure 6.1). The reason we train the DCs at this position is because only at this point we are sure that the monotonicity information should surface and be encoded *linearly*. This is due to the fact that the decoder of the LM that transforms a hidden state into a probability distribution is linear as well: if the probability of some token depends on a linguistic feature, this feature must hence be encoded linearly. The DCs are implemented using the `diagNNose` library of Jumelet (2020), and trained using 10-fold cross-validation, Adam optimization (Kingma and Ba, 2015), a learning rate of $10^{-2}$ and L1 regularization with $\lambda = 0.005$.

We train our monotonicity DCs in two separate ways. First, we divide the entire monotonicity corpus into a 90/10 train/test split, sampled *uniformly* across the different environment classes. This allows us to examine whether DM and UM environments are linearly separable in a way that is applicable to all environment classes. We refer to this classifier as the All-ENV DC.

Second, we move to a more fine-grained type of analysis. High performance of the All-ENV DC namely does not provide evidence that monotonicity is encoded the same way

**Monotonicity probing accuracy**

| | All-ENV | ADV | COND | D-NEG | S-NEG | ONLY | QNT | QUES | SMP-Q | SUP |
|---|---|---|---|---|---|---|---|---|---|---|
| **Full (exp. 1)** | $0.97_{\pm 0.01}$ | $0.93_{\pm 0.02}$ | $0.81_{\pm 0.16}$ | $0.97_{\pm 0.01}$ | $0.91_{\pm 0.02}$ | $0.94_{\pm 0.03}$ | $0.78_{\pm 0.05}$ | $0.93_{\pm 0.06}$ | $0.80_{\pm 0.08}$ | $0.66_{\pm 0.04}$ |
| **Full \ NPI (exp. 4)** | $0.95_{\pm 0.01}$ | $0.88_{\pm 0.02}$ | $0.89_{\pm 0.11}$ | $0.94_{\pm 0.01}$ | $0.86_{\pm 0.04}$ | $0.95_{\pm 0.02}$ | $0.77_{\pm 0.03}$ | $0.89_{\pm 0.05}$ | $0.78_{\pm 0.05}$ | $0.57_{\pm 0.03}$ |

DC evaluated on held-out environment class for each environment: the set of salient hidden units used by the All-ENV DC for classifying monotonicity within the *Adverbs* environment, for example, could be disjoint from the set of units used for the *Only* environment. To investigate this, we train a DC on the hidden states of all-but-one environment class, and test its performance on the excluded class. This provides a measure to what extent the monotonicity representation of DM and UM environments derived from all other environment classes *generalizes* to the held-out class, demonstrating stronger evidence that the model represent monotonicity in the same way across different environments.

**Figure 6.2:** Accuracy and standard deviation on the monotonicity diagnostic classification task, averaged over 5 seeds for each model type. The All-ENV column denotes train/test split procedure sampled uniformly over all environment class; other columns denote accuracy on one environment class that has been excluded during training.

**NPI acceptability accuracy**

| | ADV | COND | D-NEG | S-NEG | ONLY | QNT | QUES | SMP-Q | SUP |
|---|---|---|---|---|---|---|---|---|---|
| **Full (exp. 2)** | $0.88_{\pm 0.03}$ | $0.80_{\pm 0.03}$ | $0.93_{\pm 0.03}$ | $0.82_{\pm 0.02}$ | $0.84_{\pm 0.03}$ | $0.79_{\pm 0.02}$ | $0.85_{\pm 0.02}$ | $0.72_{\pm 0.01}$ | $0.76_{\pm 0.02}$ |
| **Full \ ENV∩NPI (exp. 5a)** | $0.81_{\pm 0.00}$ | $0.65_{\pm 0.04}$ | $0.89_{\pm 0.04}$ | $0.69_{\pm 0.06}$ | $0.74_{\pm 0.05}$ | $0.69_{\pm 0.01}$ | $0.83_{\pm 0.05}$ | $0.68_{\pm 0.00}$ | $0.72_{\pm 0.01}$ |

Environment class

**Figure 6.3:** Accuracy on the NPI acceptability task—based on whether the NPI was assigned a higher probability in the DM environment than in its UM counterpart.

**Results** The results of our first experiment are shown in the top row of Figure 6.2. The first column contains the average accuracy for the All-ENV DC, and it can be seen that the diagnostic classifier succeeds in this task with high accuracy (97%). This indicates that the uniform split over all environment classes is linearly separable.

Next, we consider the held-out evaluation procedure for each of the nine environment classes. It can be seen that the monotonicity signal generalizes well to five classes (*adverbs, determiner negation, only, sentential negation, and embedded questions*), all with an accuracy above 90%. The other four classes yield a higher standard deviation, indicating that these classes are encoded less consistently across initialization seeds. The accuracy for all held-out DCs is lower compared to the All-ENV DC results, indicating that the All-ENV

DC relied partly on information unrelated to a shared notion of monotonicity. The fact that the accuracy of these DCs is still so high, however, indicates that there is a substantial overlap between the way that monotonicity is encoded within the different environments.

### 6.5.2 Experiment 2: Do LMs predict the licensing conditions of NPIs?

In the next experiment we investigate the NPI acceptability judgments of the Full LMs on the corpus of Warstadt et al. (2019a). This is done by comparing the probability of an NPI conditioned on the model's representation of a DM environment ($h_t^{\downarrow}$) and a UM environment ($h_t^{\uparrow}$), where success is defined as follows:

$$P_{\text{LM}}(\text{NPI} \mid h_t^{\downarrow}) > P_{\text{LM}}(\text{NPI} \mid h_t^{\uparrow})$$

This is a common evaluation procedure in the interpretability literature (Linzen et al., 2016), and has earlier been applied in the domain of NPI licensing by Jumelet and Hupkes (2018) and Warstadt et al. (2020). Our approach is similar to the Cloze Test of Warstadt et al. (2019a), but their setup used (bi-directional) masked LMs, making it possible to directly compare the probabilities of the NPI licensor, instead of comparing the NPI probabilities. Note that we purposefully do not base NPI acceptability on comparing full sentence probabilities: in our view this type of comparison can be distracted by token probabilities not related to the NPI itself.

We split this procedure out for each of the nine environment classes. The example sentence of the Simple Questions environment in Table 6.1, for example, is evaluated as follows:

$$P_{\text{LM}}(\textit{ever} \mid \textit{Did the boy}) > P_{\text{LM}}(\textit{ever} \mid \textit{The boy did})$$

Using the full sentence probabilities for this comparison would require taking probabilities into account such as $P_{\text{LM}}(\textit{the} \mid \textit{Did})$ and $P_{\text{LM}}(\textit{boy} \mid \textit{The})$, that have no relation to NPI licensing at all.

**Results** We present the results for this experiment in the top row of Figure 6.3. The Full models demonstrate a considerable ability at predicting NPI acceptability, with the least performing class (SMP-Q, *Simple Questions*) yielding an accuracy that is still well above chance (0.72). Compared to earlier investigations on the ability of LSTM LMs in NPI licensing, our results indicate that these models are able to obtain a more sophisticated understanding of NPIs than previously thought: both Marvin and Linzen (2018) and Hu et al. (2020) report LSTM performance below chance on NPI acceptability tasks. This might in part be due to the different evaluation procedure we used (conditional vs. full-sentence probability comparison).

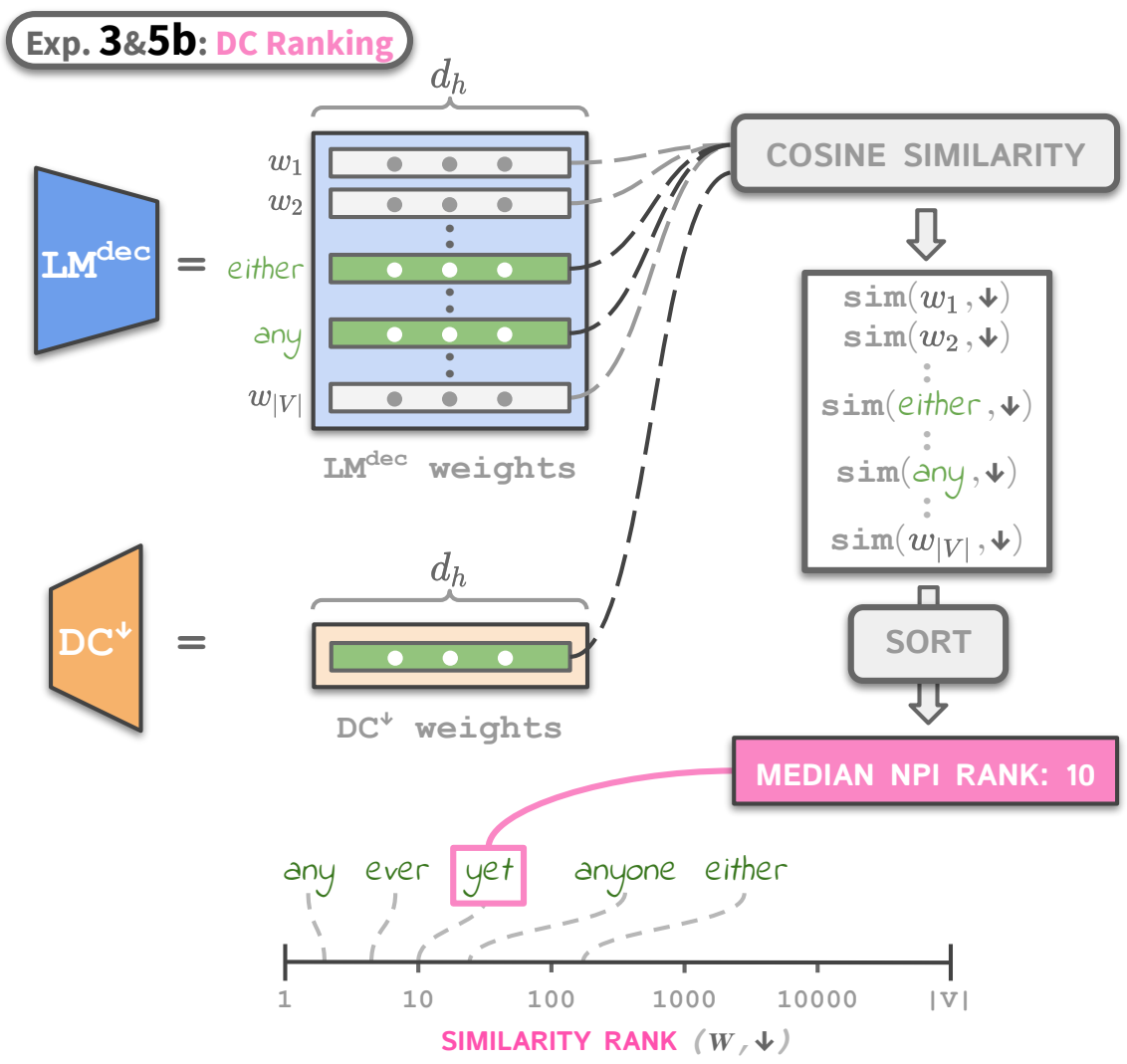

**Figure 6.4:** The *DC Ranking* experiment, in which we investigate whether the monotonicity DC and the LM decoder base their predictions on similar cues, by computing and ranking the cosine similarities between the DC weights and the decoder weights of each token.

### 6.5.3 Experiment 3: Is the LM's knowledge of DM environments and of NPI licensing related?

We have now established that our models encode a signal related to monotonicity, and are successful at predicting NPI acceptability. In our third experiment, we assess to what extent the parameters used by the LM to predict NPIs (i.e. the LM's decoder embeddings for NPIs) *overlap* with the information the DCs use to predict the monotonicity properties of a particular environment class. For this we have devised a novel **DC ranking** method, that ranks the LM's decoder weights for all tokens based on their similarity with the DC weights.

**LM decoder & monotonicity DC cosine similarity − Median NPI rank**

| | All-ENV | ADV | COND | D-NEG | S-NEG | ONLY | QNT | QUES | SMP-Q | SUP |
|---|---|---|---|---|---|---|---|---|---|---|
| Full (exp. 3) | 9 | 14 | 1121 | 16 | 28 | 449 | 206 | 82 | 14124 | 1248 |
| Full \ ENV∩NPI (exp. 5b) | 10 | 35 | 6634 | 622 | 1152 | 1418 | 1509 | 297 | 28670 | 3288 |

Environment class monotonicity DC trained on

**Figure 6.5:** Results on the median NPI rank task. A low median rank indicates that the monotonicity DC uses the same representational information as the NPI decoder.

We present a schematic overview of the method in Figure 6.4. The LM's decoder weight matrix can be interpreted as a collection of vectors corresponding to each token in the model's vocabulary. The monotonicity DC is a binary classifier, so its weights are represented by a single vector. The LM's decoder vectors are of the same dimensionality as the weight vector of the monotonicity DC, which allows us to compute the similarity between each decoder vector and the monotonicity DC. For each of the 50.000 tokens in the LM's vocabulary, we calculate the cosine similarity between the decoder weights corresponding to that token and the DC's weights. We then sort these similarity scores, which results in a ranking of tokens that are most similar to the DC.

As we are interested in finding the connection of the monotonicity DC and the LM's NPI processing in general, we compute the **median rank** over a set of 11 NPIs.[4] A low median NPI rank indicates that the LM uses the same cues for NPI prediction as the monotonicity DC, demonstrating a clear connection between NPI licensing and monotonicity.

Contrary to Experiment 1, we no longer make use of the hold-one-out training procedure, that gave insights to what extent a general monotonicity signal generalizes to a held-out environment class. Instead, we train a separate diagnostic classifier for each environment class using a train/test split made up of DM and UM environments within that class. This results in a classifier that represents the class-specific decision boundary between minimal pairs of DM and UM items and allows us to investigate to what extent these decision boundaries align with the weights of the LM decoder. Next to the environment-specific DCs we also report the DC ranking outcome for the All-ENV DC that has been trained on all environments.

**Results** The results of this experiment are presented in the top row of Figure 6.5. The first column (All-ENV) contains the result for the DC trained on all environment classes, and the median NPI rank of 9 demonstrates that the monotonicity DC aligns very closely with the NPI decoder weights of the LMs. This median rank should be interpreted within the context of the model vocabulary size: it can range upwards to 50.000, so a rank that is close to 0 signifies a tight connection between the probing task and the tokens of interest.

Moving on to the environment-specific results, it can be seen that the results vary considerably between the environment classes. The worst scoring class is again that of Simple Questions. This makes sense, as the licensing conditions for question constructions do not depend on the presence of a specific licensing token such as *not*, but on the overall structure of the whole sentence. The other environment classes lead to scores far closer to 0, indicating that for these classes monotonicity classification is closely aligned to NPI processing.

[4] We consider the following 11 single-token NPI expressions: *dared, any, anybody, anymore, anyone, anything, anywhere, ever, nor, whatsoever,* and *yet*. These are all the single-token NPIs taken from a list of NPIs that is described in §6.5.4.

Interestingly, the median rank of the All-ENV DC is lower than the ranks of all other DCs. This shows that the model has aligned its representation of NPIs to an aggregate of the monotonicity representations in the different environment classes. This allows the model to flexibly deal with NPIs in a wide range of licensing environments.

### 6.5.4 Experiment 4: Are NPIs important for learning monotonicity information?

With Experiment 3 we established that NPI processing and monotonicity are related in our LMs. Now, we investigate to what extent their representations are entangled during training. More specifically, we investigate if the signal from the presence of NPIs is indispensable for the LM to develop a notion of monotonicity, or if instead the success in categorizing monotonicity environments can be learned independently of NPIs.

We address this question by testing whether LMs can still classify the monotonicity properties of environments when they are completely deprived of NPIs during training. To do so, we train new language models on a modified corpus that does not contain any NPIs at all. To arrive at this corpus, we remove all sentences that contain at least one NPI expression from the Full corpus. We identify these expressions based on a comprehensive list of NPI expressions in English collected by Hoeksema (2012) and the list of NPIs in English compiled by Israel (2011). From this list, we manually removed expressions that have both NPI and non-NPI uses (e.g. *a thing, a bit*). The 40 NPI expressions that resulted from this procedure can be found in Appendix A.4.1. We train 5 models on this corpus and refer to them by the name Full\NPI.

In this experiment, we run the monotonicity probing procedure of Experiment 1 on the Full\NPI models. We posit that if the notion of monotonicity can be learned independently of NPIs, there should be no significant drop in performance compared to the results of the Full LMs.

**Results** We report the results of this experiment in the bottom row of Figure 6.2. Again it can be noted that the All-ENV DC, trained and tested uniformly over all environment classes, obtains a high accuracy on the task (0.95). Furthermore, none of the held-out environment DCs lead to significant drops in performance compared to the Full LMs. Based on this we conclude that even in the absence of NPI cues, LMs are still able to build up a shared robust notion of monotonicity.

### 6.5.5 Experiment 5: How robust is the connection between monotonicity and NPI processing?

This research aims to uncover whether LMs possess a robust connection between monotonicity and NPI licensing. Our findings indicate that this connection is present in our models. A major confound that has not yet been addressed, however, is the extent to which our models rely on collocational cues when judging the acceptability of an NPI. To test this, we examine whether an LM's connection between NPIs and monotonicity *generalizes* to novel environment classes in which NPIs have never been encountered during the training phase of the LM.

We have created nine modified corpora in which sentences with NPIs within a specific environment have been removed. For these different corpora, we again consider the nine NPI-licensing environments of Warstadt et al. (2019a). For each environment class we create a new corpus by removing all sentences from the Full corpus in which an NPI expression from Appendix A.4.1 is preceded by an expression belonging to that class, somewhere earlier in the sentence.[5] Note that we only remove the sentences in which the environment actually licenses an NPI; sentences in which the environment occurs without an NPI are retained. So for the *adverbs* environment, for example, we remove sentences like "*Mary rarely ate any cookies*" but not "*Mary rarely ate cookies*". For each of these nine corpora we train 3 new LMs. Models trained on these corpora are referred to by the name Full\ENV∩NPI.

We run the NPI acceptability task of Experiment 2 and the DC ranking method of Experiment 3 on the nine types of Full\ENV∩NPI models. A model with a robust connection between monotonicity and NPI processing should be able to learn for NPIs in the held-out environment that (i) the environment belongs to the class of environments in which NPIs are licensed, and that (ii) determining NPI acceptance should be done based on representational cues that are similar for monotonicity prediction.

**Results** We report the results of this experiment next to the previous results of the Full model. First, we consider the NPI acceptability task, which is reported in the bottom row of Figure 6.3. Note that each cell in this row now corresponds to a specific model type:

[5] For Simple Questions we remove a sentence if an NPI occurs in a sentence that ends with a question mark.

the ADV result, for instance, corresponds to the accuracy of the Full\ENV∩NPI models in which sentences with NPIs within adverbial environments have been removed. Our results show that the performance drops slightly for all environment classes, which can be attributed to a model's dependence on collocational cues. However, the models are still able to adequately generalize from the other environments, in which NPIs still are encountered, to the held-out environment. This demonstrates the semantic generalization capacities of the LM: it infers that the held-out environment in which NPIs have never been encountered shares some relevant properties with the other eight environment classes in which NPIs still occur.

The results for the DC ranking experiment are shown in the bottom row of Figure 6.5. Similar to the NPI acceptability results, the performance of the Full\ENV∩NPI models has dropped slightly compared to the Full models. However, if the models would no longer pick up on the connection between monotonicity in the held-out environment and NPI licensing at all, these median ranks should drop to chance, i.e. around the halfway mark of the vocabulary size (25.000). It can be seen that this is only the case for the Simple Questions environment, that was already performing poorly for the Full models. Based on this we conclude that although models depend partly on collocational cues for their connection between monotonicity and NPIs, they are still able to encode a robust connection that generalizes to novel DM environments.

## 6.6 Conclusion

Based on a series of experiments, we have established the following: (1) LMs categorise environments into DM and UM; (2) LMs are overall successful with NPI licensing; (3) LMs employ similar representational cues when processing NPIs and predicting monotonicity; (4) their categories of DM and UM environments can be learned independently of NPI occurrence; and (5) their connection between monotonicity and NPI processing is robust and not solely dependent on co-occurrence heuristics. This demonstrates that LMs have quite sophisticated knowledge of NPI licensing, which may be similar to that of humans and constitutes a vital step towards better understanding the linguistic generalisation capacities of LMs.

These results raise the question: what do LMs learn about the DM and UM environments which they succeed in finding? Do they actually learn the inferential properties of those environments, or do they rely on some other property that DM environments have in common to categorise them as such? A direction for future work would be to develop methods to probe the inferential capacities of LMs and explore how they align with the DM and UM categories they construct.

Another direction for future work would be to incorporate the recent advancements on probing-based interpretability methods in our experimental pipeline (Hewitt and Liang, 2019b; Voita and Titov, 2020). Our DC Ranking method aligns the performance of a probe with that of the language model itself, which is related to the approaches of Saphra and Lopez (2019), Elazar et al. (2021), and Lovering et al. (2021). Placing our methodology more firmly in this body of work will allow for stronger conclusions to be drawn regarding the semantic knowledge of current language models.

# Part III

# Linguistic Investigations in a Controlled Environment

6

# Overview

In the previous part of this thesis we investigated various linguistic phenomena in language models, and connected their performance to patterns in the training data. These methodologies provide deeper insights than what could be obtained from observational studies alone. However, the datasets that the models are trained on remain of high complexity and their underlying distribution is unknown. In this final part we present two studies in a training regime with full control over the generative process of the training data. This controlled paradigm provides a transparent testbed for examining questions regarding the faithfulness of interpretability methods, as well as the learning abilities and inductive biases of neural models.

**Chapter 7** We study *feature interactions* in the context of *feature attribution* methods for post-hoc interpretability. In interpretability research, getting to grips with feature interactions is increasingly recognised as an important challenge, because interacting features are key to the success of neural networks. Feature interactions allow a model to build up hierarchical representations for its input, and might provide an ideal starting point for the investigation into linguistic structure in language models. However, uncovering the exact role that these interactions play is also difficult, and a diverse range of interaction attribution methods has been proposed. In this chapter, we focus on the question which of these methods most *faithfully* reflects the inner workings of the target models. We work out a *grey box* methodology, in which we train models to perfection on a formal language classification task, using PCFGs. We show that under specific configurations, some methods are indeed able to uncover the grammatical rules acquired by a model. Based on these findings we extend our evaluation to a case study on language models, providing novel insights into the linguistic structure that these models have acquired.

**Chapter 8** We present a setup for training, evaluating and interpreting neural language models, that uses artificial, language-like data. The data is generated using a massive probabilistic grammar (based on state-split PCFGs), that is itself derived from a large natural language corpus, but also provides us complete control over the generative process. We describe and release both grammar and corpus, and test for the naturalness of our gener-ated data. This approach allows us to define closed-form expressions to efficiently compute exact lower bounds on obtainable perplexity using both causal and masked language modelling. Our results show striking differences between neural language modelling architectures and training objectives in how closely they allow approximating the lower bound on perplexity. Our approach also allows us to directly compare learned representations to symbolic rules in the underlying source. We experiment with various techniques for interpreting model behaviour and learning dynamics. With access to the underlying true source, our results show striking differences and outcomes in learning dynamics between different classes of words.

## Author Contributions

The contents of this part are based on the following two publications, which were both led by me and supervised by Willem:

1. Jaap Jumelet and Willem Zuidema. 2023a. Feature interactions reveal linguistic structure in language models. In *Findings of the Association for Computational Linguistics: ACL 2023*, pages 8697–8712, Toronto, Canada. Association for Computational Linguistics

2. Jaap Jumelet and Willem Zuidema. 2023b. Transparency at the source: Evaluating and interpreting language models with access to the true distribution. In *Findings of the Association for Computational Linguistics: EMNLP 2023*, pages 4354–4369, Singapore. Association for Computational Linguistics

*God! A red nugget! A fat egg under a dog!*

"Weird Al" Yankovic

# 7

# Feature Interactions Reveal Linguistic Structure in Language Models

This chapter tackles the issue of *faithfulness* in model explanation techniques: how can we ensure that an explanation faithfully represents the inner reasoning of a model prediction? We focus on a particular type of explanation technique that explain model predictions in terms of feature interactions, and connect these interactions to hierarchical structure. Whereas the previous chapters all focused on settings with natural language, we now take on a more controlled setting with small-scale *synthetic* languages. This way we obtain stronger guarantees about the inner workings of a model, since the solutions of the task they are trained to solve are less ambiguous than in a natural language setting.

## 7.1 Introduction

Feature attribution methods (FAMs) are a popular family of tools for explaining the behaviour of deep learning models, by explaining a prediction in terms of contributions of individual features (Ribeiro et al., 2016; Lundberg and Lee, 2017). There are many such methods proposed, and mathematical results (such as axiomatic approaches based on game theory) and theoretical frameworks (such as Covert et al. (2021)'s 'Explaining by Removing') are starting to offer a good understanding of how different methods relate to one another.

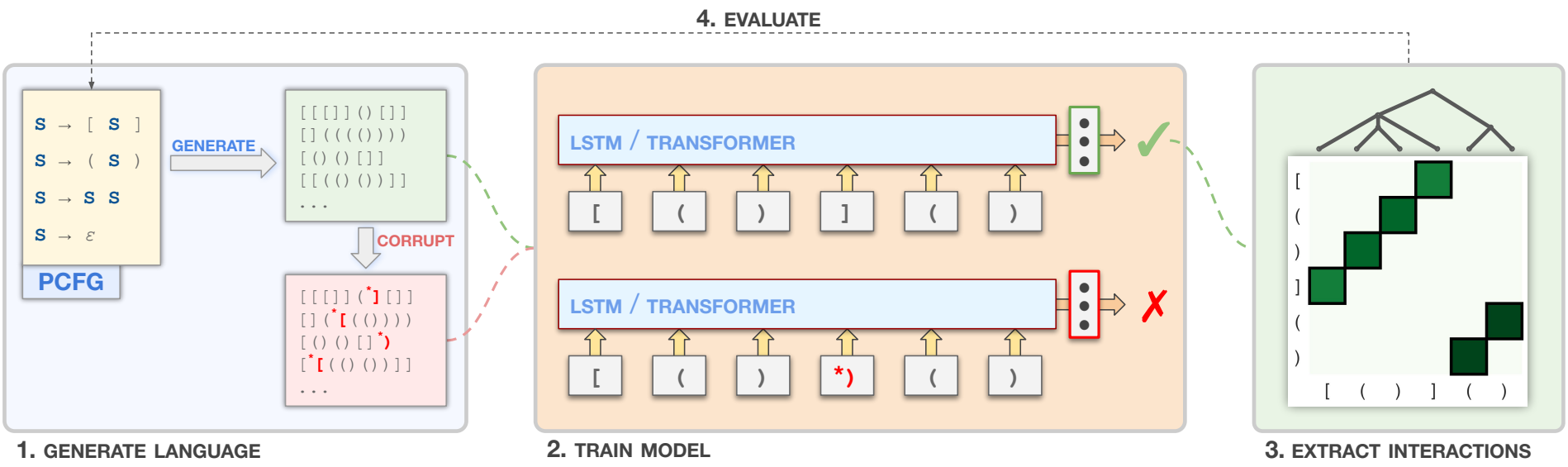

**Figure 7.1:** We generate a corpus based on a PCFG, and create negative examples by corrupting the generated corpus. Next, we train a neural model to predict whether a string is well-formed, forcing the model to obtain a comprehensive understanding of the rules of the language. Then, we extract the internal interactions by using FIDAMs described in §7.4, allowing us to directly evaluate the grammatical knowledge of the neural model.

However, there are also some important shortcomings. Perhaps most importantly, popular FAMs mostly ignore the existence of interactions between the effects of features on the prediction. This is problematic, because **Feature Interactions** are widely seen as a major factor in the success of neural networks (Goodfellow et al., 2016). This is all the more important in domains such as language and music processing, because feature interactions allow neural networks to model hierarchical representations of their input, which is considered a key design feature of language and music. To address these shortcomings, there is now an emerging literature on **feature interaction detection and attribution methods** (FIDAMs) that explain model predictions in terms of interacting features (Tsang et al., 2020; Janizek et al., 2021).

However, assessing the faithfulness of FIDAMs is even more challenging than assessing the faithfulness of feature attribution methods more generally (Jacovi and Goldberg, 2021). In this paper, we present a systematic framework to characterise FIDAMs, and derive several new FIDAMs based on that framework. We then proceed with creating an evaluation pipeline that measures a FIDAM's ability to recover the structural rules for which we have good evidence that they play an important role in the target model's performance (Figure 7.1). We first test this on a set of small-scale formal language tasks, that provide stronger faithfulness guarantees. Finally, we present a case study of a large language model on the CoLA task for linguistic acceptability.

We find that the performance of FIDAMs is very variable, and that the performance on the small-scale formal language tasks may not be predictive of the performance of methods on the large-scale natural language task. This is an illustration of what we call the **Attribution Generalisation problem**. We argue that this problem remains a key open problem in the study of explanation methods in general.

## 7.2 Related Work: Assessing Faithfulness

In this section we discuss related work on assessing the faithfulness of feature attribution methods (FAMs). A model explanation ideally provides better insights into model behaviour. However, it is important that an explanation is faithful to the reasoning of the model, and not merely plausible to a researcher. Unfortunately, attribution models can yield vastly different outcomes (Neely et al., 2022). Defining a notion of faithfulness itself is an ongoing debate, and it has been argued that we should not be aiming for a binary notion, but a graded one instead (Jacovi and Goldberg, 2021). To this end, various methodologies have been proposed to evaluate the faithfulness of explanation methods.

One research direction introduces metrics to evaluate faithfulness by quantifying the impact of features that were deemed to contribute the most by an attribution method. Hooker et al. (2019) does this by *retraining* a model on data from which the most contributing features have been removed. DeYoung et al. (2020) provide a more direct measure, by quantifying changes in model predictions when only a subset of the most contributing features is fed to model. Atanasova et al. (2020) build on this notion, introducing a range of diagnostic metrics that capture various aspects of explanation quality including faithfulness, human rationale agreement, and explanation consistency. Jain et al. (2020) ensure and evaluate faithfulness by only allowing a model access to the set of features that were deemed important by the explanation method, which has also been shown to improve model robustness (Wiegreffe et al., 2021; Ross et al., 2022).

Another line of work modifies the training data in such a way that we obtain guarantees of certain features the model must be paying attention to when making a prediction: e.g. by shuffling test data such that only part of the input resembles the statistics from the train set (Pörner et al., 2018), or by explicitly adding exploitable heuristics in the train set (Bastings et al., 2022; Adebayo et al., 2022). These two approaches could be characterised as *grey box* models: we adapt the data in such a way that we gain a degree of confidence what cues the model must be relying on, without having a full understanding of the model's internal reasoning. A *glass box* model, on the other hand, is a model whose behaviour is fully understood: it's not derived by training a model on a task, but hand-crafted. Hao (2020) utilises such models to evaluate FAMs on formal language tasks, providing more robust guarantees on model behaviour. Wilson and Frank (2023) present a setup for measuring the inductive biases for structural generalisations in neural models using formal languages.

Our own approach is related to the first line of research, making use of *grey box* models. Instead of evaluating FAMS, we evaluate FIDAMs, that provide more comprehensive insights into model reasoning. Deployment of such methods within NLP has been fairly limited, and as such evaluating their faithfulness in a language context has been an under-explored research topic.

## 7.3 A Framework for Characterising FIDAMs

Feature attribution methods typically decompose a model prediction into a sum of feature contributions (Sundararajan et al., 2017; Lundberg and Lee, 2017). A large contribution then indicates that this feature played an important role in a model's prediction. Although feature attributions can provide meaningful insights into the inner model dynamics, they paint a fairly limited picture of the model behaviour. Most importantly, **interactions** between features are lumped together, making it impossible to discern whether a large contribution of a feature stemmed from that feature alone, or from its interaction with neighbouring features. To address this, multiple methods have been proposed that decompose a model prediction into a sum of feature interactions, based on similar mathematical formalism as those of feature attributions.

**Notation** A neural network is represented as a single function $f$. The input to $f$ is denoted as $\mathbf{x}$, which consists of $N$ input features. A partial input $\mathbf{x}_S$ only consists of input features $S \subseteq N$. A value function $v(\mathbf{x}_S)$ quantifies the model output on the partial input $\mathbf{x}_S$. Padding the missing features in $\mathbf{x}_S$ with replacement features $\mathbf{x}'_{\setminus S}$ is denoted as $\mathbf{x}_S \cup \mathbf{x}'_{\setminus S}$. The attribution value of feature $i$ is denoted as $\varphi_i$, and the interaction effect of a set of features $\mathcal{I}$ is denoted as $\Gamma_{\mathcal{I}}$.

**Attribution Dimensions** Attribution methods can generally be characterised along two dimensions (Covert et al., 2021): 1) how the method deals with feature removal, and 2) how the impact of removing a feature is quantified. FIDAMs are built on the same principles as FAMs, and can be categorised along the same two dimension. By discerning these two dimensions we can separately evaluate their impact on the faithfulness of the attribution method. Furthermore, we can combine feature removal procedures with influence quantification methods in order to obtain novel attribution methods, an observation that has also been made in the context of FIDAMs by Jiang and Steinert-Threlkeld (2023), who, concurrent to our work, provide a general framework for characterising FIDAMs.

### 7.3.1 Feature Removal

It is not straight-forward to define the absence of a feature to a model's input. The main goal here is to replace the removed feature with a neutral **baseline**, that adequately represents the absence of the feature. The simplest procedure completely removes a feature from the input:

$$v(\mathbf{x}_S) = f(\mathbf{x}_S) \qquad [7.1]$$

However, this may lead to input that is no longer well-formed with respect to the original input distribution: e.g. removing '*is*' from '*This movie is great*' yields the ungrammatical input '*This movie great*'.

Methods therefore often use of a neutral input feature, the **static baseline** $\mathbf{x}'$, such as a zero-valued embedding or a pad token:

$$v(\mathbf{x}_S) = f(\mathbf{x}_S \cup \mathbf{x}'_{\setminus S}) \qquad [7.2]$$

This too, however, may lead to input that lies outside of the original input distribution (Kim et al., 2020). The reason why this is problematic is that the model may behave erratically on such modified input, posing issues to the faithfulness of the explanation.

Instead of using a static baseline, we can also opt to use a baseline that is sampled from a *background distribution* (Datta et al., 2016). There exist two approaches to this procedure (Sundararajan and Najmi, 2020; Chen et al., 2020b). The **observational conditional expectation** samples the baseline features from a distribution that is conditioned on the set of features that are still present in the input (Frye et al., 2020; Aas et al., 2021):

$$v(\mathbf{x}_S) = \mathbb{E}_{\mathbf{x}'_{\setminus S}}\left[f(\mathbf{x}_S \cup \mathbf{x}'_{\setminus S}) \mid \mathbf{x}_S\right] \qquad [7.3]$$

The **interventional conditional expectation** drops the conditional, and samples the baseline features from an independent distribution:

$$v(\mathbf{x}_S) = \mathbb{E}_{\mathbf{x}'_{\setminus S}}\left[f(\mathbf{x}_S \cup \mathbf{x}'_{\setminus S})\right] \qquad [7.4]$$

There exist two motivations for the latter approach: Lundberg and Lee (2017) drop the conditional expectation for computational reasons, allowing them to approximate the observational conditional expectation. Janzing et al. (2020) provide a perspective derived from causality theory, stating that the *intervention* of removing a feature should break the dependence between the baseline and remaining features, and hence conditioning on these features is fundamentally wrong.

The previous two methods sample baseline values for individual missing features, but we can also compute the expectation over the range of possible baselines. This yields the technique of **expected explanations** (Erion et al., 2021), in which attributions with different static baselines are averaged out over a background distribution $D$:

$$\varphi_i = \mathbb{E}_{\mathbf{x}' \sim D} \left[ \varphi_i(\mathbf{x}; \mathbf{x}') \right] \tag{7.5}$$

### 7.3.2 Quantifying Feature Influence

The simplest method of quantifying the influence of a feature is expressed as the output difference after **ablating** the feature:

$$\varphi_i = v(\mathbf{x}) - v(\mathbf{x}_{\setminus i}) \tag{7.6}$$

Note that this formulation can be combined with any of the feature removal methods: e.g. Occlusion (Zeiler and Fergus, 2014) combines this influence method with a static baseline (Eq. 7.2), whereas Kim et al. (2020) combines it with the observational conditional expectation (Eq. 7.3), employing BERT as the conditional distribution.

A more involved method leverages a technique from the field of game theory, called the **Shapley value** (Shapley, 1953). Shapley values were originally introduced in the domain of cooperative games, in which players can form coalitions to change the outcome of the game. This setup can be transferred directly to machine learning models, in which features now take up the role of the players. A Shapley value expresses the contribution of a feature as the marginal gain of including that feature in the input, averaged over all possible coalitions of features. An attractive property of Shapley values is that they are the unique solution to a set of *axioms*, providing a degree of confidence regarding the faithfulness of the attributions with respect to the model behaviour. A useful axiom is that of *completeness*, which states that all attribution values sum up to the full model output minus the baseline output:

$$\sum_i \varphi_i = f(\mathbf{x}) - f(\mathbf{x}') \tag{7.7}$$

Both these feature influence methods are based on the procedure of *perturbing* the input and capturing the change in output. A special case of this procedure is that of **gradient-based methods**, in which infinitesimal perturbations are used instead (Baehrens et al., 2010; Simonyan et al., 2014; Bhatt et al., 2020). However, it has been argued that a gra-dient does not adequately capture the contribution of a feature, as it only quantifies the rate of change at the local area around the input (Ancona et al., 2018). Furthermore, due to gradient saturation the connection between in- and output may vanish, and hence modified gradient and backpropagation procedures have been proposed instead (Bach et al., 2015; Shrikumar et al., 2017).

## 7.4 FIDAMs

7

We now address a series of interaction methods that we use in our own experiments. Research into feature interactions dates back many decades within the field of statistics. For example, two-way ANOVA (Fisher, 1925) uncovers interactions between two variables on a dependent variable by decomposing it into a sum of **main effects**, stemming from a single feature, and **interaction effects**, stemming from interactions between groups of features. The behaviour of a neural network can be explained in terms of these effects as well.

$$\mu_{ij} = \mu + \alpha_i + \beta_j + \gamma_{ij}$$

with $\mu$ denoting the grand mean, $\alpha_i$ and $\beta_j$ denoting main effects, and $\gamma_{ij}$ denoting the interaction effect between features $i$ and $j$. ANOVA is computed on a population level and is therefore unable to explain interactions within an individual sample. However, formulating main and interaction effects in an additive fashion is a fundamental assumption for many subsequent interaction methods.

**Group Ablation** The feature influence principle of Equation 7.6 can straightforwardly be extended to *groups* of features. In our experiments we will focus on pairwise interactions, but any kind of feature subset can be used here.

$$\Gamma_{i,j} = v(\mathbf{x}) - v(\mathbf{x}_{\setminus ij}) \qquad [7.8]$$

**Archipelago** Explaining model behaviour in terms of pairwise interactions will already yield a better portrayal of its internal behaviour than 'flat' attributions, but it neglects the interactions that occur within larger groups of features. Archipelago (Tsang et al., 2020) splits up the feature interaction procedure into two phases: first an interaction detection method is performed that clusters features into interaction sets, and afterwards interaction scores are assigned to these sets as a whole. Interaction detection is based on measuring the non-additive effect of pairs of features. The interaction effect that is assigned to an interaction set $\mathcal{I}$ is expressed as follows, with respect to a static baseline $\mathbf{x}'$:

$$\Gamma_{\mathcal{I}} = f(\mathbf{x}_{\mathcal{I}} \cup \mathbf{x}'_{\setminus\mathcal{I}}) - f(\mathbf{x}') \qquad [7.9]$$

Note that Archipelago expresses the interaction effect inversely compared to the Group Ablation procedure: instead of measuring the impact of removing a group of features, we now measure the impact of solely keeping this group in the input.

**Shapley(-Taylor) Interaction Index** Both the previous methods base interaction effects on direct output differences. We can modify the formulation of the Shapley value to yield interaction effects. This modification was originally introduced in the field of game theory, called the Shapley Interaction Index (SII, Owen, 1972; Grabisch and Roubens, 1999). Instead of computing the marginal gain that is achieved by a single feature, we now compute the marginal gain of *groups* of features. The Shapley-Taylor Interaction Index (STII, Sundararajan et al., 2020) is an extension of SII, satisfying additional theoretical properties.

**Hessian** Analogous to utilising the gradient for feature attributions, we can employ the second-order derivative to quantify interactions between features, which is captured by the Hessian matrix. Friedman and Popescu (2008) and Sorokina et al. (2008) consider an interaction between two variables to exist when the effect of one variable on the response depends on values of the other variable, which can be expressed in terms of the second-order partial derivative:

$$\Gamma_{i,j} = \left[\frac{\partial^2 f(\mathbf{x})}{\partial x_i \partial x_j}\right]^2$$

A common approach when using the gradient of a model as a proxy for feature importance is to multiply it with the input embeddings (Shrikumar et al., 2017; Ancona et al., 2019): in our experiments we consider an analogous method to the Hessian that we call **Hessian** $\times$ **Input**.

**Integrated Hessians** Directly using the Hessian as explanation method is prone to the same caveats as using the gradient: the interactions signal may vanish due to saturation. Integrated Hessians (IH, Janizek et al., 2021) address this issue by integrating over the Hessian manifold along a path between the input and a baseline. This is achieved by applying the method of Integrated Gradients (Sundararajan et al., 2017) to itself. An IH interaction between features $i$ and $j$ can hence be interpreted as the contribution of $i$ to the con-tribution of $j$ to the models prediction. The path integral between input and baseline is approximated via a Riemann sum interpolation.

**Other Methods** The methods explained thus far have all been incorporated in our experimental pipeline. The scope of our work focuses mainly on *pairwise* interactions, but methods that extract higher-order interactions have been proposed as well (Jin et al., 2020). Comparing such methods to linguistic structure is an exciting avenue that we leave open to future work. Other interaction methods that were not considered include two methods that preceded Archipelago: Neural Interaction Detection (Tsang et al., 2018a) and MAHE (Tsang et al., 2018b). The feature attribution method Contextual Decomposition (Murdoch et al., 2018; Jumelet et al., 2019) has been extended to extract interactions as well (Singh et al., 2019; Saphra and Lopez, 2020; Chen et al., 2020a), but these methods place the constraint that only contiguous groups of features can interact. Integrated Directional Gradients (Sikdar et al., 2021), an extension of Integrated Gradients to capture *group attributions*, could be adapted to our framework, but we leave this open for future work.

## 7.5 Evaluating FIDAMs

The final component of our framework is a methodology for evaluating the faithfulness of FIDAMs. To lay a robust foundation for such work, we propose to evaluate a range of interaction methods and baselines on smaller deep learning models (using LSTM and Transformer architectures) that have been trained to recognise formal languages, based on a probabilistic context-free grammar (PCFG).

Our models are trained on a binary language classification task, in which a model needs to learn to discern between well-formed strings and minimally corrupted counterparts. Models are trained to perfection (100% accuracy) on both train and test set. To obtain perfect performance, a model must rely solely on the grammatical rules that underlie the language, without resorting to spurious heuristics, because only these results allow completely solving the task. This way, due to the controlled nature of the task, we obtain a high degree of confidence about the model's behaviour.

The goal of our experimental approach is to recover the structure of the language *based on the trained model itself*. This is achieved by the FIDAMs outlined in §7.4. We aim to uncover whether a structural dependency between two features results in a high interaction effect. Since our models have been trained to perfection, this allows us to employ our setup as a way of measuring the **faithfulness** of a FIDAM. A method that assigns a high interaction effect to features that contain a dependency in the original grammar is able to provide a faithful reflection of a model's understanding of the task. By testing a wide range of FIDAMs and baselines we can uncover which configuration yields the most faithful explanations. A graphical overview of our approach is depicted in Figure 7.1.

**Task** The binary language classification task is set up by generating positive examples $D^+$, based on some PCFG, and negative examples $D^-$, derived from minimally corrupting the positive examples. We split the union of these two sets into a random train/test split of 80/20%. We train our models with a default cross-entropy loss, using the AdamW optimiser (Loshchilov and Hutter, 2019), a learning rate of 0.01, and a batch size of 48.

**Models** Our pipeline permits the use of any kind of neural model architecture, in our experiments we considered both LSTMs (Hochreiter and Schmidhuber, 1997) and Transformers (Vaswani et al., 2017). In our experiments we report the results of the LSTM model, but we observed similar results for Transformers: due to the black-box approach of our explanation procedure the architecture itself is not of great importance. The models are deliberately small: we use an embedding size that is equal to the number of symbols in the language it is trained on, a hidden state size of 20, and a single layer. This results in models that provide a compute-friendly test bed for evaluating the FIDAMs.

**Evaluation** We focus on *pairwise* interactions: interactions between individual pairs of features. A FIDAM that extracts pairwise interactions for an input sequence $\mathbf{x} \in \mathbb{R}^N$ returns a matrix of interaction effects $\Gamma \in \mathbb{R}^{N \times N}$. Since our goal is to uncover whether structural dependencies result in high interaction effects, we approach the evaluation of the interaction matrix as a retrieval task. By aggregating and normalising the *rank* of each interaction of interest we can quantify the performance of a FIDAM. We call this metric the **Average Relative Rank** (ARR):

$$ARR(\Gamma, \mathcal{I}) = \frac{1}{|\mathcal{I}|} \sum_{i,j \in I} \frac{R(\Gamma_i)_j}{N-1} \qquad [7.10]$$

where $\mathcal{I}$ denotes the set of interaction pairs of interest and $R(\Gamma_i)$ denotes the rank of each interaction between feature $i$ and the other features in input $\mathbf{x}$ (the lowest interaction is ranked 0, and the highest interaction is ranked $N-1$). We aggregate these scores over an evaluation set to obtain a general performance score of the FIDAM. A graphical overview of this procedure is provided in Figure 7.2.

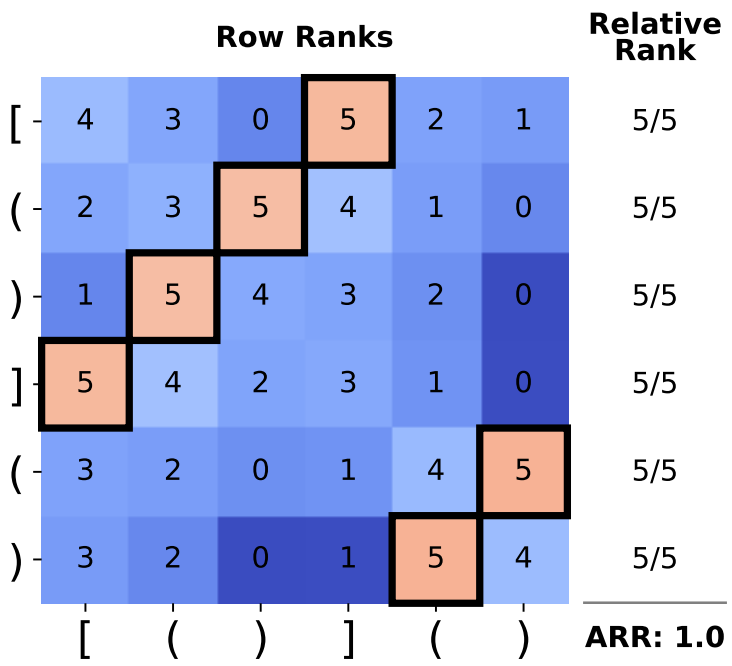

**Figure 7.2:** Example for the computation of the Average Relative Rank metric. For each row we compute the relative rank of the interaction of interest (here the Dyck language), and these row-wise relative ranks are averaged into a single score between 0 and 1. A random interaction matrix results in an ARR of around 0.5.

**Baselines** We consider a range of baselines in our experiments, based on the procedures explained in §7.3.1. For the static baselines we consider a zero-valued baseline ($\mathbf{x}' = 0$), and a baseline that uses a deterministic mapping $T$ based on the original input symbols ($\mathbf{x}' = T(\mathbf{x})$). Expected attributions are marginalised over samples from the distribution of well-formed strings $D^+$ and corrupted strings $D^-$. The interventional conditional expectation (Eq. 7.4) is computed with a corpus-wide unigram distribution ($P(x_i)$), a unigram distribution that is conditioned on the sentence position ($P(x_i|i)$), and as a joint distribution over the missing features ($P(\mathbf{x}'_{\setminus S})$), that we sample from the training corpus. The observational conditional expectation (Eq. 7.3) is computed based on the original corpus data.[1]

## 7.6 Experiments on Formal Languages

We apply the evaluation procedure of §7.5 to two formal languages: the Identity Rule language and the Dyck-2 language. In the appendix (§A.5.1) we also present results on a palindrome language.

### 7.6.1 Identity Rule

The first language we consider is a regular language consisting of strings in which the first two symbols are identical, followed by a random sequence of symbols. The language is

[1] Due to the small scale of the PCFGs considered here we can generated the complete language up to a certain length, and sample from strings that have feature overlap with the features that are still present in the partial input. For more complex tasks an auxiliary LM can be used instead.

| | *NB* | 0 | $\mathbf{x}' \sim D^+$ | $\mathbf{x}' \sim D^-$ |
|---|---|---|---|---|
| Group Ablation | – | 0.49 | **1.00** | 0.53 |
| Archipelago | – | 0.30 | 0.24 | **1.00** |
| SII | – | 0.70 | **1.00** | **1.00** |
| STII | – | **0.83** | **1.00** | **1.00** |
| Hessian | **0.93** | – | – | – |
| Hessian×Input | 0.66 | – | – | – |
| IH | – | 0.81 | **1.00** | 0.31 |

**Table 7.1:** Average Relative Rank for the Identity Rule language, columns indicate different baseline procedures. An average rank of 1 indicates that the method (correctly) assigned the interaction between the first two tokens the highest score. *NB* indicates these methods use no baseline.

formed by the following grammar:

$$\mathsf{S} \rightarrow x\ x\ \mathsf{A} \qquad x \in \{a, b, c\}$$
$$\mathsf{A} \rightarrow x\ \mathsf{A} \mid \varepsilon \qquad x \in \{a, b, c\}$$

The only interaction of interest here is between the first two symbols; all subsequent symbols are irrelevant for the prediction. An ARR score of 1.0 then indicates that for all corpus items the interaction between the first two items was the strongest out of all interactions.

We use a corpus size of 1,000 items, a maximum sequence length of 20, with 3 different input symbols. Corrupted strings are derived by altering one of the first two symbols (e.g. $\boldsymbol{aa}bcb \rightarrow \boldsymbol{ca}bcb$).

**Results** The results for an LSTM that was trained on the language are shown in Table 7.1. Due to the simplicity of the language and for brevity we only report results on three baselines. A static, zero-valued baseline provides imperfect interactions for all methods. The Hessian, that does not depend on any baseline, performs better than all other methods with that baseline. When sampling the baseline, however, multiple methods perfectly retrieve the interaction between the first two symbols for all corpus items. Interestingly, Group Ablation and IH benefit from sampling from the distribution of well-formed items, whereas Archipelago performs best when sampling from the distribution of corrupted items.

### 7.6.2 Dyck-2

The Dyck language is the language of well-nested brackets, and is a popular testbed for research on formal languages. It is a context-free language with center embedding clauses, requiring a model to keep track of a memory stack while processing a string. Earlier work on Dyck languages has shown that a wide range of neural model architectures can learn the grammar, including LSTMs (Sennhauser and Berwick, 2018), memory augmented RNNs (Suzgun et al., 2019), Transformers (Ebrahimi et al., 2020), and handcrafted RNNs (Hewitt et al., 2020; Hao, 2020). We consider the Dyck-2 language, consisting of two types of brackets. The language is formed by the following grammar:

$$\texttt{S} \rightarrow \texttt{[ S ]} \mid \texttt{( S )} \mid \texttt{S S} \mid \varepsilon$$

| | | *Static* | | *Expected* | | *Interventional* | | | *Observational* |
|---|---|---|---|---|---|---|---|---|---|
| | *No baseline* | 0 | $T(\mathbf{x})$ | $D^+$ | $D^-$ | $P(x'_i)$ | $P(x'_i \mid i)$ | $P(\mathbf{x}'_{\setminus S})$ | $P(\mathbf{x}'_{\setminus S} \mid \mathbf{x}_S)$ |
| Group Ablation | – | **0.684** | **1.000** | 0.916 | 0.884 | 0.822 | 0.821 | 0.938 | 0.956 |
| Archipelago | – | 0.466 | 0.528 | 0.250 | 0.554 | – | – | – | – |
| SII | – | 0.555 | **1.000** | **0.921** | **0.895** | 0.876 | 0.885 | 0.923 | 0.989 |
| STII | – | 0.583 | 0.999 | 0.876 | 0.820 | **0.881** | **0.906** | **0.952** | **0.991** |
| Hessian | 0.413 | – | – | – | – | – | – | – | – |
| Hessian×Input | **0.542** | – | – | – | – | – | – | – | – |
| IH | – | 0.591 | 0.837 | 0.723 | 0.665 | – | – | – | – |

**Table 7.2:** Average Relative Ranks for the Dyck language (higher indicates stronger alignment with Dyck rules), columns indicate different baseline procedures.

We use a corpus size of 15.000, a maximum sequence length of 20, and a maximum branching depth of 4. We use the same branching probabilities as Suzgun et al. (2019), which results in a uniform probability of 0.25 for each rule. Corrupted strings are derived by flipping a single bracket to any other bracket. For the baseline mapping $T(\mathbf{x})$, we map a bracket to the other bracket type, i.e. '(' $\leftrightarrow$ '[' and ')' $\leftrightarrow$ ']'. This results in a baseline that is of the same structure as the original input, but without feature overlap.

**Results** We report the results for this language in Table 7.2, computed over all our baselines for an LSTM. The zero-valued baseline again turns out to be a mediocre baseline: for none of the methods this results in a high ARR score. The method that performs best is the fixed mapping $T(\mathbf{x})$. For Group Ablation, SII, and STII this results in a perfect ARR; for IH it is the best performing baseline.

It is encouraging that a baseline exists that results in perfect ARR scores, but this mapping depends strongly on the nature of the Dyck task itself. It is, for example, unclear how this static mapping would transfer to the natural language domain. Ideally, a more general solution makes no strong assumptions about the baseline itself. The three other baseline types in Table 7.2 may provide such a solution, as these only depend on the access to the original training data. Out of these, the observational baseline performs best: for the SII and STII methods this baseline performs nearly on par with the static mapping. Obtaining this conditional distribution is challenging for more complex tasks, and it can be seen here that the interventional baseline with a joint distribution over the missing features performs well too.

## 7.7 A Natural Language Case Study: CoLA

As a case study on a larger scale natural language task, we apply our methodology to language models fine-tuned on the CoLA task (Warstadt et al., 2019b). CoLA is part of the GLUE Benchmark (Wang et al., 2019), and is defined as a binary classification task of determining the linguistic acceptability of a single input sentence. The task consists of linguistically valid sentences, and sentences that contain either a syntactic, semantic, or morphological violation. A model that performs well on this task must have a thorough grasp of grammatical structure, and as such it provides a useful test bed for our FIDAM evaluation procedure.

In the previous experiments there was a degree of certainty about the structure that must be encoded by the model. In the natural language domain, however, we do not have such certainty, and should therefore be careful of making strong claims about faithfulness. Furthermore, natural language is highly multi-faceted and can not be captured by a single hierarchical structure that covers all these facets. Nonetheless, we consider it valuable to test our setup on a natural domain in order to see if interesting differences between FIDAMs arise, and whether particular facets of language such as syntactic dependency structure can be extracted.

### 7.7.1 Experimental Setup

For our experiment we consider the RoBERTa-base model (Liu et al., 2019) which obtains a Matthew's Correlation Coefficient score of 69.70 on the in-domain validation split. We filter out sentences that contain words that are split into multiple subwords by the tokenizer, since this leads to issues with aligning the interactions of multiple subwords to the dependency graph that is used for evaluation. Furthermore, we limit sentences to a max length of 14 in order to allow the STII and SII methods to be computed exactly without approximations. This resulted in a subset of around 60% of the original in-domain validation split that we will use in our experiment.

We evaluate the FIDAM scores on the dependency parse tree of the sentence, that we obtain with the parser of `spaCy` (Honnibal et al., 2020). The ARR score is computed based on the interaction of each token with its *parent* token. We omit the interaction of the token that has the ROOT node as its parent. An example of this procedure can be found in Appendix A.5.2. Do note that our evaluation procedure is one of many possibilities: we make the assumption that a token should interact strongly with its parent, but other interactions are likely to play a role within the model as well. We leave a more detailed investigation into using different types of linguistic structure open for future work.

We again consider the FIDAMs of Group Ablation, STII/SII, and Integrated Hessians. We leave out Archipelago, since its procedure of assigning features to a single interaction set is not feasible with our setup in which multiple child tokens might be interacting with the same parent token. Due to computational constraints we were unable to compute the full Hessian matrix of the language model, whose computation scales quadratically in the number of input *neurons* (Bishop, 2007, §5.4). For the static baselines we again consider the zero-valued baseline, as well as the `<pad>` token. The interventional baselines are obtained by computing simple count-based distributions over a sample of 100.000 sentences from the Google Books corpus. The distributions are based on the tokenization of the model's tokenizer, and allow for computationally efficient sampling. We leave the incorporation of an observational baseline for future work, where an auxiliary masked LM might provide a useful conditional probability distribution.

### 7.7.2 Results

The results for the experiment are shown in Table 7.3. As expected, due to reasons outlined at the start of this section, none of the methods reaches ARR scores that are close to 1. Nonetheless, it is encouraging to see that various method/baseline combinations attain ARR scores that are far above chance level, indicating that there exists a strong degree of alignment between feature interactions and dependency structure. Contrary to the Dyck results, using a zero-valued baseline yields some of the highest ARR scores, which indicates that within RoBERTa's embedding space this baseline represents a better neutral value.

A closer inspection of these results shows that the ARR scores are strongly negatively correlated to sentence length: for Group Ablation with a `<pad>` baseline, for example, we obtain a Spearman correlation of -0.38 ($p << 0.001$, regression plot in Appendix A.5.3). This is not surprising: as the sentence length increases, the chance of a token's largest interaction being with its parent decreases. Another correlation of interest is between the ARR score and the model's prediction of a sentence's acceptability. A high correlation would indicate that the FIDAM's alignment with dependency structure are indicative of a model's performance. For this we obtain a Spearman correlation of 0.14 ($p = 0.036$): a relatively weak result that indicates that the structure our FIDAM extracted is only partly driving the model's comprehension of the sentence structure.

| | *Static* | | *Interventional* | |
|---|---|---|---|---|
| | 0 | `<pad>` | $P(x'_i)$ | $P(\mathbf{x}'_{\setminus S})$ |
| Group Ablation | 0.702 | **0.757** | 0.518 | 0.491 |
| SII | **0.746** | 0.668 | **0.714** | **0.696** |
| STII | 0.741 | 0.708 | 0.704 | 0.658 |
| IH | 0.577 | 0.516 | – | – |

**Table 7.3:** Average Relative Ranks for the dependency tree recovery of RoBERTa fine-tuned on CoLA.

## 7.8 Discussion & Conclusions

In this paper, we have presented a framework for characterising FIDAMs and evaluating their faithfulness. For the characterisation we set out two dimensions, feature removal and feature influence, along which existing FIDAMs can be characterised, by extending the 'Explaining by Removing' framework of Covert et al. to also apply to FIDAMs. This allows us to place each of the known FIDAMs in a two-dimensional grid, and to define novel variants of these models. As such, many of the methods that we incorporated in our experiments are novel FIDAMs, such as combining Archipelago with expected explanations and STII with an observational baseline.

To assess the faithfulness of FIDAMs, we made use of formal language theory and 'grey box models'. We use formal grammars to generate multiple datasets, each with known feature interactions, and train deep learning models to perfection on those datasets. Using FIDAMs, we can then extract the learned feature interactions based on the model itself, and compare these interactions to the dependencies in the original grammar. We demonstrate that only specific combinations of FIDAMs and baselines are able to retrieve the correct interactions, while methods such as Archipelago and Integrated Hessians consistently fail to do so.

Finally, we tested our methodology on a natural language case study using a model fine-tuned on the CoLA task for linguistic acceptability. Our results on the formal language tasks either did not turn out to be predictive of this experiment or, alternatively, the results *were* predictive but the LMs made less use of dependency graph information than we might have expected. This illustrates the challenge of the Attribution Generalisation problem, and the open question remains how we can transfer faithfulness guarantees from a synthetic, controlled context to the domain of natural language and LLMs.

We do show, however, that under certain configurations feature interactions align to some degree with the (syntactic) dependency structure of a sentence. This paves the way for revealing linguistic structure in a more direct way than, for instance, can be achieved with Structural Probes (Hewitt and Manning, 2019). Investigating whether different methods and baseline configurations are able to retrieve different aspects of structure is an exciting next step that we look forward to exploring in more detail. This could be examined, for instance, through the lens of contrastive explanations Yin and Neubig (2022), a procedure that demonstrates that different baselines can reveal different aspects of linguistic structure. Furthermore, investigating the role that attention plays in modelling interactions could be a fruitful line of work, for instance by incorporating *context mixing* methods to our pipeline, such as *Value Zeroing* (Mohebbi et al., 2023) and *ALTI* (Ferrando et al., 2022).

*Am I the only one around here who gives a shit about the* ***rules****?*

Walter Sobchak

# 8

# Transparency at the Source: Evaluating and Interpreting Language Models With Access to the True Distribution

One of the goals in Chapter 7 was to determine the faithfulness of explanation methods based on their performance on 'grey-box models', for which we have a better understanding of their inner workings. It remains challenging, however, to determine whether the performance of a model on a grey-box model generalises faithfully to large-scale black-box models: performance on a toy domain does not necessarily transfer to the natural language domain. With these challenges in mind we introduce in this chapter a methodology in which we still have full control over the distribution of the training data, but at a scale that is much closer to natural language.

## 8.1 Introduction

When we train a Language Model on large natural language corpora, we are in effect estimating a probability distribution over possible next tokens or masked tokens. The true distribution is unknown, so we cannot directly quantitatively measure how good our estimation is, or qualitatively assess whether our LM has discovered 'true' underlying patterns in the data. The best we can do is to measure perplexity on a new sample from the same unknown distribution, and compare that perplexity to the perplexity we obtain with other

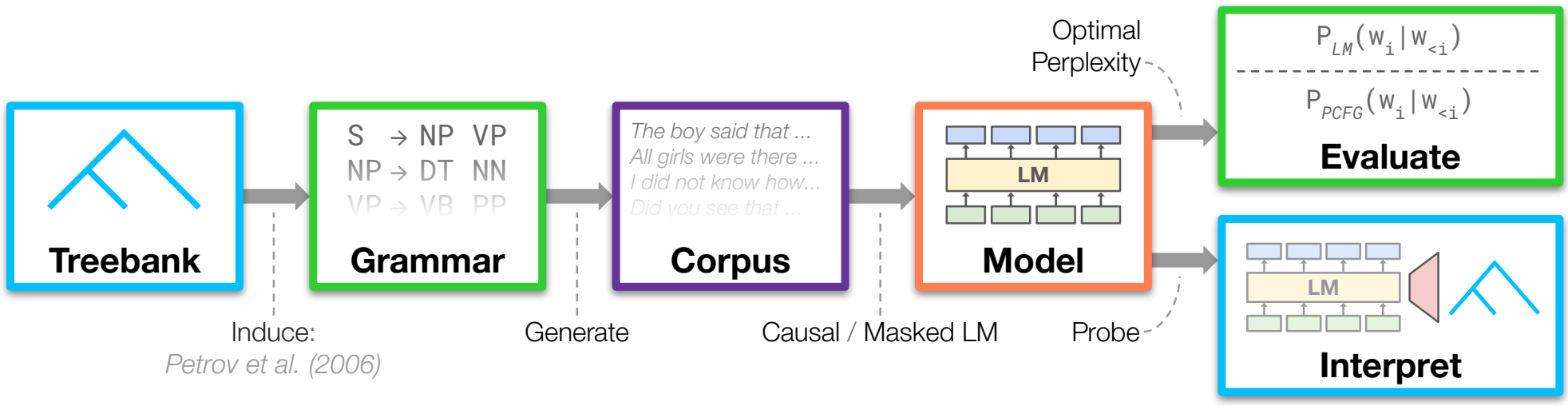

**Figure 8.1:** Conceptual overview of our experimental pipeline. First, we induce a massive probabilistic grammar from a natural language treebank. From this grammar we generate a corpus, that is used to train various language models on. Then, with access to the true distribution of our grammar, we evaluate and interpret our models.

8

model architectures or training regimes (Brown et al., 1992), or with the expected perplexity given a (compute-optimal) scaling law (Kaplan et al., 2020; Hoffmann et al., 2022).

This approach has been enormously successful, but it leaves a number of interesting questions unanswered. First, one consequence of the absence of an explicit stochastic source is that it is impossible to exactly determine the *optimal perplexity* that a language model can obtain. Second, when designing interpretability methods to assess whether an LM has learned specific linguistic rules, we lack a *gold standard*. If an interpretability method fails to find evidence that the LM has learned a specific linguistic construction we cannot be sure whether this is a failure of the LM, a failure of the interpretability method, or whether our assumption that knowledge of this linguistic construction as an essential component of English fluency is wrong.

One approach to address these problems is to move to artificially generated data, but such data is often of trivial complexity compared to the richness of natural language, in terms of the vocabulary size or the number of grammatical rules.

In this paper, we explore an alternative approach. We also use generated data to train language models on, but to make sure the data approximates natural language in complexity, we use a *massive probabilistic grammar*, that is itself derived from a large natural language corpus. To obtain the grammar, we use an automatically parsed section from the The Pile corpus (Gao et al., 2020), and use the state-split framework (Petrov et al., 2006) – one of the most successful statistical parsing frameworks from before the rise of deep learning – to obtain a statistical grammar with more than 2 million rules. In §8.3.1 we describe the procedure for obtaining this grammar in detail.

This setup allows us to compute the exact lower bound on perplexity, although that computation turns out to still be nontrivial. This is due to the computational complexity of the problem that makes a naive approach infeasible, even on modern hardware. One key contribution from this paper is a closed-form expression to efficiently compute masked token probabilities for PCFGs, complementing the classic closed form for causal language modelling (Stolcke, 1995). Furthermore, our setup provides a gold standard to qualitatively assess results from interpretability methods against. We conduct a wide range of experiments, showing the naturalness of our generated data, determining the impact of model and data properties on language model performance, and interpreting model behaviour and learning dynamics using various interpretability tools. Figure 8.1 presents a schematic overview of our methodology.

## 8.2 LMs and Synthetic Data

Various approaches have controlled the structure of the training data to investigate particular properties of neural models or the learning process itself. A major line of work in this direction does this to investigate the impact of typological features of a language on model performance. For example, Cotterell et al. (2018) and Mielke et al. (2019) train LMs on parallel, translated corpora, which allows them to isolate the impact of typological differences between languages. Ravfogel et al. (2019) investigate typological differences as well, but create synthetic differences by modifying English sentences on typological features such as subject-object-verb order, an approach that has been pursued earlier by Wang and Eisner (2016). In all these papers the starting point of the training data is an (English) language corpus of which the underlying generative process is unknown.

There is a small set of papers where all data is derived from artificial languages, that are under full control. Dankers et al. (2022) generate naturalistic data for MT evaluation, but for models that were trained on natural data. Papadimitriou and Jurafsky (2023) investigate the inductive biases of LMs by pretraining models on data derived from simple grammars that isolate abstract linguistic properties such as nested recursion and Zipfian power laws. White and Cotterell (2021) provide a setup using simple PCFGs, in which typological features can be controlled. Hopkins (2022) observe that the simplistic nature of the PCFGs used in their work may not be reflective enough of natural language, and proposes Pitman-Yor processes to obtain artificial language that are more natural language-like. The grammars in these papers, however, remain fairly simple, and vocabulary size, sentence length and other quantities stay limited compared to those seen in natural languages. Our setup provides a massive PCFG that is derived from natural language, with an efficient procedure for comparing the grammar's distributions to those of trained LMs.

## 8.3 PCFGs and Perplexity

In our work we make use of large-scale PCFGs to generate data on which we train various language models. In this section we give a brief background on PCFGs and how to evaluate modelling performance on PCFG-generated data using perplexity.

### 8.3.1 State-split PCFGs

There is a rich history in NLP of modelling language using interpretable, hierarchical grammatical formalisms. One of the most common formalisms is the Probabilistic Context-Free Grammar (PCFG), which provided the backbone for various successful parsing methods (Klein and Manning, 2003; Collins, 2003; Charniak and Johnson, 2005). A PCFG consists of simple weighted rewriting rules and can be used for both parsing and generation, by sampling and expanding rules based on the distribution described by the PCFG.

The context-freeness assumption of PCFGs, however, makes it challenging to capture non-syntactic dependencies between phrases.[1] One successful approach that addresses this is the **state splitting** procedure of Petrov et al. (2006), colloquially known as the '*Berkeley Parser*'. The general idea of state splitting is that we can split the non-terminal symbols in a PCFG into fine-grained subsymbols, that are specialised towards specific linguistic phenomena using automatic learning. The splitting process is repeated multiple times, in an alternating split-and-merge strategy: if the splitting of a symbol does not result in a considerable improvement of the data likelihood the subsymbols are merged back. This results in an optimal trade-off between fine-grained non-terminal categories that still generalise beyond the training data. Furthermore, by keeping track of the history of split states we can project the final, fine-grained grammar back to the simpler, coarser base grammar. This then allows for highly efficient *coarse-to-fine* parsing, making it possible to parse sentences with large-scale grammars which would be intractable with a normal PCFG (Petrov and Klein, 2007).

### 8.3.2 Perplexity Lower Bound

The performance of a language model is often evaluated in terms of its **perplexity** on a test set (Jelinek et al., 1977; Jurafsky and Martin, 2024). Perplexity is defined as the normalised

[1] For example, a simple PCFG may encode that *dog* is a noun and *bark* a verb, but from a single `S→NP VP` rule it can not be deduced that *dogs* are more likely to *bark* than *cats*.

inverse probability of test set $\mathcal{D}$:

$$
\begin{aligned}
PPL(\mathcal{D}) &= P(\mathcal{D})^{\frac{1}{|\mathcal{D}|}} \\
&= \exp\left(\frac{\ln P(\mathcal{D})}{|\mathcal{D}|}\right)
\end{aligned}
$$

Since the 'true' distribution over natural language is unknown, there is no way of determining the optimal perplexity that is attainable by a model.[2] In our setup, however, we *do* have access to the generative process behind our data. It therefore becomes possible to compute an exact lower bound for the perplexity, allowing us to express exactly how accurately a LM modelled the distribution set out by the PCFG.

The perplexity lower bound depends on the language modelling objective of interest: for masked language modelling it will be different than for causal language modelling. For both objectives we need to define a way to extract the required probability distributions out of the PCFG. We describe how this is possible in the next two sections.

### Causal Language Modelling

When doing causal language modelling, we aim to model the distribution $P(w_i|w_{<i})$: the probability of token $w_i$ conditioned on its prior context. The perplexity of a sentence can directly be decomposed into a product of such conditional probabilities, by continuously expanding the joint distribution over the entire string:

$$PPL(\mathbf{w}) = \left(\prod_i P(w_i|w_{<i})\right)^{\frac{1}{|\mathbf{w}|}}$$

Extracting a token distribution from a PCFG $G$, $P_G(w_i|w_{<i})$, is a topic that has been explored in detail since the 1990s (Jelinek and Lafferty, 1991; Hale, 2001). An efficient procedure is that of Stolcke (1995), who adapts the Earley parsing algorithm to compute the probabilities of prefix strings $w_{1i}$. The conditional token probability can then easily be recovered as the fraction of two subsequent prefix strings: $P_G(w_i|w_{<i}) = P_G(w_{1i})/P_G(w_{1(i-1)})$. In our experiments we make use of the implementation of Stolcke's procedure by Luong et al. (2013), called `EarleyX`.

[2] Shannon (1951) famously provides an approximation of the perplexity of the English language on a character level, but this does not yield an exact lower bound.

### Masked Language Modelling

When masked language modelling, we aim to approximate the probability distribution of $P(w_i|w_{\backslash i})$, which is achieved by masking out the token at position $i$ with a special mask token. Unlike in causal language modelling, a sentence's probability can not be decomposed into masked token probabilities. To aggregate the performance of a masked LM, we therefore need to make use of the pseudo-log-likelihood (*PLL*) (Salazar et al., 2020). The pseudo-log-likelihood expresses the log probability of a sentence as a sum of masked token probabilities:

$$\psi\text{-}LL(\mathbf{w}) = \sum_i \ln P(w_i|w_{\backslash i})$$

We can then use the *PLL* to compute the *pseudo-perplexity* as follows:

$$\psi\text{-}PPL(\mathbf{w}) = \exp\left(\frac{\psi\text{-}LL(\mathbf{w})}{|\mathbf{w}|}\right)$$

To compute this quantity for a PCFG we need to find a way to extract masked token probabilities from the grammar: $P_G(w_i|w_{\backslash i})$. Since such a procedure does not yet exist, we propose a novel method.[3] Fortunately, it turns out to be possible to do this efficiently by employing the **inside-outside** algorithm (Baker, 1979; Manning and Schütze, 2001).

**Inside-Outside** The algorithm defines two probability distributions. The **inside** probabilities $\beta$ define the probability of a substring $w_{pq}$, given that the substring has non-terminal $N^j$ as its parent:

$$\beta_j(p,q) = P_G(w_{pq}|N^j_{pq}) \qquad [8.1]$$

The **outside** probabilities $\alpha$ define the joint probability of generating the non-terminal $N^j$ that spans position $p$ to $q$, as well as all the words outside the substring $w_{pq}$:

$$\alpha_j(p,q) = P_G(w_{1(p-1)}, N^j_{pq}, w_{(q+1)m}) \qquad [8.2]$$

Using the in- and outside probabilities, we can directly express the masked token probability in a closed-form as follows:

$$P_G(w_i|w_{\backslash i}) = \sum_j \beta_j(i,i) \cdot \frac{\alpha_j(i,i)}{\sum_k \alpha_k(i,i)} \qquad [8.3]$$

[3] Concurrent to our work, Zhao et al. (2023) find a similar formulation for expressing masked token PCFG probabilities.

We provide a detailed derivation in Appendix A.6.1. Our formulation allows for a highly efficient procedure of extracting these probabilities from a PCFG, since we can make use of the optimised coarse-to-fine parsing methodology of Petrov and Klein (2007) to obtain the in- and outside probabilities.

## 8.4 PCFG Corpus

Our approach can roughly be divided into three parts (Figure 8.1): i) **data generation** using state-split PCFGs, ii) **language modelling** using Transformer LMs, and iii) **model evaluation**, via probing-based interpretability tasks with a focus on *learning dynamics*. In this section we explain the data generation procedure, and do an analysis of our generated corpus.

### 8.4.1 Data Generation

**Grammar** To generate data we make use of the state-split PCFG procedure of Petrov et al. (2006)[4], as explained in §8.3.1. The treebank we use for learning a PCFG is a parsed version of the BookCorpus (Zhu et al., 2015; Conneau et al., 2018), which was parsed using the Stanford Parser (Manning et al., 2014), and is nowadays part of The Pile corpus under MIT License (Gao et al., 2020). We learn the PCFG on a sample of 500,000 sentences, which is the maximum corpus size we could learn on our system with 32GB RAM memory. Learning is done for 5 split-merge cycles, filtering rules with a probability below $10^{-8}$. Part of the learning procedure is transforming the base grammar to a X-bar style grammar, which ensures that each rule has at most two children.

The resulting PCFG consists of 54,497 unique tokens and 2.5M rules: 2.22M terminal rules, 272k binary non-terminal rules, and 13k unary non-terminal rules. The original treebank contained 96 non-terminal symbols: after split-merging we obtain 718 split non-terminals. We plot a fine-grained plot of the number of times each of the original 96 non-terminals has been split in Appendix A.6.2.

**Data** Having learned the PCFG from a treebank, we can sample from it to generate our training corpus data. For training our models we generate 7.5 million sentences (97 million tokens), and development/test/evaluation corpora of 50k/50k/10k sentences, respectively. Sentence lengths are capped between 6 and 25, and duplicate sentences are allowed (but with no overlapping sentences between the train/development/test corpora): if we would not allow this, the data distribution that the model has access to would not reflect the original PCFG distribution. We provide a more detailed analysis of the final grammar and corpora in the following section, and a sample of sentences in Appendix A.6.3.

[4] https://github.com/slavpetrov/berkeleyparser

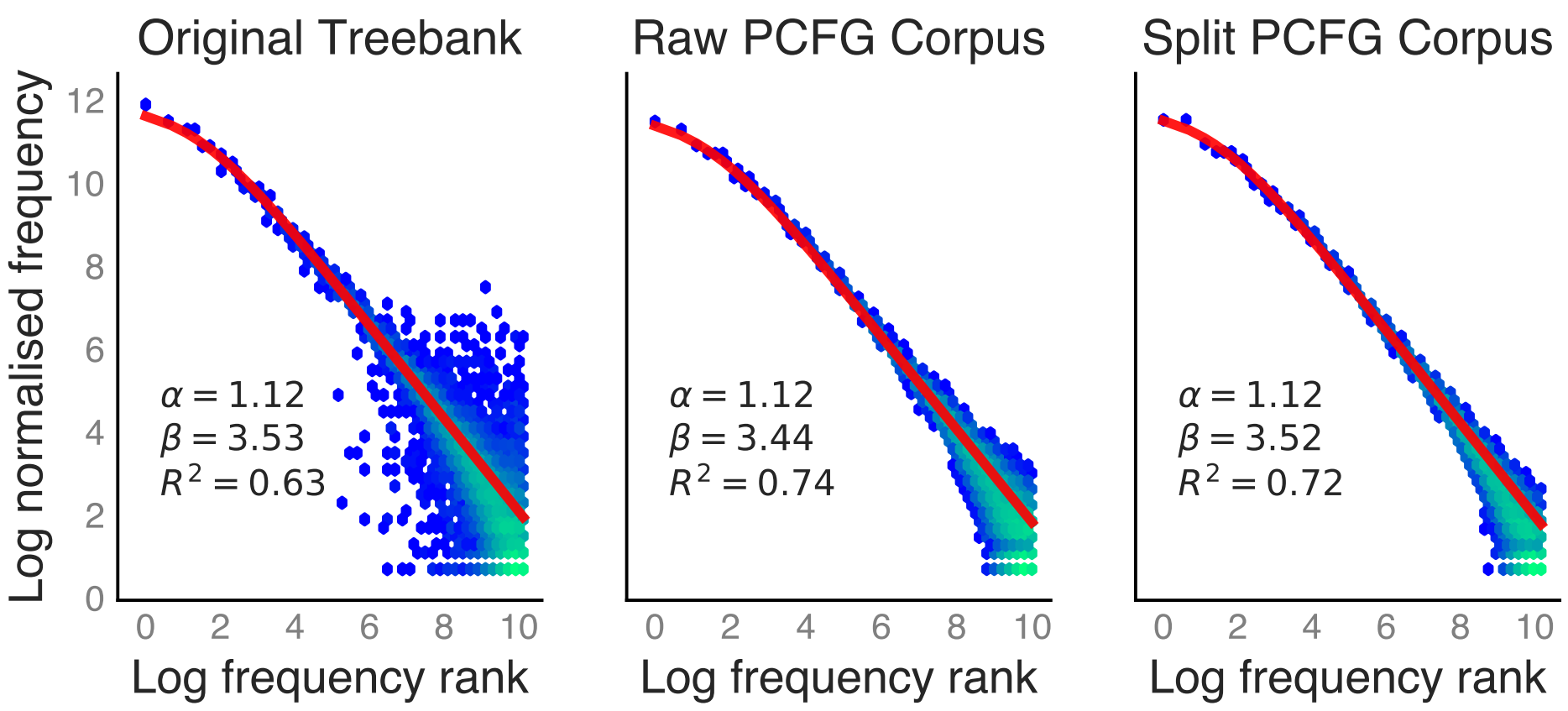

**Figure 8.2:** Relationship between a token's frequency rank and its frequency, which is near linear on a log-log plot. Rank and frequency are computed over two disjoint splits of the corpora, following the procedure of Piantadosi (2014).

### 8.4.2 Naturalness of PCFG Corpus

In this subsection we briefly investigate whether the corpus generated from our grammar follows similar patterns as natural language. We compare the properties of our fine-grained PCFG to both the original Treebank, and a simple count-based PCFG that is derived directly from the original Treebank distribution. We refer to these three sources as *Split PCFG*, *Treebank*, and *Raw PCFG*, respectively.

**Zipf's Law** One important linguistic feature that has been studied in detail within quantitative linguistics is that the frequency of a word is logarithmically proportional to its frequency rank. This feature is referred to as *Zipf's law*, and has been shown to be a universal language feature (Zipf, 1936). Zipf's law, in the updated form of Mandelbrot (1953), states that the $r^{\text{th}}$ most common word in a corpus has a frequency that is proportional to

$$f(r) \propto \frac{1}{(r+\beta)^\alpha} \qquad [8.4]$$

We can investigate this feature for our generated data, and compare this against the frequency distribution of the original treebank. We follow the procedure of Piantadosi (2014), who argues that frequency and rank should not be computed on the same corpus, since that way a token's rank is always spuriously correlated to its frequency. Instead, we can compute the rank and frequency on two independent corpora, and investigate the *Zipfian* relationship based on that. We estimate the $\alpha$ and $\beta$ parameters of Eq. 8.4 using the MLE implementation of Vogelmann (2020).

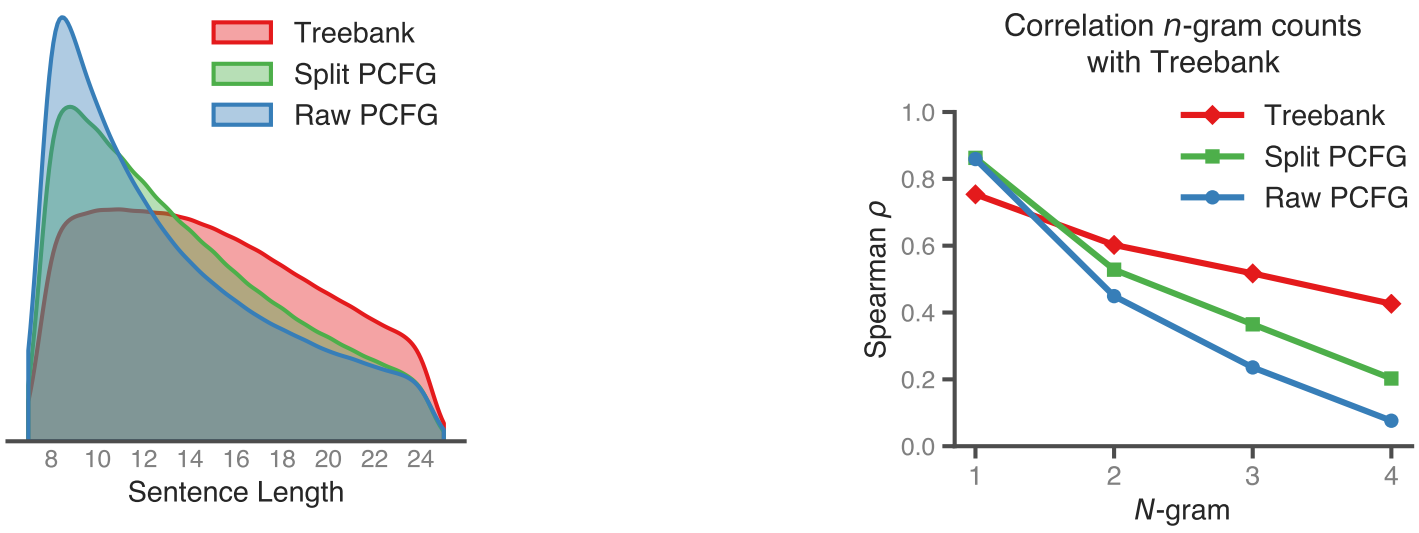

**Figure 8.3:** Distribution over sentence length for the three corpora, and the correlation of $n$-gram distributions with respect to the original Treebank data.

We plot our results in Figure 8.2. It can be seen that all three corpora follow Zipf's law with almost equal parameterisation of $\alpha$ and $\beta$. Interestingly, the generated corpora yield considerably lower residuals compared to the original treebank data. This demonstrates that in a natural language for infrequent tokens there exists greater uncertainty between the relation of frequency rank and frequency. Nevertheless, it is encouraging to see that our generated corpus follows a similar Zipfian distribution, likely an important property for the learnability of language (Kurumada et al., 2013; Hendrickson and Perfors, 2019).

**Sentence Length** We investigate the distribution over sentence lengths in the corpora, and plot the histograms in Figure 8.3a. Both the PCFG corpora are skewed towards shorter sentences, whereas the original Treebank data is more uniformly spread out. The *Split PCFG* is more similar to the Treebank distribution than the *Raw PCFG*. We could opt to to subsample from our *Split PCFG* corpus to obtain a similar sentence length distribution as the Treebank corpus, but in that case the corpus distribution would no longer align directly with the PCFG itself.

***N*-grams** One aspect of natural language that state-split PCFGs aim to cover is that dependencies between words are often semantically correlated, as explained in §8.3.1. This phenomenon is known as *selectional preference*, and has been used by Hopkins (2022) to investigate the naturalness of their training data. There are many ways that selectional pref-erence can be investigated, but we focus on a simple heuristic here. For increasing $n$ we compute the Spearman correlation between a corpus' frequency distribution over $n$-grams and the distribution of the original Treebank data. A high correlation then indicates that the corpus contains $n$-grams of a similar (natural) structure, and a low correlation indicates that the original $n$-gram dependencies are less present in the data. We compute this also for the Treebank itself, by splitting the corpus in half.

The results are shown in Figure 8.3b. For $n = 1$, the PCFG corpora both obtain a correlation that is higher than the Treebank with itself. This may be related to the large residual that we saw in the Zipf distribution of Figure 8.2, which showed that the unigram distribution of the Treebank corpus contains more variance in the tail than the PCFG corpora. For $n > 1$, it can be seen that the *Split PCFG* obtains higher Spearman correlation than the *Raw PCFG*, which demonstrates it has improved on the modelling of selectional preference.

8

**Other Frameworks** It is encouraging to see that our generated data generally follows similar patterns as the original natural data. However, we would like to stress that our usage of state-split PCFGs is certainly not the only option to obtain a generative process that models natural language well. Other frameworks that could be explored in future work include non-parametric Bayesian methods (Teh, 2006; O'Donnell, 2015), data-oriented parsing (Bod et al., 2003; van Cranenburgh et al., 2016), or $n$-gram based models (Pauls and Klein, 2012).

## 8.5 Language Modelling

We now move on to our experiments on language modelling the PCFG corpus that has been defined in the previous section. The underlying PCFG of the data generation process allows us to place the LM's performance against an optimal baseline.

### 8.5.1 Metrics

We compute the perplexity scores on the evaluation corpus for both the language models and the PCFG, using the methods of §8.3.2.[5] We skip over tokens that are not part of the model tokenizer (i.e. those mapped to `<unk>`), since the language models did not have access to the true distribution of the original token. Based on the log probabilities of the language model and those of the PCFG, we can also compute their correlation. This allows us to determine to what extent the distribution of the LM is aligned with the true data distribution of the PCFG. We report two metrics: Spearman's $\rho$, a non-parametric rank correlation, and the $R^2$ coefficient, which measures the proportion of variation in the PCFG probabilities that is predictable from the LM probabilities.

[5] Because the EarleyX algorithm has not been optimised for coarse-to-fine parsing, we were unfortunately forced to compute the causal PCFG probabilities on a smaller subset. The results reported are on a sample of 1100 token probabilities. Concurrent to our work, more efficient prefix probability algorithms have been introduced by Nowak and Cotterell (2023) and Opedal et al. (2023); we leave the incorporation of these procedures open for future work.

### 8.5.2 Model Architecture

To emulate the learning behaviour of well-known Transformer LMs, we train LMs with similar architectures. For masked language modelling we train BERT, RoBERTa, and DeBERTa architectures (Devlin et al., 2019; Liu et al., 2019; He et al., 2021) of a much smaller size, inspired by the BabyBERTa models of Huebner et al. (2021). For causal language modelling we train GPT-2 and OPT architectures (Radford et al., 2019; Zhang et al., 2022). In our experiments we use model architectures with the following configuration: 8 layers, 8 heads, 256-dimensional hidden states, a batch size of 64, and a learning rate of $5 \cdot 10^{-4}$. Although a more optimal configuration may be found through a more elaborate hyperparameter search, it already provides a compute-friendly setup for training highly adequate models, as we will see in subsequent experiments. For training the LMs we use the `transformers` library (Wolf et al., 2020), training the models for 1 epoch with a cosine learning rate scheduler. For tokenization we use a whitespace tokenizer, with a minimum token frequency of 5 in the training corpus (else it is mapped to an `<unk>` token), which results in a vocabulary of 23,921 tokens. We use the same tokenizer for all models, to ensure a fair comparison across models.

### 8.5.3 Language Model Performance

We report the perplexity scores for our language models and PCFG in Table 8.1a. We provide a regression plot of the results for the GPT-2 model in Figure 8.1b. The considerable difference in PCFG perplexity between masked and causal language modelling shows that the amount of possible continuations is, unsurprisingly, far greater for causal language modelling. Despite being a task with greater uncertainty, however, the causal LMs acquire distributions that are closer aligned to the true PCFG distribution than the masked LMs. It can be seen that a higher Spearman's $\rho$ and $R^2$ yields a lower perplexity for all models, which demonstrates that models improve their performance by aligning their distribution

| | Architecture | Size | ($\psi$)-PPL | $R^2$ | $\rho$ |
|---|---|---|---|---|---|
| *Masked* | PCFG | | 63.9 | | |
| | BERT | 9.5M | 71.9 | 92.7 | 96.9 |
| | RoBERTa | 9.5M | 71.8 | 92.7 | 96.9 |
| | DeBERTa | 10.7M | **71.1** | 93.0 | 97.0 |
| *Causal* | PCFG | | 183.1 | | |
| | GPT-2 | 12.7M | **192.8** | 97.0 | 98.4 |
| | OPT | 15.6M | 194.2 | 96.7 | 98.3 |
| *TB* | DeBERTa | | *41.1* | | |
| | GPT-2 | | *115.0* | | |

**(a)** Results across models.

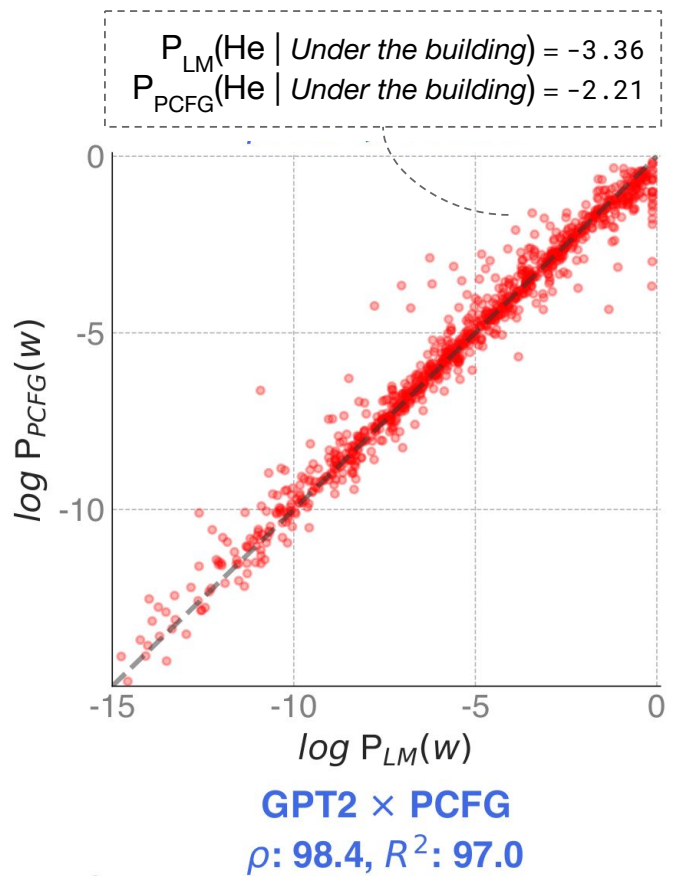

**(b)** Item-level probabilities for GPT2.

**Table 8.1:** Perplexity scores for various masked and causal LM architectures. $R^2$ and Spearman's $\rho$ are computed with respect to the PCFG log probabilities. Size denotes the total number of model parameters. *TB* denotes performance of LMs trained on the corpus of the original treebank data.

closer to the true distribution. As the DeBERTa and GPT-2 architectures obtain the best performance, we will use these architectures in subsequent experiments to evaluate and interpret their performance and behaviour.

**Treebank** We also trained masked and causal LMs on the sample of 500k sentences from the original treebank that had been used for inducing the state-split PCFG. This experiment serves as another baseline for the naturalness of the PCFG-generated data: the treebank data contains all possible linguistic cues that a LM could rely on, and a worsened performance on synthetic data indicates that some of these cues have been lost during grammar induction. We can see that this is the case in Table 8.1: the performance of both LMs is considerably better than the LMs trained on PCFG-generated data.

### 8.5.4 Corpus & Model Size

To investigate the impact of the amount of training data and the impact of model size, we conduct an additional experiment where we investigate how model performance shifts as these two aspects are varied. For the impact of corpus size we test models on increasing training sets: ranging from 10.000 to 7.5 million sentences. The model architectures is the same as outlined in §8.5.2, across all corpus sizes. Each model is trained for the same amount of steps (15,000), but for smaller corpus sizes we set the maximum number of epochs to 50, to reduce overfitting. In order to compare causal and masked LM performance directly, we report the *relative perplexity*, which is defined as the LM perplexity divided by the PCFG perplexity lower bound.

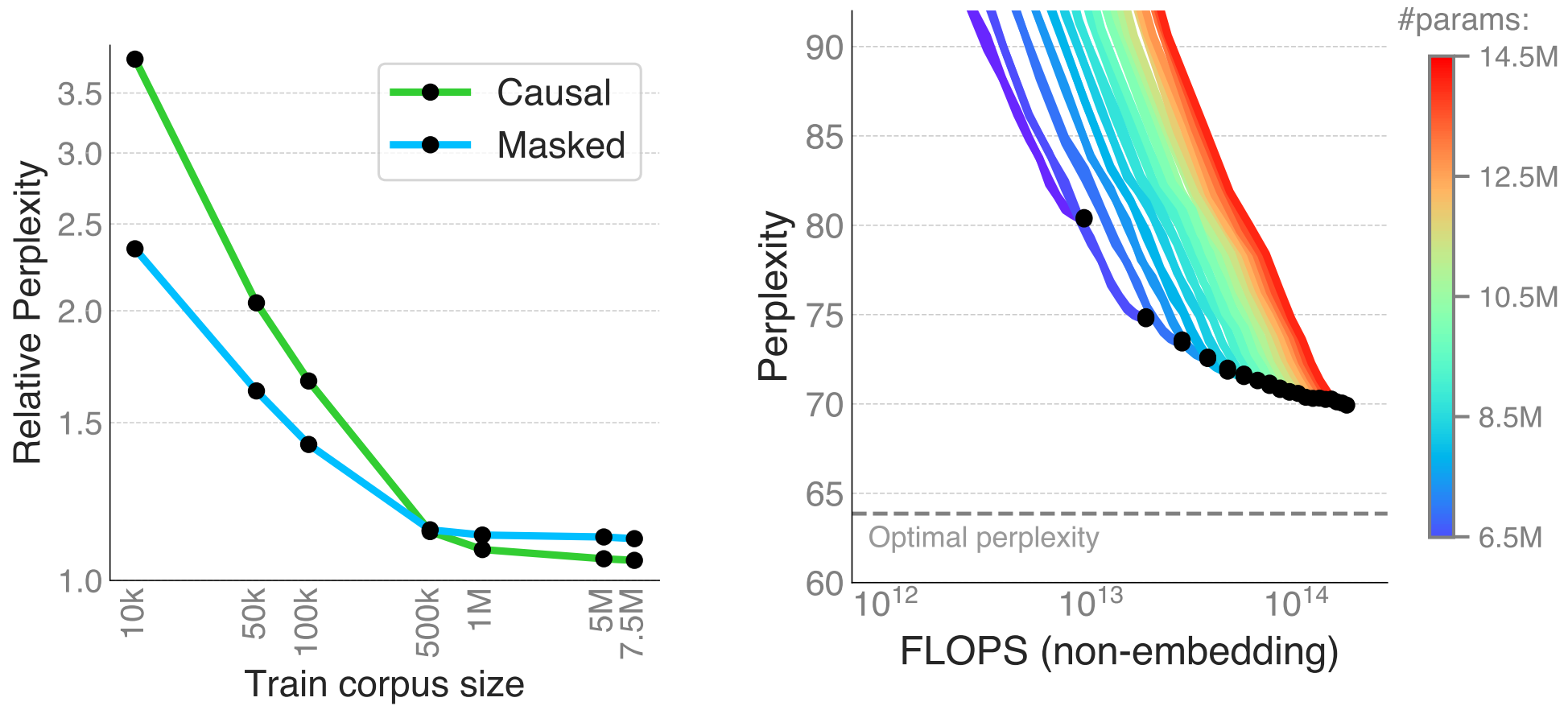

**Figure 8.4:** Language modelling performance expressed as a function of training size (a) and model size (b). Relative perplexity is computed with respect to the PCFG's perplexity lower bound, averaged over 3 seeds.

For the impact of model size we train DeBERTa models with an increasing amount of layers, ranging from 1 to 20. All models are trained on the maximum corpus size of 7.5 million sentences for 1 epoch. We report model performance as a function of FLOPS across intermediate checkpoints, similar to how this is done for evaluating scaling laws of massive LMs (Kaplan et al., 2020; Hoffmann et al., 2022). FLOPS are approximated using the `fvcore` library.

**Results** We report the results in Figure 8.4. Interestingly, the causal LMs converge slower to the final performance than masked LMs, but ultimately reach a better relative perplexity. The corpus of 500,000 sentences appears to be the inflection point: the masked LMs appear to reach a plateau at this point, whereas the causal LMs continue to improve with more training data.

For the model size experiment it can be seen that performance improves as model size increases. A considerable gap between the optimal perplexity and model perplexity remains, however. The models we consider here are still relatively small, and it appears that larger scale models may be beneficial for the task, without risk of overfitting yet. Nonetheless, it remains interesting to see that our models follow a similar pattern as seen for large-scale LMs (Hoffmann et al., 2022), with a similar near-linear decrease in perplexity in log space.

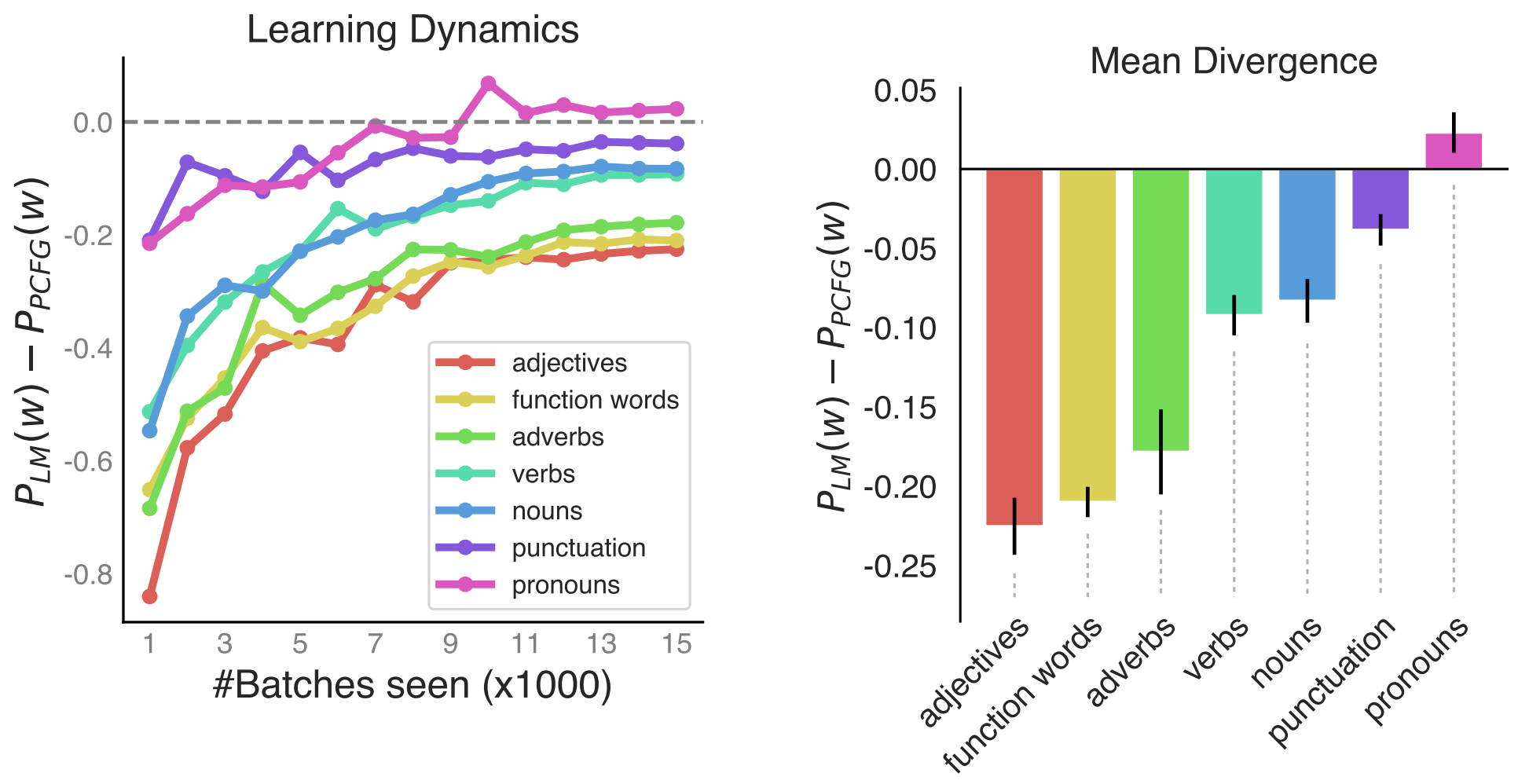

**Figure 8.5:** The LM probability divergence from the PCFG distribution, aggregated by general part-of-speech classes, for the DeBERTa model.

## 8.6 Model Interpretability

We investigate the behaviour and learning dynamics of the DeBERTa model from §8.5.3. Our setup provides an interesting test bed for interpretability purposes: since we have transparent access to the underlying structure of our data, we can make stronger assumptions about the cues the model must be relying on than would be possible in natural language. We conduct two experiments, one on learning dynamics and another on Part-of-Speech probing.

### 8.6.1 Learning Dynamics

The LMs of §8.5.3 obtain a distribution that aligns closely to the true PCFG distribution. We investigate how this distribution forms during training by aggregating token probabilities to a general POS class, mapping the 43 POS tags in the original data to one of 7 base classes. We report the *divergence* of the LM probabilities from the PCFG, expressed as the mean difference in log probabilities.

**Results** We report the results in Figure 8.5. We can see that different POS classes are learned at different stages during training. Punctuation and pronouns obtain a low divergence at the start of training already, which is maintained across checkpoints. A possible explanation for this might be that punctuation is less dependent of context, and more on the position in the sentence: quotation marks occur mostly at the start and end of the sentence, and most sentences end with a period symbol.

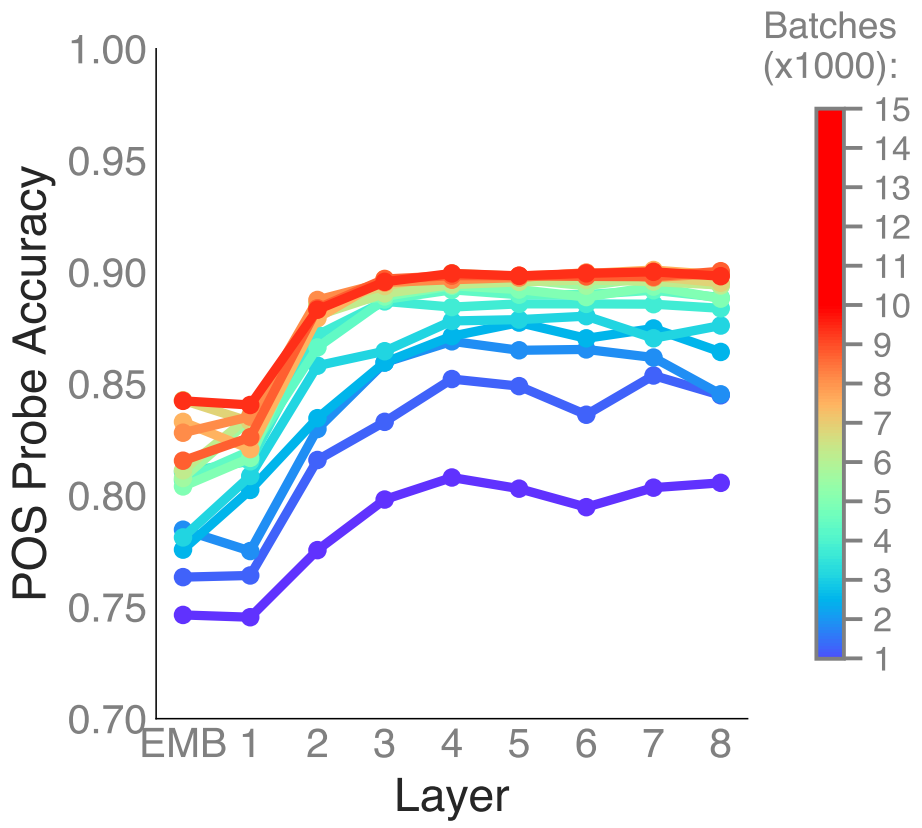

| Metric | $\rho$ |
|---|---|
| $P_{\text{pcfg}}$ | 0.54 |
| $P_{\text{lm}}$ | 0.59 |
| $\lvert P_{\text{lm}} - P_{\text{pcfg}} \rvert$ | -0.51 |

**Figure 8.6:** Left: POS Probing performance across layers (x-axis) and split out for multiple checkpoints through training. Right: Spearman correlation between POS probing probabilities of the true class and the token probabilities of the PCFG, LM, and probability divergence.

### 8.6.2 POS Probing

Probing is an interpretability technique in which auxiliary classifiers are trained on top of the representations of a LM, to find out whether abstract features are encoded (Adi et al., 2017; Hupkes et al., 2018). We conduct a probing experiment on the representations of the DeBERTa model, classifying the part-of-speech (or pre-terminal) node above each token in the syntax tree.[6] We split the train and test sections based on the underlying token of a POS tag: this ensures that the probe finds a robust signal that does not simply map a token's identity to its most common POS tag, but that generalises based on other tokens with that POS tag (Hewitt and Liang, 2019a).

**Results** We report the results in Figure 8.6. Performance reaches a plateau at 40% of training, after which it settles at an accuracy of around 90%. We can see that POS information is encoded consistently from around layer 3 in the model, and that this performance does not deteriorate in upper layers, unlike pretrained LMs like BERT (Tenney et al., 2019a), where upper layers appear to encode sentence-level semantic features. This indicates that the PCFG nature of our data relies more on syntactic cues, which is likely the effect of a loss of *naturalness*, as explored in §8.4.2.

[6] Note that our procedure right now does not take structural ambiguity into account, we leave the incorporation of the full POS distribution open for future work.

Additionally, we also investigate the correlation between the probabilities of the POS probe and the data distribution. We would expect that if a model relies on the POS signal extracted by our probe for its token predictions, that in cases where these predictions diverge from the PCFG distribution, the POS performance is lower as well. We compute the Spearman correlation for both the LM and PCFG distributions, as well as the divergence. Both LM and PCFG distributions obtain a positive correlation: this shows that the POS probe performs better on tokens with a higher probability. The divergence metric, however, yields a *negative* correlation: this shows that when the LM's probabilities diverge more from the PCFG distribution, the POS probing performance drops as well. We can conclude from this that representing POS information is a prerequisite for competent language model performance.

## 8.7 Conclusions and Future Work

In this paper, we train language models on artificial data from an explicit, known source, but in a setup where that data approximates natural language data in richness and complexity. This approach allowed us to compute an exact lower bound on perplexity, and evaluate how well different LM architectures can approach this optimum. We observed striking differences in the learning dynamics between masked and causal language models, where the latter converge much more slowly, but ultimately approximate the optimum more closely. And we observe large differences between word types in how well the trained LMs approximate the true probabilities from the underlying source grammar.

Our proposed methodology for evaluating and interpreting LMs thus involves inducing a rich probabilistic grammar, and using that grammar to generate a language-like training sets for neural language models. It did become clear, however, that our grammar induction has led to a sacrifice of various linguistic cues, which is demonstrated by a lower selectional preference (§8.4.2) and higher perplexity with respect to the original treebank (§8.5.3). We leave the exploration of which linguistic cues exactly are lost open for future work, for example by evaluating the LMs on the BLiMP benchmark for linguistic phenomena (Warstadt et al., 2020).

We believe our methodology holds promises beyond the specific modelling choices we have made in this paper. In the future we hope to see studies using different grammar induction procedures and focusing on other interpretability techniques, including the ambitious efforts, under the label 'mechanistic interpretability', to reverse engineer the behaviour of LLMs. Furthermore, we believe our work will be of interest to work on NLG evaluation and uncertainty estimation, where having access to the true data distribution is important (Pimentel et al., 2023; Zhang et al., 2023; Giulianelli et al., 2023).

Perhaps the most important relevance of our work, and such alternative approaches, is in relation to a key open problem in interpretability research: how to evaluate the faithfulness of attribution and probing techniques. Prior research has attempted to define controlled environments that provide a certain degree of certainty about *how* the model is solving a task. Examples include the training of models on small-scale artificial data to perfection (Hao, 2020; Jumelet and Zuidema, 2023a), or by explicitly providing shortcuts that the model *must* rely on (Bastings et al., 2022; Ebert et al., 2023). Scaling this up to more realistic, natural data, however, remains an open problem. Our setup, with direct access to a natural-like distribution, now provides much stronger guarantees about what patterns in the data the model is relying on, and allows for a gradual increase of the complexity of the data distribution.

*I'm finished!*

Daniel Plainview, *There Will Be Blood*

# 9 Conclusion

THE PRIMARY QUESTION WE INVESTIGATE in this thesis is whether language models possess a deep understanding of grammatical structure comparable to that of humans. This inquiry arises from a deep fascination with the impressive capabilities of language models, which can generate fluent, grammatical text despite lacking explicit supervision on the underlying grammar of their training data. Our research was divided into three parts. In Part I, we explored the presence of abstract linguistic information in language model predictions through *structural priming*, examining the similarities between model behaviour and human findings. In Part II, we presented three studies into model performance across various linguistic phenomena, explaining observed patterns in terms of training distribution and conducting a large-scale corpus intervention study to assess the generalisation capacities of LMs. Finally, in Part III, we investigated structure-building in a fully controlled environment, using it as a testbed for assessing the faithfulness of various interpretability techniques. In this concluding chapter, we reflect on the broader contributions and implications of our findings.

## 9.1 Language Model Predictions Are Affected by Abstract Structure

Inspired by influential studies from psycholinguistics, we adopted a behavioural approach for *finding structure in language models* by applying the structural priming paradigm to them. Through extensive experiments on a carefully crafted evaluation corpus, we demonstrated that, like humans, language model predictions are influenced by prior context at an abstract, structural level. We designed various evaluation conditions to examine the impact of factors such as semantic similarity, lexical overlap, recency, and cumulativity on priming effects, situating our findings within the broader context of psycholinguistic research. Moreover, structural priming offers a robust framework for investigating the interplay between syntax and semantics in language model predictions. Our experiments revealed that structural predictions are significantly influenced by the degree of semantic similarity and lexical overlap in prior context, providing strong evidence that syntactic and semantic properties are deeply intertwined in language model representations.

Our approach in Chapter 2 was *observational*, setting the stage in Chapter 3 for a more detailed, *correlational* investigation into the factors driving the observed priming effects at the item level. Drawing inspiration from earlier corpus linguistics research, we demonstrated how various abstract factors explain these priming effects. A key factor that we identified is the *inverse frequency effect*: the less likely a construction is for the model, the larger the subsequent priming effect. Inverse frequency effects support an *implicit learning* account of human language processing, where significant prediction errors lead to overcompensation in future predictions. It remains a promising direction to leverage such psycholinguistic concepts for explaining more general LM behaviour, such as in-context learning (Chan et al., 2022; Zhou et al., 2024).

A natural follow-up to the *correlational* approach of Chapter 3, which we were not able to do in the time frame of this dissertation project, would be to conduct a *causal* study of structural priming effects, following the FiCT methodology of Chapter 5. What kind of patterns need to be present in a training corpus for structural priming to occur? And how do structural priming effects arise during training? Another exciting direction to studying priming in LMs would take on a stronger focus on interpretability: can we locate *where* in the model priming patterns are being activated? Finding the model circuits that are responsible for priming effects can provide useful insights into whether these circuits play a role in other behaviour as well.

Our findings in Chapter 3 also raise intriguing questions about how we should relate language model predictions to human language *production* patterns on the one hand, and *comprehension* patterns on the other. While recent studies suggest that the mechanisms driving human language production and comprehension are more similar than previously assumed (Tooley, 2023), we must remain cautious in our expectations regarding how well language models can reflect comprehension phenomena when they are only exposed to hu-man produced data. Although processing difficulties from comprehension can manifest in production data—such as the observed inverse frequency effect—many other phenomena may only emerge in language model behaviour through indirect means. Using language models as tools to explain cognitive processes remains a promising research direction, and we hope that our studies on structural priming will inspire further exploration in this area.

## 9.2 Language Models (Can) Generalise From Indirect Evidence

Assessing the grammatical abilities of language models is typically done on a phenomenon-by-phenomenon basis. By evaluating models across a wide range of linguistic phenomena, we can determine their grammatical competence and identify constructions that remain challenging. We contribute to this approach by presenting three studies into the generalisation patterns of various phenomena, focusing in particular on *adjective order* and *negative polarity items*. This work aligns with broader research exploring the trade-off between memorisation and generalisation in neural models (e.g. Dankers et al., 2023; Leong and Linzen, 2024). Do LMs build up grammatical representations that extend beyond mere memorisation of co-occurrences? After all, we expect a model that 'understands' subject-verb agreement to generalise to subject-verb combinations it has never encountered before (Gulordava et al., 2018).

Our findings present a nuanced account of LM generalisation. On the one hand, the FiCT methodology from Chapter 5 demonstrates that LMs can generalise to constructions they have not seen during training. For example, our models that never encountered subjects with prepositional modifiers were still able to correctly predict the correct number of the verb. For negative polarity items, which are licensed by downward monotone environments, we demonstrate that models can generalise to novel context-NPI combinations using a general encoding of monotonicity. These findings suggest that LMs *do* create robust linguistic generalisations. On the other hand, our findings in Chapter 4 reveal that a significant amount of LM behaviour can be directly explained by the underlying data distribution, and that out-of-domain generalisation is limited.

The question that remains then is how to reconcile these opposing conclusions. Could the ostensible generalisations of Chapter 5 and 6 be explained by pattern memorisation, given a more thorough analysis of the training distribution? Or is there something unique about the bi-gram nature of adjective order that prevents the model from forming abstract generalisations? These findings underscore the importance of evaluating LM behaviour with rigorous control over the data distribution, instead of treating the latter as another 'black box'. The methodology outlined in Chapter 8 may offer a valuable framework for investigating this issue: by maintaining full control over the data distribution, we can measure the trade-off between generalisation and memorisation more precisely, which will allow us to conclude under which conditions abstract generalisation emerges.

## 9.3 Interpretability Should Be Evaluated with Care

In the final part of the thesis we investigated how *hierarchical structure* is represented in language models. For this endeavour, we can no longer rely on the simple *Targeted Syntactic Evaluations* used in earlier chapters and we therefore shifted our focus to more complex interpretability methods. In particular, we posited that *feature interactions* are a vital component for constructing hierarchical structure. By conducting extensive experiments on a wide range of feature interaction detection and attribution methods (FIDAMs) we were able to determine the conditions under which hierarchical structure could be faithfully extracted from model predictions.

This desire of *faithful* explanations proved challenging and led us in Chapter 7 to focus on small, synthetic toy languages where we could train our models to near perfection. The underlying assumption here is that an explanation for a model of a language should directly reflect the hierarchical rules that constitute that language, as a perfect model of that language cannot rely on spurious heuristics. Through this approach, we demonstrated that various FIDAMs can accurately retrieve the correct hierarchical structure. However, the question remains: how can we transfer findings from this controlled environment to larger-scale settings that better reflect the complexity of natural language? Do faithful explanations in a toy domain suggest that the same method will generate more faithful explanations in natural language contexts?

In Chapter 8, we proposed a methodology for scaling up synthetic languages. By leveraging large-scale probabilistic context-free grammars, we generated corpora that approximate the complexity of natural language while maintaining full control over the underlying data distribution. We showed that LMs can model this grammar with high precision and that probing methods can successfully extract a grammar's non-terminal states from model representations. A natural next step for future work is to connect this framework with the feature interactions discussed in Chapter 7. Additionally, it would be valuable to adjust the complexity of the grammar to assess how the faithfulness of interpretability changes as the model's accuracy in representing the underlying data distribution decreases.

Our decision to focus on feature interactions for extracting hierarchical structure from model representations may have been suboptimal. When we initiated that study in 2021, mechanistic interpretability was still in its infancy, and feature attribution methods dominated interpretability research. Since then, numerous impressive studies have demonstrated the utility of interpretability methods that closely align with model components, and significant progress has been made in expanding the different facets of model behaviour that can be explained by these tools. While the higher-level abstractions offered by attribution methods still hold strong explanatory value, the faithfulness issues we encountered may be less severe when using interpretability tools at Marr's *implementational level*.

## 9.4 On the Role of Linguistics in NLP and NLP in Linguistics

Finally, we reflect back on the interplay between linguistics and NLP that we outlined in Chapter 1 and Figure 1.1. We begin by examining the role of linguistics in NLP. While the engineering pursuit of optimising NLP models has largely moved away from operationalising linguistic intuitions in favour of a fully data-driven approach, linguistic concepts remain crucial in model evaluations (Opitz et al., 2024). Similarly, many experiments in this thesis leverage linguistic concepts for interpreting language model behaviour, such as monotonicity, adjective order, and structural priming. Psycholinguistic assessments of language models may also help us predict how language models will perform in novel contexts and reveal unintended stereotypical biases. As such, linguistics remains a vital part of *grounding* language models to actual language use.

The opposite direction, where NLP developments inform linguistic theory, is less explored. What can language models tell us about the study of language? Are their generalisations indicative of human language processing? We identify several promising areas where NLP findings could benefit linguistics.

For instance, LMs can serve as confirmation tools for linguistic theories. Since linguistic theory often seeks to define abstractions over linguistic structure, it is encouraging to find that language models form similar abstractions, suggesting that the proposed abstraction may be the 'correct' one (Tenney et al., 2019a). Beyond confirming hypotheses, LMs can also be used to *generate* novel hypotheses. An exciting example is provided by Lakretz et al. (2021), who analysed the behaviour of LMs on complex subject-verb agreement patterns to hypothesise how humans might process these patterns, later confirming their hypotheses through human studies. In a follow-up to our work in Chapter 3, Sinclair et al. (2024) em-ploy a similar workflow, leveraging structural priming effects in LMs to investigate human priming effects derived from eye-tracking and EEG studies.

Another promising direction is connecting LM behaviour to studies of human language acquisition. Although the vast scale at which current LMs are trained no longer mirrors human language use, recent research has increasingly employed LMs trained on smaller scales to investigate questions related to language acquisition (Warstadt et al., 2023). By using LMs as test subjects in acquisition studies, we can assess the learnability of complex constructions with much greater control than is possible in human studies (Warstadt, 2022; Yedetore et al., 2023). Our work on FiCT, discussed in Chapter 5, aligns with this approach, demonstrating that linguistic abilities can be acquired from indirect evidence. As such, LMs may provide a valuable perspective on long-standing questions regarding the *poverty of the stimulus* and related topics in language acquisition (Berwick et al., 2011; Pearl, 2022).

The study of language is multifaceted, and this thesis primarily focuses on the subfields of formal linguistics and psycholinguistics. However, techniques from NLP have also proven highly useful in other language-related disciplines, such as neurolinguistics, phonology, and sociolinguistics. For instance, research at the intersection of NLP and neurolinguistics has explored brain decoding tasks, where dense representations from language models are used to 'reconstruct' high-dimensional representations from brain scans (Abnar et al., 2019; Toneva and Wehbe, 2019; Schrimpf et al., 2021). These approaches can also be applied to investigate our research goal of grammatical structure acquisition, albeit at a different level of abstraction. This line of research again exemplifies how NLP can inform cognitive theories, with various studies showing that language model representations can reflect brain representations at a fine-grained level (Stanojević et al., 2022; Fresen et al., 2024).

We would like to conclude with a note of language inclusivity. The predominant focus on English in NLP research has led to systems that are highly optimized for English but may be suboptimal for other languages, which has contributed to a language technology landscape that is not inclusive of many low-resource languages (Joshi et al., 2020; Bird, 2022; Blasi et al., 2022). While this disparity is largely driven by differences in available training resources, linguistic factors have also been shown to play a significant role. Some languages, for instance, are intrinsically harder to model than others with contemporary deep learning methods due to typological and morphological differences (Mielke et al., 2019; Bisazza et al., 2021). Resolving these issues requires both a deep understanding of the inductive biases of NLP systems and a cross-lingual analysis of learnability and acquisition patterns.

In other words, the interplay between NLP and linguistics will remain crucial for creating fair and inclusive language systems. Ultimately this may also lead to the development of more efficient learning algorithms that are significantly less data hungry than the current paradigm, allowing NLP systems to become more efficient at modelling low-resource languages.

# A

# Supplementary Materials

## A.1 Human-like Structural Priming Effects

### A.1.1 Sentence-level asymmetry effects

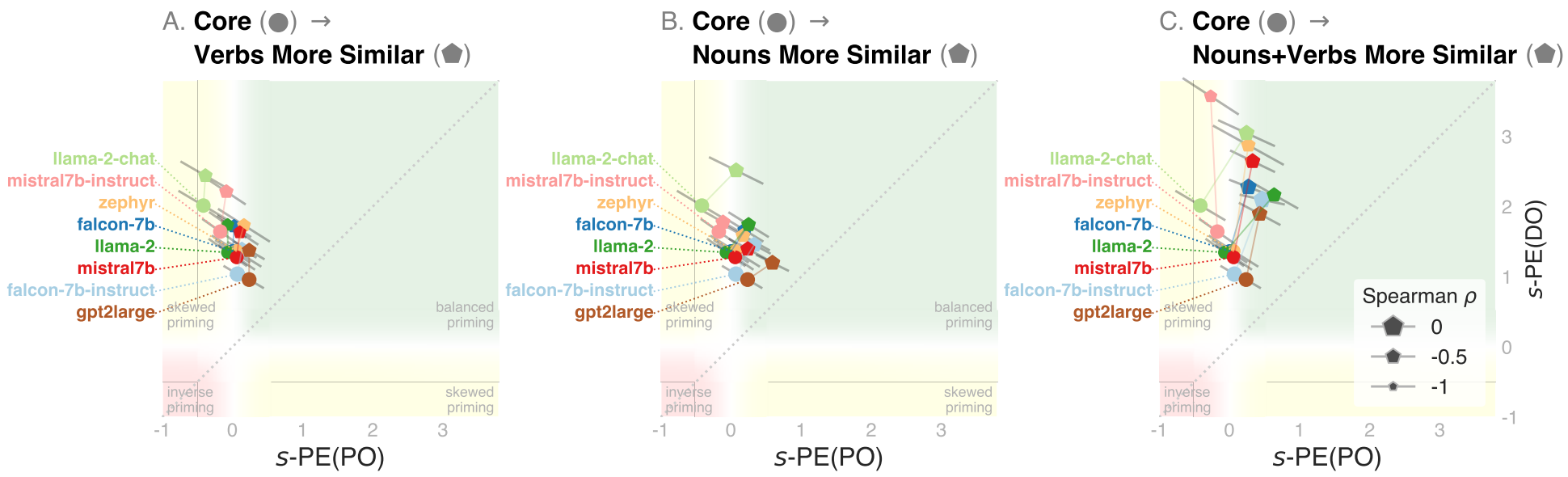

The increased semantic similarity conditions can be seen to primarily have an effect on boosting the PE score of the dominant structure (DO), while leaving the PE score of the opposite structure unaffected (PO). Furthermore, it can be seen that the the effect of increasing the similarity of nouns and verbs is not *linearly additive*: increasing similarity of both nouns and verbs has a far greater impact than the individual conditions combined.

### A.1.2 Token-level PE for Increased Semantic Similarity

The token-level PE scores for the 3 conditions with increased semantic similarity are shown below.

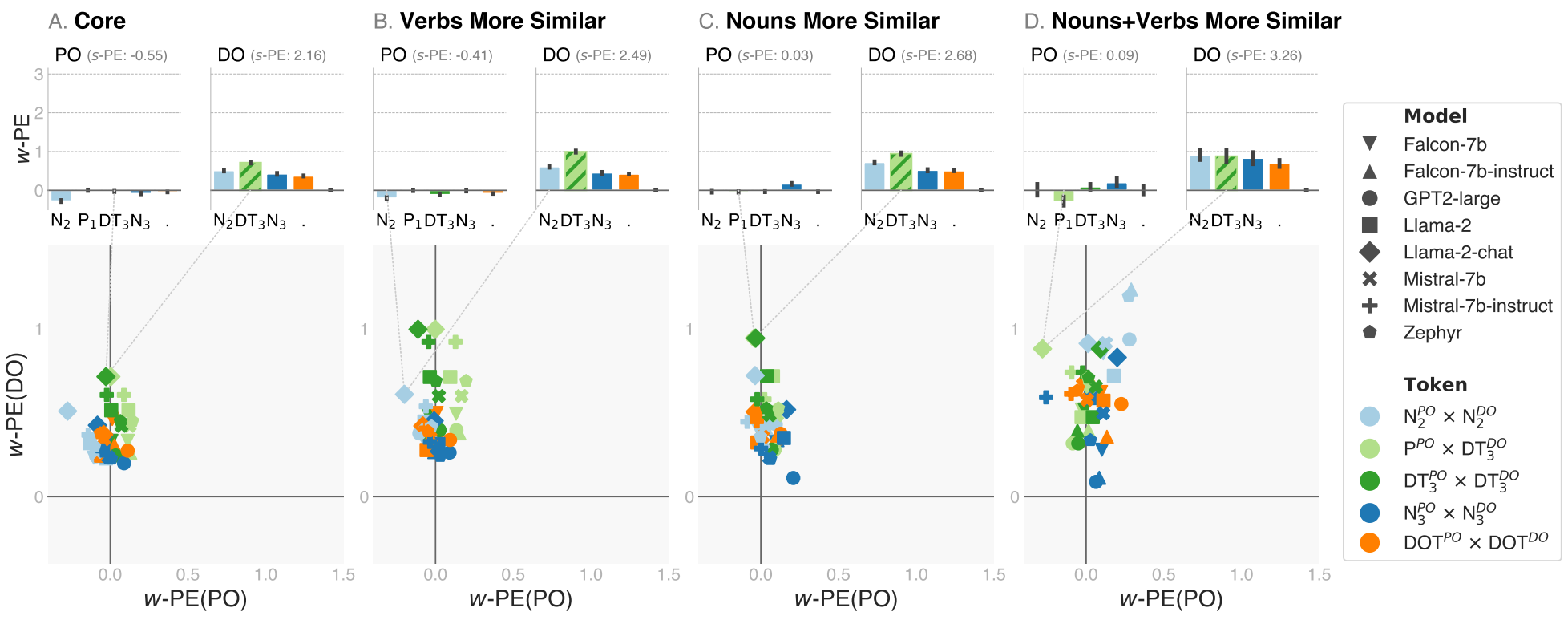

## A.1.3 PE Correlations

We plot the Spearman correlations across LMs for i) the congruent conditional log probabilities, ii) the incongruent conditional log probabilities, and iii) the Priming Effect scores below.

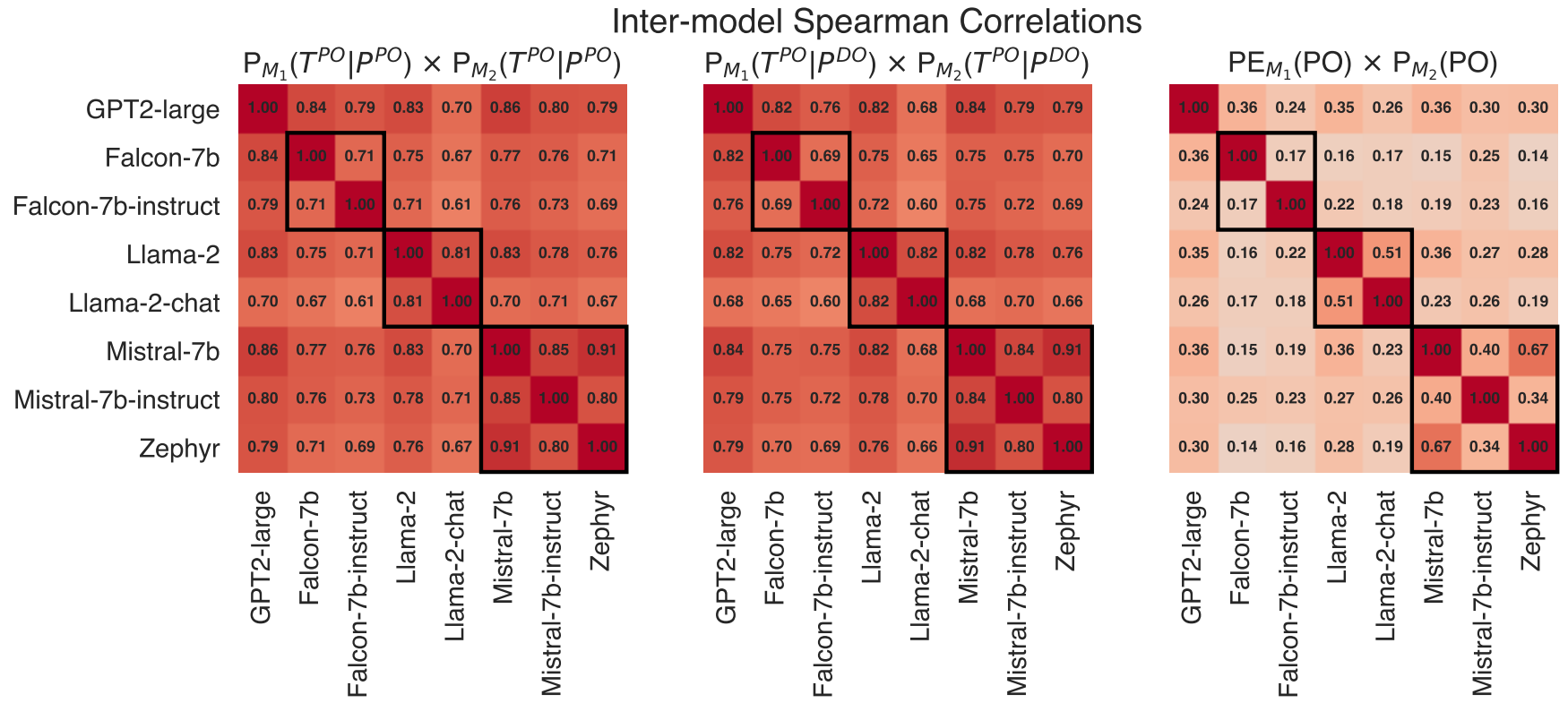

## A.1.4 Preference Order Correlation

The Spearman correlation between LMs and human structural preference order are shown below. Surprisingly, there exists a high degree of variation in preference order, both within models and across models and human preference. Only *Falcon-7b* and its instruction-tuned variant retain a high preference overlap; all the other aligned LMs diverge quite strongly from their base model. None of the LMs have a significant correlation with respect to the reported human preference order, which is in contrast to Hawkins et al. (2020)'s positive findings of strong correlations between model and human preferences. We leave a more thorough investigation of these differences open for future work.

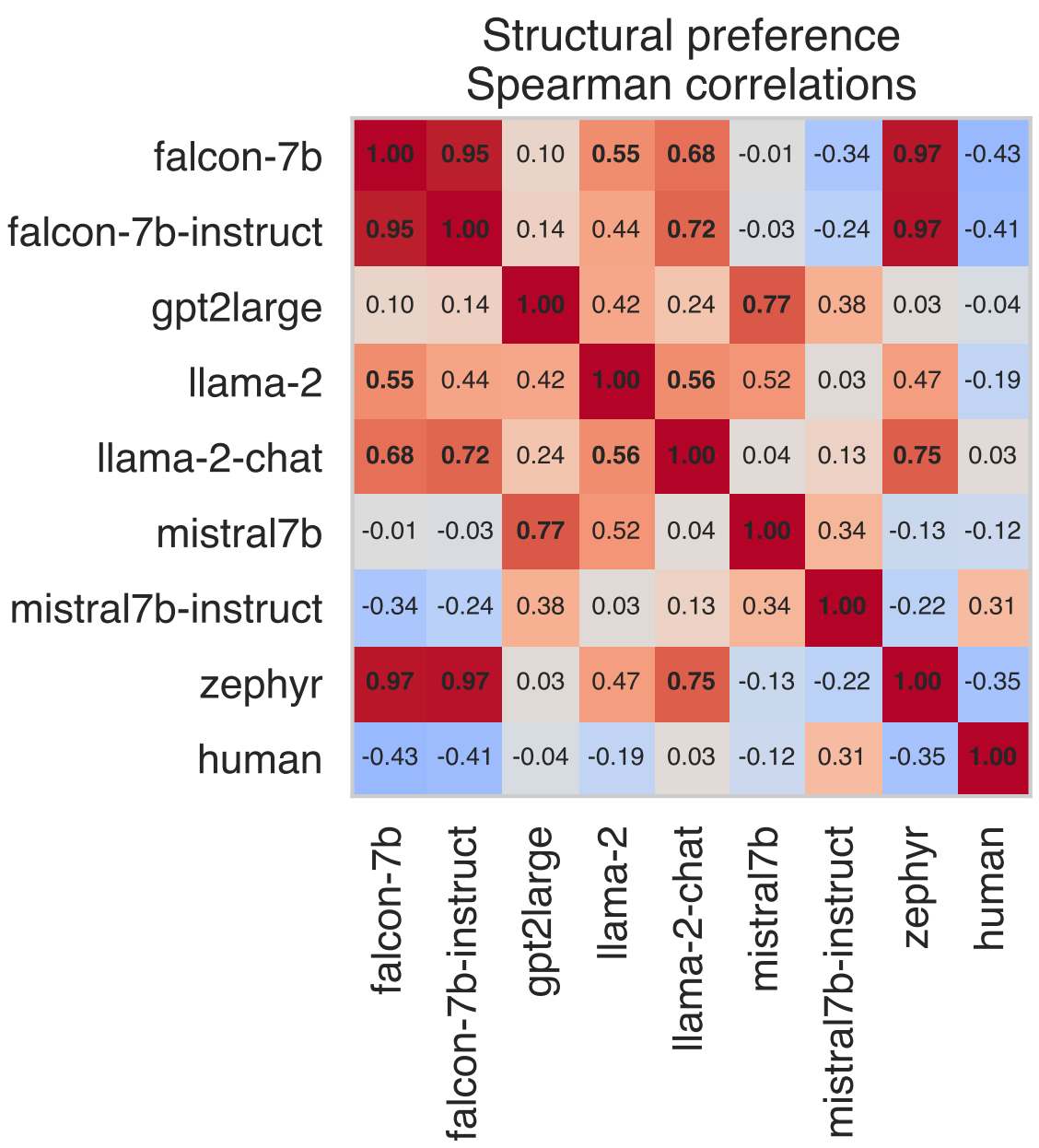

### A.1.5 Linear Mixed-effect Model Summary

The LMM results including coefficients, standard error, $z$-score and $p$-values are shown in Table A.1 and A.2.

| | | | | |
|---|---|---|---|---|
| 0 | Model: | MixedLM | Dependent Variable: | $s_\partial$-PE(PO) |
| 1 | No. Observations: | 30000 | Method: | REML |
| 2 | No. Groups: | 8 | Scale: | 1.8203 |
| 3 | Min. group size: | 3655 | Log-Likelihood: | -51620.4121 |
| 4 | Max. group size: | 3858 | Converged: | No |
| 5 | Mean group size: | 3750.0 | | |

| | Coef. | Std.Err. | z | P>\|z\| | [0.025 | 0.975] |
|---|---|---|---|---|---|---|
| Intercept | -0.072 | 0.111 | -0.652 | 0.515 | -0.289 | 0.145 |
| sim($n_1$) | 0.041 | 0.010 | 4.260 | 0.000 | 0.022 | 0.060 |
| sim($n_2$) | 0.097 | 0.009 | 10.381 | 0.000 | 0.079 | 0.115 |
| sim($n_3$) | 0.109 | 0.009 | 11.652 | 0.000 | 0.091 | 0.127 |
| sim($v$) | 0.010 | 0.009 | 1.092 | 0.275 | -0.008 | 0.027 |
| sim($s$) | -0.000 | 0.017 | -0.008 | 0.994 | -0.034 | 0.034 |
| $N_1$ overlaps | -0.148 | 0.052 | -2.850 | 0.004 | -0.250 | -0.046 |
| $N_2$ overlaps | 0.696 | 0.056 | 12.510 | 0.000 | 0.587 | 0.805 |
| $N_3$ overlaps | 0.473 | 0.050 | 9.405 | 0.000 | 0.374 | 0.571 |
| Det. overlaps | 1.010 | 0.026 | 38.149 | 0.000 | 0.958 | 1.062 |
| Verb overlaps | 1.491 | 0.046 | 32.578 | 0.000 | 1.401 | 1.581 |
| Prep. overlaps | 1.025 | 0.027 | 38.574 | 0.000 | 0.973 | 1.077 |
| $-$P(prime$_{po}$) | 0.390 | 0.023 | 16.788 | 0.000 | 0.345 | 0.436 |
| $-$P(prime$_{do}$) | -0.259 | 0.025 | -10.504 | 0.000 | -0.307 | -0.210 |
| $-$P(target$_{po}$) | 0.044 | 0.023 | 1.885 | 0.059 | -0.002 | 0.089 |
| $-$P(target$_{do}$) | 0.055 | 0.025 | 2.223 | 0.026 | 0.006 | 0.103 |
| PO-pref($v^p$) | -0.129 | 0.010 | -13.377 | 0.000 | -0.148 | -0.110 |
| PO-pref($v^t$) | -0.029 | 0.010 | -3.041 | 0.002 | -0.048 | -0.010 |
| Group Var | 0.097 | 0.061 | | | | |

**Table A.1:** Raw LMM results for predicting $s_\partial$-PE(PO).

| | | | | | |
|---|---|---|---|---|---|
| 0 | Model: | MixedLM | Dependent Variable: | $s_{\partial}$-PE(DO) |
| 1 | No. Observations: | 30000 | Method: | REML |
| 2 | No. Groups: | 8 | Scale: | 2.2337 |
| 3 | Min. group size: | 3655 | Log-Likelihood: | -54688.2103 |
| 4 | Max. group size: | 3858 | Converged: | No |
| 5 | Mean group size: | 3750.0 | | |

| | Coef. | Std.Err. | z | P>\|z\| | [0.025 | 0.975] |
|---|---|---|---|---|---|---|
| Intercept | 1.339 | 0.111 | 12.047 | 0.000 | 1.121 | 1.557 |
| sim($n_1$) | 0.071 | 0.011 | 6.701 | 0.000 | 0.050 | 0.092 |
| sim($n_2$) | 0.021 | 0.010 | 2.018 | 0.044 | 0.001 | 0.041 |
| sim($n_3$) | 0.122 | 0.010 | 11.826 | 0.000 | 0.102 | 0.142 |
| sim($v$) | 0.171 | 0.010 | 17.197 | 0.000 | 0.151 | 0.190 |
| sim($s$) | -0.044 | 0.019 | -2.331 | 0.020 | -0.082 | -0.007 |
| $N_1$ overlaps | 0.456 | 0.057 | 7.966 | 0.000 | 0.344 | 0.568 |
| $N_2$ overlaps | -0.146 | 0.061 | -2.384 | 0.017 | -0.266 | -0.026 |
| $N_3$ overlaps | 0.717 | 0.055 | 12.947 | 0.000 | 0.608 | 0.826 |
| Det. overlaps | 1.671 | 0.029 | 57.289 | 0.000 | 1.614 | 1.728 |
| Verb overlaps | 1.535 | 0.050 | 30.433 | 0.000 | 1.436 | 1.634 |
| Prep. overlaps | 0.244 | 0.029 | 8.349 | 0.000 | 0.187 | 0.302 |
| $-$P(prime$_{po}$) | -0.333 | 0.026 | -12.989 | 0.000 | -0.383 | -0.283 |
| $-$P(prime$_{do}$) | 0.416 | 0.027 | 15.332 | 0.000 | 0.363 | 0.469 |
| $-$P(target$_{po}$) | -0.045 | 0.026 | -1.759 | 0.079 | -0.095 | 0.005 |
| $-$P(target$_{do}$) | 0.311 | 0.027 | 11.452 | 0.000 | 0.258 | 0.365 |
| PO-pref($v^p$) | 0.231 | 0.011 | 21.761 | 0.000 | 0.210 | 0.252 |
| PO-pref($v^t$) | 0.215 | 0.011 | 20.262 | 0.000 | 0.194 | 0.236 |
| Group Var | 0.097 | 0.031 | | | | |

**Table A.2:** Raw LMM results for predicting $s_{\partial}$-PE(DO).

## A.2 Adjective Order

### A.2.1 CAP🧢 Examples

We provide a sample of CAP examples below, categorised by the contextual and isolated AOP scores of Pythia-12b. For this sample we selected the most salient examples of the four quadrants in Figure 4.7.

| | CAP Sentence | AOP(●) | AOP(●\|$c$) |
|---|---|---|---|
| | **CAP order always preferred** | | |
| 1. | *In solution, lead(II) hydroxide is a somewhat weak base, forming lead(II) ion, Pb, under **weakly acidic** conditions.* | 16.5 | 18.8 |
| 2. | *It has the **only deep** water port in Cambodia.* | 16.9 | 15.4 |
| 3. | *The greatest social reforms in Denmark are certainly the work of the **last half**-century.* | 15.4 | 16.3 |
| 4. | *Icterids are unusual in songbirds because they have **considerable sexual** dimorphism.* | 16.9 | 14.5 |
| | **Swapped order always preferred** | | |
| 5. | *Couches are usually bought in a set together with cushions, which give them a bouncey and **decorative comfortable** touch.* | −7.1 | −6.8 |
| 6. | *No, its 10 feet tall, and it's a **red big** monster, demon like.* | −7.3 | −5.9 |
| 7. | *It does this by following the **grammatical basic** rules of syntax.* | −5.8 | −6.0 |
| 8. | *The **daily average** attendance from January through November 2010 was 22,133 people a day.* | −5.8 | −5.7 |
| | **Context improves AOP** | | |
| 9. | *Being a spherical 3-manifold, it is the only homology 3-sphere, besides the 3-sphere itself, with a **finite fundamental** group.* | 0.9 | 16.8 |
| 10. | *Since the 2015 Styria **municipal structural** reform, it is part of the municipality Birkfeld.* | −0.5 | 15.3 |
| 11. | *I won it at a **fair last** night.* | 0.2 | 15.4 |
| 12. | *I say that the man who does not play whist lays up a **sad old** age for himself.* | 2.6 | 17.5 |
| | **Context worsens AOP** | | |
| 13. | *This is an ability which may have been present in their **last common** ancestor in the Archaean.* | 12.3 | 4.0 |
| 14. | *Jacob, you've had a **complex partial** seizure, which can cause psychosis, including religious psychosis.* | 7.3 | 0.3 |
| 15. | *He was also a right-arm **medium fast** bowler with three wickets in test matches to his credit.* | 6.2 | −0.6 |
| 16. | *He wheeled round sharply, and distinguished her lying with **helpless outspread** arms on the couch.* | 4.0 | −2.8 |

### A.2.2 Correlations to Cognitive Predictors

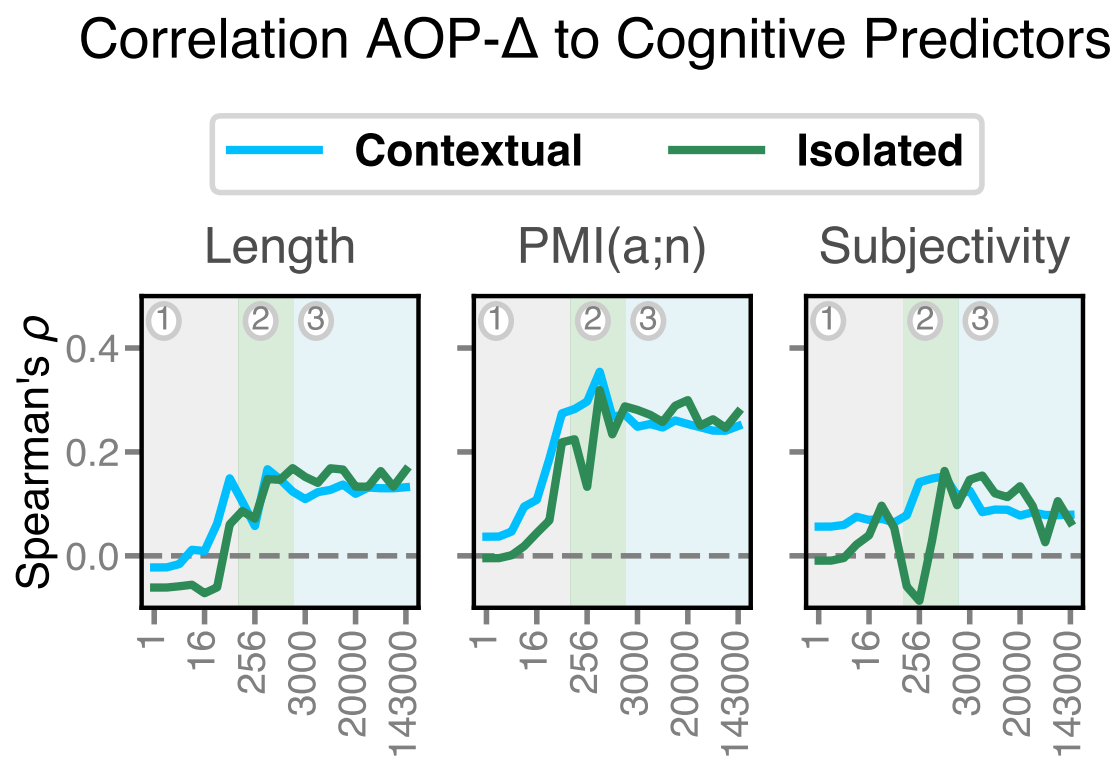

Correlations during training of AOP-Δ scores (Pythia-1.4b), with scores derived from various cognitive predictors.

### A.2.3 One-shot AOP

Below we plot the AOP-Δ scores for the 27 adjective pairs that are seen exactly once in the first 10% of training. By investigating these curves at an item-level, we gain insight whether the AOP-Δ score is directly affected by an item occurrence, or dependent on external factors that drive AOP indirectly. If AOP acquisition is purely driven by co-occurrence statistics, we would expect to see a clear spike in AOP-Δ at the checkpoint where an adjective pair has just been encountered. If, however, AOP is driven more strongly by a general, abstract signal of adjective order, then the AOP-Δ scores will be less impacted by adjective occurrence alone.

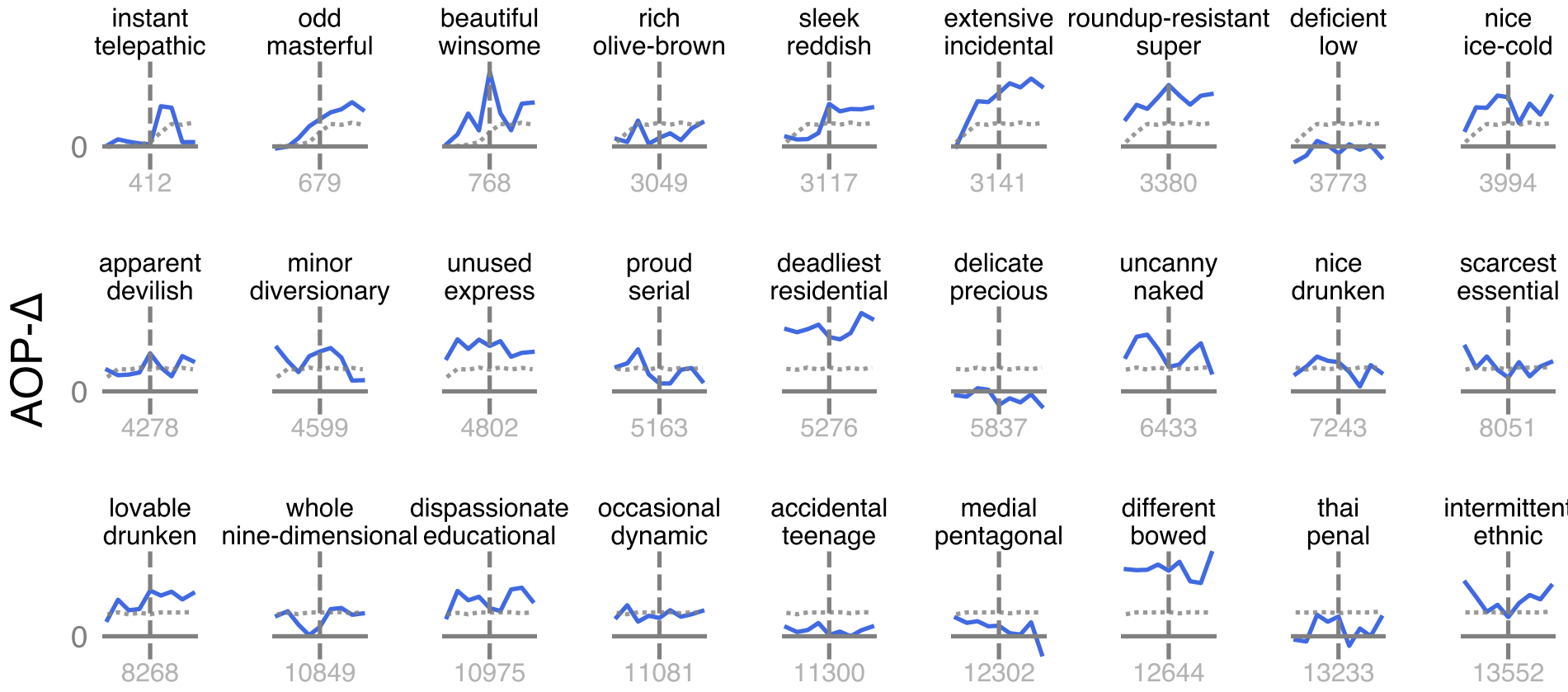

We observe the latter scenario for these 27 adjective pairs. While there are various cases where AOP-Δ is maximal at the point of occurrence (e.g. *beautiful winsome* and *sleek reddish*), in the majority of cases the AOP-Δ curve is not affected by it. There could be various reasons for this. On a macro-scale, the granularity of the checkpoints could be too coarse for the effect to still be visible: from batch 1,000 checkpoints are released every thousand batches, and as such the amount of data seen between two successive checkpoints is over 2 billion token. On a micro-scale, the amount of tokens in a batch could simply be too large for a single bi-gram occurrence to have an immediate effect on AOP. However, this does not explain the trend we observe in Figure 4.5; and numerous works on memorisation in LMs have shown that single-occurrence sequences can be memorised in Pythia (Lesci et al., 2024; Prashanth et al., 2024).

### A.2.4 Context Improvement Ratios

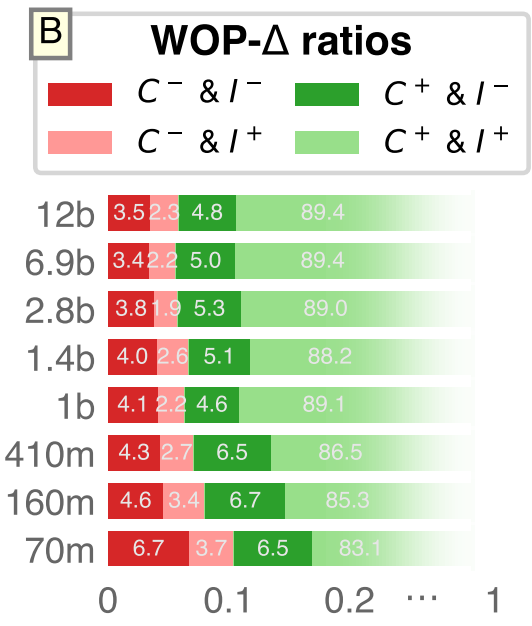

The relative number of items that are affected by context with respect to isolated AOP-Δ scores for each model size. These ratios correspond to the four quadrants of Figure 4.7A.

# A.3 Filtered Corpus Training

## A.3.1 Training Hyperparameters

Selected training hyperparameters. Any omitted values were set to the defaults associated with version 4.30.2 of the `transformers` package.

| | |
|---|---|
| adam_beta1 | 0.9 |
| adam_beta2 | 0.999 |
| adam_epsilon | 1e-08 |
| dataloader_num_workers | 8 |
| evaluation_strategy | epoch |
| fp16 | True |
| gradient_accumulation_steps | 1 |
| ignore_data_skip | True |
| learning_rate | 5e-05 |
| lr_scheduler_type | linear |
| num_train_epochs | 40 |
| per_device_train_batch_size | 32 |
| per_device_eval_batch_size | 32 |
| optim | adamw_torch |
| seed | 0,1,2,3,4 |
| save_strategy | epoch |

## A.3.2 Full Result Tables

Figure A.1 contains the mean accuracies (across random seeds) on all BLiMP benchmarks for both models and every filtered corpus. Figure A.2 contains the paradigm-level $P\Delta$ scores for the FULL and Filtered models, and various Pearson correlations.

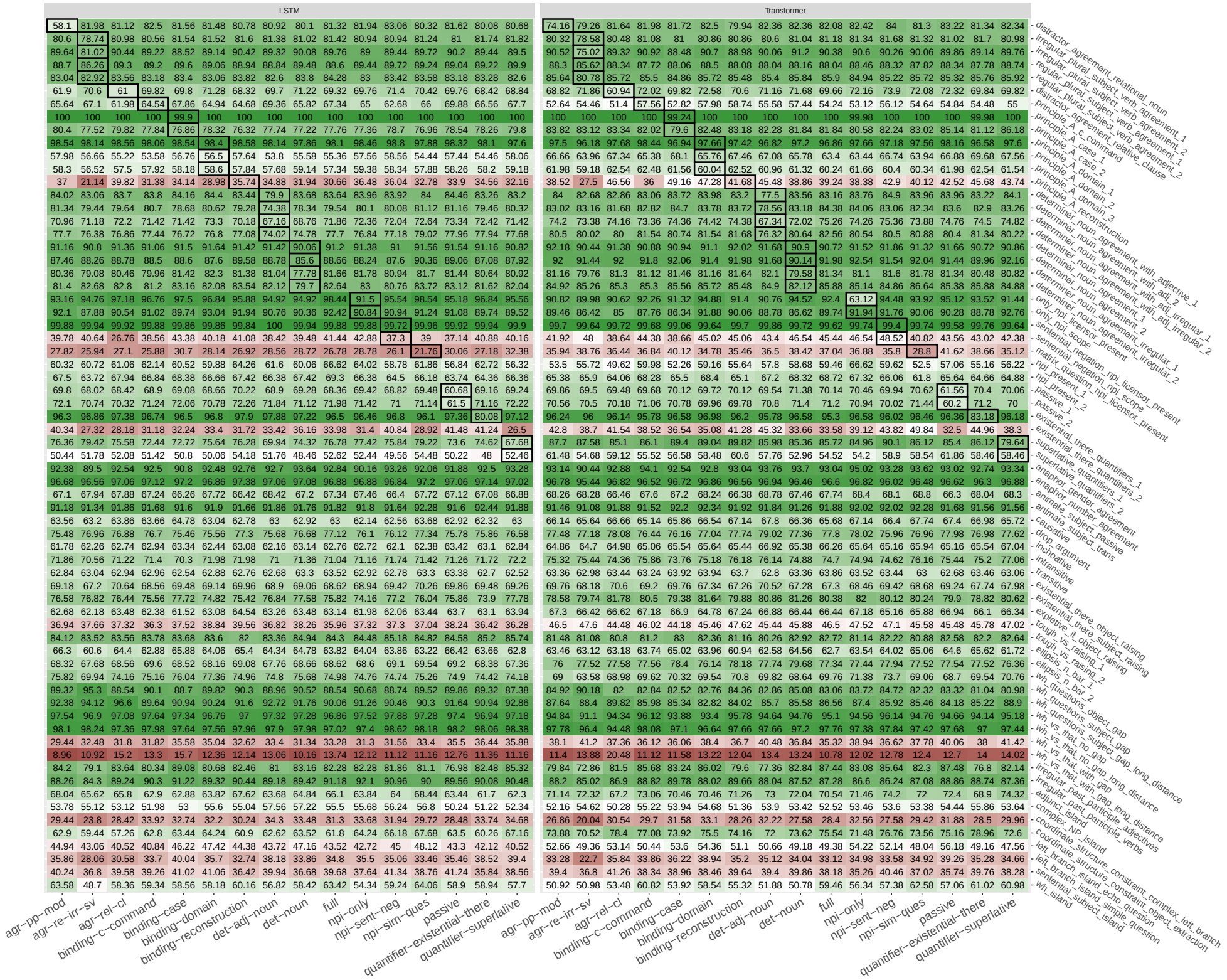

**Figure A.1:** Complete BLiMP benchmark accuracy results for all models, averaged across the five starting seeds for a given training corpus and benchmark. Boxes with bold outlines correspond to benchmarks targeted by the model's corpus filter (i.e. where $F = F(B)$).

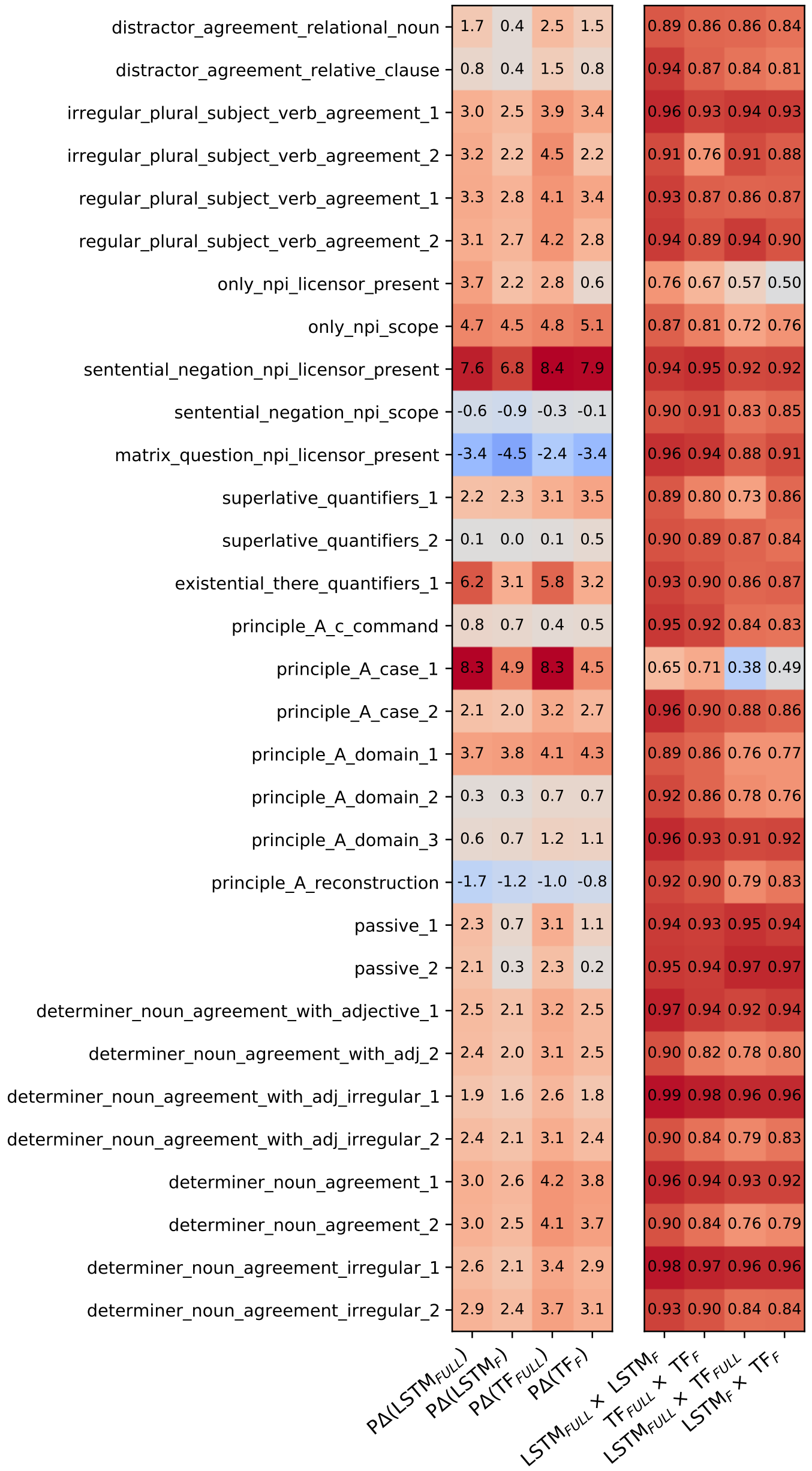

**Figure A.2:** *P*Δ scores for the LSTMs and Transformers (first four columns), and the Pearson correlations between these *P*Δ scores (last four columns).

## A.4 Monotonicity

### A.4.1 Filtered NPIs

We here present the full list of NPIs that were used for filtering sentences from the Full corpus, resulting in the Full\NPI corpus. The method for selecting these expressions is described in §6.5.4.

*A damn, any, any longer, any old, anybody, anymore, anyone, anything, anything like, anytime soon, anywhere, anywhere near, as yet, at all, avail, by much, can possibly, could possibly, ever, in any, in days, in decades, in minutes, in years, just any, just yet, let alone, much help, nor, or anything, set foot, squat, such thing, that many, that much, that often, the slightest, whatever, whatsoever, yet.*

This resulted in a reduction of 75.062 sentences out of the 3.052.726 sentences in the original Full corpus (2.46%).

# A.5 Feature Interactions

## A.5.1 Palindromes

One additional language we investigated is the context-free language of palindromes. In order to process a palindrome, a model needs to keep track of the dependency between each token in the first half of the string with its counterpart in the second half. Palindromes can contain a special symbol in the middle of a string to demarcate the two string halves, making it less ambiguous for the model at which point it should track whether the palindrome is well-formed. In our experiments, however, we found our models to perform well on both forms of palindromes. Furthermore, following Suzgun et al. (2019), we use a homomorphic mapping $h$ for the second half of the string, allowing the model to use separate embeddings for symbols occurring in the first and second half of a string:

$$\mathsf{S} \rightarrow x\ \mathsf{S}\ h(x) \mid \varepsilon \qquad x \in \{a, b, c, \cdots\}$$

We use a corpus size of 5.000, 10 different input symbols, and a maximum sequence length of 18. For the fixed baseline mapping $T(\mathbf{x})$ we map a symbol onto another random symbol, preserving the grammaticality of the palindrome (e.g. $abBA \rightarrow cdDC$).

| | *Static* | | *Expected* | | *Interventional* | | | *Observational* |
|---|---|---|---|---|---|---|---|---|
| | 0 | $T(\mathbf{x})$ | $D^+$ | $D^-$ | $P(x'_i)$ | $P(x'_i \vert i)$ | $P(\mathbf{x}'_{\setminus S})$ | $P(\mathbf{x}'_{\setminus S} \vert \mathbf{x}_S)$ |
| Group Ablation | 0.450 | **0.980** | **1.000** | **0.943** | 0.777 | **0.836** | **1.000** | 0.939 |
| Archipelago | 0.356 | 0.452 | 0.439 | 0.717 | – | – | – | – |
| SII | 0.472 | 0.933 | **1.000** | 0.892 | **0.804** | 0.817 | **1.000** | **1.000** |
| STII | 0.472 | 0.921 | 0.999 | 0.917 | 0.760 | 0.792 | **1.000** | 0.999 |
| Hessian | **0.523** | – | – | – | – | – | – | – |
| Hessian × Input | **0.523** | – | – | – | – | – | – | – |
| IH | 0.505 | 0.637 | 0.693 | 0.535 | – | – | – | – |

**Results** The results for this language, trained with an LSTM, are shown above. Again, the zero-valued baseline performs poorly, with most methods scoring ARRs even below chance level. The fixed baseline mapping again performs well for Group Ablation, SII, and STII, although it is not the best performing baseline this time. These three FIDAMs obtain perfect performance when using the expected baselines over a distribution of well-formed palindromes, which also holds for the interventional baseline with a joint distribution over the missing features. This is in contrast to the Dyck results, where the observational baseline resulted in better ARR scores for all three of these methods.

### A.5.2 ARR Example

An example of a sentence with a high ARR (0.93), for the Group Ablation method with a <pad> baseline:

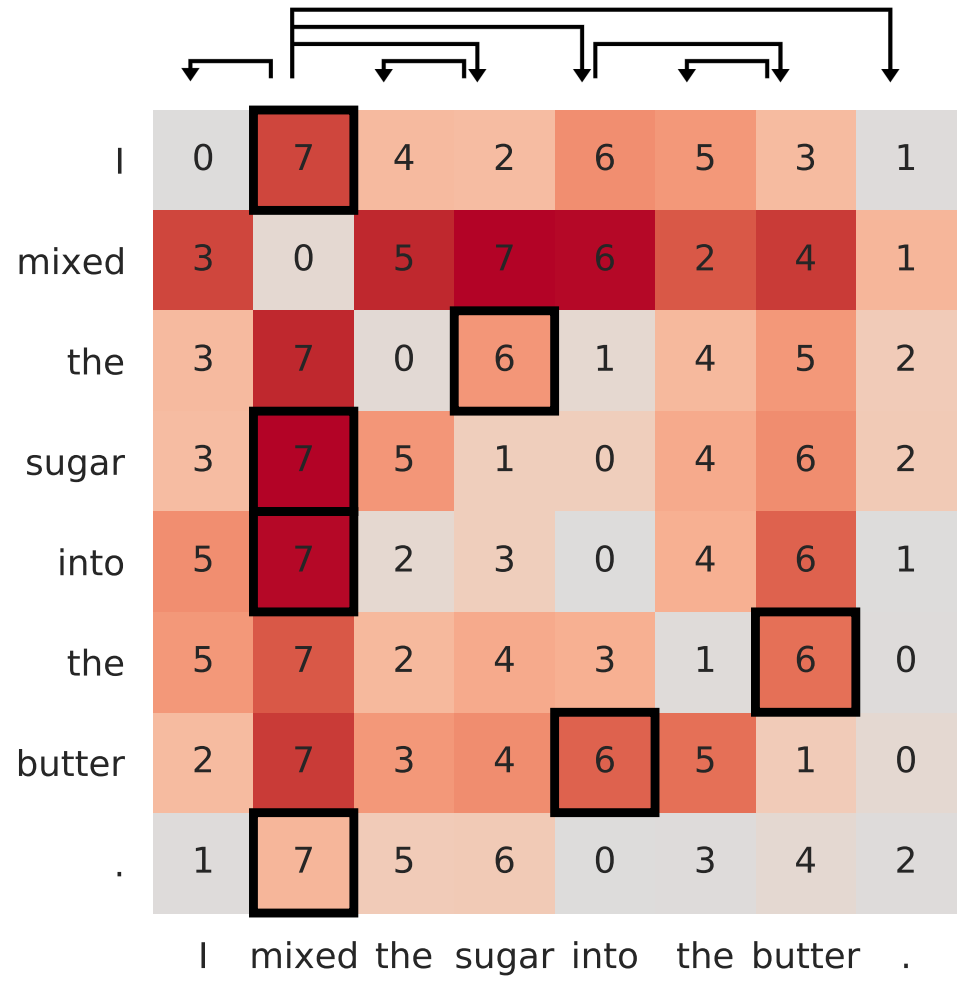

### A.5.3 Correlation CoLA ARR and sentence length

Correlation between sentence length and ARR, shown here for Group Ablation with a <pad> baseline. Spearman's $\rho = -0.38$ ($p << 0.001$):

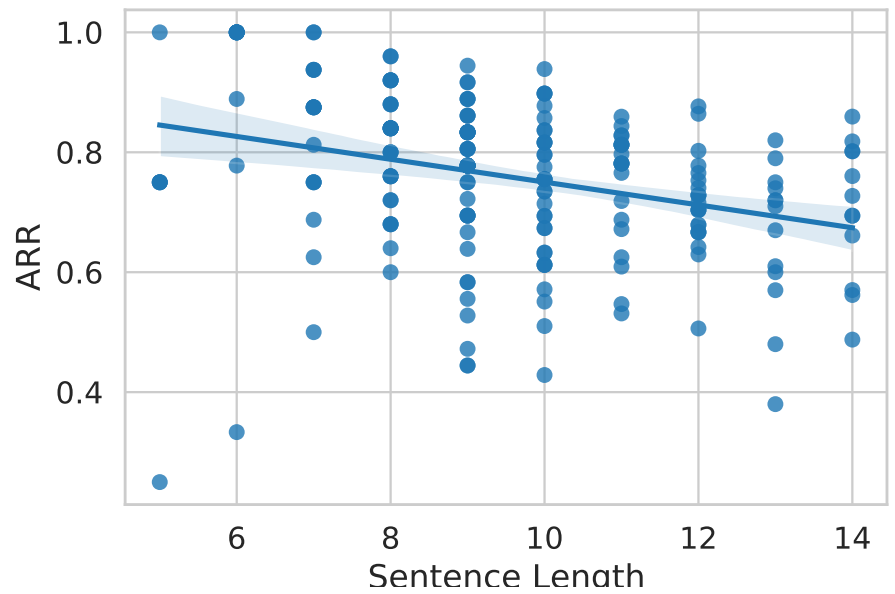

## A.6 Transparency at the Source

### A.6.1 Masked Language Modelling in PCFGs

In this section we demonstrate how the inside and outside probabilities can be used to compute masked token probabilities.

Firstly, the masked token probability for $w_i$ can be expressed as the marginalisation over all non-terminal symbols $N^j_{ii}$ that are the parent of $w_i$:

$$P_G(w_i|w_{\setminus i}) = \sum_j P_G(w_i, N^j_{ii}|w_{\setminus i}) \tag{A.1}$$

which can be rewritten to

$$\sum_j P_G(w_i|N^j_{ii}, w_{\setminus i}) \cdot P_G(N^j_{ii}|w_{\setminus i}) \tag{A.2}$$

The first term is equal to $P_G(w_i|N^j_{ii})$: due to the context-free nature of PCFGs the probability of token $w_i$ solely depends on its parent non-terminal token, and not on neighbouring tokens in the sentence. This term is therefore equal to the inside probability $\beta_j(i,i)$:

$$\begin{aligned} P_G(w_i|N^j_{ii}, w_{\setminus i}) &= P_G(w_{ii}|N^j_{ii}) && \text{[A.3]} \\ &= \beta_j(i,i) && \text{[A.4]} \end{aligned}$$

The second term can be expressed in terms of outside probabilities. In the special case of a substring of length 1 ($w_{ii}$), the words outside of the substring are equivalent to the masked token context $w_{\setminus i}$:

$$\alpha_j(i,i) = P_G(N^j_{ii}, w_{\setminus i}) \tag{A.5}$$

We can expand this as

$$\alpha_j(i,i) = P_G(N^j_{ii}|w_{\setminus i}) \cdot P_G(w_{\setminus i}) \tag{A.6}$$

The first term, $P_G(N^j_{ii}|w_{\setminus i})$, is equal to the second term in Eq. A.2. The second term, $P_G(w_{\setminus i})$, can be marginalised out over all non-terminals $N^k_{ii}$; i.e. over all potential non-terminal parents of token $w_i$. By doing this we are able to express $P_G(N^j_{ii}|w_{\setminus i})$ in terms of

(normalised) outside probabilities:

$$P_G(w_{\backslash i}) = \sum_k P_G(w_{\backslash i}, N^k_{ii}) \quad \text{[A.7]}$$

$$= \sum_k \alpha_k(i,i) \quad \text{[A.8]}$$

$$P_G(N^j_{ii}|w_{\backslash i}) = \frac{\alpha_j(i,i)}{P_G(w_{\backslash i})} \quad \text{[A.9]}$$

$$= \frac{\alpha_j(i,i)}{\sum_k \alpha_k(i,i)} \quad \text{[A.10]}$$

Where Eq. A.9 is a simple reordering of Eq. A.6. Plugging this all in Eq. A.2 we obtain the masked token probability distribution expressed in terms of inside-outside probabilities:

$$P_G(w_i|w_{\backslash i}) = \sum_j \beta_j(i,i) \cdot \frac{\alpha_j(i,i)}{\sum_k \alpha_k(i,i)} \quad \text{[A.11]}$$

### A.6.2 Number of Non-terminal Splits

We plot the number of times each non-terminal type has been split in our final grammar in Figure A.3. The more often a rule is split, the more fine-grained its distribution can become. It can be seen that in general open-class, pre-terminals lead to more state splits, although common non-terminals such as NP, VP, and S have been split often as well.

### A.6.3 Sample of Corpus Sentences

A sample of sentences taken from the evaluation corpus. We provide an example of the parse tree of the first sentence in Figure A.4.

*Her mouth wasn't very close .*

*It met Calvin's lips and closed the wind .*

*The human ceased me to be average .*

*The Captain arrived and went apt of some not to think .*

*Liz spotted us through Bonjour and Monday , traveling around each of Ryland against the doorway .*

*After my injured procedure , you shrugged , backstabbing into the wall .*

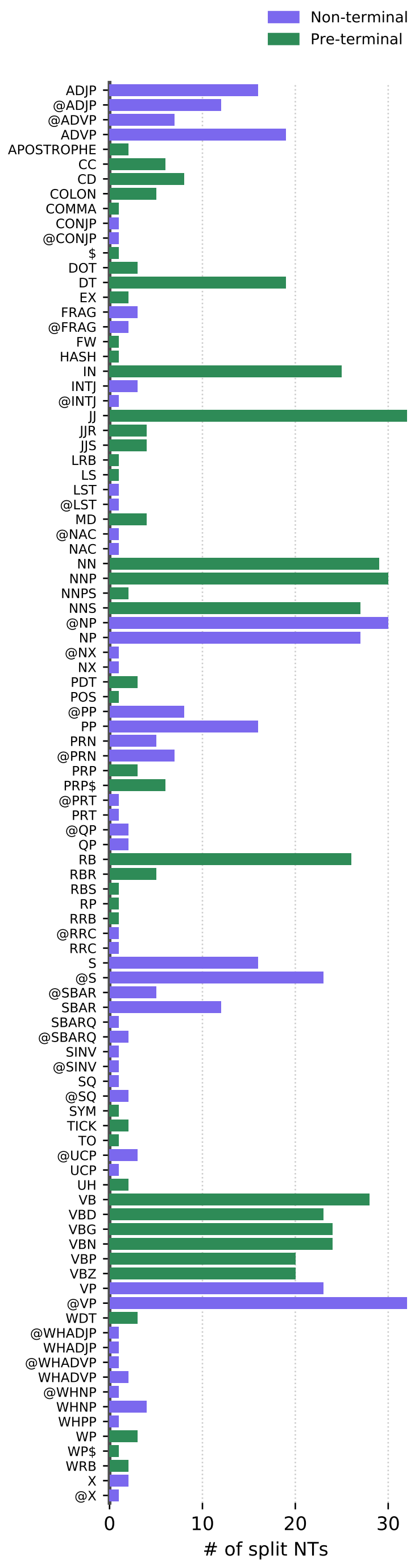

**Figure A.3:** Number of splits per non-terminal. The '@' rules are the result of the X-Bar binarization procedure.

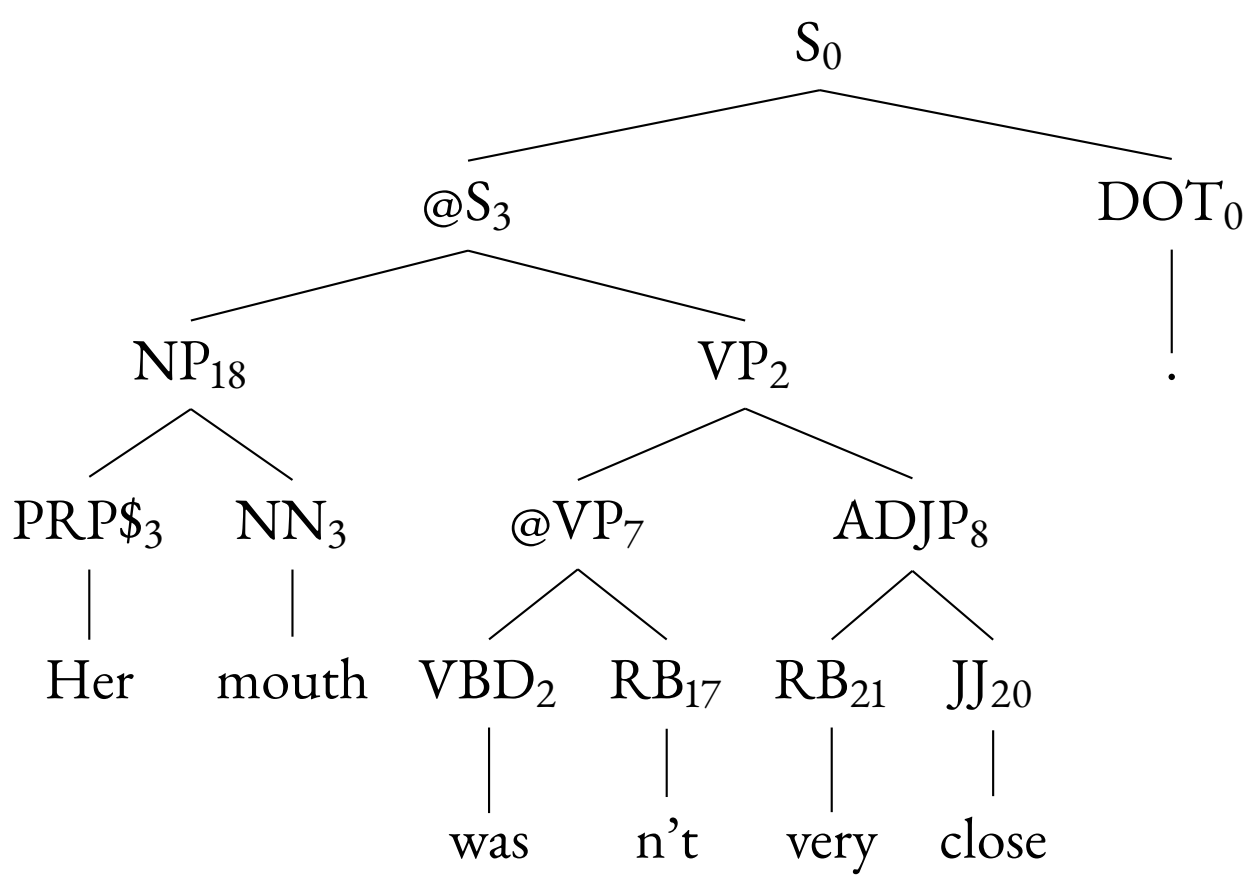

**Figure A.4:** Example tree from the evaluation corpus. The @ non-terminals are the result of the X-Bar binarization procedure of Petrov et al. (2006).

*Josh kept departing thirty hundred alibis no king .*

*As I separates or eventually killed the one top of a giant length , Travis left .*

*A general car was trespassed at tunnels on the image of the misplaced formation .*

*I think that does find a perfect visit to activity .*

# B

# List of Publications

The work described in this thesis originates from the following papers, in their order of publication:

1. Jaap Jumelet, Milica Denic, Jakub Szymanik, Dieuwke Hupkes, and Shane Steinert-Threlkeld. 2021. Language models use monotonicity to assess NPI licensing. In *Findings of the Association for Computational Linguistics: ACL-IJCNLP 2021*, pages 4958–4969, Online. Association for Computational Linguistics

2. Arabella Sinclair*, Jaap Jumelet*, Willem Zuidema, and Raquel Fernández. 2022. Structural persistence in language models: Priming as a window into abstract language representations. *Transactions of the Association for Computational Linguistics*, 10:1031–1050

3. Jaap Jumelet and Willem Zuidema. 2023a. Feature interactions reveal linguistic structure in language models. In *Findings of the Association for Computational Linguistics: ACL 2023*, pages 8697–8712, Toronto, Canada. Association for Computational Linguistics

4. Jaap Jumelet and Willem Zuidema. 2023b. Transparency at the source: Evaluating and interpreting language models with access to the true distribution. In *Findings of the Association for Computational Linguistics: EMNLP 2023*, pages 4354–4369, Singapore. Association for Computational Linguistics

During my PhD I have also contributed to the following papers:

# Structuur Ontrafelen uit Taalmodellen

## Samenvatting

Wanneer we spreken, schrijven of luisteren, creëren we constant voorspellingen die gefundeerd zijn op onze kennis van de grammatica van een taal. Kinderen vergaren deze grammaticale kennis in luttele jaren, wat hen in staat stelt nieuwe constructies te begrijpen zonder deze ooit eerder gehoord te hebben. Taalmodellen vormen representaties van taal door stap voor stap het volgende woord in een zin te voorspellen en deze hebben de laatste jaren een enorme maatschappelijke invloed gehad. De onderzoeksvraag die centraal staat in deze thesis is of zulke modellen een diepgaand begrip van grammaticale structuur bezitten dat vergelijkbaar is met dat van mensen. Deze vraag bevindt zich op het kruispunt van natuurlijke taalverwerking, taalkunde en interpreteerbaarheid. Om tot een antwoord te komen, ontwikkelen we nieuwe technieken voor interpreteerbaarheid die ons een beter begrip verschaffen van de complexe aard van grootschalige taalmodellen. We benaderen onze onderzoeksvraag vanuit drie richtingen. Ten eerste onderzoeken we de aanwezigheid van abstracte linguïstische informatie door middel van *structural priming*, een belangrijk paradigma in de psycholinguïstiek voor het onthullen van grammaticale structuur in menselijke taalverwerking. Vervolgens onderzoeken we verscheidene linguïstische verschijnselen, zoals *adjectiefvolgorde* en *negatief polaire uitdrukkingen*, en verbinden een model zijn begrip van deze verschijnselen aan de tekstdistributie waarop deze is getraind. Tot slot introduceren we een gecontroleerde testomgeving voor het bestuderen van hiërarchische structuren in taalmodellen met behulp van verschillende synthetische talen met toenemende complexiteit. Ook onderzoeken we de rol van *feature interactions* bij het modelleren van deze structuur. Onze bevindingen geven een gedetailleerd verslag van de grammaticale kennis die is ingebed in de representaties van taalmodellen en bieden verschillende richtingen voor het onderzoeken van fundamentele linguïstische vraagstukken met behulp van computationele methoden.

# Finding Structure in Language Models

## Abstract

When we speak, write or listen, we continuously make predictions based on our knowledge of a language's grammar. Remarkably, children acquire this grammatical knowledge within just a few years, enabling them to understand and generalise to novel constructions that have never been uttered before. Language models are powerful tools that create representations of language by incrementally predicting the next word in a sentence, and they have had a tremendous societal impact in recent years. The central research question of this thesis is whether these models possess a deep understanding of grammatical structure similar to that of humans. This question lies at the intersection of natural language processing, linguistics, and interpretability. To address it, we will develop novel interpretability techniques that enhance our understanding of the complex nature of large-scale language models. We approach our research question from three directions. First, we explore the presence of abstract linguistic information through *structural priming*, a key paradigm in psycholinguistics for uncovering grammatical structure in human language processing. Next, we examine various linguistic phenomena, such as *adjective order* and *negative polarity items*, and connect a model's comprehension of these phenomena to the data distribution on which it was trained. Finally, we introduce a controlled testbed for studying hierarchical structure in language models using various synthetic languages of increasing complexity and examine the role of *feature interactions* in modelling this structure. Our findings offer a detailed account of the grammatical knowledge embedded in language model representations and provide several directions for investigating fundamental linguistic questions using computational methods.

# Titles in the ILLC Dissertation Series

ILLC DS-2020-06: **Dieuwke Hupkes**
*Hierarchy and interpretability in neural models of language processing*

ILLC DS-2020-07: **Ana Lucia Vargas Sandoval**
*On the Path to the Truth: Logical & Computational Aspects of Learning*

ILLC DS-2020-08: **Philip Schulz**
*Latent Variable Models for Machine Translation and How to Learn Them*

ILLC DS-2020-09: **Jasmijn Bastings**
*A Tale of Two Sequences: Interpretable and Linguistically-Informed Deep Learning for Natural Language Processing*

ILLC DS-2020-10: **Arnold Kochari**
*Perceiving and communicating magnitudes: Behavioral and electrophysiological studies*

ILLC DS-2020-11: **Marco Del Tredici**
*Linguistic Variation in Online Communities: A Computational Perspective*

ILLC DS-2020-12: **Bastiaan van der Weij**
*Experienced listeners: Modeling the influence of long-term musical exposure on rhythm perception*

ILLC DS-2020-13: **Thom van Gessel**
*Questions in Context*

ILLC DS-2020-14: **Gianluca Grilletti**
*Questions & Quantification: A study of first order inquisitive logic*

ILLC DS-2020-15: **Tom Schoonen**
*Tales of Similarity and Imagination. A modest epistemology of possibility*

ILLC DS-2020-16: **Ilaria Canavotto**
*Where Responsibility Takes You: Logics of Agency, Counterfactuals and Norms*

ILLC DS-2020-17: **Francesca Zaffora Blando**
*Patterns and Probabilities: A Study in Algorithmic Randomness and Computable Learning*

ILLC DS-2021-01: **Yfke Dulek**
*Delegated and Distributed Quantum Computation*

ILLC DS-2021-02: **Elbert J. Booij**
*The Things Before Us: On What it Is to Be an Object*

ILLC DS-2021-03: **Seyyed Hadi Hashemi**
*Modeling Users Interacting with Smart Devices*

ILLC DS-2021-04: **Sophie Arnoult**
*Adjunction in Hierarchical Phrase-Based Translation*

ILLC DS-2021-05: **Cian Guilfoyle Chartier**
*A Pragmatic Defense of Logical Pluralism*

ILLC DS-2021-06: **Zoi Terzopoulou**
*Collective Decisions with Incomplete Individual Opinions*

ILLC DS-2021-07: **Anthia Solaki**
*Logical Models for Bounded Reasoners*

ILLC DS-2021-08: **Michael Sejr Schlichtkrull**
*Incorporating Structure into Neural Models for Language Processing*

ILLC DS-2021-09: **Taichi Uemura**
*Abstract and Concrete Type Theories*

ILLC DS-2021-10: **Levin Hornischer**
*Dynamical Systems via Domains: Toward a Unified Foundation of Symbolic and Non-symbolic Computation*

ILLC DS-2021-11: **Sirin Botan**
*Strategyproof Social Choice for Restricted Domains*

ILLC DS-2021-12: **Michael Cohen**
*Dynamic Introspection*

ILLC DS-2021-13: **Dazhu Li**
*Formal Threads in the Social Fabric: Studies in the Logical Dynamics of Multi-Agent Interaction*

ILLC DS-2021-14: **Álvaro Piedrafita**
*On Span Programs and Quantum Algorithms*

ILLC DS-2022-01: **Anna Bellomo**
*Sums, Numbers and Infinity: Collections in Bolzano's Mathematics and Philosophy*

ILLC DS-2022-02: **Jan Czajkowski**
*Post-Quantum Security of Hash Functions*

ILLC DS-2022-03: **Sonia Ramotowska**
*Quantifying quantifier representations: Experimental studies, computational modeling, and individual differences*

ILLC DS-2022-04: **Ruben Brokkelkamp**
*How Close Does It Get?: From Near-Optimal Network Algorithms to Suboptimal Equilibrium Outcomes*

ILLC DS-2022-05: **Lwenn Bussière-Carae**
*No means No! Speech Acts in Conflict*

ILLC DS-2022-06: **Emma Mojet**
*Observing Disciplines: Data Practices In and Between Disciplines in the 19th and Early 20th Centuries*

ILLC DS-2022-07: **Freek Gerrit Witteveen**
*Quantum information theory and many-body physics*

ILLC DS-2023-01: **Subhasree Patro**
*Quantum Fine-Grained Complexity*

ILLC DS-2023-02: **Arjan Cornelissen**
*Quantum multivariate estimation and span program algorithms*

ILLC DS-2023-03: **Robert Paßmann**
*Logical Structure of Constructive Set Theories*

ILLC DS-2023-04: **Samira Abnar**
*Inductive Biases for Learning Natural Language*

ILLC DS-2023-05: **Dean McHugh**
*Causation and Modality: Models and Meanings*

ILLC DS-2023-06: **Jialiang Yan**
*Monotonicity in Intensional Contexts: Weakening and: Pragmatic Effects under Modals and Attitudes*

ILLC DS-2023-07: **Yiyan Wang**
*Collective Agency: From Philosophical and Logical Perspectives*

ILLC DS-2023-08: **Lei Li**
*Games, Boards and Play: A Logical Perspective*

ILLC DS-2023-09: **Simon Rey**
*Variations on Participatory Budgeting*

ILLC DS-2023-10: **Mario Giulianelli**
*Neural Models of Language Use: Studies of Language Comprehension and Production in Context*

ILLC DS-2023-11: **Guillermo Menéndez Turata**
*Cyclic Proof Systems for Modal Fixpoint Logics*

ILLC DS-2023-12: **Ned J.H. Wontner**
*Views From a Peak: Generalisations and Descriptive Set Theory*

ILLC DS-2024-01: **Jan Rooduijn**
*Fragments and Frame Classes: Towards a Uniform Proof Theory for Modal Fixed Point Logics*

ILLC DS-2024-02: **Bas Cornelissen**
*Measuring musics: Notes on modes, motifs, and melodies*

ILLC DS-2024-03: **Nicola De Cao**
*Entity Centric Neural Models for Natural Language Processing*

ILLC DS-2024-04: **Ece Takmaz**
*Visual and Linguistic Processes in Deep Neural Networks: A Cognitive Perspective*

ILLC DS-2024-05: **Fatemeh Seifan**
*Coalgebraic fixpoint logic Expressivity and completeness result*

ILLC DS-2024-06: **Jana Sotáková**
*Isogenies and Cryptography*

ILLC DS-2024-07: **Marco Degano**
*Indefinites and their values*

ILLC DS-2024-08: **Philip Verduyn Lunel**
*Quantum Position Verification: Loss-tolerant Protocols and Fundamental Limits*

ILLC DS-2024-09: **Rene Allerstorfer**
*Position-based Quantum Cryptography: From Theory towards Practice*

ILLC DS-2024-10: **Willem Feijen**
*Fast, Right, or Best? Algorithms for Practical Optimization Problems*

ILLC DS-2024-11: **Daira Pinto Prieto**
*Combining Uncertain Evidence: Logic and Complexity*

ILLC DS-2024-12: **Yanlin Chen**
*On Quantum Algorithms and Limitations for Convex Optimization and Lattice Problems*